\documentclass[10pt,twocolumn,letterpaper]{article}
\usepackage{floatrow}
\newfloatcommand{capbtabbox}{table}[][\FBwidth]
\usepackage{blindtext}

\usepackage{iccv}
\iccvfinalcopy
\usepackage{times}
\usepackage{epsfig}
\usepackage{graphicx}
\usepackage{amsmath}
\usepackage{tabularray}
\usepackage[dvipsnames]{xcolor}
\usepackage{caption}
\usepackage[accsupp]{axessibility}
\usepackage{multirow}
\usepackage{subcaption}
\usepackage{makecell}
\usepackage{amssymb}
\usepackage[noend]{algpseudocode}
\usepackage{ifthen}
\usepackage[nothing]{algorithm}
\algrenewcommand{\algorithmicrequire}{\textbf{Input:}}
\algrenewcommand{\algorithmicensure}{\textbf{Output:}}
\algnewcommand\And{\textbf{and} }
\usepackage{multicol}
\usepackage{bm}
\usepackage{comment}
\usepackage{booktabs}
\usepackage{authblk}

\DeclareMathOperator*{\argmin}{arg\,min}

\newcommand*{\B}[1]{\ifmmode\bm{#1}\else\textbf{#1}\fi}
\newcommand{\ourmethod}{SHIFT3D}
\usepackage[draft]{pgf}
\usepackage[pagebackref=true,breaklinks=true,letterpaper=true,colorlinks,bookmarks=false]{hyperref}
\iccvfinalcopy 
\def\iccvPaperID{5287} 

\ificcvfinal\fi

\newcommand{\mycomment}[1]{}

\newboolean{roi}
\setboolean{roi}{false}

\begin{document}

\title{\ourmethod: Synthesizing Hard Inputs For Tricking 3D Detectors}

\author[1]{Hongge Chen\thanks{hongge.chen@getcruise.com}}
\author[1]{Zhao Chen}
\author[1]{Gregory P. Meyer}
\author[1]{Dennis Park}
\author[1]{\authorcr Carl Vondrick}
\author[1]{Ashish Shrivastava}
\author[1]{Yuning Chai\thanks{yuning.chai@getcruise.com}}
\affil[1]{Cruise LLC}
\maketitle
\ificcvfinal\thispagestyle{empty}\fi

\begin{abstract}
We present \ourmethod{}, a differentiable pipeline for generating 3D shapes that are structurally plausible yet challenging to 3D object detectors. In safety-critical applications like autonomous driving, discovering such novel challenging objects can offer insight into unknown vulnerabilities of 3D detectors. By representing objects with a signed distanced function (SDF), we show that gradient error signals allow us to smoothly deform the shape or pose of a 3D object in order to confuse a downstream 3D detector. Importantly, the objects generated by \ourmethod{} physically differ from the baseline object yet retain a semantically recognizable shape. Our approach provides interpretable failure modes for modern 3D object detectors, and can aid in preemptive discovery of potential safety risks within 3D perception systems before these risks become critical failures. 


\end{abstract}

\section{Introduction}


As 3D computer vision models become ubiquitous in real-world applications, their reliability becomes a critical safety concern should they fail in undetected or unaddressed ways. Autonomous vehicles, for example, rely heavily on 3D vision, and failures on the road within these systems can lead to collisions and other dangerous events.

Like most statistical models, contemporary 3D detectors are typically susceptible to failure on rare or unknown events encountered in the real world. Most solutions in the field tend to be \textit{reactive}, with more data being collected a posteriori~\cite{emam2021active, jiang2022improving, Wu_2021_ICCV} or models being ad-hoc retrained to target specific already-labeled data~\cite{chen2022gradtail, liu2019large}. However, when a single failure can lead to loss of life on the road, there is an urgent need for detecting these failures in a \textit{proactive} way. Therein lies one of the central challenges of deploying safe AI in practice: a model trained on a finite dataset can perform poorly on data it has not seen, but by definition we cannot just collect unknown data to better understand what a model does not know.

\begin{figure}[t]
\centering
    \includegraphics[width=\linewidth,trim={1cm 14cm 3cm 2cm},clip
    ]{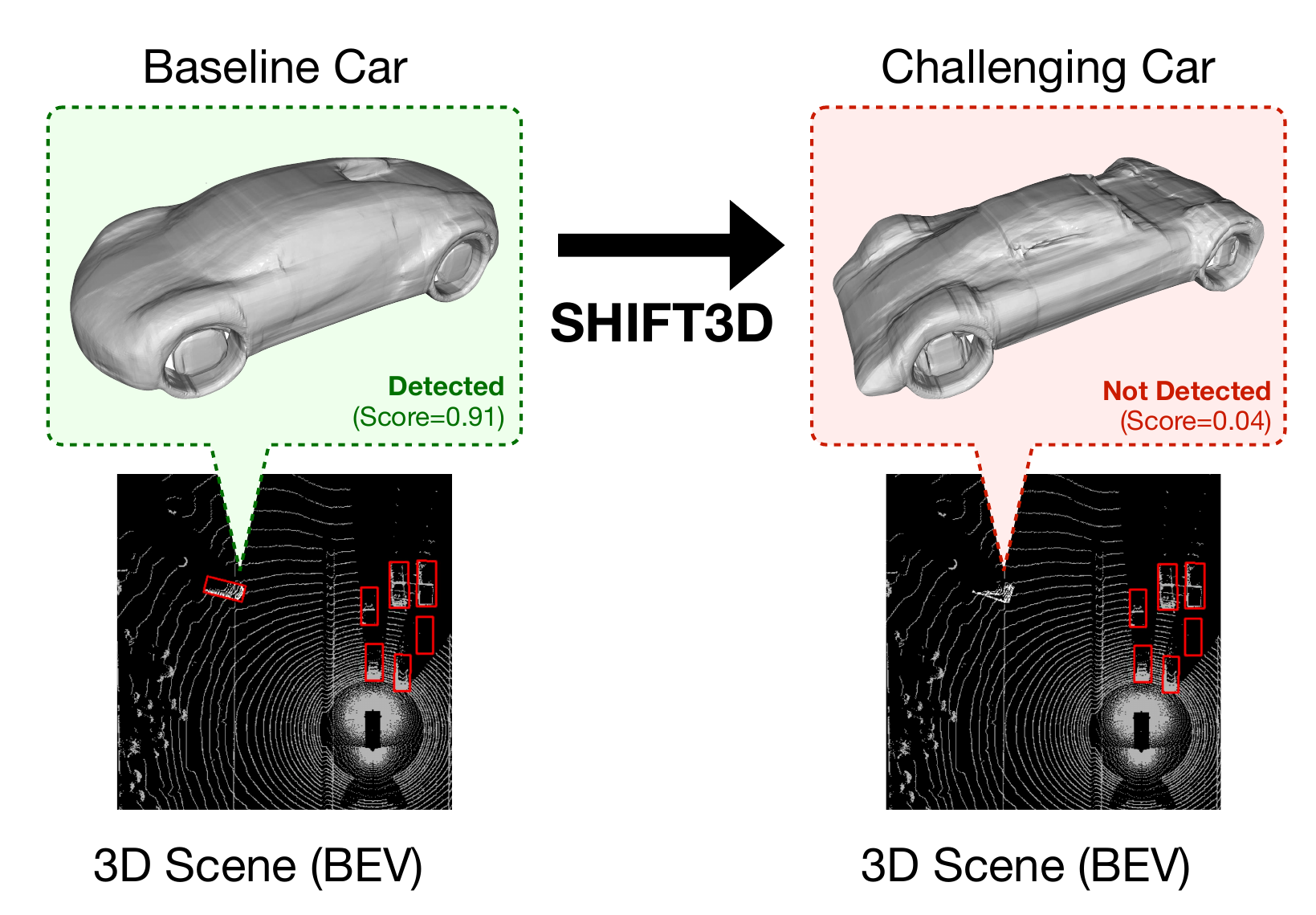}
\caption{\ourmethod{} creates challenging 3D objects from a baseline real object and inserts them with realistic occlusions into a point cloud scene. We also synthesize occlusion patterns, therefore the synthetic objects can be placed in various locations in the scene.}
\end{figure}

In this paper, we propose \ourmethod{} (\textbf{S}ynthesizing \textbf{H}ard \textbf{I}nputs \textbf{f}or \textbf{T}ricking \textbf{3}D \textbf{D}etectors), an approach to proactively synthesizing hard examples for 3D object detection through generative modeling. \ourmethod{} works by perturbing pre-existing objects into hard examples for 3D vision models, and as we demonstrate, works  well in the autonomous driving setting. The perturbative approach provides multiple benefits for us: we can explore new parts of the input distribution which are not explicitly represented by the training data, and the perturbation in the shape space allows us to interpret precisely what changes in shape are most salient in making the example challenging. Since interpretability is a key benefit of this process, it is also important for these perturbations to be \textit{naturalistic}, with changes easily recognizable by the human eye. We use \textit{naturalistic} primarily to contrast with the traditional \textit{adversarial} class of perturbations~\cite{szegedy2013intriguing,xie2017adversarial,goodfellow2014explaining,moosavi2016deepfool}, where minuscule changes produce data that challenges the deep model, but which are imperceptible to the human eye and thus not very helpful for diagnosing model scalability issues in the real world.

Like traditional methods of adversarial attack~\cite{goodfellow2014explaining}, we use gradients as our primary signal to generate challenging input deformations which then confuse the downstream model. 
However, unlike traditional methods of adversarial attack, our method creates 3D objects with observable changes within the object topology that reflect meaningful challenges to the downstream model. To ensure this property, we perform gradient perturbations not on the 3D object directly but only in an appropriate latent space, and allow a decoder to interpret the perturbed latent space into a final 3D object. In this work, the decoder is a pre-trained signed distance function (SDF) network \cite{park2019deepsdf}, which is a popular model to decode a latent vector into a 3D surface. Additionally, we can also perturb the pose of the object without shape deformation. Our synthesized examples are also robust to background scene choice and to insertion location, making it easy to turn a challenging \ourmethod{} object into a fully realized challenging scene.

\ourmethod{} operates as follows: a baseline object is taken from some reference mesh and we obtain a latent encoding $\B z$ for the baseline object from a pretrained DeepSDF~\cite{park2019deepsdf} model. The DeepSDF object is then placed in a new point cloud at pose $\B \theta$ (position and orientation), with postprocessing to add occlusions so the insertion will look physically plausible. This new scene is sent through a pretrained 3D object detector, which assigns a detection score to the inserted object. This pipeline allows us to differentiate the detection score all the way back to the DeepSDF shape encoding $\B z$, along with the pose parameters $\B \theta$. We then find the gradients that \textit{maximize} the detection error and apply these perturbations to the latent encoding as well as to $\B \theta$, which we then decode into a final 3D object. Whether we change $\B z$ (shape) or $\B \theta$ (pose), we will demonstrate that these final 3D objects have reliably low detection scores and visually look substantially different from the baseline object. 

Our main contributions within this work are as follows:

\begin{itemize}
\item We introduce \ourmethod{}, a pipeline to generate challenging 3D objects and insert them with realistic occlusions into a point cloud scene.
\item We show that \ourmethod{} can be made fully differentiable through an implicit differentiation setup, which can extend perturbations beyond the shape of the 3D object and to its placement geometry as well. 
\item We test \ourmethod{} in the autonomous driving setting, and show that the objects generated by \ourmethod{} consistently mislead our LiDAR-based 3D detector and are robustly transferable between different locations within a scene and different scenes as well.
\end{itemize}

\section{Related Works}
\vspace{-0.1cm}
Perception is a critical aspect of ensuring safety and security in autonomous driving systems. As the use of LiDAR sensors increases in autonomous vehicles, point cloud data has become a common input representation. However, recent research has shown that adversarial attacks are effective against point cloud models as well. Various techniques have been proposed to perturb, add, or remove points from the point cloud data~\cite{xiang2019generating,yang2019adversarial,wicker2019robustness,liu2020adversarial,liu2019extending,zheng2019pointcloud,kim2021minimal}. 

3D adversarial attacks on autonomous vehicle systems usually focus on more structurally natural or physically realizable perturbations. LidarAdv~\cite{cao2019adversarial} proposed creating meshes that can deceive LiDAR detectors by using mesh vertices to generate adversarial objects, which were 3D-printed. Similar techniques were employed in~\cite{tu2020physically}, where mesh objects were rendered on top of vehicles in LiDAR log data, which enabled the vehicle to evade detection. Going a step further,~\cite{cao2021invisible} created adversarial objects that could deceive both cameras and LiDAR in a multi-sensor fusion-based perception system. Unlike our proposed method, all the studies thus far on generating adversarial 3D objects for autonomous vehicles produce objects without any semantic meaning, or use mesh-based perturbations on existing real objects, creating traditional adversarial examples that are unrecognizable by the naked eye.

There has also been lines of work investigating adversarial attacks by leveraging generative models~\cite{joshi2019semantic,zhou2020lg,song2018constructing,xiao2018generating}, and generating adversarial images from scratch using rendering engines~\cite{zeng2019adversarial,venkateshsemantic,xiao2019meshadv,alcorn2019strike}. However, these works are mostly in image space. To the best of our knowledge, \ourmethod{} is the first work that can generate plausible-looking challenging objects and realistically render them into a full 3D point cloud scene. 

\section{Method}
\vspace{-0.1cm}



\begin{figure*}[th]
    \centering
    \includegraphics[width=0.9\textwidth]{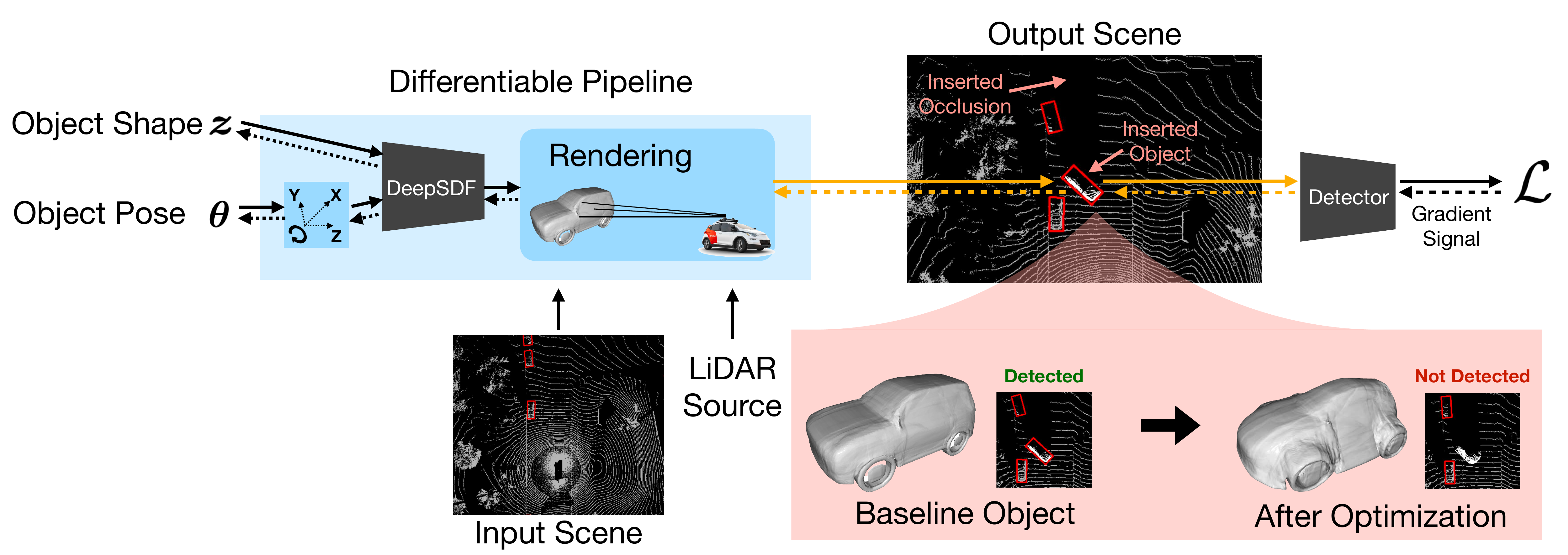}
    \caption{Overview of insertion process of \ourmethod{} objects into a LiDAR scene. We use a DeepSDF model to locate points on an object's surface given its shape ($\B z$) and pose ($\B \theta$). We reconstruct beams by connecting input sensor position to existing LiDAR points in the input scene. If a beam intersects with the \ourmethod{} object, the corresponding point is moved to the object's surface. \ourmethod{} objects are optimized to deceive a detector model. Starting with an baseline shape or pose, we render the object in the input scene and then minimize detection model's score. We alternate between rendering the object and minimizing the score for several steps until an adversarial object is produced. }
    \label{fig:pipeline}
\end{figure*}
\ourmethod{} manipulates the shape and pose of an object in a way that makes it difficult to detect for a 3D vision model, and thus we must create a pipeline that can take an object's shape and pose and render it in a given scene. 
Importantly, this pipeline will be differentiable with respect to both the shape and pose parameters, which allows us to differentiate the detector loss all the way back to these parameters. These gradients allow perturbations in these parameters that then generate adversarial examples.

To ensure that the pipeline is differentiable, we start with a Sign Distance Function (SDF)~\cite{park2019deepsdf} network $g$. We can use $g$ to find a latent shape embedding $\B z$ that decodes to a given shape, and then insert that shape with pose $\B \theta$ into a scene with realistically rendered occlusions.
The latter is possible because $g$ allows us to query every pre-existing point in the original scene and determine whether that point is now behind the inserted object (i.e. occluded). Occluded points are moved to the inserted object's surface. 

We then minimize the detection score of the object detector, $f$. In this work, we assume we are in the white-box setting where we have access to the detector model weights, and thus can directly differentiate through the detector. 
Unlike traditional adversarial algorithms that directly perturb the 3D points of the object, we only modify the shape latent parameters, $\B z$, or the pose parameters, $\B \theta$; this process helps ensure that the perturbations look natural. 

Next, we describe the SDF model, rendering, and \ourmethod{} optimization in more detail.


\vspace{-0.1cm}\subsection{Object Representation using DeepSDF Model}
SDF is an implicit representation of the object's shape and is learned using a deep neural network~\cite{park2019deepsdf} and a dataset of 3D objects. 
The DeepSDF function, $g$, is trained to evaluate to zero for the points on the object's surface.
That is, given a shape vector $\boldsymbol{z} \in \mathbf{R}^{d_{\B z}}$ of an object and a point $\boldsymbol{x} \in \mathbf{R}^3$ in space, the function $g(\boldsymbol{z}, \boldsymbol{x}) = 0$ if $\boldsymbol{x}$ is on the object's surface, $g(\boldsymbol{z}, \boldsymbol{x}) < 0$ if $\boldsymbol{x}$ is inside the object, and $g(\boldsymbol{z}, \boldsymbol{x}) > 0$ if $\boldsymbol{x}$ is outside the object.
In other words, an object's surface in 3D space is precisely the level set of all points $\boldsymbol{x}$ for a given object shape $\boldsymbol{z}$ where $g(\boldsymbol{z}, \boldsymbol{x}) = 0$.

\vspace{-0.1cm}\subsection{Object Rendering in LiDAR scene}
\label{sec:rendering}
 To render an object with a given pose into a LiDAR scene, we first extend LiDAR beams from the source to points in the original point cloud. If any beam intersects the inserted object, that point is now behind the object, and so we move the corresponding point to the surface of the object. We assume that no object points are dropped due to sensor noise or object material type.
See Figure~\ref{fig:pipeline} for an overview of the rendering process.

Assume an object has shape $\B z$ and pose $\B \theta$.
Let $\hat{\B x}_i$ be a LiDAR point in the input scene and and $\B x_i$ be the corresponding point in the output scene after rendering the object. We can reconstruct the laser beams in the scene by drawing a straight line connecting the sensor with each $\hat{\B x}_i$.
Let $\hat{\B x}_i^{(j)}$ be the $j^\text{th}$ sample point on the line.
To determine whether a sampled point, $\hat{\B x}_i^{(j)}$, lies inside the object, we evaluate it using the SDF function $g$.
Since the SDF function is trained in the object coordinate frame, we need to transfer the point  $\hat{\B x}_i^{(j)}$ into the object coordinate frame using the pose of the object.
Let this transformation from sensor coordinate to object coordinate be represented by $T(\cdot; \B \theta)$, which is parameterized by the pose of the object or the relative transform between the sensor and the object.
To put it concisely, the point, $\hat{\B x}_i$, is occluded by the inserted object if $g(\B z, T(\hat{\B x}_i^{(j)}; \B \theta)) < 0$ for any $j$. If this condition is met, then we need to move $ \hat{\B x}_i$ to the object's surface. To do this, we perform a binary search to find the point $\B x_i$, that lies between $\B \hat{\B x}_i^{(j)}$ and the LiDAR sensor, and is on the object's surface within some tolerance $\epsilon$; i.e. , $\left|g(\B z, T(\B x_i; \B \theta))\right| < \epsilon$. On the other hand, if $g(\B z, T(\hat{\B x}_i^{(j)}; \B \theta)) > 0$ holds for all $j$, $ \hat{\B x}_i$ remains in its original position, i.e. $\B x_i \gets \hat{\B x}_i$. Note that we only use laser beams reconstructed from existing LiDAR points in the input scene, and do not create new laser beams. More etailed information of rendering are presented in Section~\ref{sec:supp_rendering} of the supplementary material.

\ifthenelse{\boolean{roi}}{
  Casting beams induced by every LiDAR point in the scene can be both computationally expensive and unnecessary. To optimize this process, a smaller region of interest (ROI) is defined within the scene, outside of which no beams will intersect with the object. By evaluating only the points inside this region and leaving the remaining points unchanged, we significantly reduce the number of points processed, resulting in a faster rendering (see ROI in Section \ref{sec:exp_setup}).  
}{}

\vspace{-0.1cm}\subsubsection{Realism Constraints on Object Pose} \label{sec:physical_constraints}
We can also improve rendering quality by ensuring that the rendered object satisfies the following physical constraints:

\textbf{The SDF object does not overlap with the scene objects.} To prevent the SDF object from overlapping with existing objects in the input scene, we avoid object positions that would result in any point in the scene having an SDF value less than $-\epsilon_\text{overlap}<0$.

\textbf{The SDF object touches the ground.} 
To ensure that the SDF object is in contact with the ground, we enforce that at least one point in the original scene has SDF value greater than, $\epsilon_{\text{float}}>0$.


\vspace{-0.1cm}\subsection{Natural Adversarial Perturbation of the Object}
The rendered scene can be represented with a set of $n$ 3D points $\mathcal X = \{\B x_i\}$, $i=1, \dots, n$.
The detector takes $\mathcal X$ as input and produces a set of bounding boxes and their corresponding detection scores.
Let the set of proposed bounding boxes and their scores be denoted by $\mathcal Y = \{(\B y_j, p_j)\}$, where $\B y_j$ denotes the bounding box coordinates for the $j$-th bounding box, $p_j$ is the corresponding detection score.
The number of bounding boxes proposed depends on the scene size and the anchor resolution in the detector architecture. We then render the object in the scene with a given $\B z$ and $\B \theta$, and compute the gradients w.r.t. $\B z$ or $\B \theta$ to minimize the detection score. After updating $\B z$ or $\B \theta$, we re-render the object. We alternate between updating the parameters and re-rendering until we achieve a desired detection score.

Note that though the rendering pipeline involves beam casting with binary search, which is not a differentiable operation, once a specific point $\B x_i$ is found and is affixed to the \ourmethod{} object's surface, the rest of the pipeline is fully differentiable w.r.t. $\B z$ and $\B \theta$. We now discuss the differentiation process and loss function in detail. 

\vspace{-0.1cm}\subsubsection{Adversarial Loss Functions}
To fool the detector, we define the adversarial loss:
\begin{equation}
\mathcal{L}_{\text{adv}}=\sum_{(\B y_i, p_i) \in \mathcal Y}-\text{IoU}(\B y^*,\B y_i)\log(1-p_i),
\label{eq:adv_loss}
\end{equation}
where $\B y^*$ is the inserted object's ground-truth bounding box, and IoU represents standard intersection over union function.
A similar adversarial function has been used in \cite{tu2020physically} and \cite{xie2017adversarial}.
Our adversarial loss function is designed to suppress all bounding box proposals that overlap with the ground truth, with the degree of suppression weighted by the amount of overlap.
During optimization, we only back-propagate through the the detection confidence, $p_i$, to focus on minimizing the detection score rather than the bounding box parameters.

Given an input shape, $\B z_0$, we regularize the perturbed shape, $\B z$, to be close to $\B z_0$.
Specifically, we add an $\ell_2$ regularization to $\mathcal L_{\text{adv}}$ and it prevents the perturbed shape from significantly diverging from the original shape.
\begin{equation}
\min_{\B z}\mathcal{L}(\B z)=\min_{\B z}\mathcal{L}_{\text{adv}}(\B z)+\lambda \lVert \B z - \B z_0 \rVert_2^2,
\label{eq:loss_z}
\end{equation}
where $\lambda$ is a hyper-parameter that controls the strength of regularization. 

When optimizing object pose, $\B\theta$, we apply hard constraints to restrict the $\B \theta$ value within an $\ell_2$ norm ball $\mathcal{B}(\B \theta_0)$ centered at the input pose, $\B \theta_0$.
This hard constraint helps to ensure that the object does not move too far away from the baseline position.
The optimization problem for the pose parameters can be written as,
\begin{equation}
\min_{\B \theta\in\mathcal{B}(\B \theta_0)}\mathcal{L}(\B \theta)=\min_{\B \theta\in\mathcal{B}(\B \theta_0)}\mathcal{L}_{\text{adv}}(\B \theta).
\label{eq:loss_theta}
\end{equation}
We note that during this process, any change made to the object's pose will also necessitate a corresponding modification of its ground truth $y^*$ in $\mathcal{L_{\text{adv}}}$ at each step.

\vspace{-0.1cm}\subsubsection{Optimization}
To optimize the loss functions in Equations~\eqref{eq:loss_z} and \eqref{eq:loss_theta}, we need to back-propagate through both the detector model, $f$, and the the object's SDF function, $g$.
The gradients $d\mathcal{L}/d\B z$ and $d\mathcal{L}/d\B \theta$ can easily be computed using implicit differentiation, and are given by,
\begin{equation}
\frac{d\mathcal{L}}{d\B z} = \sum_{i=1}^nm_i\frac{d\B x_i}{d\B z}\frac{d\mathcal{L}}{d\B x_i}, \label{eq:chain_rule_z}
\end{equation}
\begin{equation}
\frac{d\mathcal{L}}{d\B \theta} = \sum_{i=1}^nm_i\frac{d\B x_i}{d\B \theta}\frac{d\mathcal{L}}{d\B x_i},
    \label{eq:chain_rule_theta}
\end{equation}
where the mask $m_i$ has value of $1$ if $\B x_i$ is on object's surface and $0$ otherwise.
$m_i$ helps remove any gradients from background points that do not contribute to the object's shape or pose.
The gradients $d\B x_i/d\B z$ and $d\B x_i/d\B \theta$ are $d_{\B z}\times 3$ and $d_{\B \theta}\times 3$ Jacobian matrices, respectively. 
For the points on the object's surface, these gradients can easily be determined through implicit differentiation, since they satisfy the constraints $g(\B z, \B x_i)=0$. 

\vspace{-0.1cm}\subsubsection{Gradients for Adversarial Shape}\label{sec:diff_z}
To compute $d\B x_i/d \B z$ in Eq.~\eqref{eq:chain_rule_z}, we parameterize the LiDAR points as $\bm{x}_i=k_i\B e_i + \B s$, where $\B s \in \mathbb R^3$ denotes the LiDAR sensor position, $\B e_i \in \mathbb R^3$ is a unit vector in the direction of $\B x_i$ positioned at $\B s$,  and $k_i$ is the distance of the point from the sensor.
With this parameterization, we can represent $\frac{d\B x_i}{d \B z} = \frac{dk_i}{d\B z}\B e_i^T$, and Eq.~\eqref{eq:chain_rule_z} can be written as,
\begin{equation}
\frac{d \mathcal{L}}{d \B z}=\sum_{i=1}^{n}m_i\left(\B e_i\cdot\frac{d \mathcal{L}}{d \B x_i}\right)\frac{d k_i}{d \B z}.
\label{eq:dl_dz}
\end{equation}
Since the points $\bm{x}_i$ on the object ($m_i = 1$) are subject to the constraint $g(\B z, \B x_i)=0$, we can use implicit differentiation to calculate $d k_i/d \B z$, and write Eq.~\eqref{eq:dl_dz} as, 
\begin{flalign}
   & \frac{d \mathcal{L}}{d \B z} =-\sum_{i=1}^{n}m_i\left(\B e_i\cdot\frac{d \mathcal{L}}{d \B x_i}\right)\left(\frac{\partial g}{\partial k_i}\right)^{-1}\frac{\partial g(\cdot, \B x_i)}{\partial \B z} \qquad \qquad \qquad \qquad \qquad \nonumber \\
   &=-\sum_{i=1}^{n}m_i\left(\B e_i\cdot\frac{d \mathcal{L}}{d \B x_i}\right)\left(\B e_i\cdot\frac{\partial g(\B z, \cdot)}{\partial \B x_i}\right)^{-1}\frac{\partial g(\cdot, \B x_i)}{\partial \B z}.
\label{eq:z_final}
\end{flalign}

\vspace{-0.5cm}\subsubsection{Gradients for Adversarial Pose }\label{sec:diff_theta}
Similar to shape gradient, $dk_i/d \B z$ , we can compute $d k_i/d\B \theta$ and use implicit differentiation to update Eq.~\eqref{eq:chain_rule_theta} as,
\begin{flalign}
&\frac{d \mathcal{L}}{d \B\theta}=\sum_{i=1}^{n}m_i\left(\frac{d\B s}{d\B\theta}+k_i\frac{d\B e_i}{d\B\theta}+\frac{dk_i}{d\B\theta}\B e_i^T\right)\frac{d \mathcal{L}}{d \B x_i}. \nonumber \\
&  =\sum_{i=1}^{n}m_i\left[\frac{d\B s}{d\B\theta}+k_i\frac{d\B e_i}{d\B\theta}-\left(\frac{\partial g}{\partial k_i}\right)^{-1}\frac{\partial g}{\partial \B\theta}\B e_i^T\right]\frac{d \mathcal{L}}{d \B x_i} \nonumber \\
&=\sum_{i=1}^{n}m_i\left(\frac{d\B s}{d\B\theta}+k_i\frac{d\B e_i}{d\B\theta}\right)\left[\mathbf{I}-\left(\B e_i\cdot\frac{\partial g}{\partial \B x_i}\right)^{-1}\frac{\partial g}{\partial \B x_i}\B e_i^T\right]\frac{d \mathcal{L}}{d \B x_i}.
\label{eq:theta_final}
\end{flalign}
We present the overall procedure for adversarial shape and pose generation in Algorithm~\ref{algo:adv_z_theta}. Note that for pose generation, after gradient descent, we also need to project the resulting $\B \theta$ back to $\mathcal{B}(\B \theta_0)$ to satisfy the hard constraints. Details of calculating gradients in \eqref{eq:z_final} and \eqref{eq:theta_final} are discussed in Section~\ref{sec:auto_diff} of the supplementary materials.

\begin{algorithm}
\caption{Pseudocode for adversarial perturbations of an object's shape or pose in a given LiDAR scene.}\label{algo:adv_z_theta}
\begin{algorithmic}[1]
\Require Object SDF model $g(\cdot, \cdot)$, initial shape, $\B z_0$, or pose parameters, $\B \theta_0$, detector model $f$, LiDAR sensor positions $\B s$, input scene $\{ \hat{\B x}_i\}$, maximum iterations $N_{\text{iter}}$, and learning rate $\alpha$.
\Ensure Adversarial shape, $\B z_{\text{adv}}$, or pose, $\B \theta_{\text{adv}}$, parameters.
\State Calculate LiDAR beam length $\hat{k}_i\gets \lVert\hat{\B x}_i-\B s\rVert_2$.
\State Calculate LiDAR beam directions $\B e_i\gets(\hat{\B x}_i-\B s)/\hat{k}_i$.
\State $\B z \gets \B z_0$ or $\B \theta \gets \B \theta_0$; $\mathcal{L}_{\text{min}}\gets\infty$.
\State Render object: calculate rendered points $\{\B x_i\}$, and set $m_i \gets 1$ when points on the object. (Section ~\ref{sec:rendering}).
\While{iteration $< N_{\text{iter}}$ }
\State Calculate $d\mathcal{L}/d\B z$ using Eq.~\eqref{eq:z_final} or $d\mathcal{L}/d\B \theta$ using Eq.~\eqref{eq:theta_final}.
\State $\B z \gets \B z - \alpha \cdot d\mathcal{L}/d\B z$ or   $\B \theta \gets \text{Proj}_{\mathcal{B}(\B \theta_0)}[\B \theta - \alpha \cdot d\mathcal{L}/d\B \theta]$.
\State Re-render object with the updated $\B z$ or $\B \theta$ in the input scene and update $\{\B x_i\}$ and $m_i$.
\State Forward propagate to obtain $\mathcal{L}$.
\If{$\B x_i$ satisfy all realism constraints}
\State $\mathcal{L}_{\text{min}}\gets\min\{\mathcal{L}_{\text{min}},\ \mathcal{L}\}$.
\If{$\mathcal{L}=\mathcal{L}_{\text{min}}$}
    \State $\B z_{\text{adv}} \gets \B z$ \; \; or \; \; $\B \theta_{\text{adv}} \gets \B \theta$.
\EndIf
\EndIf
\EndWhile
\end{algorithmic}
\end{algorithm}









\section{Experiments}
\vspace{-0.1cm}

We first describe our datasets and models, followed by an overview of our experimental setup.
We then dive into results, highlighting that \ourmethod{} outputs are effective at confusing detectors and robust along various axes, while demonstrating the potential insights \ourmethod{} can provide for improving detection models in the real world.

\vspace{-0.1cm}\subsection{Datasets and Models}\label{sec:exp_data_model}
We utilize the Waymo Open Dataset (WOD) \cite{sun2020scalability}, which provides LiDAR point clouds and 3D bounding box annotations of objects detected by LiDAR sensors mounted on autonomous vehicles. The large scale and high quality of WOD, coupled with its extensive geographical coverage, make it an excellent benchmark for evaluating a variety of scenarios. 
Following the approach of \cite{tu2020physically}, we focus on the ``Vehicle" category in WOD and restrict our evaluation to 2D bounding boxes in a bird's eye view.

Our object detection models are based on the PointPillars architecture~\cite{lang2019pointpillars} and SST~\cite{fan2022embracing}, which operate solely on point cloud data without utilizing auxiliary inputs such as intensity. Specifically, PointPillars partitions the input points into discrete bins (i.e., pillars) from a bird's eye view, and for each pillar, extracts features using PointNet. SST uses a single-stride sparse transformer to maintain the original resolution from the beginning to the end of the network. 
We trained our model on the WOD training set and use the validation set of WOD as input scenes for rendering objects. On the vanilla detection task, our PointPillars and SST models achieve comparable performance with the baseline in~\cite{sun2020scalability} (refer to Section~\ref{sec:supp_experiments} of the supplementary material for more on baseline detection metrics, such as AP and APH).

Our DeepSDF model is trained on the ``Automobile" category of ShapeNet~\cite{chang2015shapenet}. 
We used the same training procedure and model architecture as the original DeepSDF paper~\cite{park2019deepsdf}, where the latent shape encoding $\B z$ has a dimension of $d_{\B z} = 256$. 
We selected 5 representative vehicle objects from the ``Automobile" category as our baseline objects: Coupe, Sports Car, SUV, Convertible, and Beach Wagon. 
The meshes of these objects reconstructed by DeepSDF are depicted in the third column of Figure~\ref{fig:z_figures}. To ensure that the dimensions of each object are realistic and consistent with the natural vehicles in the scene, we manually scaled each baseline object.

\vspace{-0.1cm}\subsection{Setup}\label{sec:exp_setup}

We insert our baseline SDF objects into $500$ randomly chosen WOD validation set scenes. 
To ensure diversity in object placement, we randomly positioned the inserted objects between $15$ and $50$m from the autonomous vehicle. 
The object's heading was randomly chosen between 0 and $2\pi$. 
Keeping objects at least $15$m away reduces the number of points rendered on the object and avoids out-of-memory issues. To describe the pose transformations, we use a $6$-dimensional vector $\B \theta$ consisting of the coordinates $(x, y, z)$ and the angles of yaw, pitch, and roll.  

\ifthenelse{\boolean{roi}}{}{
We draw a sphere with a $7$m radius centered at the location of the inserted object. If we emit a beam from the LiDAR source to any point, and that beam lies completely outside this sphere, we ignore that point as it is far enough away from the inserted object to not affect the rendering process. For additional implementation details, please refer to Section~\ref{sec:supp_rendering} in supplementary material.
}



\ifthenelse{\boolean{roi}}{
  To define the region of interest (ROI) for a scene, we follow a three-step process. First, we draw a sphere with a $7$m radius centered at the location of the baseline objects in each scene. Next, we create a cone with its apex at the LiDAR sensor, adjusting the vertex angle to ensure that the cone's curved surface is tangent to the sphere. Finally, we exclude from the ROI the truncated section of the cone near its apex that cannot be reached by any point within the sphere. If there are multiple LiDAR sensors on the autonomous vehicle, we construct an ROI for each of them and take their union to obtain the final ROI.
}{
}

\vspace{-0.1cm}\subsection{Hyper-parameters}~\label{sec:hyperparams}
We run $40$ steps of adversarial perturbation. 
During optimization, we record the result with lowest loss that also satisfies all constraints specified in Section~\ref{sec:physical_constraints}. 
We pick a learning rate of 0.01 for all experiments after extensive hyperparameter search. 
In adversarial shape optimization, we select a $\lambda$ of either 1 or 10, depending on which results in the lowest loss value $\mathcal{L}_{\text{adv}}(\B z)$. 
However, in cases where different settings of $\lambda$ produce very similar loss values, we will choose the larger $\lambda$ as it keeps the perturbed shape closer to the original. 
Refer to Section~\ref{sec:supp_experiments} of the supplementary material for detailed discussion on hyper-parameters.
To achieve realistic object placement, we set $\epsilon_{\text{overlap}}=-0.02$ and $\epsilon_{\text{float}}=0.02$, which are defined in Section~\ref{sec:physical_constraints}.
Additionally, we exclude cases where the inserted SDF objects may be occluded by other objects in the scene, by disallowing inserted objects that result in fewer than $300$ on-object points post rendering. Each object is placed randomly with the above realism constraints into $500$ scenes. Note that our focus is on the detector’s performance under varying conditions, including rare ones. So we consciously opted not to enforce additional traffic constraints we placing the inserted objects.

\begin{figure*}[h!]
       \centering

\begin{subfigure}[b]{0.196\textwidth}
\includegraphics[width=\textwidth,]{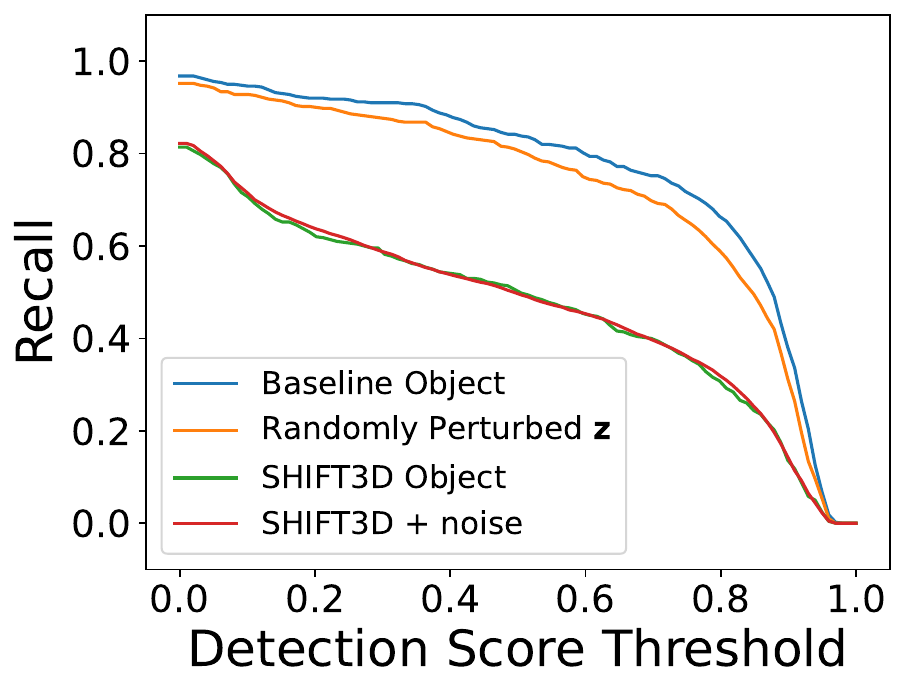}
\caption{Coupe}
\end{subfigure}
\begin{subfigure}[b]{0.196\textwidth}
\includegraphics[width=\textwidth,]{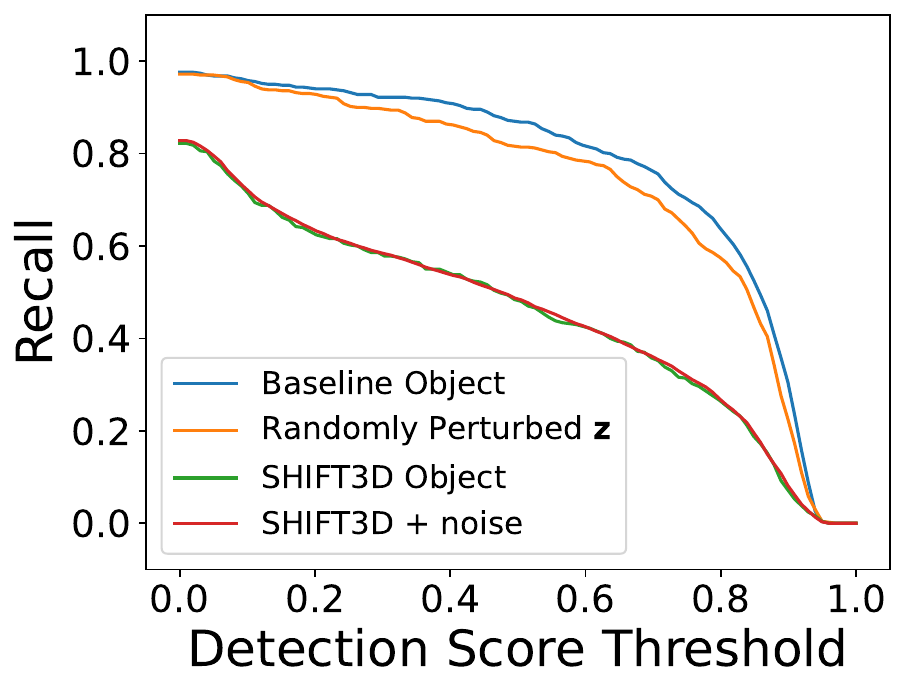}
\caption{Sports Car}
\end{subfigure}
\begin{subfigure}[b]{0.196\textwidth}
\includegraphics[width=\textwidth,]{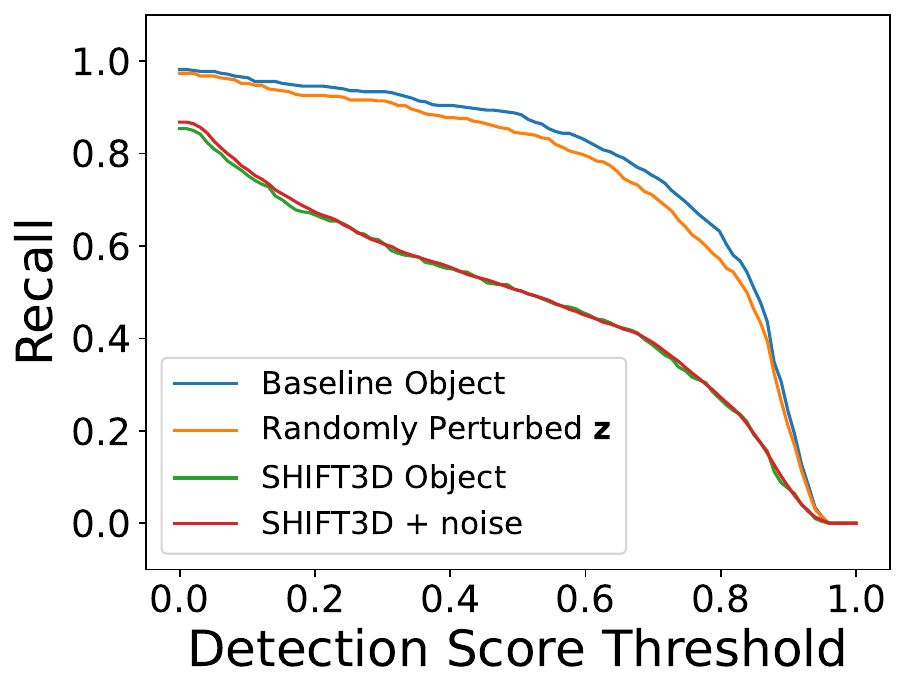}
\caption{SUV}
\end{subfigure}
\begin{subfigure}[b]{0.196\textwidth}
\includegraphics[width=\textwidth,]{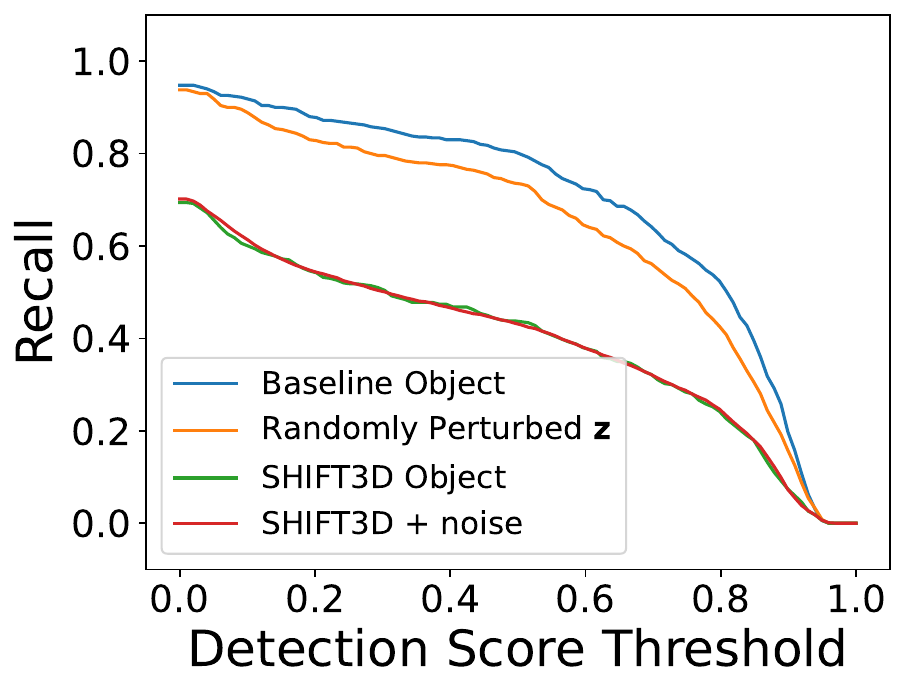}
\caption{Convertible Car}
\end{subfigure}
\begin{subfigure}[b]{0.196\textwidth}
\includegraphics[width=\textwidth,]{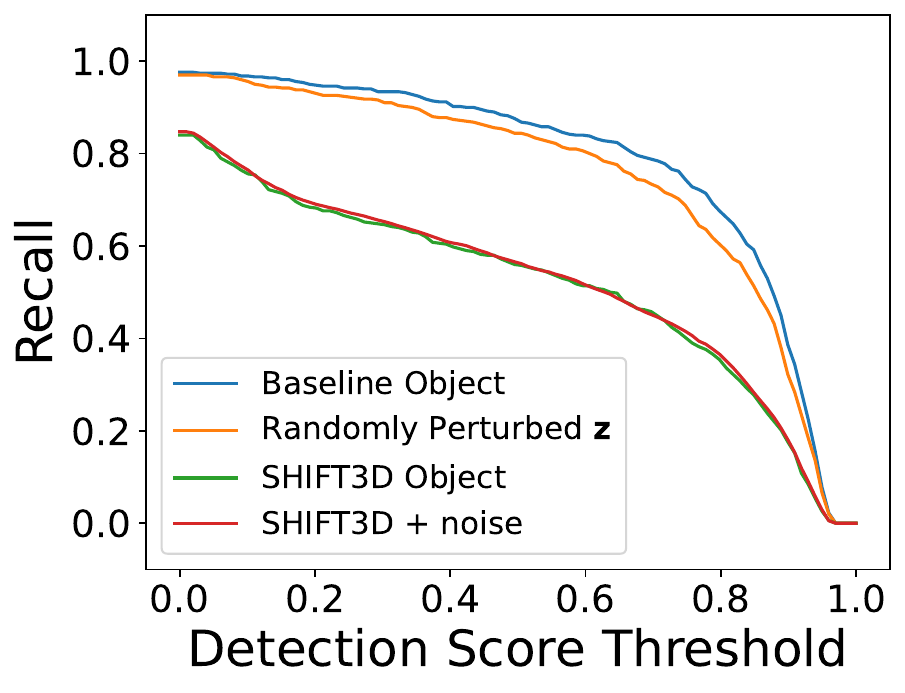}
\caption{Beach Wagon}
\end{subfigure}
\caption{ Threshold-recall curves to evaluate adversarial \emph{shape} generation for different vehicles in the ``Automobile" category. 
Each curve represents the recall rate of the detector at different detection threshold values, computed from $500$ scenes with the corresponding vehicle present. 
The proposed method for generating adversarial shapes shows significantly lower recall rates compared to random perturbation, demonstrating its effectiveness in deceiving the detector.
}
\label{fig:recall_threshold_curve}
    \end{figure*}

\begin{table}[h!]
  \centering
  \begin{tabular}{cccccc}
  \toprule
    &\multicolumn{4}{c}{Area Under Curve (AUC)}\\
    Method & Coupe & \thead{Sports \\ Car} & SUV & \thead{Conv. \\ Car} & \thead{Beach \\ Wagon} \\\midrule

    Baseline & 0.755 & 0.756 &  0.755&  0.681 &   0.779\\
    Random & 0.716 &0.721 &  0.728 &  0.625 &0.746\\\midrule
     \thead{\text{\ourmethod{}} \\ \text{+noise}}& 0.469 &0.451 &  0.473  & 0.392 &0.519\\
    \text{\ourmethod{}} & \textbf{0.466} & \textbf{0.448}&    \textbf{0.469}& \textbf{0.390}  &\textbf{0.514}\\
    
    \bottomrule
  \end{tabular}
  \caption{The AUC for the curves in Figure~\ref{fig:recall_threshold_curve}, which demonstrates that \ourmethod{} produces challenging examples that confuse a 3D detector far more successfully than random perturbations. Small perturbations of \ourmethod{} objects do not appreciably affect their ability to confuse detectors. 
  }
  \label{tab:auc_z}
\end{table}

\vspace{-0.1cm}\subsection{Evaluation Metrics}\label{sec:exp_metrics}
We consider an inserted 3D object to be detected if a predicted bounding box with detection score greater than a threshold exists with box center less than 5m from the center of the added object. 
If multiple boxes meet these criteria, we select the box with the highest detection score. 
To assess the performance of the detector, we generate a \emph{threshold-recall curve} by sweeping the detection score threshold from $0$ to $1$ and calculating the recall rate for the added SDF objects (the precision of detecting inserted objects is always $100\%$), with AUC being then calculated for each curve as a useful overall detector performance metric. Refer to the supplementary material for detailed discussion on metrics. 

\vspace{-0.1cm}\subsection{Adversarial Shape Generation Results}

\begin{figure*}
       \centering

\begin{subfigure}[b]{0.18\textwidth}
\includegraphics[width=\textwidth, trim={18mm 0 2.7cm 0},clip]{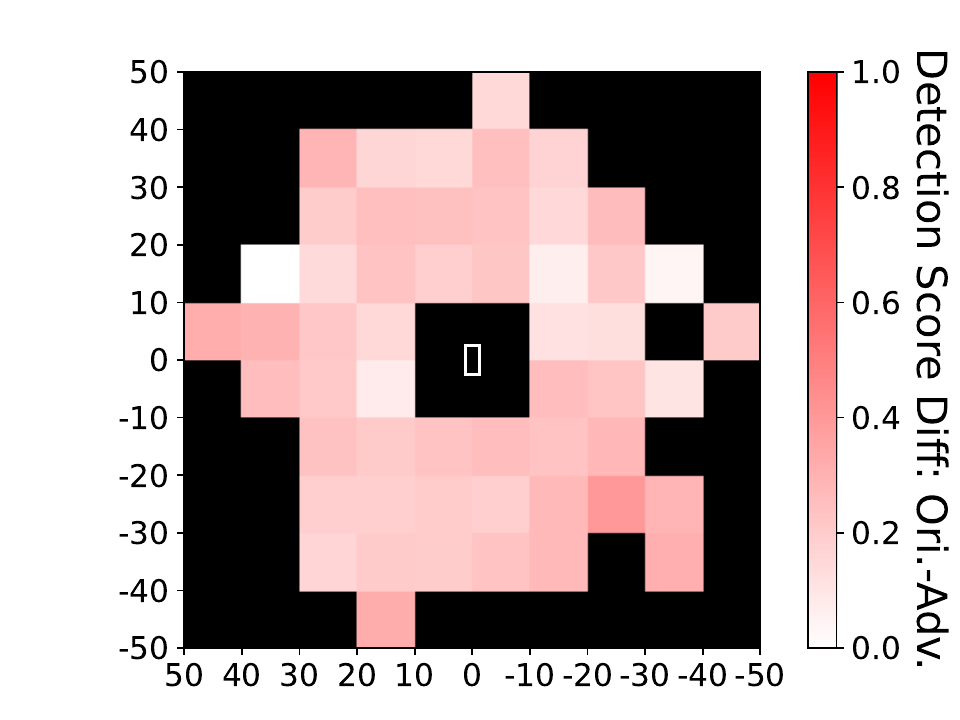}
\caption{Coupe}
\end{subfigure}
\begin{subfigure}[b]{0.18\textwidth}
\includegraphics[width=\textwidth, trim={18mm 0 2.7cm 0},clip]{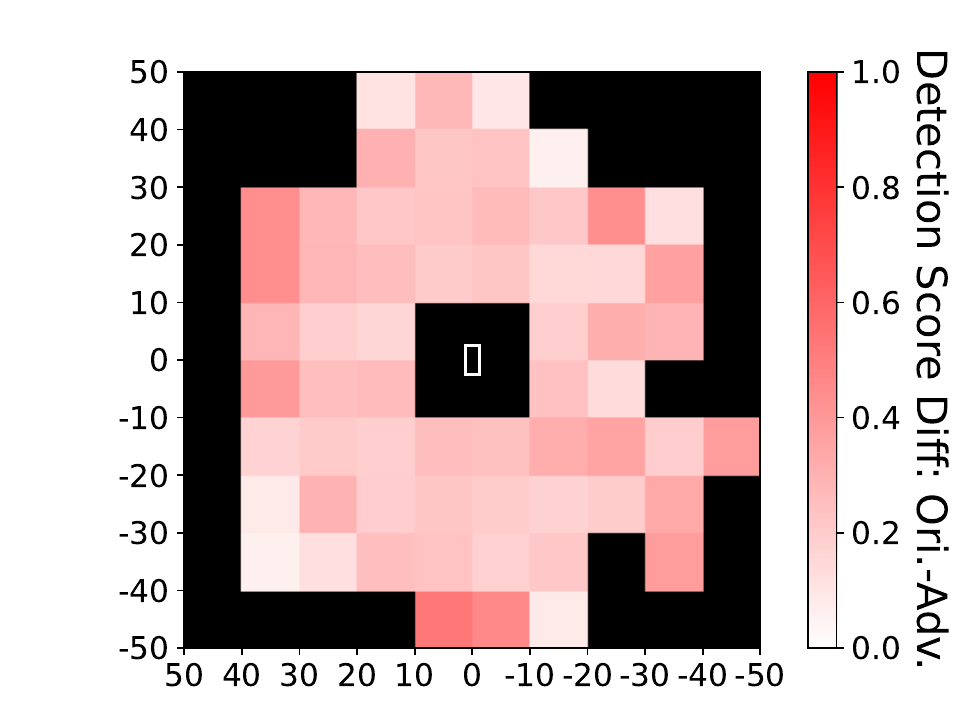}
\caption{Sports Car}
\end{subfigure}
\begin{subfigure}[b]{0.18\textwidth}
\includegraphics[width=\textwidth, trim={18mm 0 2.7cm 0},clip]{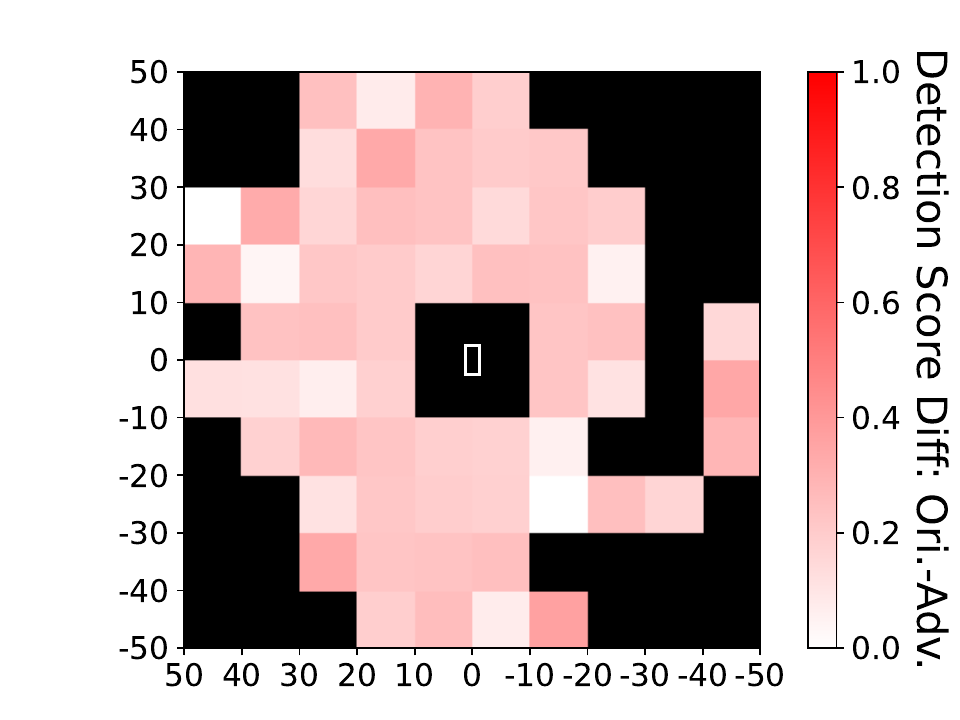}
\caption{SUV}
\end{subfigure}
\begin{subfigure}[b]{0.18\textwidth}
\includegraphics[width=\textwidth, trim={18mm 0 2.7cm 0},clip]{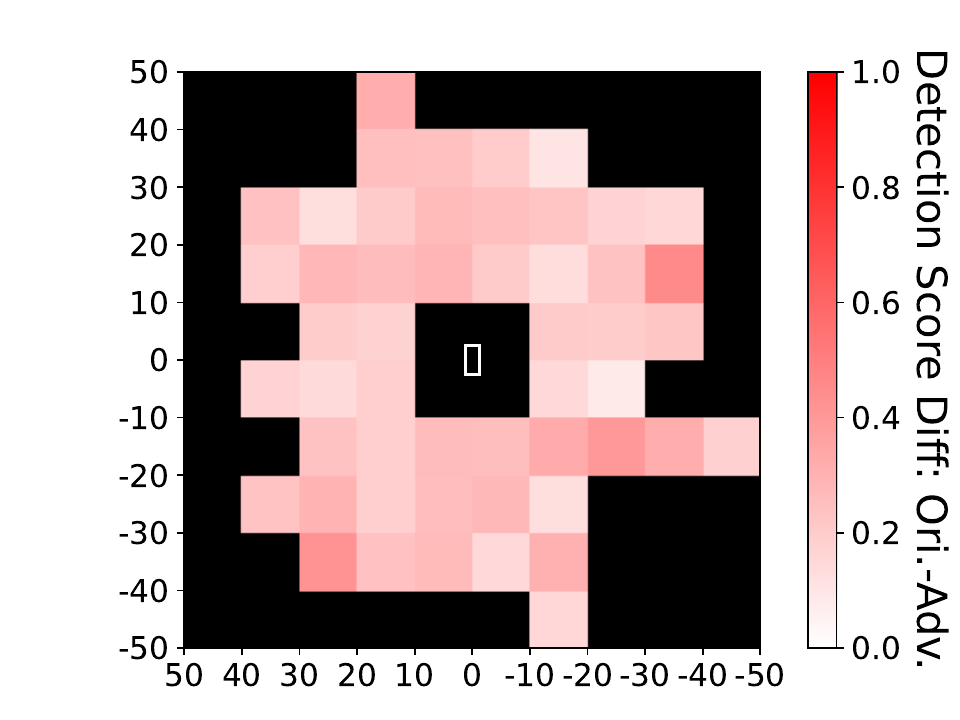}
\caption{Convertible Car}
\end{subfigure}
\begin{subfigure}[b]{0.215\textwidth}
\includegraphics[width=\textwidth, trim={18mm 0 0 0},clip]{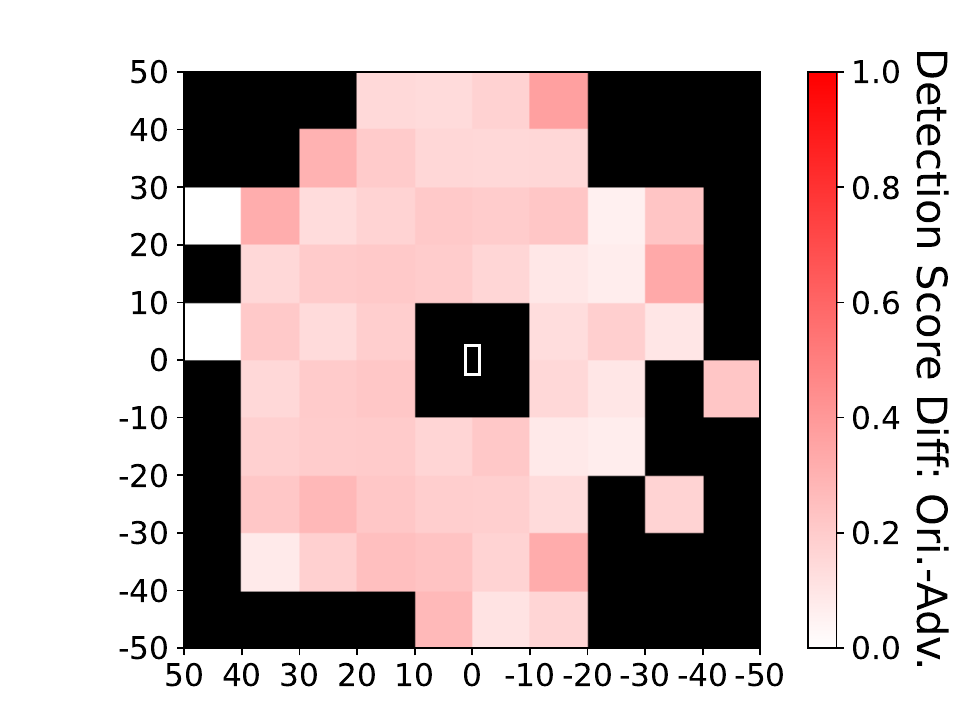}
\caption{Beach Wagon}
\end{subfigure}
\caption{
Evaluation of the robustness of adversarial shapes generated for all vehicles across various poses and locations in the scene from a bird's eye view. 
The plot depicts the detection score reduction compared to the original shape in the same pose, where black cells indicate no placement. 
The rectangle at the center represents the autonomous vehicle. 
This figure illustrates the transferability and robustness of the adversarial shapes across different locations in the input scene scene.
}
\label{fig:bev_score}
    \end{figure*}

\begin{figure*}
\centering

\begin{subfigure}[b]{0.196\textwidth}
\includegraphics[width=\textwidth,]{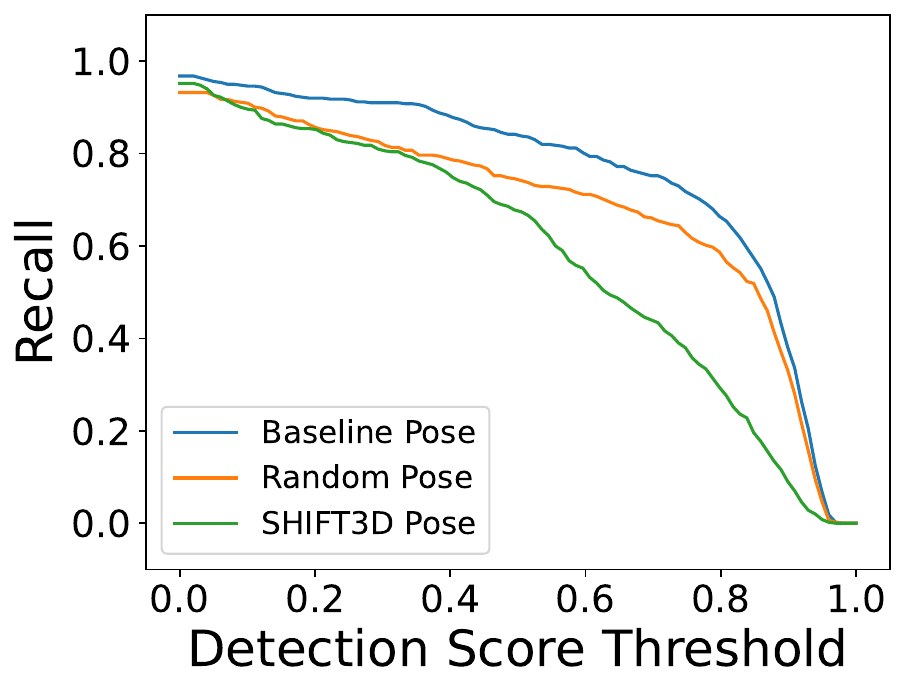}
\caption{Coupe}
\end{subfigure}
\begin{subfigure}[b]{0.196\textwidth}
\includegraphics[width=\textwidth]{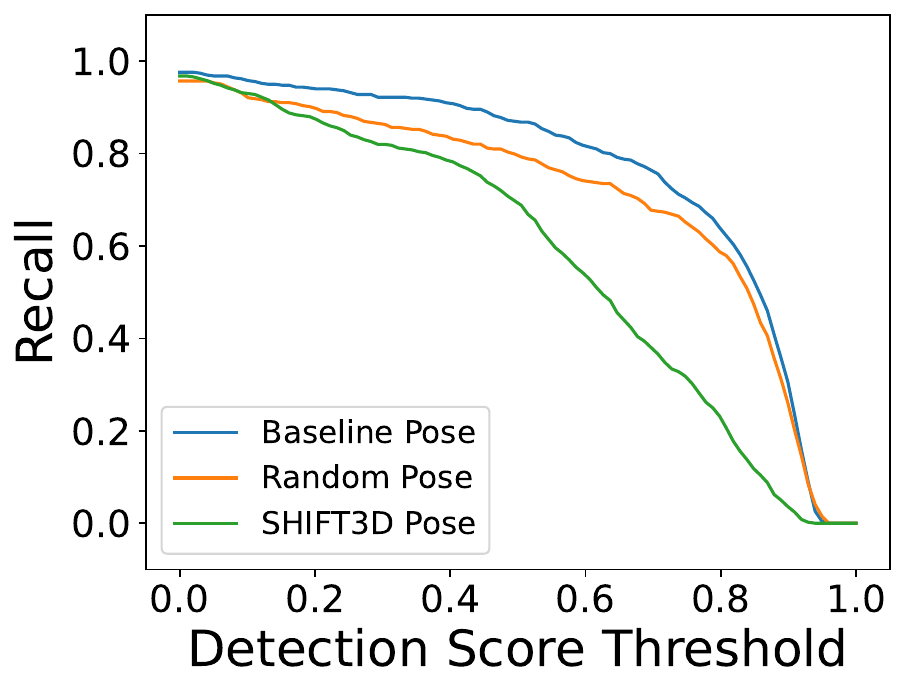}
\caption{Sports Car}
\end{subfigure}
\begin{subfigure}[b]{0.196\textwidth}
\includegraphics[width=\textwidth,]{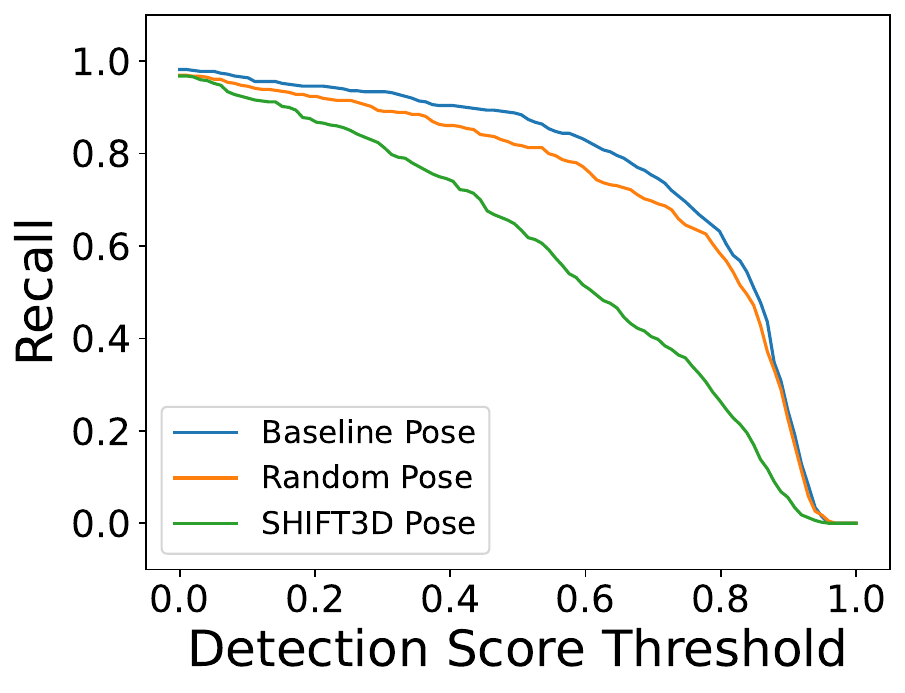}
\caption{SUV}
\end{subfigure}
\begin{subfigure}[b]{0.196\textwidth}
\includegraphics[width=\textwidth,]{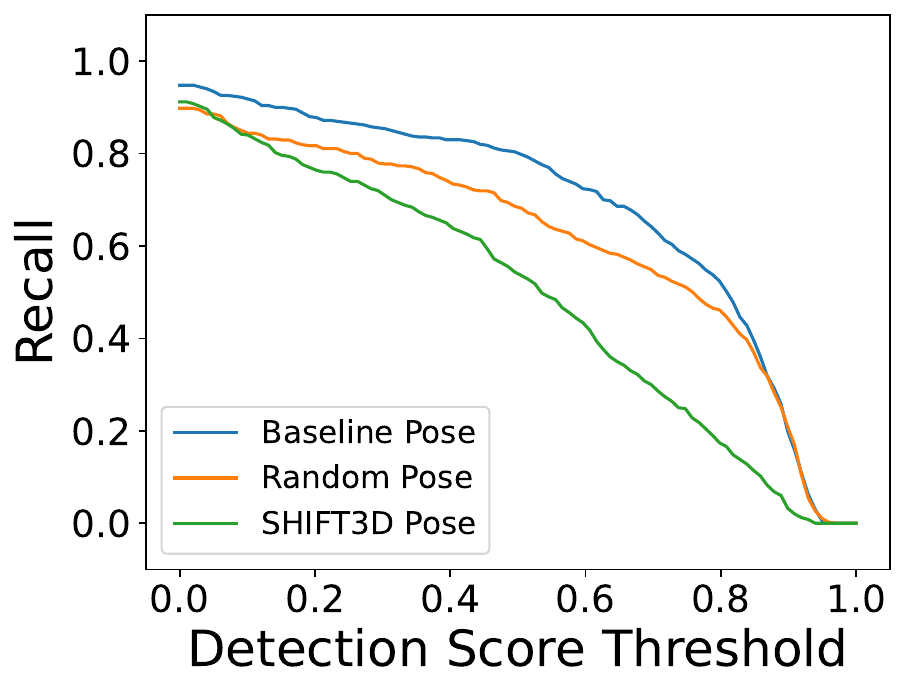}
\caption{Convertible Car}
\end{subfigure}
\begin{subfigure}[b]{0.196\textwidth}
\includegraphics[width=\textwidth,]{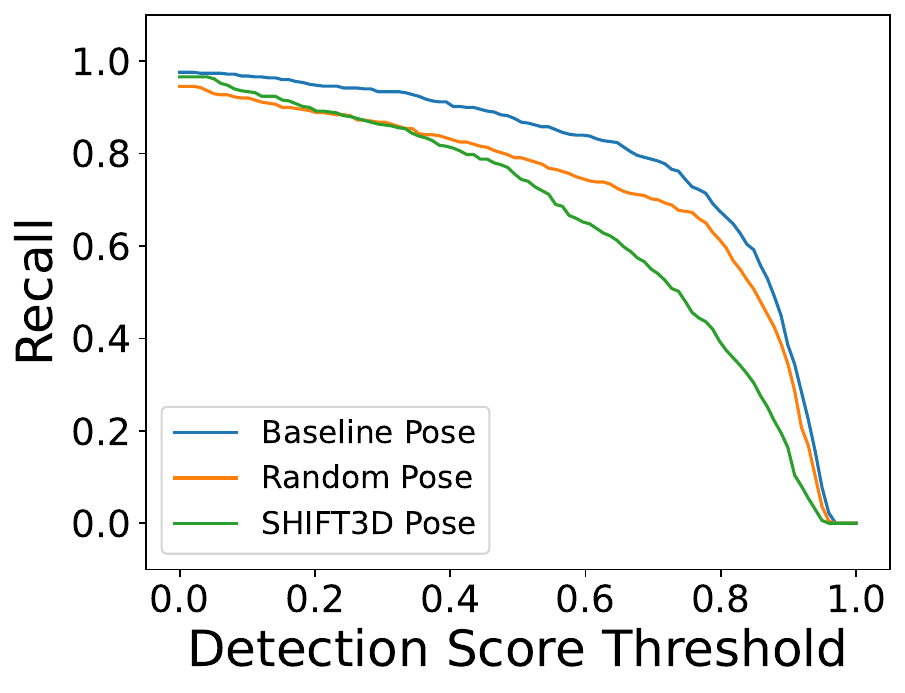}
\caption{Beach Wagon}
\end{subfigure}
\caption{Threshold-recall curves evaluate adversarial \emph{pose} for ``Automobile" category vehicles. 
Similar to the curves in Figure~\ref{fig:recall_threshold_curve}, each curve shows the detector recall rate at different thresholds, from $500$ scenes with the vehicle present. 
Our method exhibits significantly lower recall rates, outperforming random pose in deceiving the detector.}
\label{fig:recall_threshold_curve_theta}
\end{figure*}
We first present our experimental results on PointPillars, while the results on SST are shown in Section~\ref{sec:sst} of the supplementary material. 

Figure~\ref{fig:recall_threshold_curve} presents the threshold-recall curve for each baseline object and its corresponding adversarial objects across the 500 scenes. 
Since the generated adversarial objects are scene-specific, their shapes vary across different scenes. To quantify the overall reduction in recall, we compute the area under the threshold-recall curves (AUC) and report the numerical values in Table~\ref{tab:auc_z}. 

We compare the performance of our method against objects generated by randomly perturbing the latent shape encoding. 
For each adversarial object, we compute the $\ell_2$ distance between the original encoding $\B z_0$ and the encoding $\B z_{\text{adv}}$ generated by \ourmethod{}, and generate another object by randomly selecting another encoding $\B z_\text{{rand}}$ on a $d_{\B z}$-dimensional sphere with a radius of $\lVert \B z_0 -\B z_\text{{adv}}\rVert_2$. 
The results presented in Figure~\ref{fig:recall_threshold_curve} and Table~\ref{tab:auc_z} clearly show that the recall rate of the adversarial objects is significantly lower compared to both the randomly perturbed and baseline objects. 
These findings demonstrate that our method is substantially more effective at generating adversarial samples.
To analyze the robustness of \ourmethod{} outputs, we sample 10 additional $\B z$ encodings around $\B z_{\text{adv}}$ by adding Gaussian noise with standard deviation 0.01, which is approximately 5\% of the average value of $\lVert \B z_0 -\B z_\text{{adv}}\rVert_2$. 
As shown in Figure~\ref{fig:recall_threshold_curve} and Table~\ref{tab:auc_z}, such examples have almost identical Threshold-recall curves and area under curve values as those of $\B z_{\text{adv}}$. 


To assess the transferability of \ourmethod{} outputs across different locations in a scene, we randomly placed each of the 500 objects in 10 distinct scenes while satisfying the physical placement constraints described in Section \ref{sec:physical_constraints}. 
We applied random yaw rotations during the placements and maintained consistency by subjecting the baseline objects to identical rotation and translation operations. 
We then divided the resulting adversarial and baseline objects into $10m\times 10m$ cells and computed the average reduction in detection scores. 
As shown in Figure~\ref{fig:bev_score}, we found that the detection scores of the adversarial objects consistently decreased across all locations, indicating that the adversarial shapes are not location-specific and remain robust when placed in different positions in a scene.

\begin{figure*}
  \centering
  \caption{Visualizations of challenging objects created by \ourmethod{} and their scenes. \textcolor{red}{Red} boxes are detector's output. 
  }
  \label{fig:z_figures}
  \begin{tabular}{c|cc|cc}
    \toprule
  Input Scene  & Baseline Object  &  \thead{Baseline Object \\ in the Scene}  & \thead{\ourmethod{} Object \\ and Their Semantic Patterns}  & \thead{\ourmethod{} Object \\ in the Scene}\\  
    \midrule
    \begin{subfigure}[b]{0.18\linewidth}
      \centering
      \includegraphics[width=\linewidth,trim={78cm 67cm 82cm 98cm},clip]{supp_figures/z_figures/e6a809d5-e969-4e0b-97c9-cc598afc8648_input.pdf}
    \end{subfigure}
    &
    \begin{subfigure}[b]{0.18\linewidth}
      \centering
      \includegraphics[width=\linewidth]{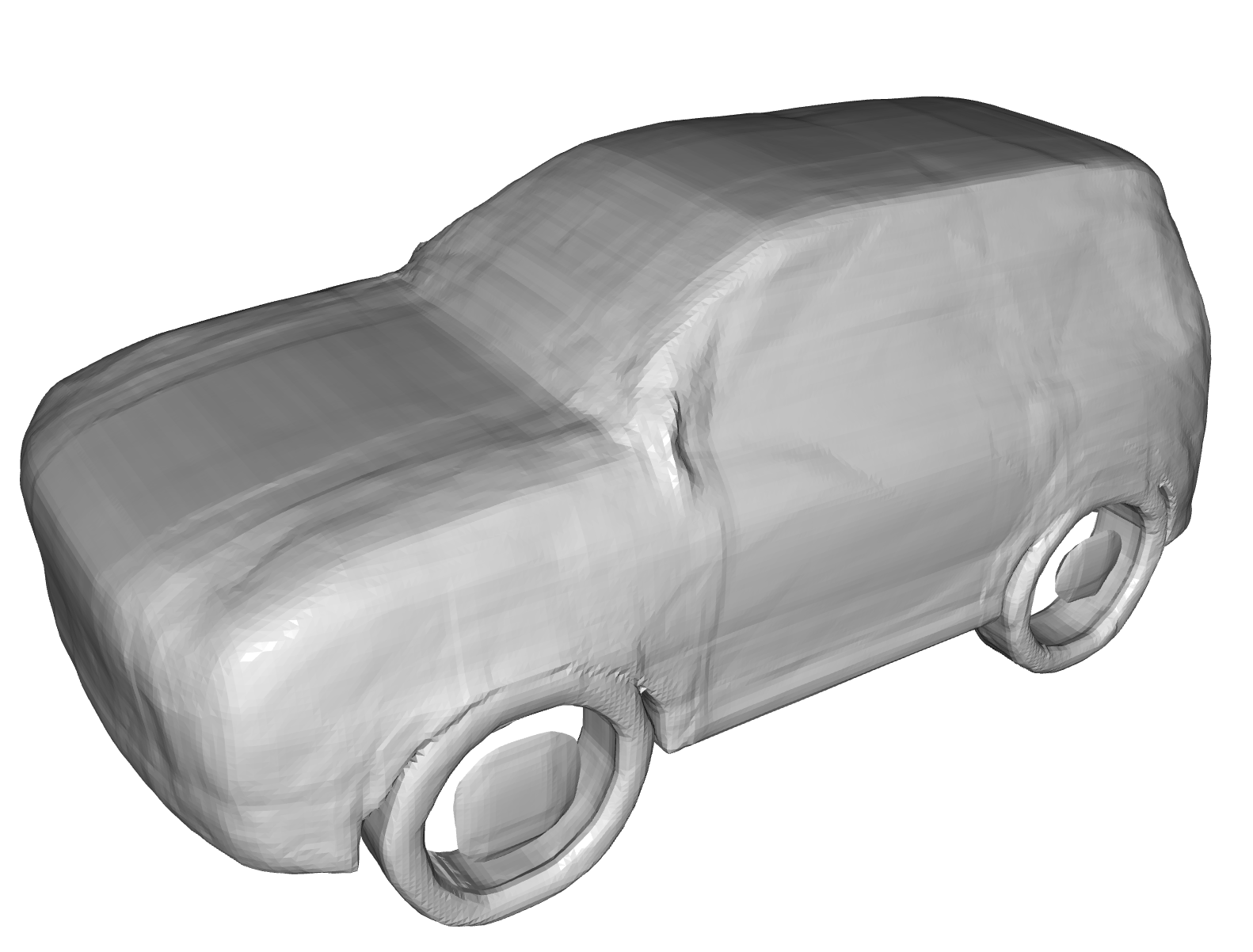}
    \end{subfigure}
    &
    \begin{subfigure}[b]{0.18\linewidth}
      \centering
      \includegraphics[width=\linewidth,trim={78cm 67cm 82cm 98cm},clip]{supp_figures/z_figures/e6a809d5-e969-4e0b-97c9-cc598afc8648_ori.pdf}
    \end{subfigure}
    &
    \begin{subfigure}[b]{0.18\linewidth}
      \centering
      \includegraphics[width=\linewidth]{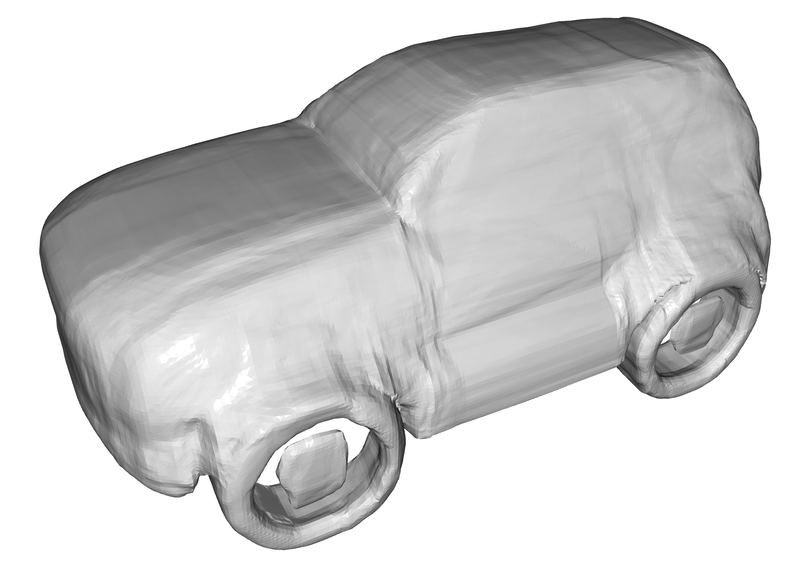}
    \end{subfigure}
    &
    \begin{subfigure}[b]{0.18\linewidth}
      \centering
      \includegraphics[width=\linewidth,trim={78cm 67cm 82cm 98cm},clip]{supp_figures/z_figures/e6a809d5-e969-4e0b-97c9-cc598afc8648_adv_14.pdf}
    \end{subfigure}
    \\
    &  SUV & \textcolor{ForestGreen}{Detection Score: 0.74}  &  smaller windshield &\textcolor{OrangeRed}{Detection Score: 0.07}\\
\mycomment{
\begin{subfigure}[b]{0.18\linewidth}
      \centering
      \includegraphics[width=\linewidth,trim={73cm 85cm 87cm 80cm},clip]{supp_figures/z_figures/5336fe7b-48b7-4910-b721-917b35a5e936_input.pdf}
    \end{subfigure}
    &
    \begin{subfigure}[b]{0.18\linewidth}
      \centering
      \includegraphics[width=\linewidth]{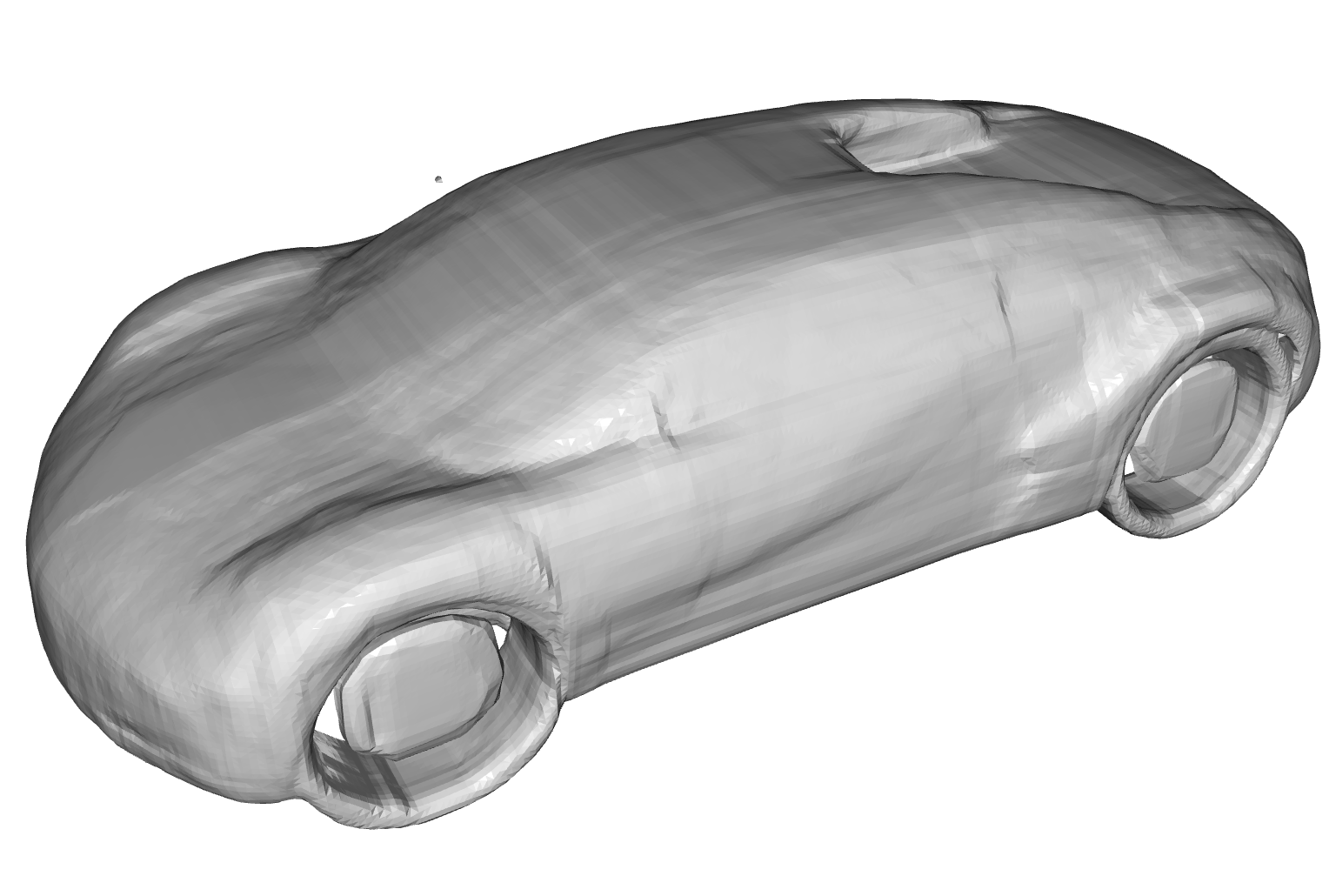}
    \end{subfigure}
    &
    \begin{subfigure}[b]{0.18\linewidth}
      \centering
      \includegraphics[width=\linewidth,trim={73cm 85cm 87cm 80cm},clip]{supp_figures/z_figures/5336fe7b-48b7-4910-b721-917b35a5e936_ori.pdf}
    \end{subfigure}
    &
    \begin{subfigure}[b]{0.18\linewidth}
      \centering
      \includegraphics[width=\linewidth]{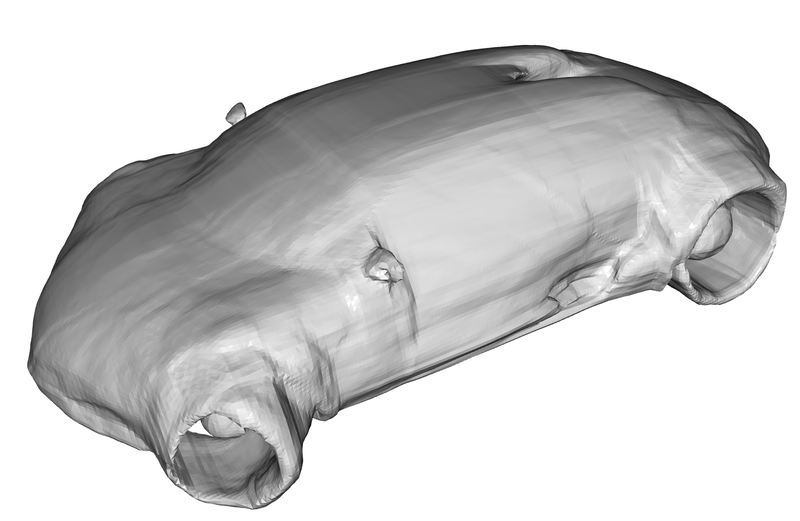}
    \end{subfigure}
    &
    \begin{subfigure}[b]{0.18\linewidth}
      \centering
      \includegraphics[width=\linewidth,trim={73cm 85cm 87cm 80cm},clip]{supp_figures/z_figures/5336fe7b-48b7-4910-b721-917b35a5e936_adv_1.pdf}
    \end{subfigure}
    \\
    &  Sports Car & \textcolor{ForestGreen}{Detection Score: 0.85}  &  \thead{front wheel rotated\\ sharper front}&\textcolor{OrangeRed}{Detection Score: 0.09}\\
}
    \midrule
\begin{subfigure}[b]{0.18\linewidth}
      \centering
      \includegraphics[width=\linewidth,trim={75cm 95cm 85cm 70cm},clip]{supp_figures/z_figures/4c9edbd6-faee-4afa-9b95-6ccb3b26e68b_input.pdf}
    \end{subfigure}
    &
    \begin{subfigure}[b]{0.18\linewidth}
      \centering
      \includegraphics[width=\linewidth]{figures/threshold_recall/dcaSportsCar.png}
    \end{subfigure}
    &
    \begin{subfigure}[b]{0.18\linewidth}
      \centering
      \includegraphics[width=\linewidth,trim={75cm 95cm 85cm 70cm},clip]{supp_figures/z_figures/4c9edbd6-faee-4afa-9b95-6ccb3b26e68b_ori.pdf}
    \end{subfigure}
    &
    \begin{subfigure}[b]{0.18\linewidth}
      \centering
      \includegraphics[width=\linewidth]{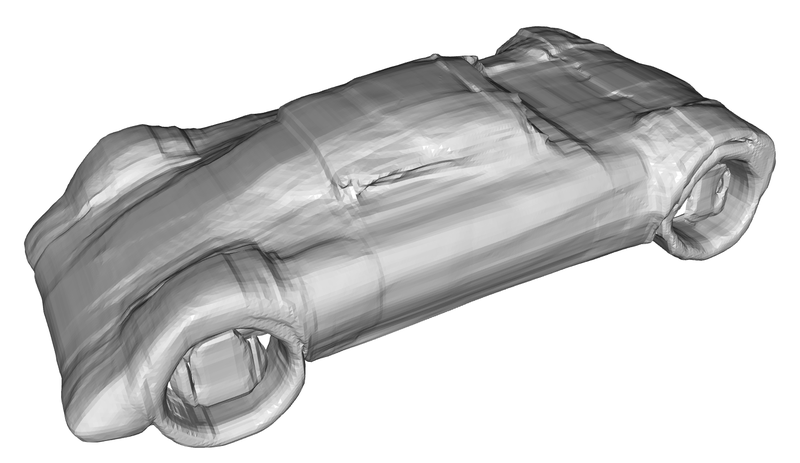}
    \end{subfigure}
    &
    \begin{subfigure}[b]{0.18\linewidth}
      \centering
      \includegraphics[width=\linewidth,trim={75cm 95cm 85cm 70cm},clip]{supp_figures/z_figures/4c9edbd6-faee-4afa-9b95-6ccb3b26e68b_adv_3.pdf}
    \end{subfigure}
    \\
    &  Sports Car & \textcolor{ForestGreen}{Detection Score: 0.91}  & \thead{lower chassis\\ square front}   &\textcolor{OrangeRed}{Detection Score: 0.04}\\
    \midrule
\begin{subfigure}[b]{0.18\linewidth}
      \centering
      \includegraphics[width=\linewidth,trim={78cm 95cm 82cm 70cm},clip]{supp_figures/z_figures/1fbfb867-70db-467d-9dd5-a728f25cc24d_input.pdf}
    \end{subfigure}
    &
    \begin{subfigure}[b]{0.18\linewidth}
      \centering
      \includegraphics[width=\linewidth]{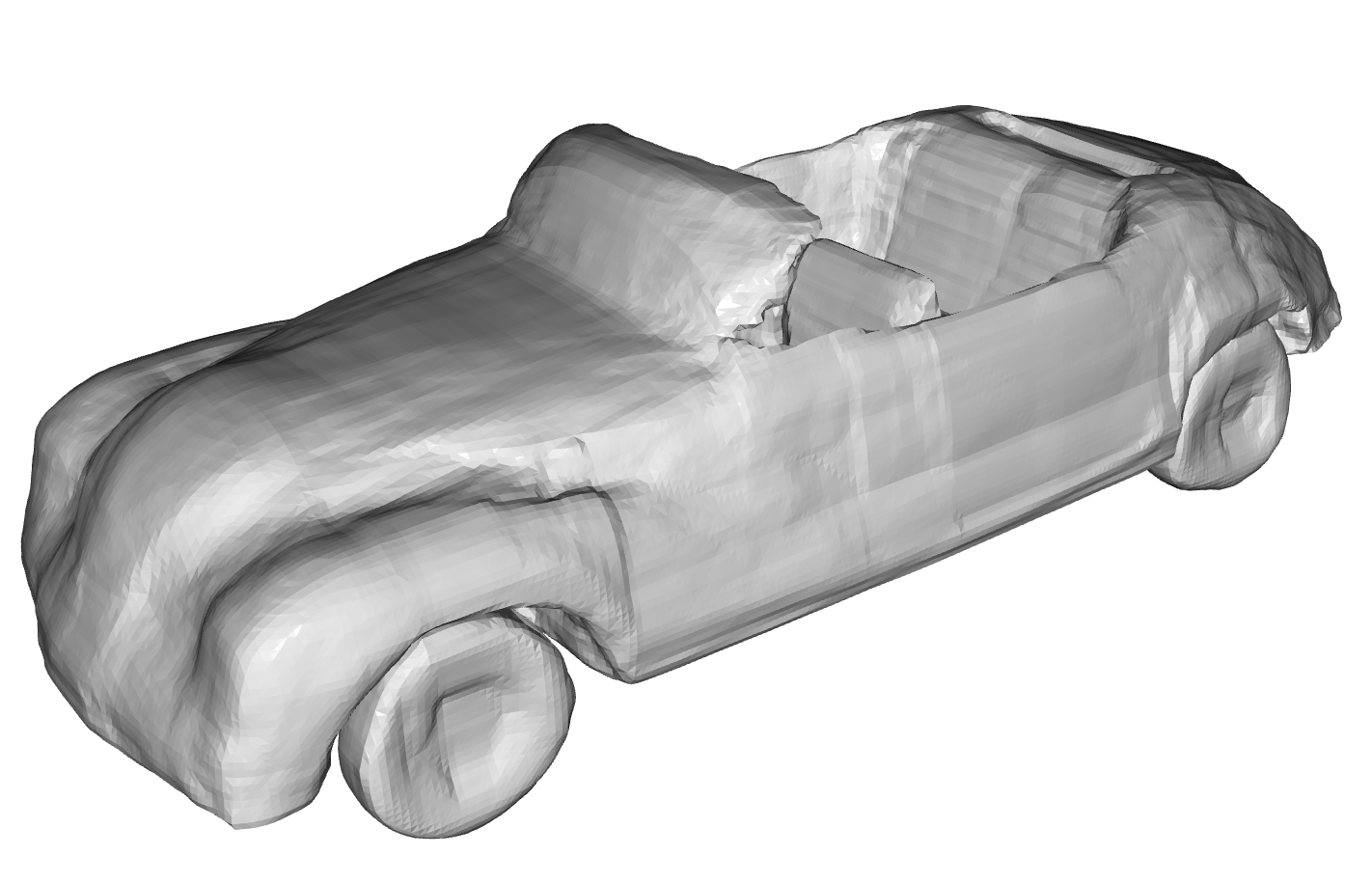}
    \end{subfigure}
    &
    \begin{subfigure}[b]{0.18\linewidth}
      \centering
      \includegraphics[width=\linewidth,trim={78cm 95cm 82cm 70cm},clip]{supp_figures/z_figures/1fbfb867-70db-467d-9dd5-a728f25cc24d_ori.pdf}
    \end{subfigure}
    &
    \begin{subfigure}[b]{0.18\linewidth}
      \centering
      \includegraphics[width=\linewidth]{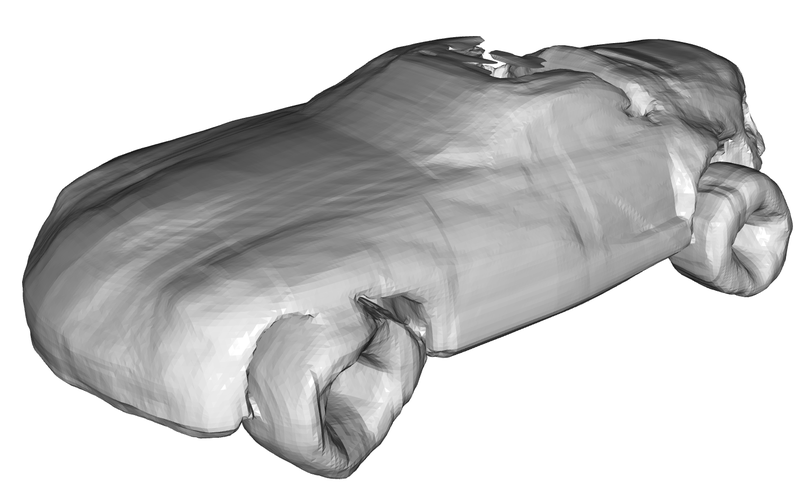}
    \end{subfigure}
    &
    \begin{subfigure}[b]{0.18\linewidth}
      \centering
      \includegraphics[width=\linewidth,trim={78cm 95cm 82cm 70cm},clip]{supp_figures/z_figures/1fbfb867-70db-467d-9dd5-a728f25cc24d_adv_1.pdf}
    \end{subfigure}
    \\
    &  Convertible Car & \textcolor{ForestGreen}{Detection Score: 0.71}  & \thead{lower chassis\\ round front}   &\textcolor{OrangeRed}{Detection Score: 0.07}\\
 

    \midrule
\begin{subfigure}[b]{0.18\linewidth}
      \centering
      \includegraphics[width=\linewidth,trim={75cm 90cm 85cm 75cm},clip]{supp_figures/z_figures/cb0b36df-f9b4-4089-8d97-d994519c8bfe_input.pdf}
    \end{subfigure}
    &
    \begin{subfigure}[b]{0.18\linewidth}
      \centering
      \includegraphics[width=\linewidth]{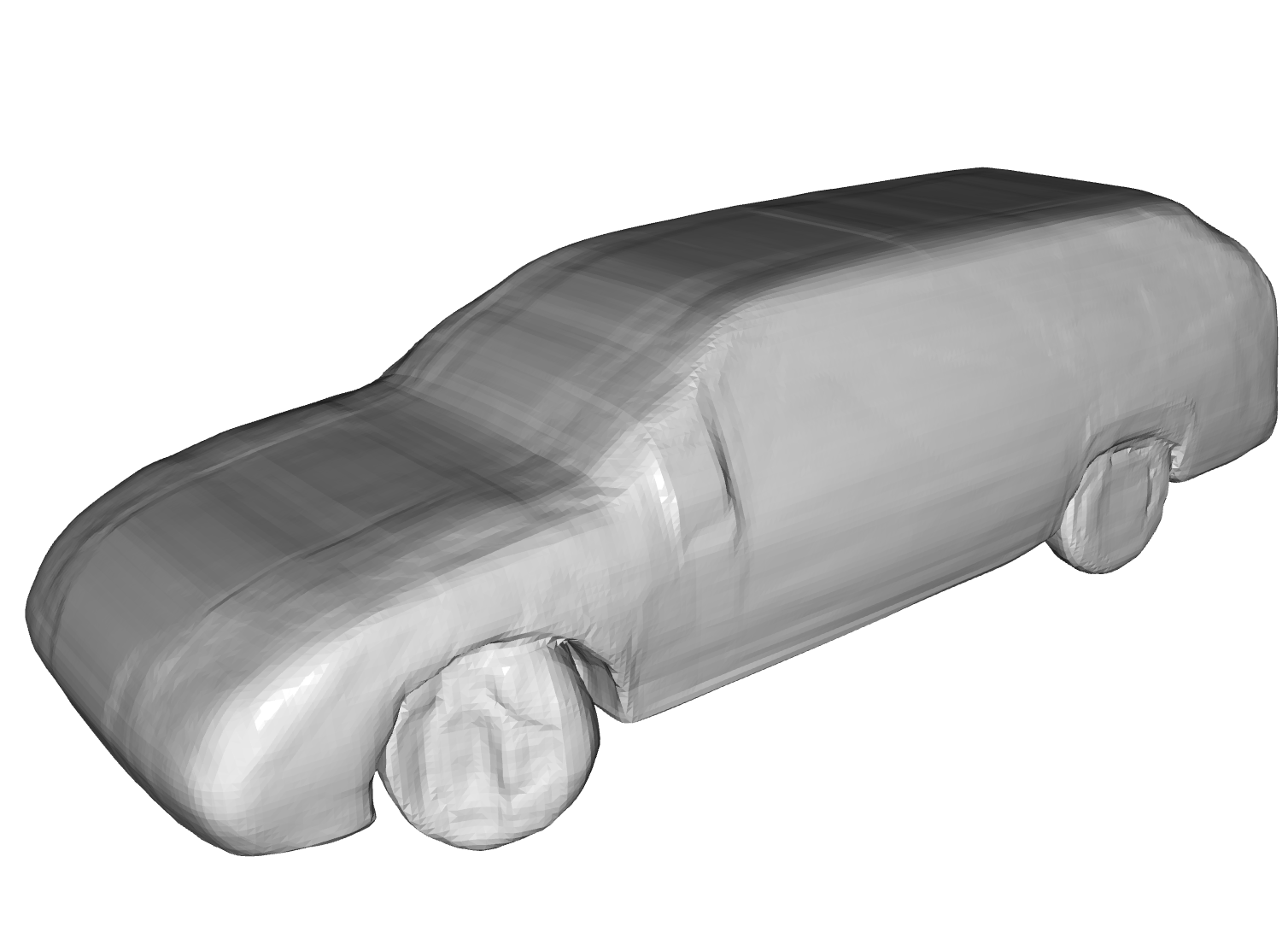}
    \end{subfigure}
    &
    \begin{subfigure}[b]{0.18\linewidth}
      \centering
      \includegraphics[width=\linewidth,trim={75cm 95cm 85cm 70cm},clip]{supp_figures/z_figures/cb0b36df-f9b4-4089-8d97-d994519c8bfe_ori.pdf}
    \end{subfigure}
    &
    \begin{subfigure}[b]{0.18\linewidth}
      \centering
      \includegraphics[width=\linewidth]{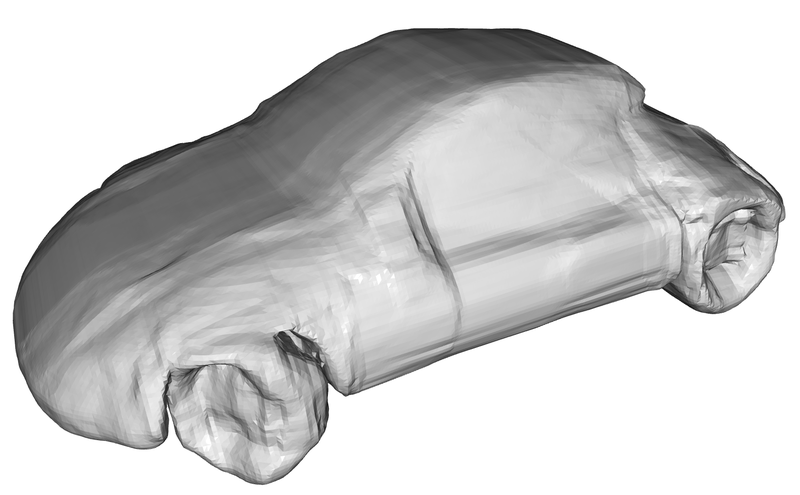}
    \end{subfigure}
    &
    \begin{subfigure}[b]{0.18\linewidth}
      \centering
      \includegraphics[width=\linewidth,trim={75cm 95cm 85cm 70cm},clip]{supp_figures/z_figures/cb0b36df-f9b4-4089-8d97-d994519c8bfe_adv_4.pdf}
    \end{subfigure}
    \\
    & Beach Wagon  & \textcolor{ForestGreen}{Detection Score: 0.93}  & \thead{shorter chassis\\ hatchback}   &\textcolor{OrangeRed}{Detection Score: 0.10}\\
    \midrule
\begin{subfigure}[b]{0.18\linewidth}
      \centering
      \includegraphics[width=\linewidth,trim={80cm 70cm 80cm 95cm},clip]{supp_figures/z_figures/5e079dc1-c014-4e4a-823f-a005fb7be70a_input.pdf}
    \end{subfigure}
    &
    \begin{subfigure}[b]{0.18\linewidth}
      \centering
      \includegraphics[width=\linewidth]{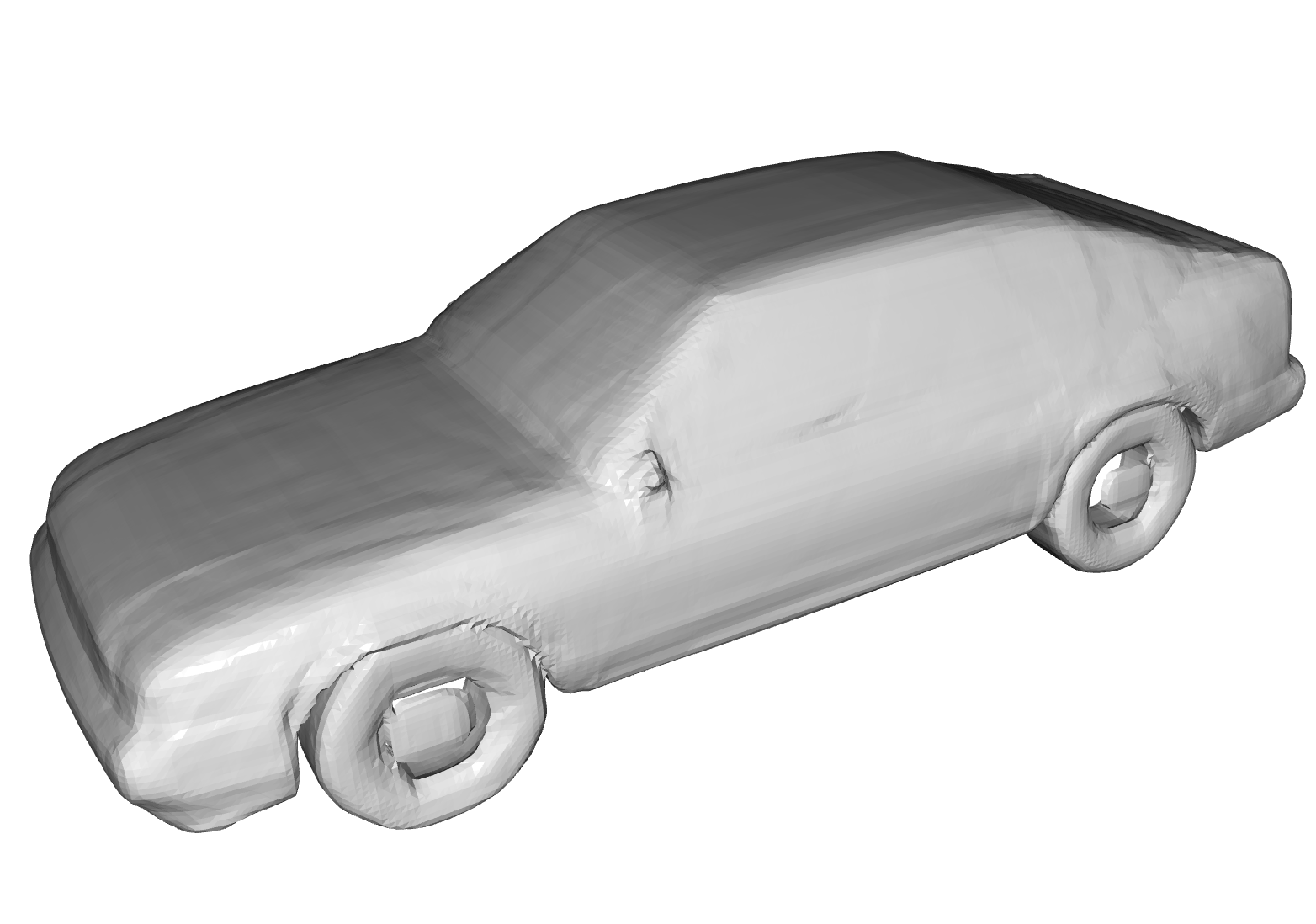}
    \end{subfigure}
    &
    \begin{subfigure}[b]{0.18\linewidth}
      \centering
      \includegraphics[width=\linewidth,trim={80cm 70cm 80cm 95cm},clip]{supp_figures/z_figures/5e079dc1-c014-4e4a-823f-a005fb7be70a_ori.pdf}
    \end{subfigure}
    &
    \begin{subfigure}[b]{0.18\linewidth}
      \centering
      \includegraphics[width=\linewidth]{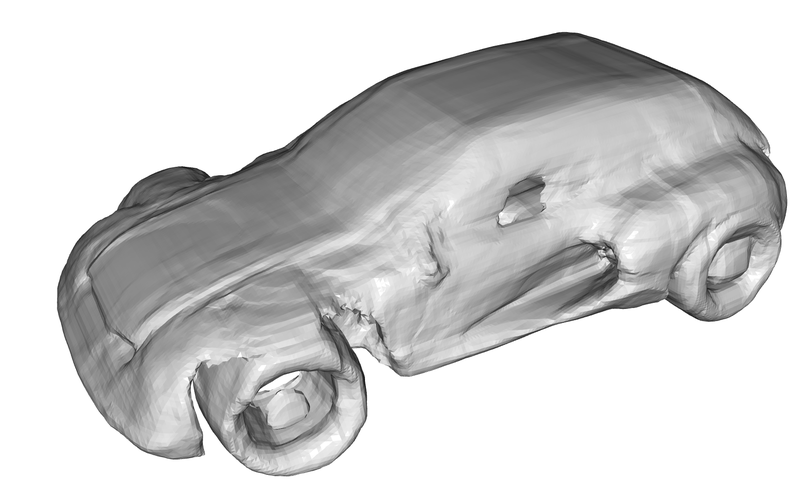}
    \end{subfigure}
    &
    \begin{subfigure}[b]{0.18\linewidth}
      \centering
      \includegraphics[width=\linewidth,trim={80cm 70cm 80cm 95cm},clip]{supp_figures/z_figures/5e079dc1-c014-4e4a-823f-a005fb7be70a_adv_5.pdf}
    \end{subfigure}
    \\
    &  Coupe & \textcolor{ForestGreen}{Detection Score: 0.93}  & \thead{shorter chassis\\ wider}   &\textcolor{OrangeRed}{Detection Score: 0.22}\\
   \bottomrule
  \end{tabular}
\end{figure*}

In Figure~\ref{fig:z_figures} we present visualizations for the challenging objects generated by \ourmethod{}, together with their corresponding point cloud scenes. We also point out the semantic patterns of the \ourmethod{} objects. Crucially, as we visualize more \ourmethod{} objects, we can begin to detect semantic patterns; for example, vehicles presenting with lower chassis will very often fool the 3D detector. These insights highlight a key benefit of using \ourmethod{}, as we now have concrete candidates for proactive model improvements. To validate these semantic features present in the log data, we identify and test natural objects in the WOD log data that are most similar to the objects generated by \ourmethod{}. Those objects are retrieved by reconstruct object shapes with DeepSDF and select the object whose latent shape vector $\B z_{\text{natural}^*}$ is the closest one to a \ourmethod{} generated latent shape vector $\B z_{\text{\ourmethod{}}}$. Detailed information about this experiment is presented in Section~\ref{sec:supp_retrieval} of the supplementary material.

Visualization of more objects generated by \ourmethod{} is presented in Section~\ref{sec:more_shape_vis} of the supplementary material. In Figure~\ref{fig:coupe_interpolation} of the supplementary, we also present visualizations of adversarial objects and the corresponding detection scores at various steps of the optimization process in \ourmethod{}.
We also test \ourmethod{} on an SST detector and our results are presented in Section~\ref{sec:sst} in the supplementary materials. Additionally, we report transferability of \ourmethod{} between SST and PointPillars models in Section~\ref{sec:supp_transfer}.


\vspace{-0.1cm}\subsection{Adversarial Pose Generation Results}


We present threshold-recall curves for both original and \ourmethod{} poses of objects across $500$ scenes in Figure~\ref{fig:recall_threshold_curve_theta}, along with comparisons to randomly generated poses.
For pose generations in both \ourmethod{} and random, we limit displacement to under $4$ meters on the $xy$-plane. We manually verified that a buffer zone of $3$ meters around the object is enough to prevent any part of it from exceeding the \ifthenelse{\boolean{roi}}{ROI boundaries.}{the 7m sphere we draw around the object as in Section~\ref{sec:exp_setup}.} We also limit rotations to under $\pm 0.1$ rads on the pitch and roll axes, and do not limit rotation along the yaw axis, while ensuring that the constraints in Section~\ref{sec:physical_constraints} are satisfied. Our results, presented in Figure~\ref{fig:recall_threshold_curve_theta} and Table~\ref{tab:auc_theta}, demonstrate a significant reduction in recall performance for adversarial objects, even when placed in alternate poses.
In contrast, random pose perturbations with the same constraints result in a minimal reduction in recall. Visualizations for \ourmethod{} adversarial pose generation are presented in Section~\ref{sec:pose_visualization} in the supplementary material.

\begin{table}[t]
  \centering
  \begin{tabular}{cccccc}
  \toprule
  
     &\multicolumn{4}{c}{Area Under Curve (AUC)}\\
    Method & Coupe & \thead{Sports \\ Car} & SUV & \thead{Conv. \\ Car} & \thead{Beach \\ Wagon} \\\midrule
    Baseline & 0.755 & 0.756 &0.755 &0.681 &0.779 \\
    Random & 0.683 & 0.705 &0.719 &0.611 &0.713\\\midrule
    \ourmethod{} & \textbf{0.580} &\textbf{0.574}  & \textbf{0.571}& \textbf{0.489} & \textbf{0.648}\\\bottomrule
  \end{tabular}
  \caption{
  AUC metric of the curves in Figure~\ref{fig:recall_threshold_curve_theta} quantifies the proposed method's improvements in adversarial \emph{pose} generation. 
  Random perturbations have little effect on detection score, while our method results in lower AUC values for all vehicles, successfully deceiving detection performance.
  }
  \label{tab:auc_theta}
\end{table}

\vspace{-0.1cm}\subsection{Improving Model Robustness Using Objects Generated by \ourmethod{}}
\label{sec:finetune}
\begin{figure}
    \centering
    \includegraphics[width=0.5\textwidth]{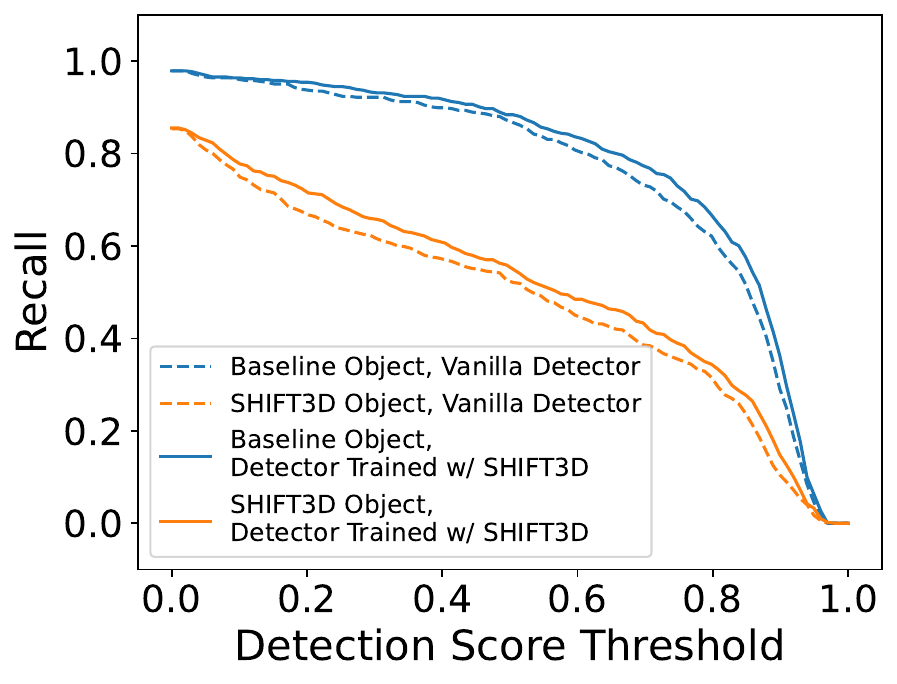}
    \caption{Threshold-recall curves before and after fine-tuning the 3D detector with objects generated by \ourmethod{}.}
    \label{fig:fine_tune_curve}
\end{figure}

We also conducted a pilot study which aims to investigate the effectiveness of using objects generated by \ourmethod{} for data augmentation and improving the robustness of our detector model. We randomly select 2000 scenes containing objects generated by our method from our five baseline vehicles and use them to fine-tune our PointPillar model with a learning rate of $10^{-7}$ for one epoch. We evaluate the performance of the detector model by testing it on another 500 scenes, each containing an object from the 5 baseline vehicles. We compare the performance of the finetuned detector model under \ourmethod{} against the vanilla detector and plot the threshold-recall curve in Figure~\ref{fig:fine_tune_curve}. Additionally, we report the AUC values in Table~\ref{tab:fine_tune_auc}. 
Our results show a clear improvement in the AUC metric for both baseline and \ourmethod{} generated objects. Additionally, we evaluate the performance of the fine-tuned model on the vanilla WOD detection task and present the results in Table~\ref{tab:natural_metrics} in Section~\ref{sec:supp_natural_metrics} of the supplementary material. Our result reveals that the performance metrics of the fine-tuned model are nearly identical to those of the vanilla detector when evaluated on the natural WOD data. This suggests that our fine-tuning approach does not introduce any significant regression. Furthermore, we may observe a slight improvement in the model's performance on the natural data, although this effect, if present, appears to be rather small.


\begin{table}[th]
  \centering
  \begin{tabular}{ccc}
  \toprule
  
     &\multicolumn{2}{c}{Area Under Curve (AUC)}\\
    Models & Vanilla Detector & Finetuned Detector \\\midrule
    Baseline & 0.751 & \textbf{0.774} \\
    \ourmethod{} & 0.482 & \textbf{0.514} \\\bottomrule
  \end{tabular}
  \caption{AUC metric values before and after fine-tuning the 3D detector with objects generated by \ourmethod{}.}
  \label{tab:fine_tune_auc}
\end{table}

\mycomment{

\begin{figure}
\begin{floatrow}
\ffigbox{
  \includegraphics[width=0.3\textwidth]{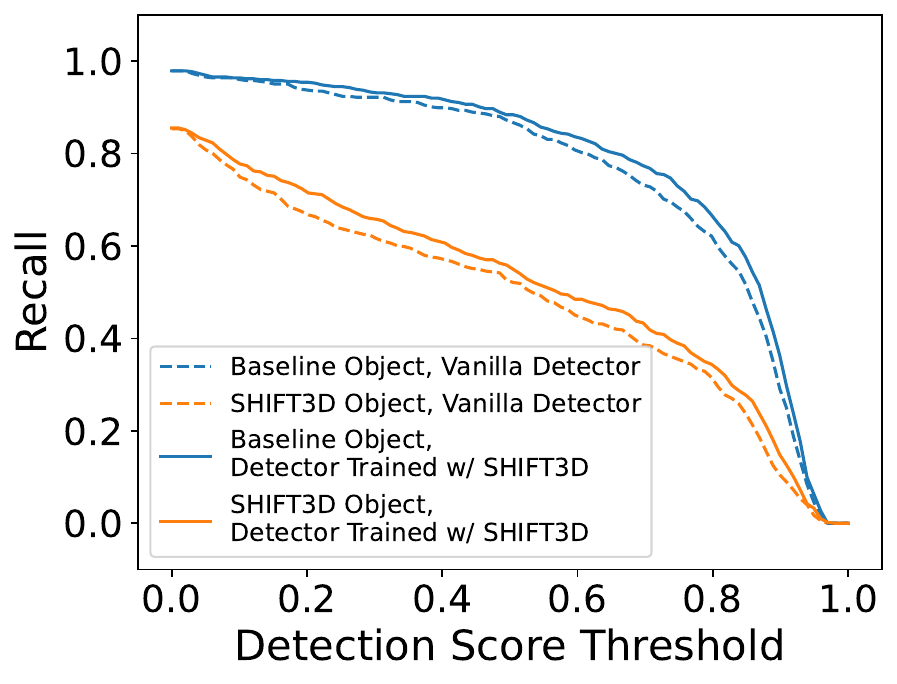}
}{
  \caption{Threshold-recall curves before and after fine-tuning the 3D detector with objects generated by \ourmethod{}.}
}
\capbtabbox{%
  \begin{tabular}{ccc}
  \toprule
  
     &\multicolumn{2}{c}{Area Under Curve (AUC)}\\
    Models & Vanilla Detector & Finetuned Detector \\\midrule
    Baseline & 0.751 & \textbf{0.774} \\
    \ourmethod{} & 0.482 & \textbf{0.514} \\\bottomrule
  \end{tabular}
}{%
  \caption{AUC metric values before and after fine-tuning the 3D detector with objects generated by \ourmethod{}.}%
}
\end{floatrow}
\end{figure}

}
\vspace{-0.5cm}
\section{Conclusion}
\vspace{-0.1cm}

\ourmethod{} is a differentiable pipeline for generating challenging yet natural examples that provide preemptive insights into failure modes in 3D vision systems. We demonstrated that all model inputs generated by \ourmethod{} reliably confuse 3D detectors in an autonomous vehicles setting, and do so regardless of the insertion location or reference scene. Through diagnostic information provided by \ourmethod{}, we will be better equipped to catch failures within 3D vision models before they appear, and circumvent scenarios where single missed objects lead to catastrophic results. 


\clearpage
{\small
\bibliographystyle{ieee_fullname}
\bibliography{iccv_bib}
}

\appendix
\twocolumn[
\begin{@twocolumnfalse} 
\begin{center}

{\LARGE	 \textbf{Supplementary Material for}\\
\textbf{\ourmethod: Synthesizing Hard Inputs For Tricking 3D Detectors}}\\
\vspace{0.7cm}
{\large Hongge Chen\textsuperscript{1}, Zhao Chen\textsuperscript{1}, Gregory P. Meyer\textsuperscript{1}, Dennis Park\textsuperscript{1},\\
Carl Vondrick\textsuperscript{1}, Ashish Shrivastava\textsuperscript{1}, and Yuning Chai\textsuperscript{1}\\
\vspace{0.3cm}
\textsuperscript{1}Cruise LLC}

\end{center}
\end{@twocolumnfalse}
]

\section{Detailed Information of Rendering and Gradient Calculation}
\label{sec:supp_rendering}
\subsection{Rendering}

This section details how we cast rays towards SDF objects that have been inserted into a point cloud scene. 
As noted in the main paper, the first step is to determine the object's bounding box center and draw a sphere with a $7$-meter radius centered at that point. 
When generating adversarial poses, we restrict displacement to less than $4$ meters on the $xy$-plane to ensure that no portion of the object extends beyond this $7$-meter sphere.

\begin{figure*}[h]
    \centering
    \includegraphics[width=0.6\textwidth]{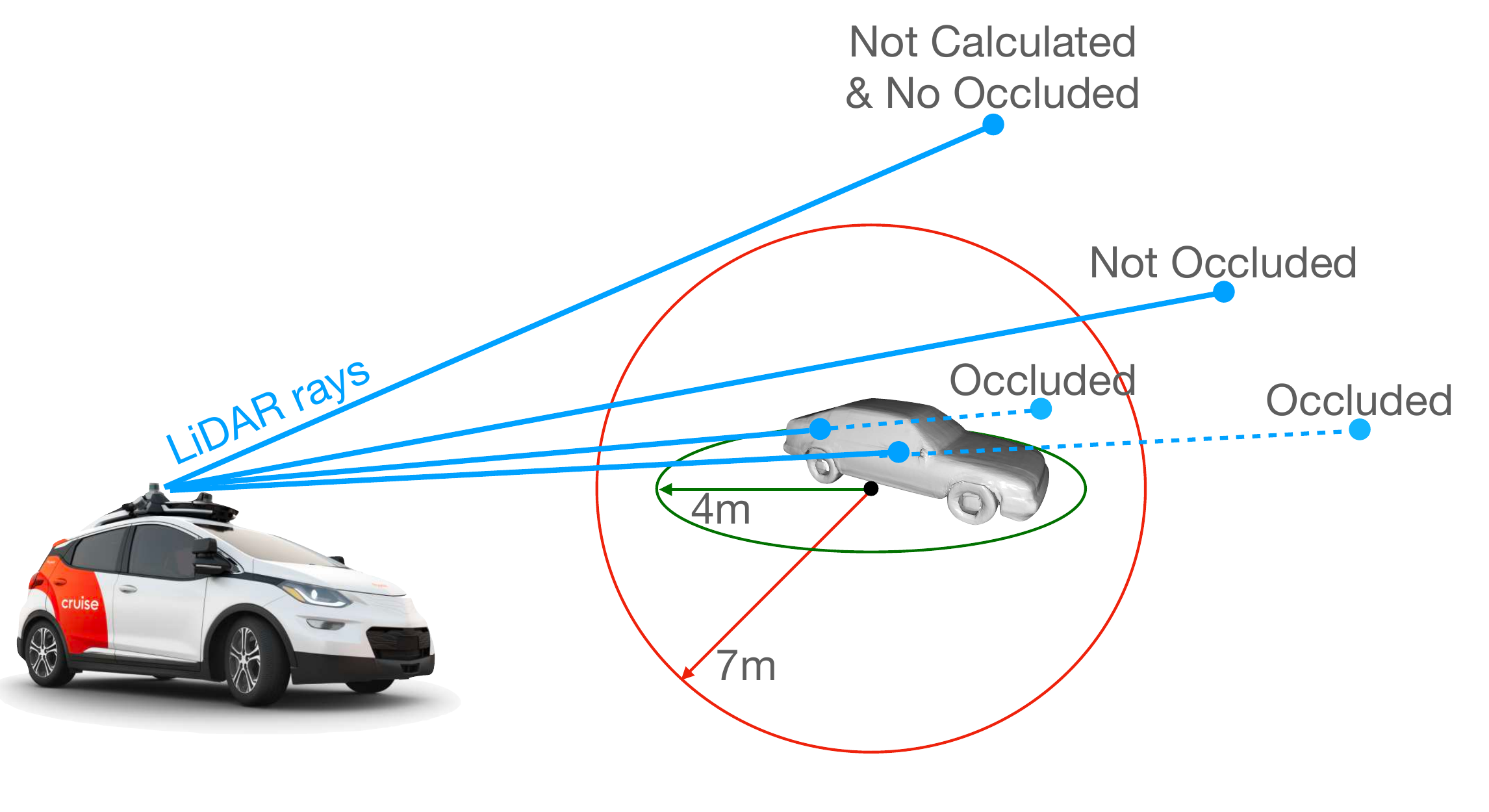}
    \caption{Rendering an SDF object with realistic occlusion by moving background points to the object's surface. We evaluate points only if the beam from the LiDAR source to that point intersects with a $7$-meter radius sphere drawn around the inserted object. When changing the object's pose, we limit displacement to under 4 meters on the $xy$-plane so that there is no part of the object exceeding the $7$-meter sphere.}
    \label{fig:roi}
\end{figure*}

A LiDAR point $\hat{\B x}_i$ in the input scene is evaluated for potential adjustment (due to now being occluded by the inserted object) only if a straight line (the laser beam) connecting it to the LiDAR source intersects with the $7$-meter sphere centered at the object insertion location.
We refer to these points as $\{\hat{\B x}_i | i \in \mathcal{I}_{\text{calc}}\}$. 
Any LiDAR points that do not meet these criteria are assumed to remain unoccluded after the object insertion, and are thus kept unchanged ($\B x_i \gets \hat{\B x}_i$). Figure~\ref{fig:roi} illustrates this process.


To process the LiDAR points in $\{\hat{\B x}_i | i\in\mathcal{I}_{\text{calc}}\}$, we sample $J+1$ points on all beams.
To calculate the step size for the $i^{\text{th}}$ beam, we first compute its beam length $k_i = \lVert\hat{\B x}_i - \B s\rVert_2$. 
Then, the step size for this beam is set to be $k_i / J$.
Using this step size, we sample points ${\hat{\B x}_i^{(j)}}$ along the beam between the point, $\hat{\B x}_i$, and the LiDAR source, $\B s$, where $j=0,1,\dots,J$, and endpoints $\hat{\B x}_i^{(0)} = \B s$ and $\hat{\B x}_i^{(J)} = \hat{\B x}_i$.
The value $J$ is chosen to make sure that none of the beams have a step size greater than $0.1$ meters.

Next, we evaluate the signed distance function $g(\B z, T(\hat{\B x}_i^{(j)};\B\theta))$ for all the sampled points. 
If for all $j$, $g(\B z, T(\hat{\B x}_i^{(j)};\B\theta))>0$, then $\hat{\B x}_i$ is not occluded and we set $\B x_i\gets \hat{\B x}_i$ and $m_i\gets 0$. 

Otherwise, $\hat{\B x}_i$ is occluded and needs to be replaced by a $\B x_i$ on the surface of the inserted SDF object, and $m_i\gets 1$.
To this end, we find the first $j$ such that $g(\B z, T(\hat{\B x}_i^{(j)};\B\theta))<0$, which is denoted as $j_{\text{in}}$.  Note that since we place the object at least 15m away from the AV, we know that $j_{\text{in}}>0$ always holds. Finally, we run a binary search of $30$ steps between $\hat{\B x}_i^{(j_{\text{in}})}$ and $\hat{\B x}_i^{(j_{\text{in}}-1)}$ to find the $\B x_i$ such that $|g(\B z, T(\B x_i;\B\theta))|<\epsilon=10^{-6}$. $\B x_i$ is on the surface of the SDF object and added to the output scene, while the original $\hat{\B x}_i$ is occluded and removed from the output scene.

\subsection{Pose Transformation}
To describe the pose transformations $T(\cdot;\B \theta)$, we use a 
6-dimensional vector $\B \theta$ consisting of the coordinates $(x_{\B\theta}, y_{\B\theta}, z_{\B\theta})$ and the angles of yaw ($\delta_{\B\theta}$), pitch ($\beta_{\B\theta}$), and roll ($\gamma_{\B\theta}$). The transformation matrix is given as
\begin{equation}
    T(\B x;\B \theta)= R(\delta_{\B\theta})R(\beta_{\B\theta})R(\gamma_{\B\theta})\left[\B x-(x_{\B\theta}, y_{\B\theta}, z_{\B\theta})^T\right],
\end{equation}
where $R(\delta_{\B\theta,})$, $R(\beta_{\B\theta})$, and $R(\gamma_{\B\theta})$ are $3\times 3$ 3D rotation matrices of yaw, pitch, and roll.

\subsection{Gradient Calculation Using Automatic Differentiation Tools}
\label{sec:auto_diff}
To compute Eq.~\eqref{eq:z_final} using automatic differentiation tools, we  first calculate $(\B e_i\cdot\frac{d \mathcal{L}}{d \B x_i})$ and $(\B e_i\cdot\frac{\partial g(\B z, \cdot)}{\partial \B x_i})^{-1}$ for all $\B x_i$, utilizing two back-propagation operations on $\mathcal{L}$ and $g$. Subsequently, we detach the resulting two tensors from the computational graph and treat them as coefficients. Finally, we perform another back-propagation operation with respect to $\B z$ on a weighted sum of SDF values. A similar three-step strategy is also applied to Eq.~\eqref{eq:theta_final}.

\section{Detailed Information of Experiments}
\label{sec:supp_experiments}
\subsection{Metrics of Detectors}
\label{sec:supp_natural_metrics}
We first train our PointPillars and SST detector models exclusively on point cloud data from the Waymo Open Dataset training set. We did not incorporate any auxiliary inputs such as intensity. We show the performance of our model on the vanilla WOD detection task in Table~\ref{tab:natural_metrics} and compare it with the baseline model in~\cite{sun2020scalability}.

\begin{table}[ht!]
  \centering
  \begin{tabular}{c|c|c|c}
  \toprule
    Model   &   \thead{Metric \\ (Vehicle)} & \thead{Overall BEV \\ (LVL\_1/LVL\_2)}   &\thead{ Overall 3D \\ (LVL\_1/LVL\_2 )}\\\midrule
    
    \multirow{2}{*}{Baseline in~\cite{sun2020scalability}}&  APH & 79.1/71.0  &62.8/55.1\\
    &  AP & 80.1/71.9  &63.3/55.6\\\midrule

    \multirow{2}{*}{Our PointPillars}&  APH &  81.0/73.3 &60.3/52.6\\
    &  AP & 82.3/74.6  &61.0/53.3\\\midrule

    \multirow{2}{*}{Our SST}&  APH & 89.2 /81.8 &74.8/66.4\\
    &  AP & 90.0/ 82.5 &75.3/66.8\\\midrule

    Our PointPillars&  APH &  81.2/73.5 &60.5/52.8\\
    after Fine-tuning&  AP & 82.6/74.9  &61.2/53.5\\
    
    \bottomrule
  \end{tabular}

  \caption{Our PointPillars and SST detectors achieves comparable APH and AP for vehicles as the baseline model reported in~\cite{sun2020scalability}. The fine-tuned PointPillars detector model shown in the last row is discussed in Section~\ref{sec:finetune}.}
  \label{tab:natural_metrics}
\end{table}

\subsection{Hyper-Parameters}
In Section~\ref{sec:hyperparams}, we provided a summary of the hyper-parameters utilized in our experiments. Here, we provide additional details regarding the hyper-parameters used in the generation of adversarial shape and pose.

We conducted a hyper-parameter search to determine the optimal learning rate $\alpha$ for generating both adversarial shape and adversarial pose. Specifically, we considered values of $\alpha$ in the set \{0.0001, 0.001, 0.01, 0.1\}. In addition, for adversarial shape generation experiments, we also searched for the optimal value of $\lambda$ in the set \{0.1, 1, 10\}.

To select the optimal value of $\alpha$, we employed the threshold-recall curve's AUC metric as the target. Our experiments revealed that $\alpha=0.01$ resulted in the lowest AUC value for both shape and pose experiments. Moreover, for adversarial shape generation, the lowest AUC value was obtained with $\alpha=0.01$ regardless of the value of $\lambda$.

To select the optimal value of $\lambda$, we chose the value that produced the lowest loss value $\mathcal{L}_{\text{adv}}(\B z)$. In instances where different values of $\lambda$ produced similar loss values, we chose the larger value of $\lambda$ to minimize the perturbation. Our experiments determined that $\lambda=10$ was optimal for Coupe and Sports Car, while $\lambda=1$ was optimal for SUV, Convertible Car, and Beach Wagon.

\section{Additional Visualizations for Adversarial Shape Generation}
\label{sec:more_shape_vis}
\begin{figure*}[htbp]

  \centering
  \caption{Visualizations of random selected challenging objects generated by \ourmethod{}}
  \label{fig:shape_vis_big_table}
  \begin{tabular}{c|c|c|c|c}
    \toprule
    Sports Car & SUV & \thead{Convertible\\Car} &\thead{Beach\\Wagon}&Coupe\\\midrule

\begin{subfigure}[b]{0.17\linewidth}
      \centering
      \includegraphics[width=\linewidth,trim={15cm 5cm 20cm 12cm},clip]{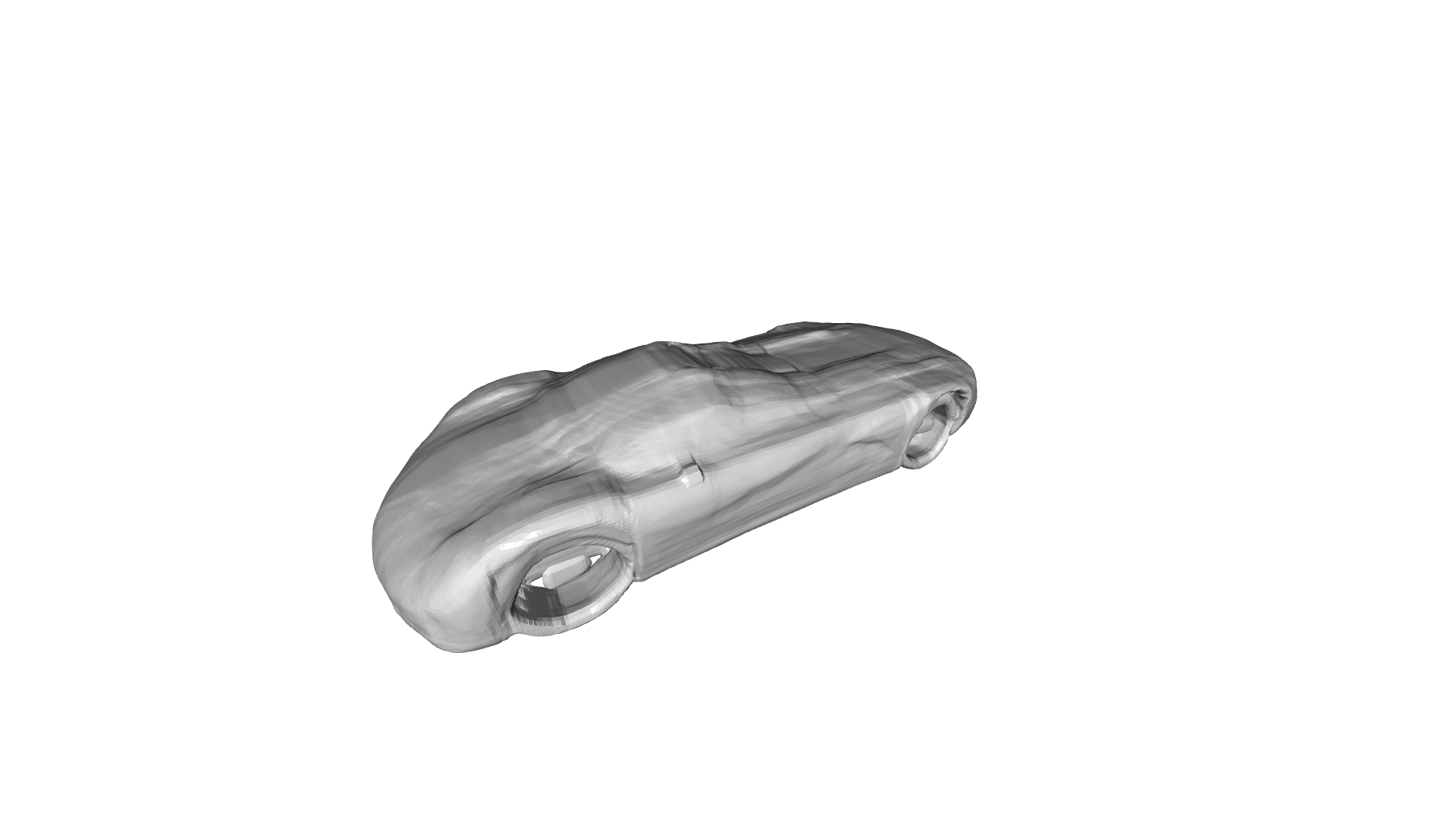}
    \end{subfigure}

&

\begin{subfigure}[b]{0.17\linewidth}
      \centering
      \includegraphics[width=\linewidth,trim={15cm 4cm 20cm 10cm},clip]{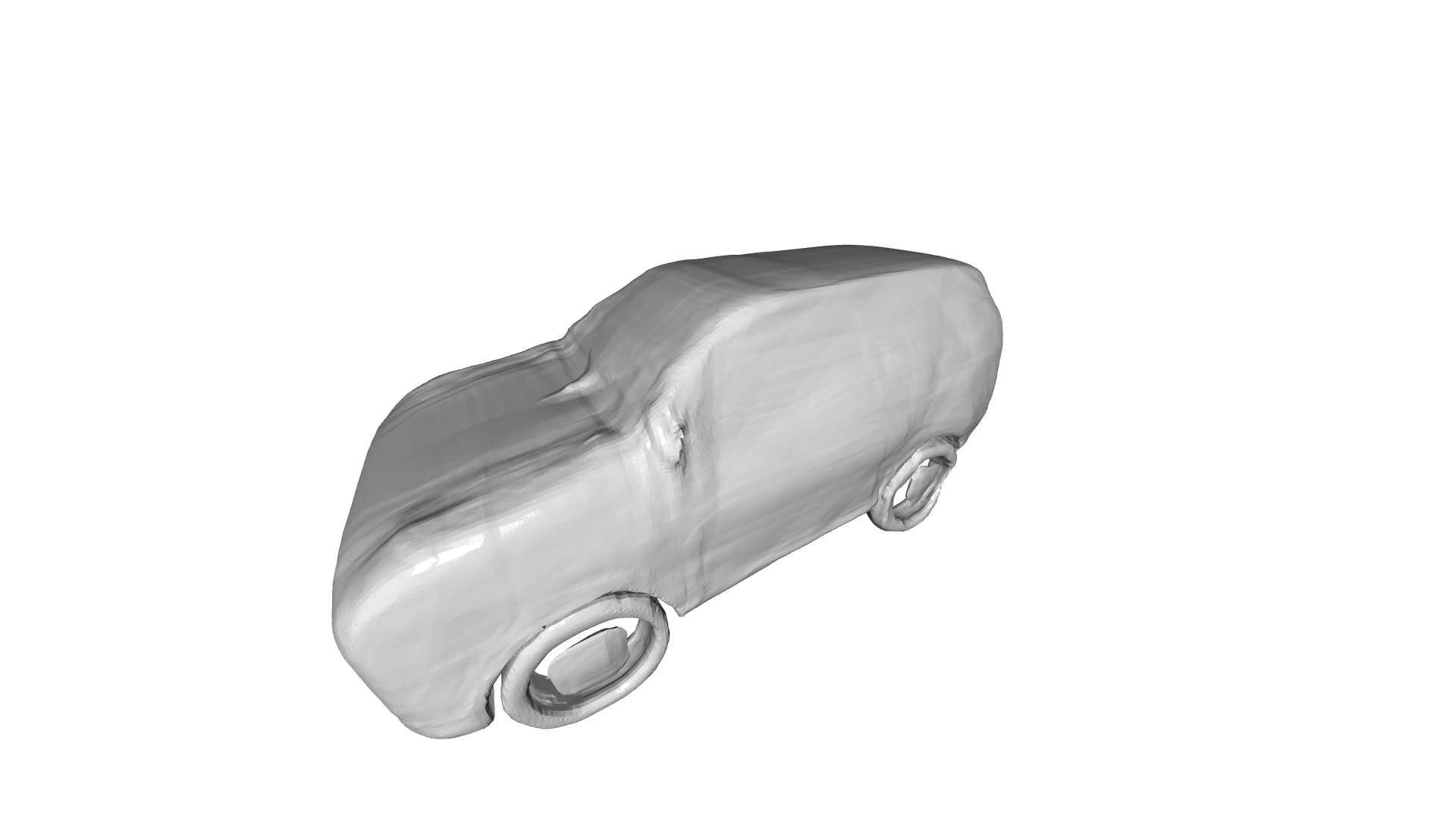}
    \end{subfigure}

&

\begin{subfigure}[b]{0.17\linewidth}
      \centering
      \includegraphics[width=\linewidth,trim={15cm 5cm 20cm 12cm},clip]{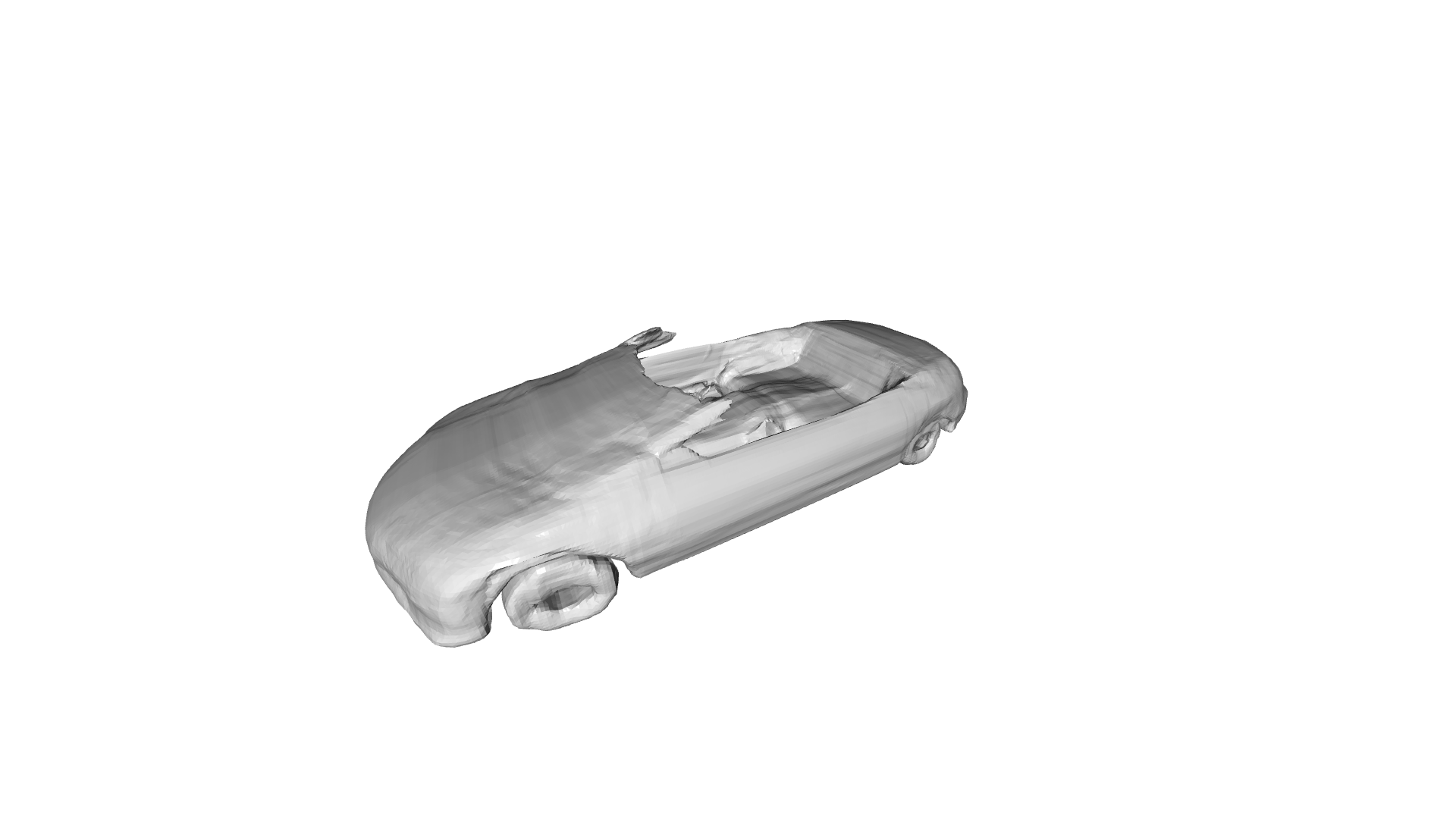}
    \end{subfigure}

&

\begin{subfigure}[b]{0.17\linewidth}
      \centering
      \includegraphics[width=\linewidth,trim={15cm 5cm 20cm 12cm},clip]{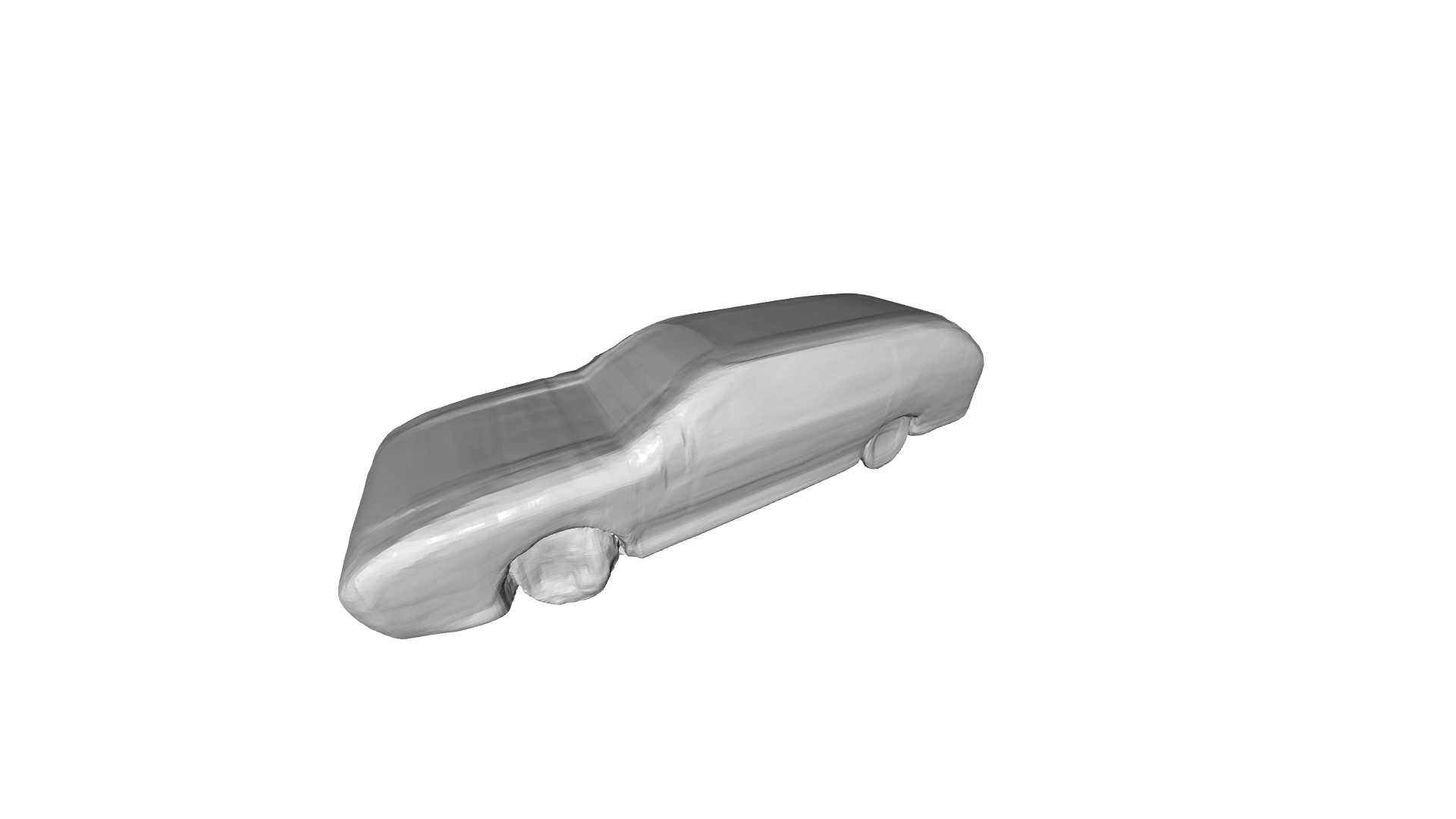}
    \end{subfigure}

&

\begin{subfigure}[b]{0.17\linewidth}
      \centering
      \includegraphics[width=\linewidth,trim={15cm 5cm 20cm 12cm},clip]{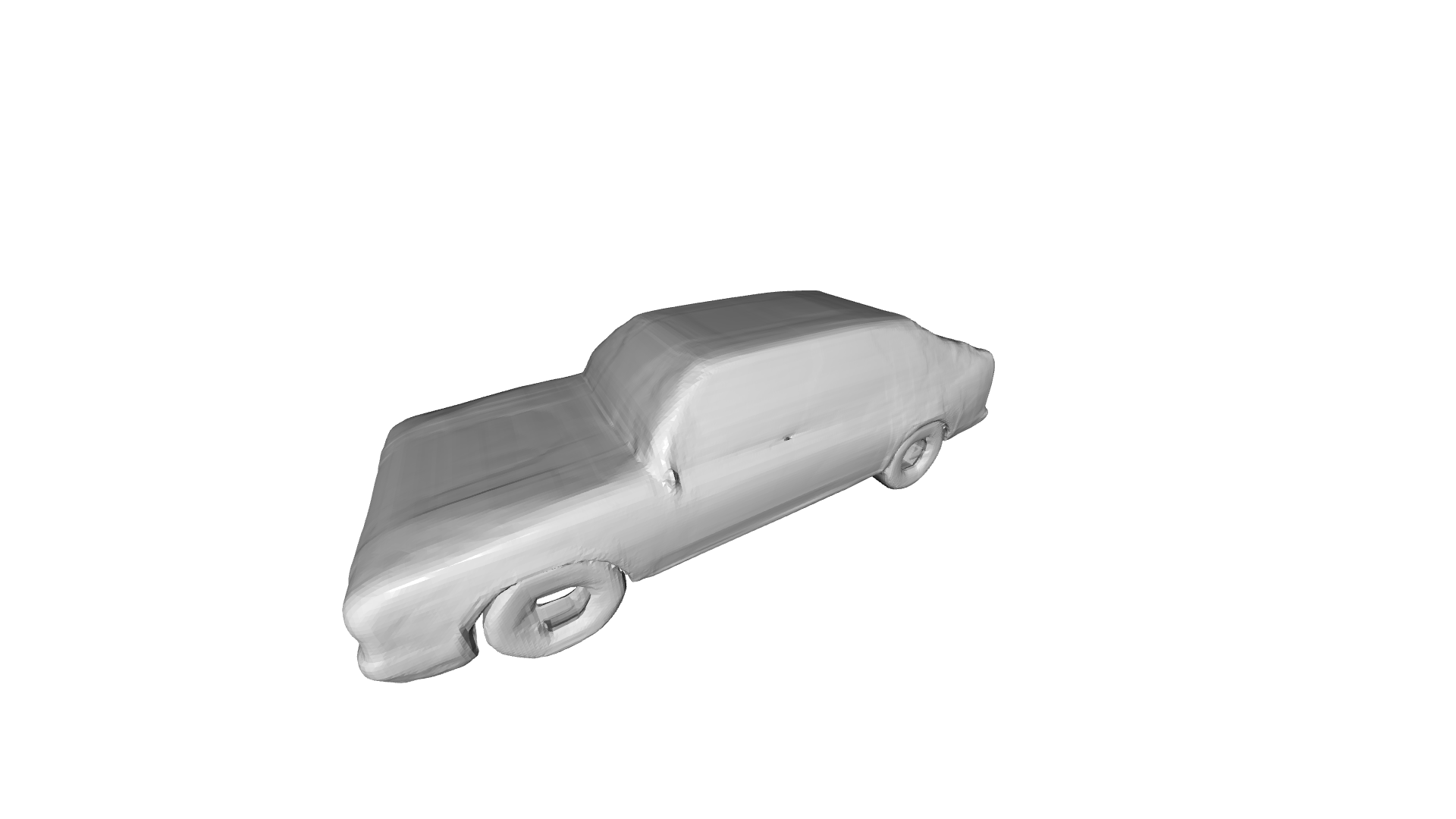}
    \end{subfigure}

\\
\begin{subfigure}[b]{0.17\linewidth}
      \centering
      \includegraphics[width=\linewidth,trim={15cm 5cm 20cm 12cm},clip]{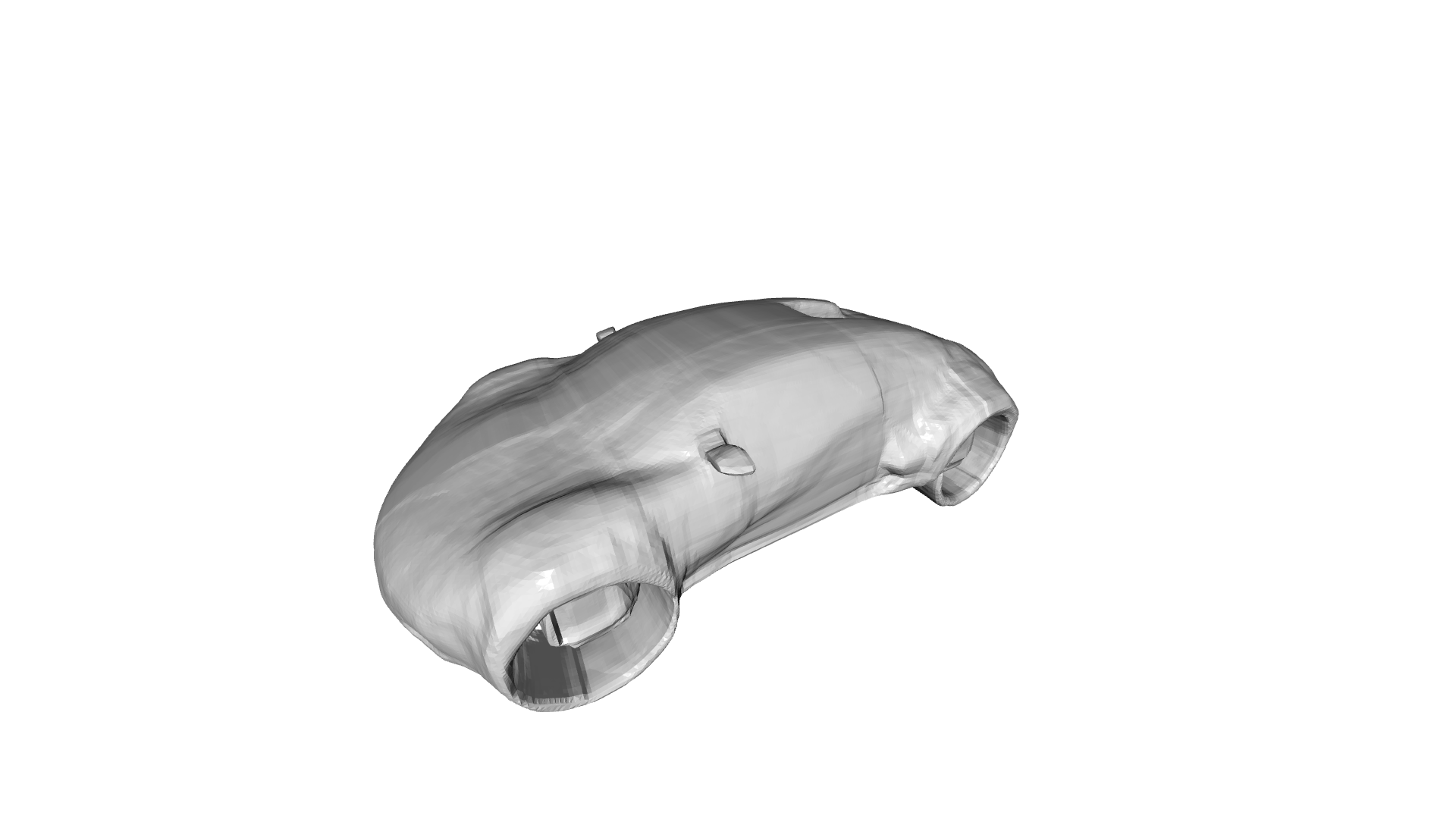}
    \end{subfigure}

&

\begin{subfigure}[b]{0.17\linewidth}
      \centering
      \includegraphics[width=\linewidth,trim={15cm 5cm 20cm 12cm},clip]{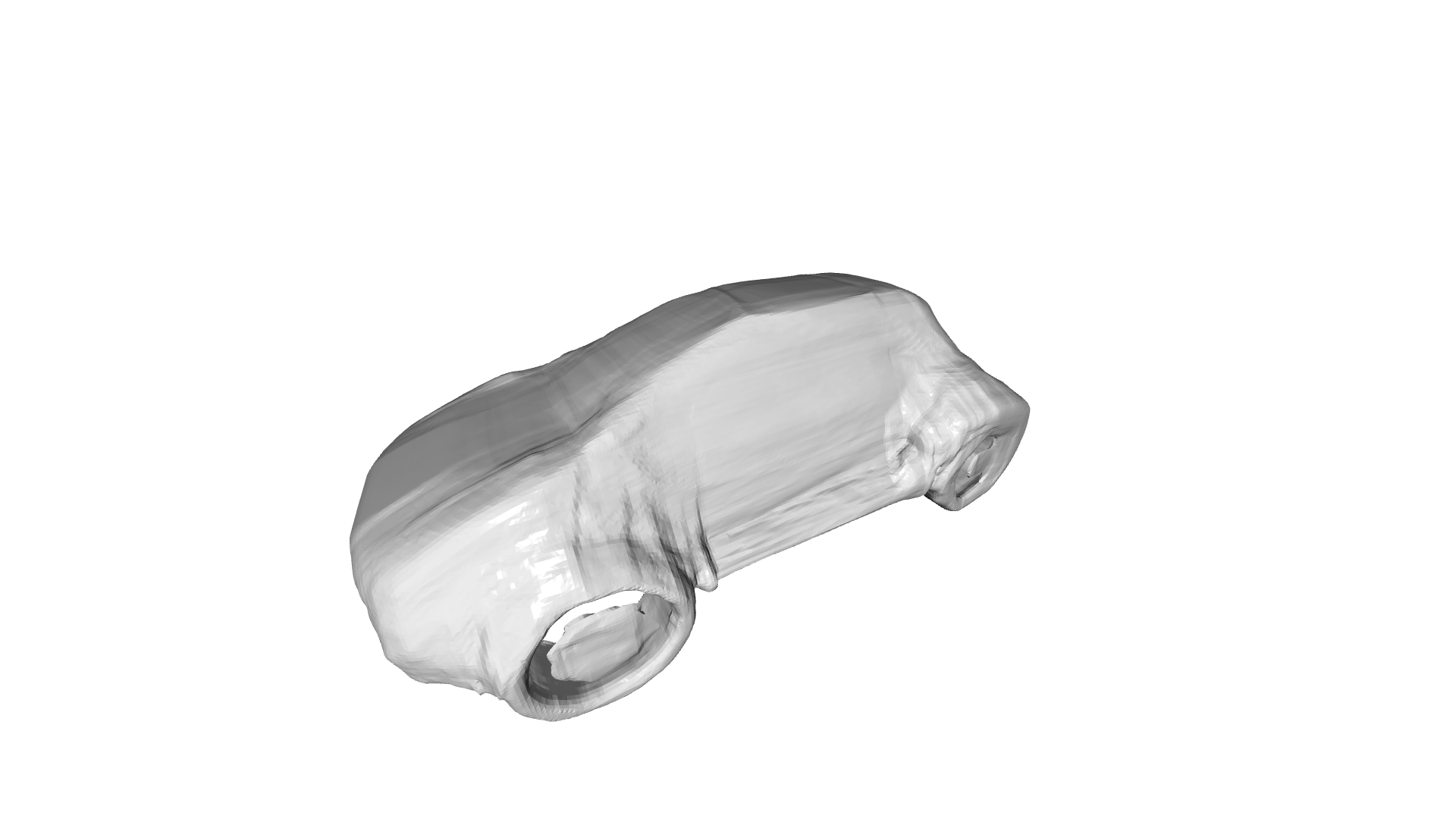}
    \end{subfigure}

&

\begin{subfigure}[b]{0.17\linewidth}
      \centering
      \includegraphics[width=\linewidth,trim={15cm 5cm 20cm 12cm},clip]{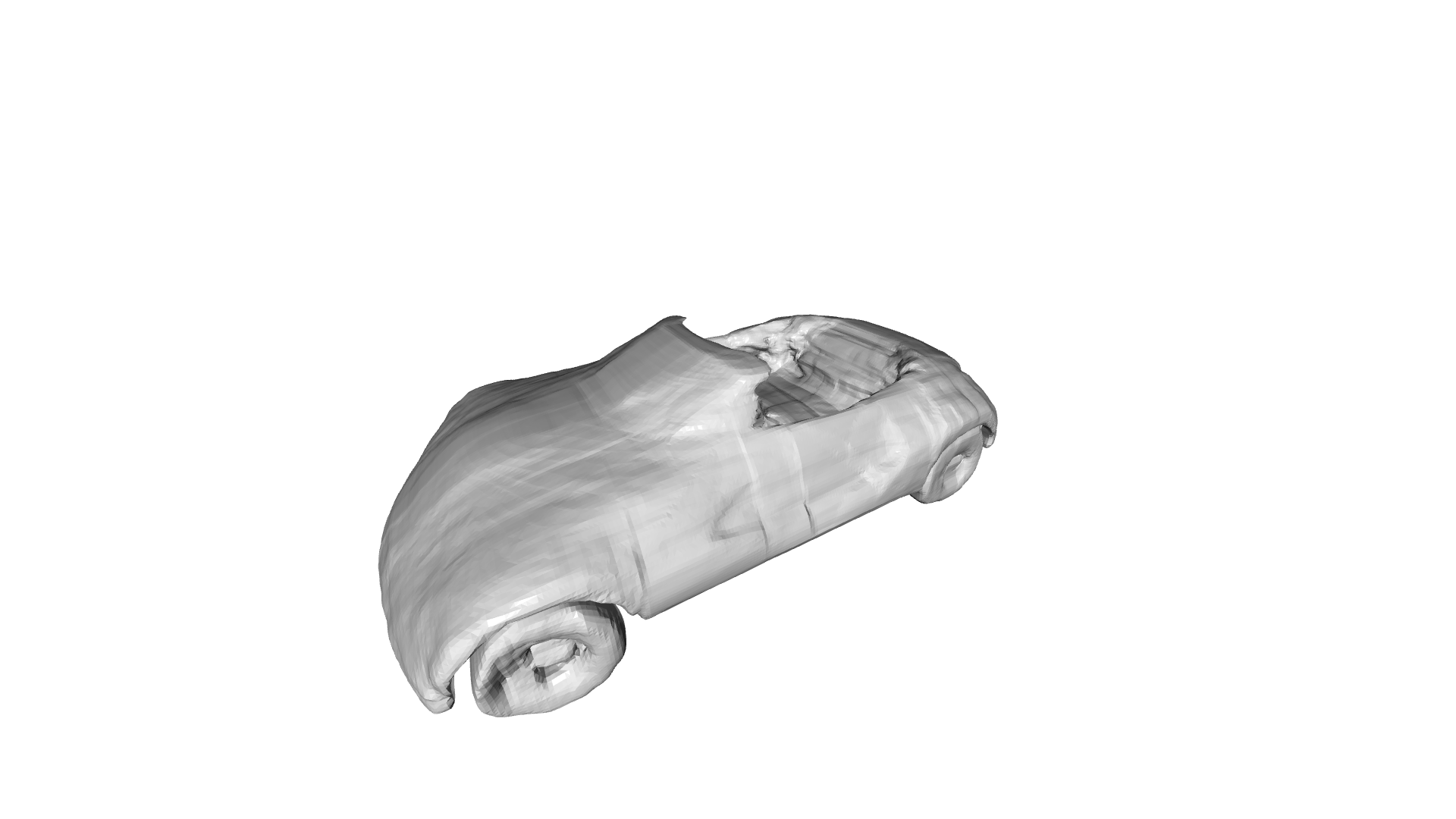}
    \end{subfigure}

&

\begin{subfigure}[b]{0.17\linewidth}
      \centering
      \includegraphics[width=\linewidth,trim={15cm 5cm 20cm 12cm},clip]{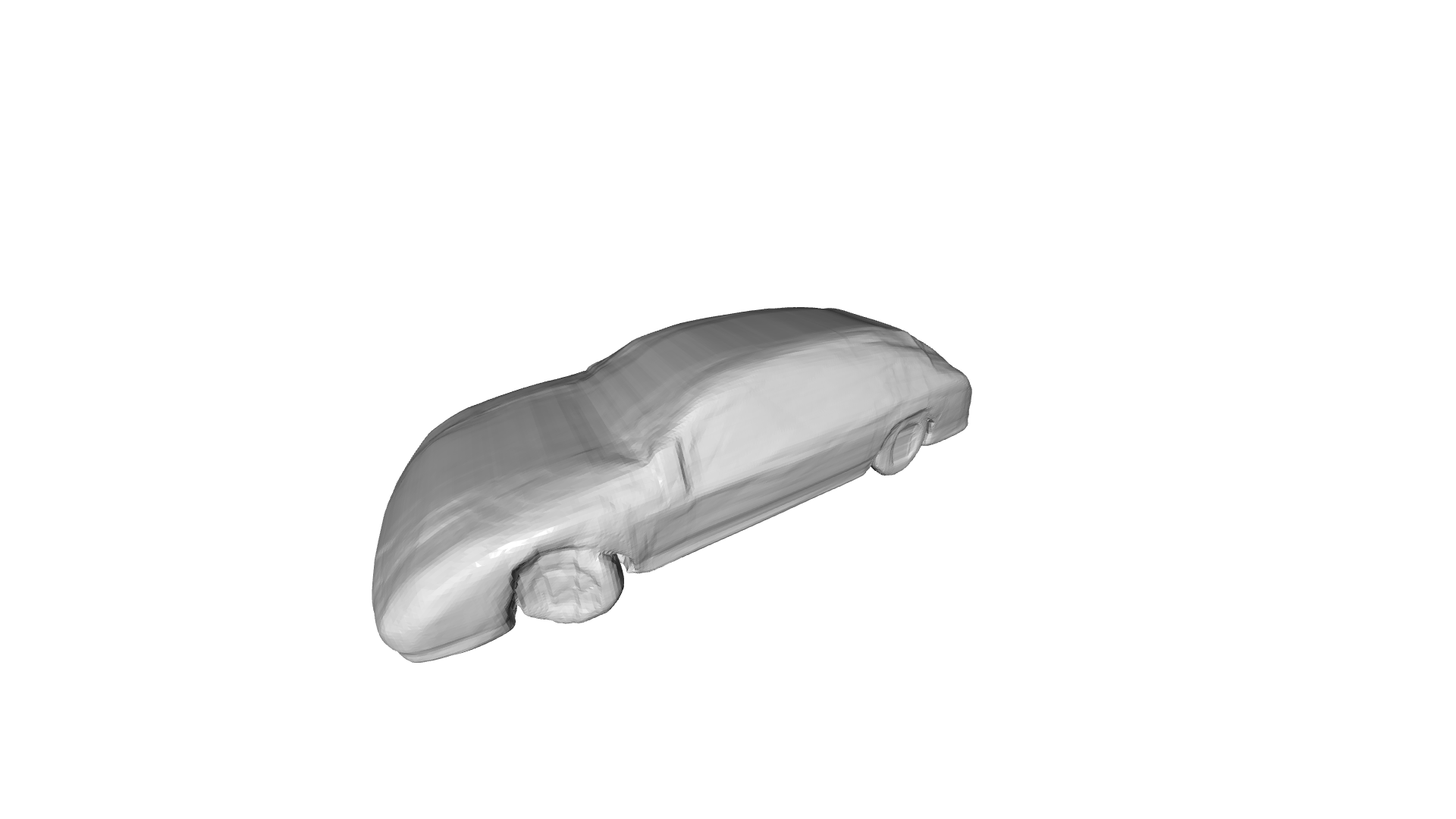}
    \end{subfigure}

&

\begin{subfigure}[b]{0.17\linewidth}
      \centering
      \includegraphics[width=\linewidth,trim={15cm 5cm 20cm 12cm},clip]{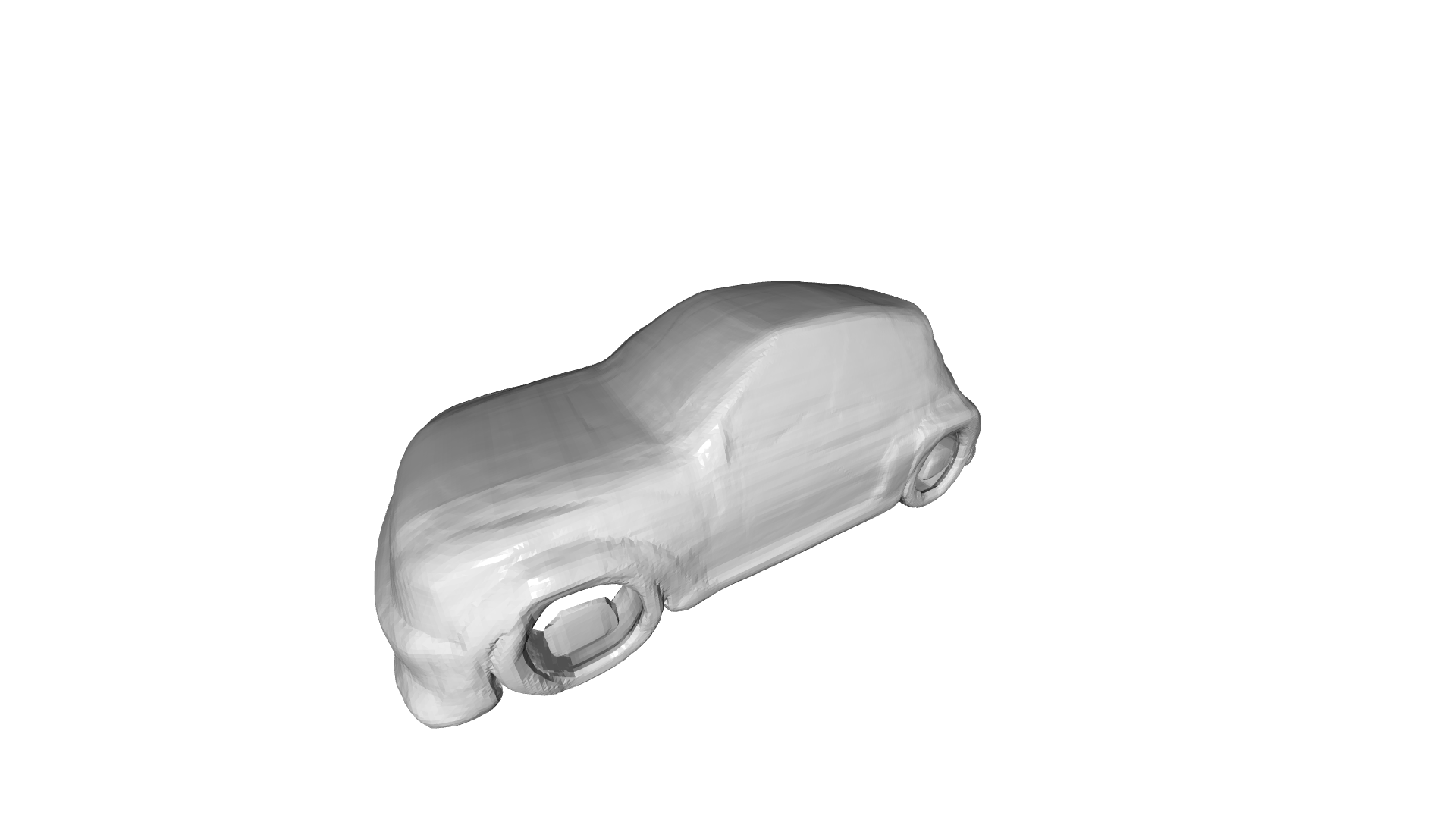}
    \end{subfigure}

\\
\begin{subfigure}[b]{0.17\linewidth}
      \centering
      \includegraphics[width=\linewidth,trim={15cm 5cm 20cm 12cm},clip]{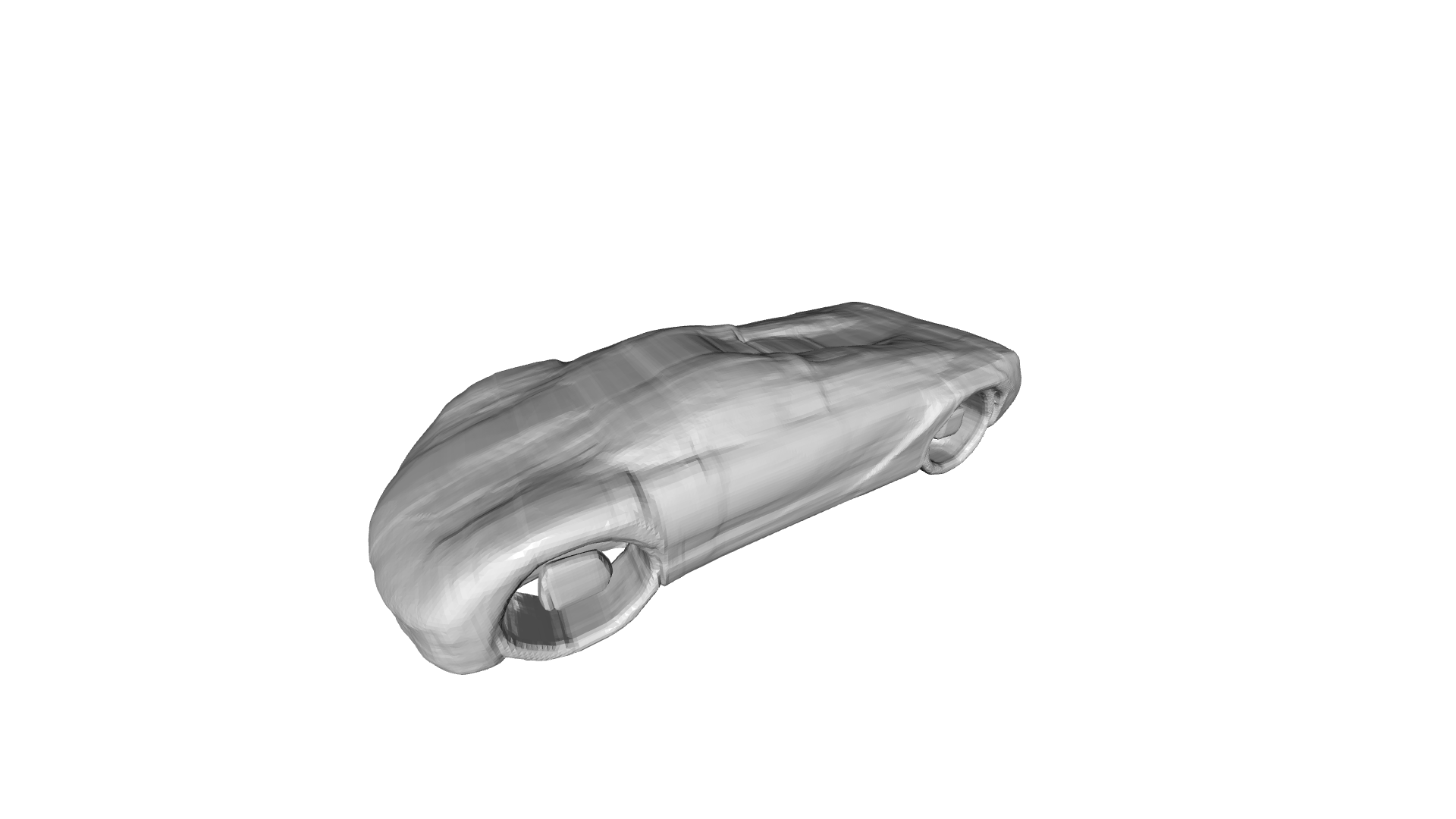}
    \end{subfigure}

&

\begin{subfigure}[b]{0.17\linewidth}
      \centering
      \includegraphics[width=\linewidth,trim={15cm 5cm 20cm 12cm},clip]{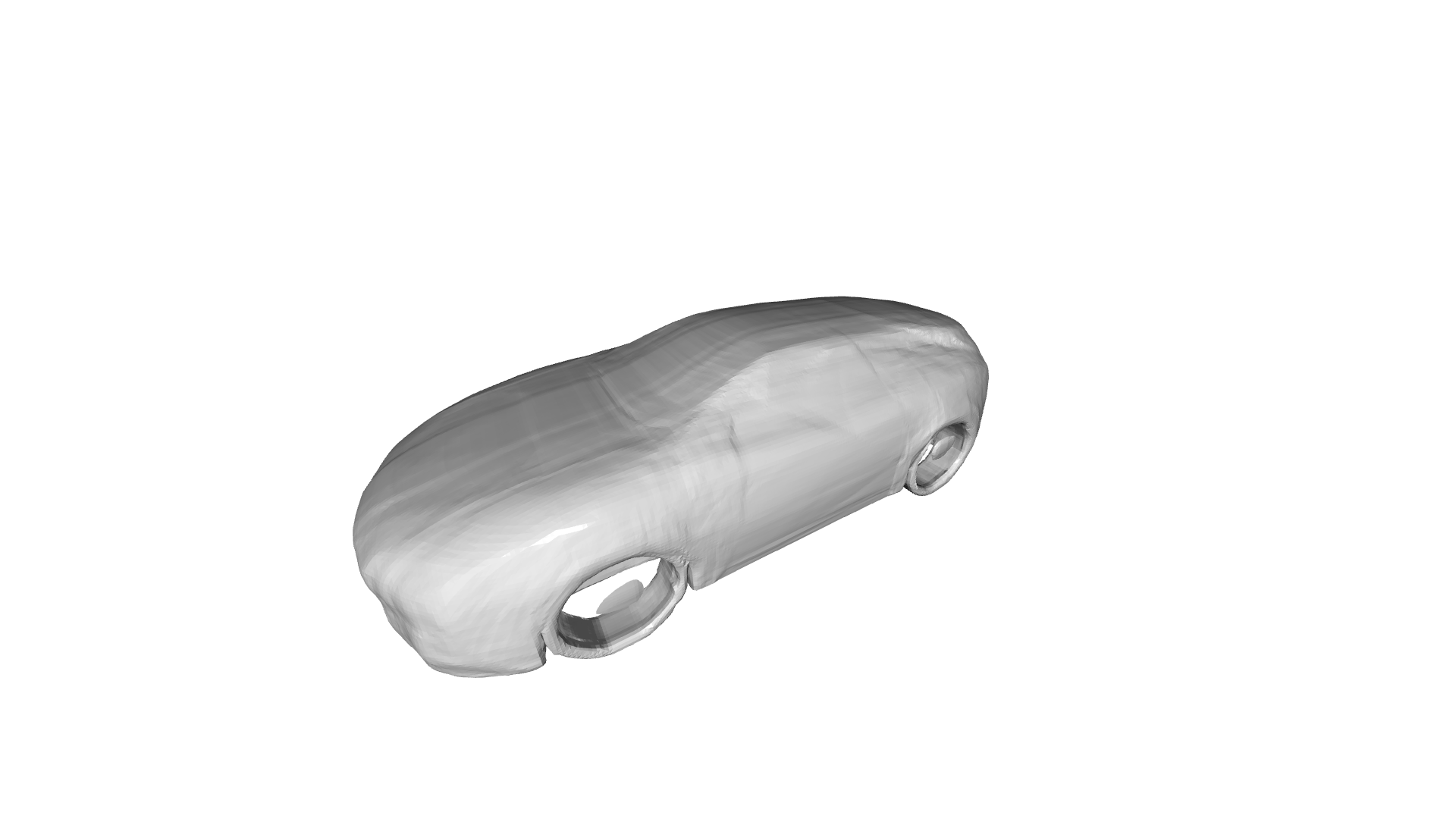}
    \end{subfigure}

&

\begin{subfigure}[b]{0.17\linewidth}
      \centering
      \includegraphics[width=\linewidth,trim={15cm 5cm 20cm 12cm},clip]{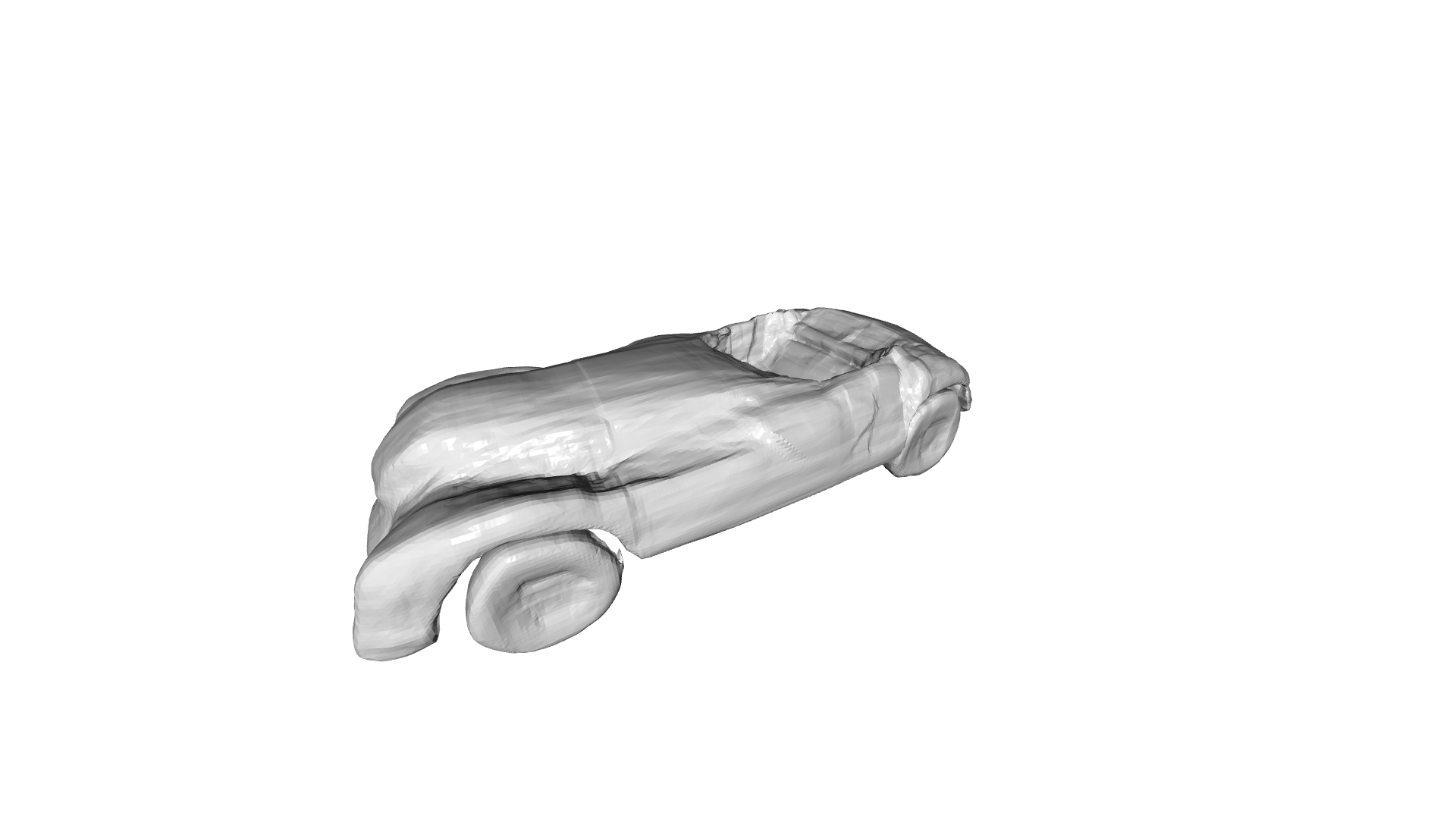}
    \end{subfigure}

&

\begin{subfigure}[b]{0.17\linewidth}
      \centering
      \includegraphics[width=\linewidth,trim={15cm 5cm 20cm 12cm},clip]{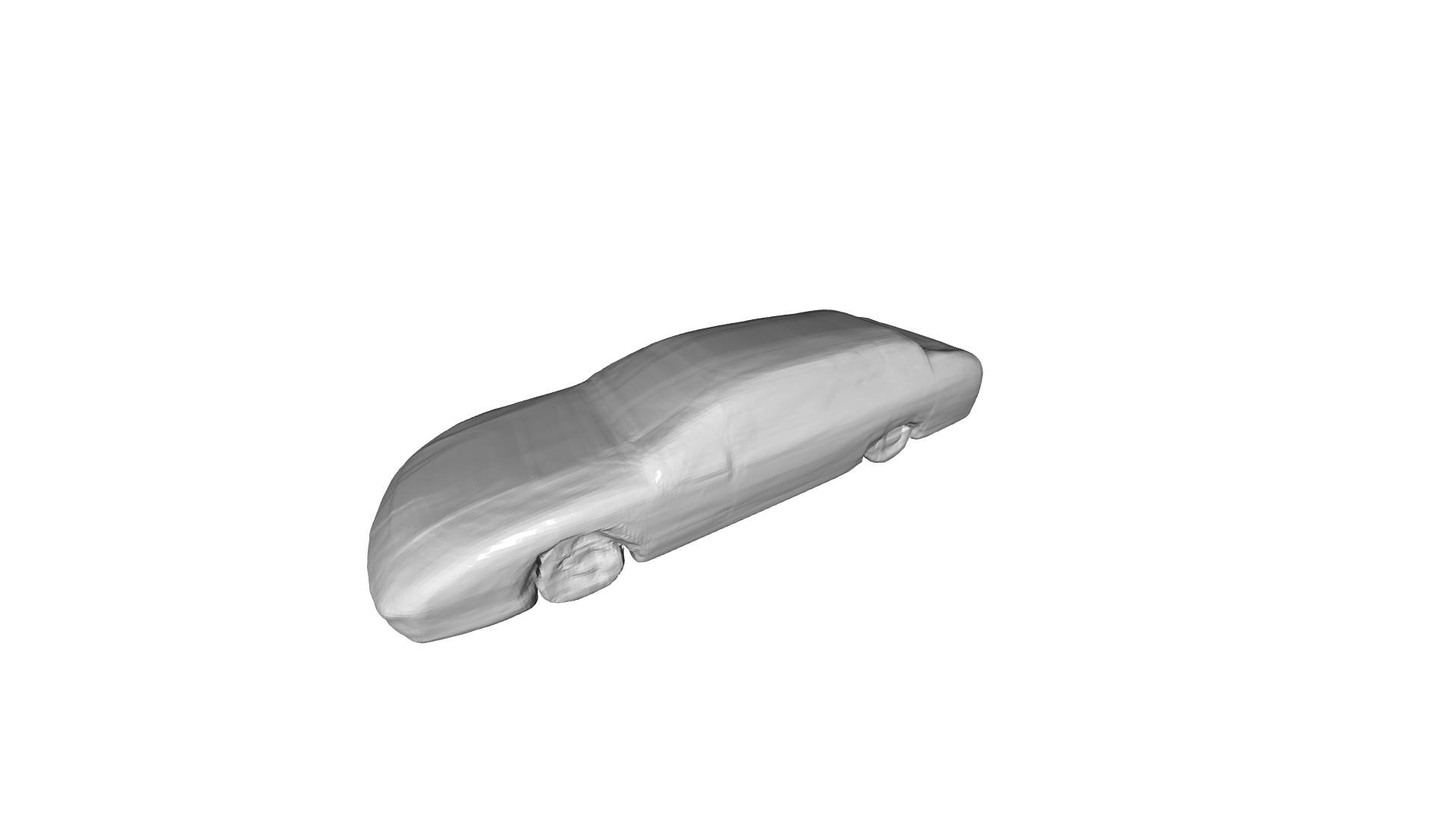}
    \end{subfigure}

&

\begin{subfigure}[b]{0.17\linewidth}
      \centering
      \includegraphics[width=\linewidth,trim={15cm 5cm 20cm 12cm},clip]{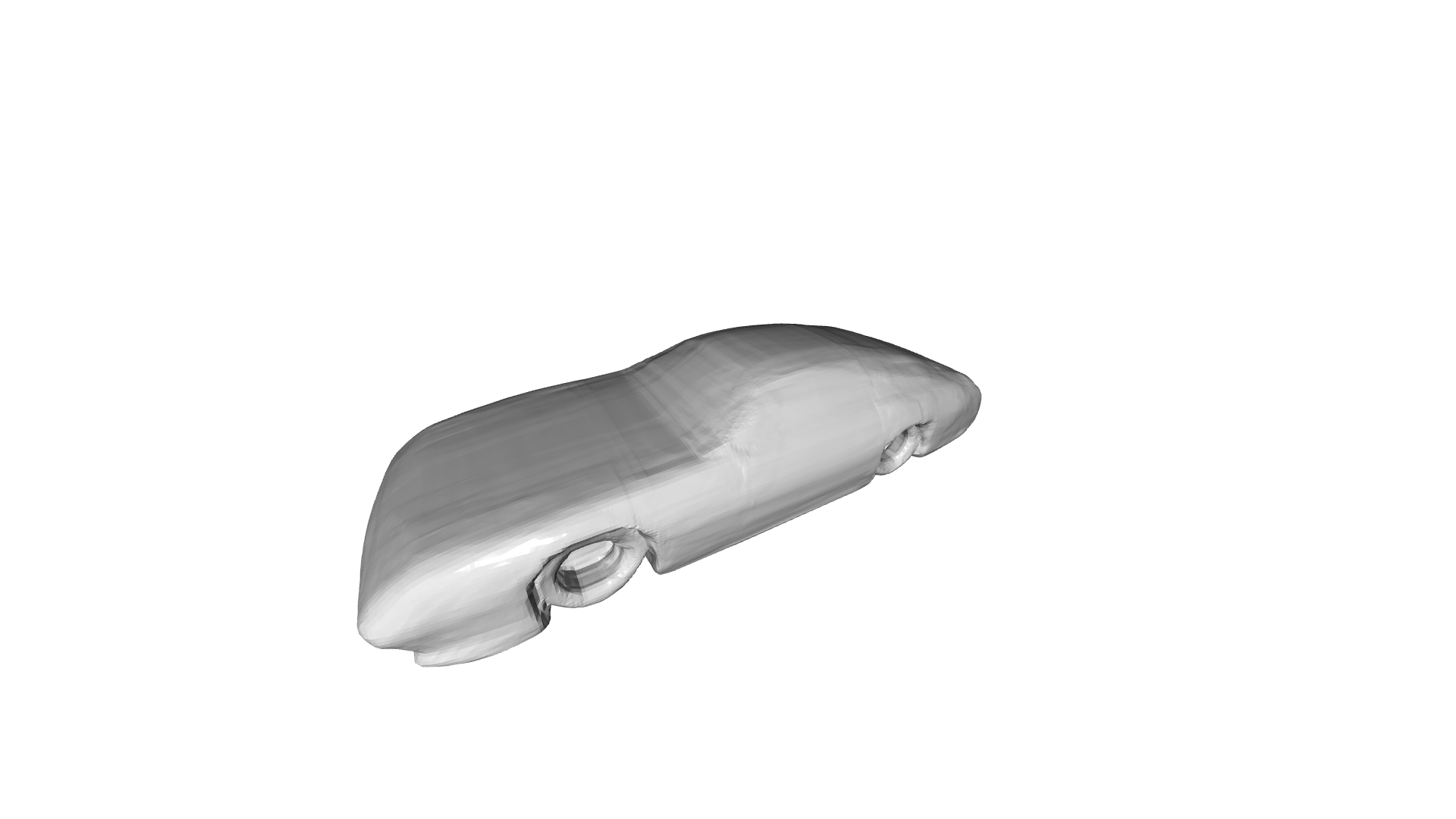}
    \end{subfigure}

\\
\begin{subfigure}[b]{0.17\linewidth}
      \centering
      \includegraphics[width=\linewidth,trim={15cm 5cm 20cm 12cm},clip]{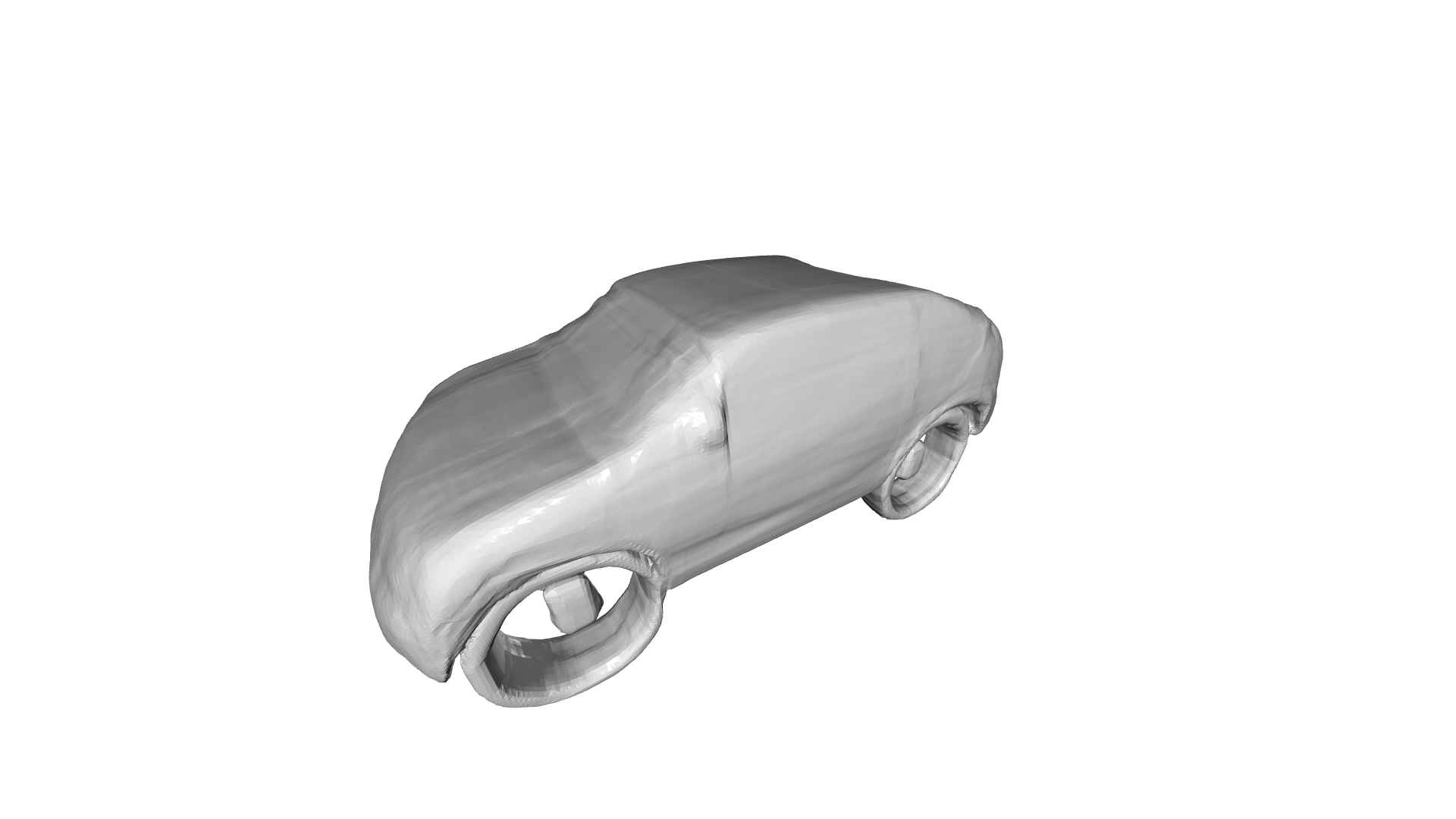}
    \end{subfigure}

&

\begin{subfigure}[b]{0.17\linewidth}
      \centering
      \includegraphics[width=\linewidth,trim={15cm 5cm 20cm 12cm},clip]{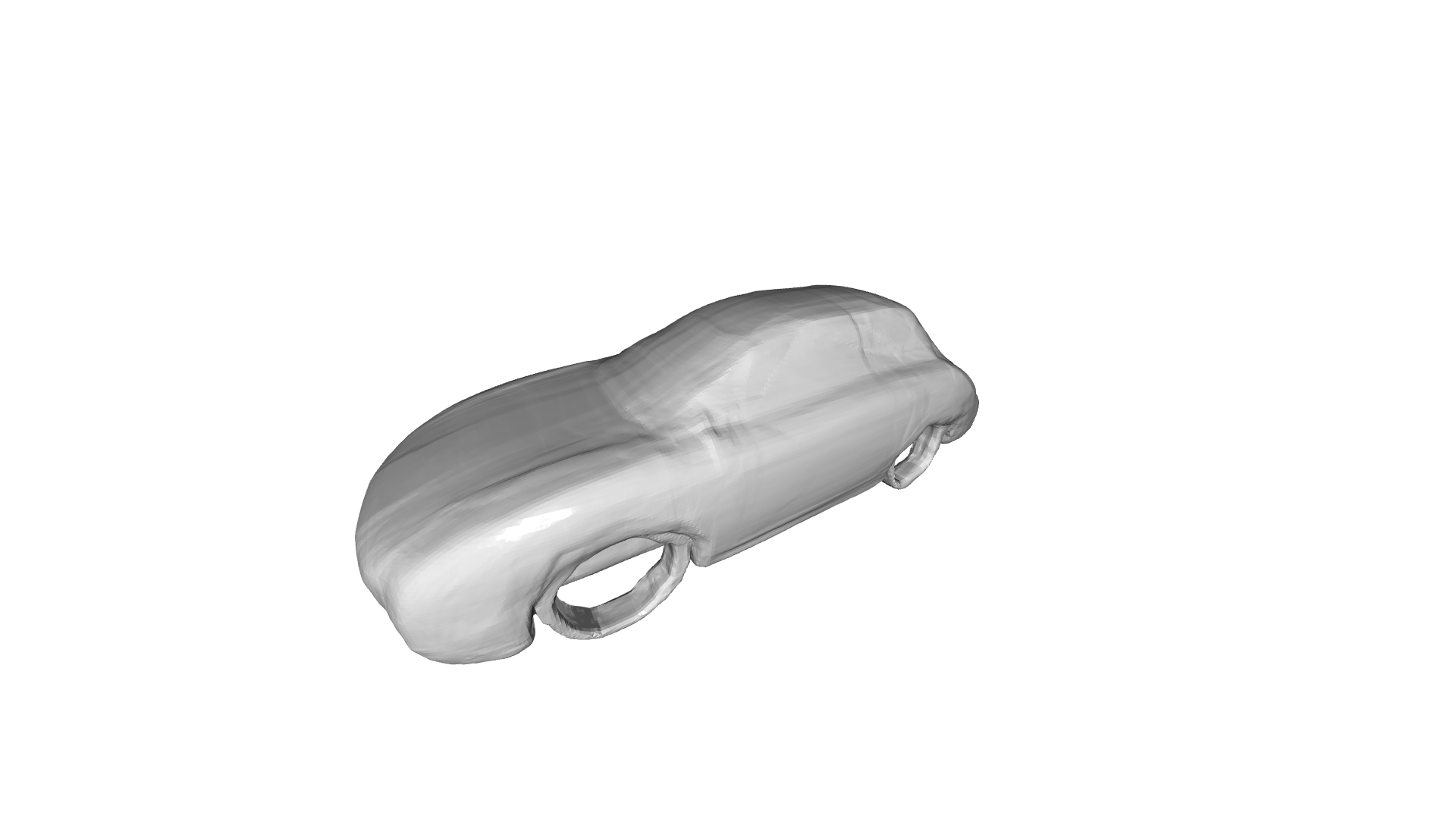}
    \end{subfigure}

&

\begin{subfigure}[b]{0.17\linewidth}
      \centering
      \includegraphics[width=\linewidth,trim={15cm 5cm 20cm 12cm},clip]{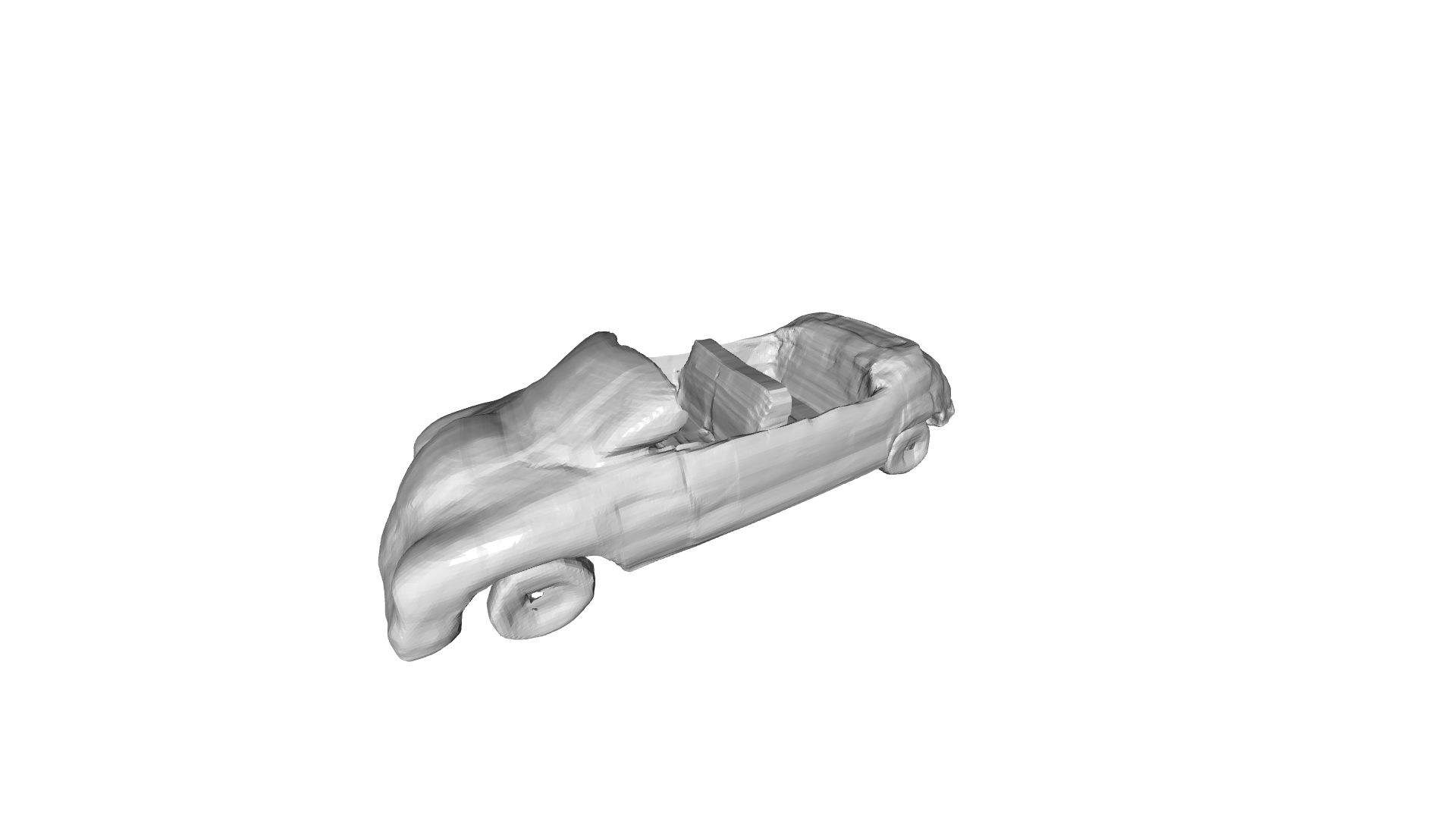}
    \end{subfigure}

&

\begin{subfigure}[b]{0.17\linewidth}
      \centering
      \includegraphics[width=\linewidth,trim={15cm 5cm 20cm 12cm},clip]{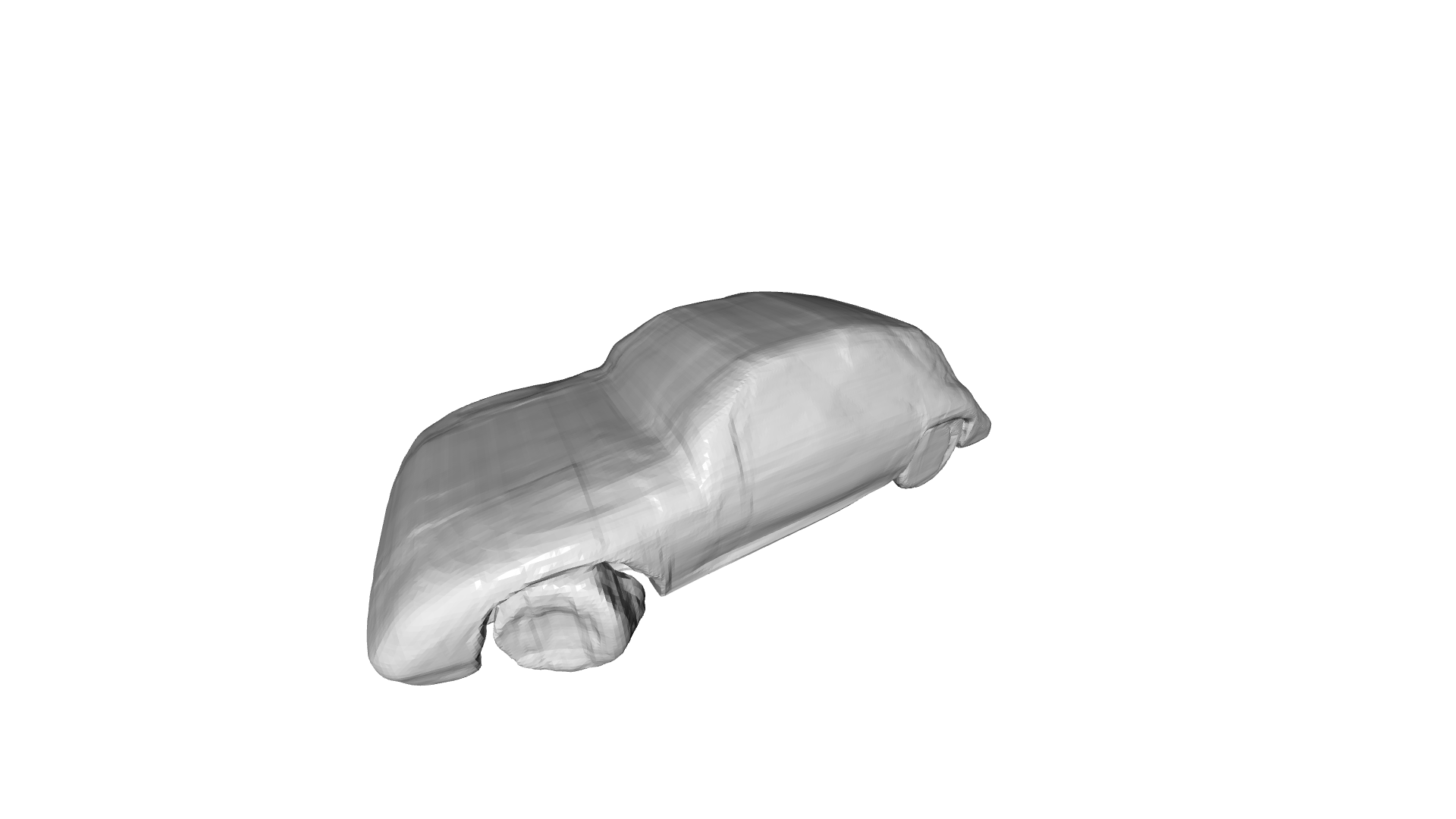}
    \end{subfigure}

&

\begin{subfigure}[b]{0.17\linewidth}
      \centering
      \includegraphics[width=\linewidth,trim={15cm 5cm 20cm 12cm},clip]{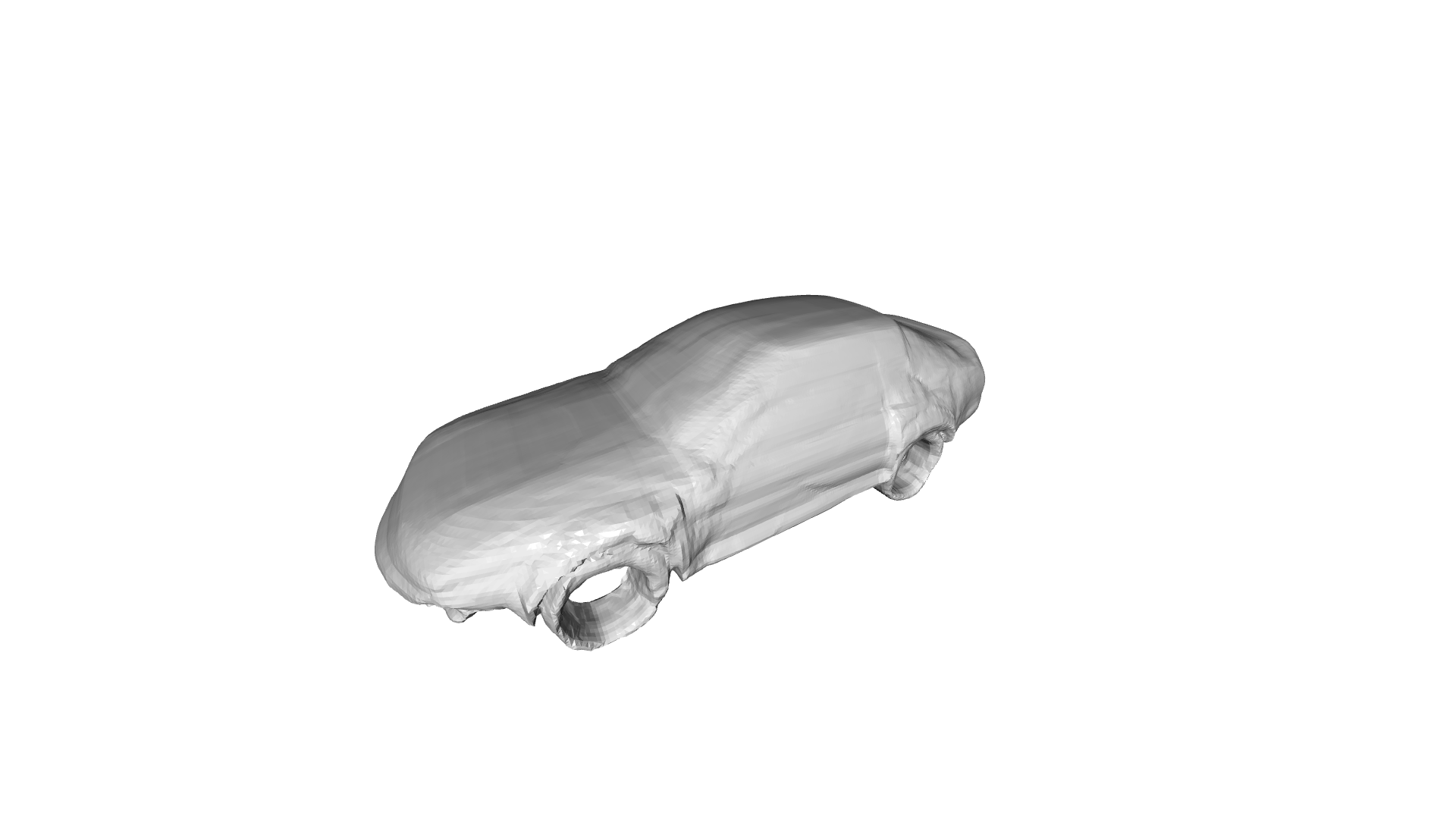}
    \end{subfigure}

\\
\begin{subfigure}[b]{0.17\linewidth}
      \centering
      \includegraphics[width=\linewidth,trim={15cm 5cm 20cm 12cm},clip]{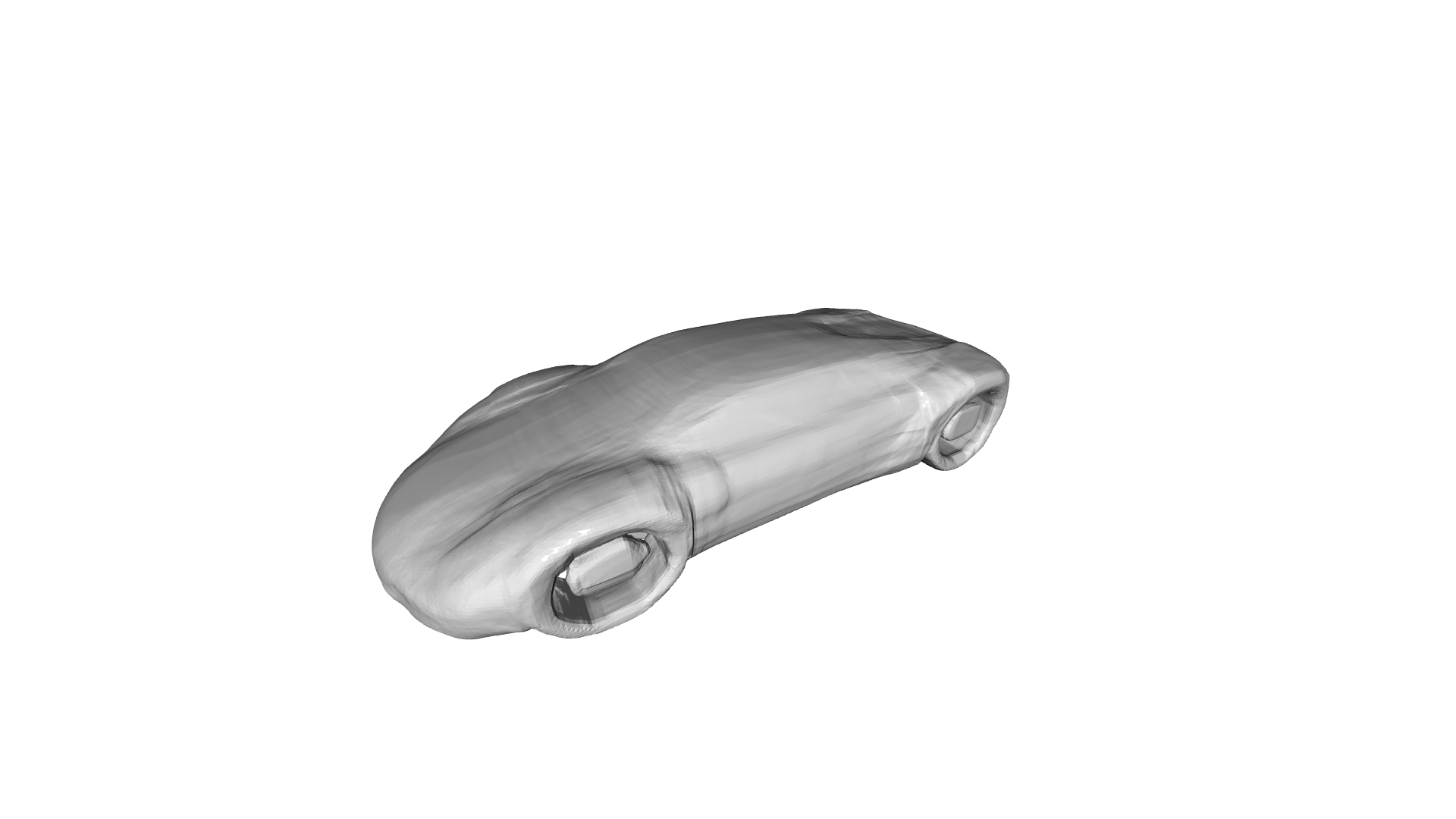}
    \end{subfigure}

&

\begin{subfigure}[b]{0.17\linewidth}
      \centering
      \includegraphics[width=\linewidth,trim={15cm 5cm 20cm 10cm},clip]{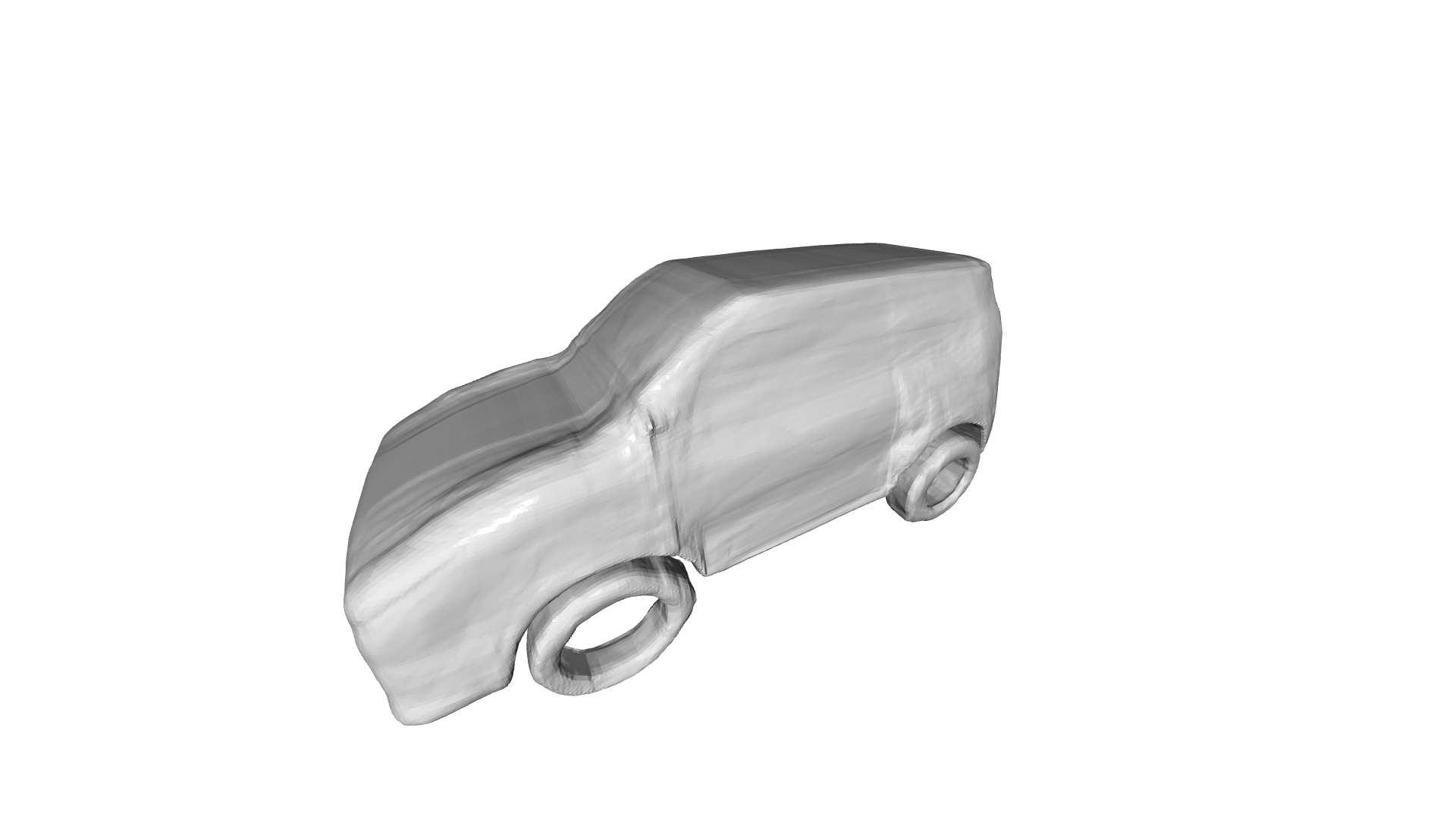}
    \end{subfigure}

&

\begin{subfigure}[b]{0.17\linewidth}
      \centering
      \includegraphics[width=\linewidth,trim={15cm 5cm 20cm 12cm},clip]{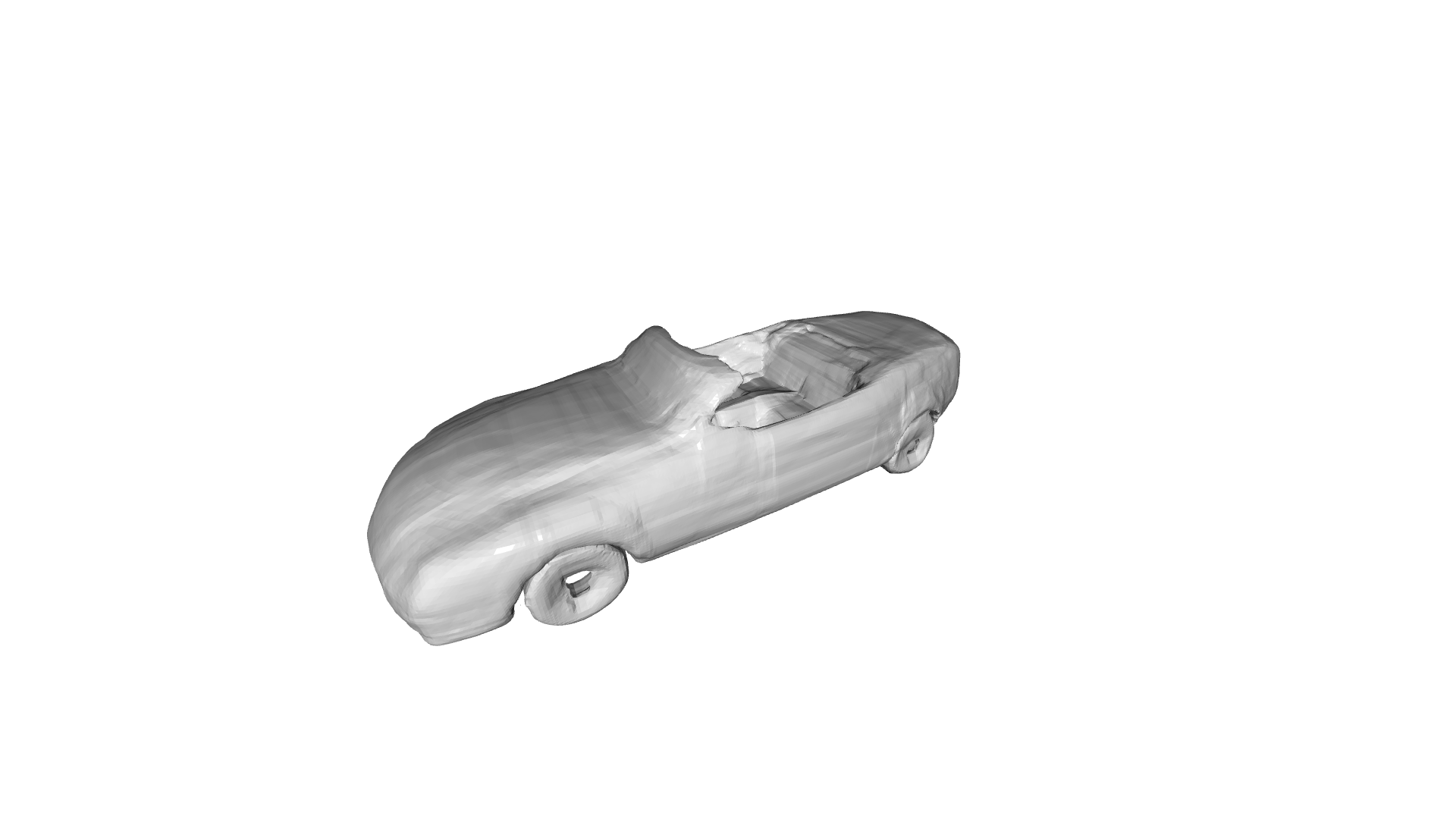}
    \end{subfigure}

&

\begin{subfigure}[b]{0.17\linewidth}
      \centering
      \includegraphics[width=\linewidth,trim={15cm 5cm 20cm 12cm},clip]{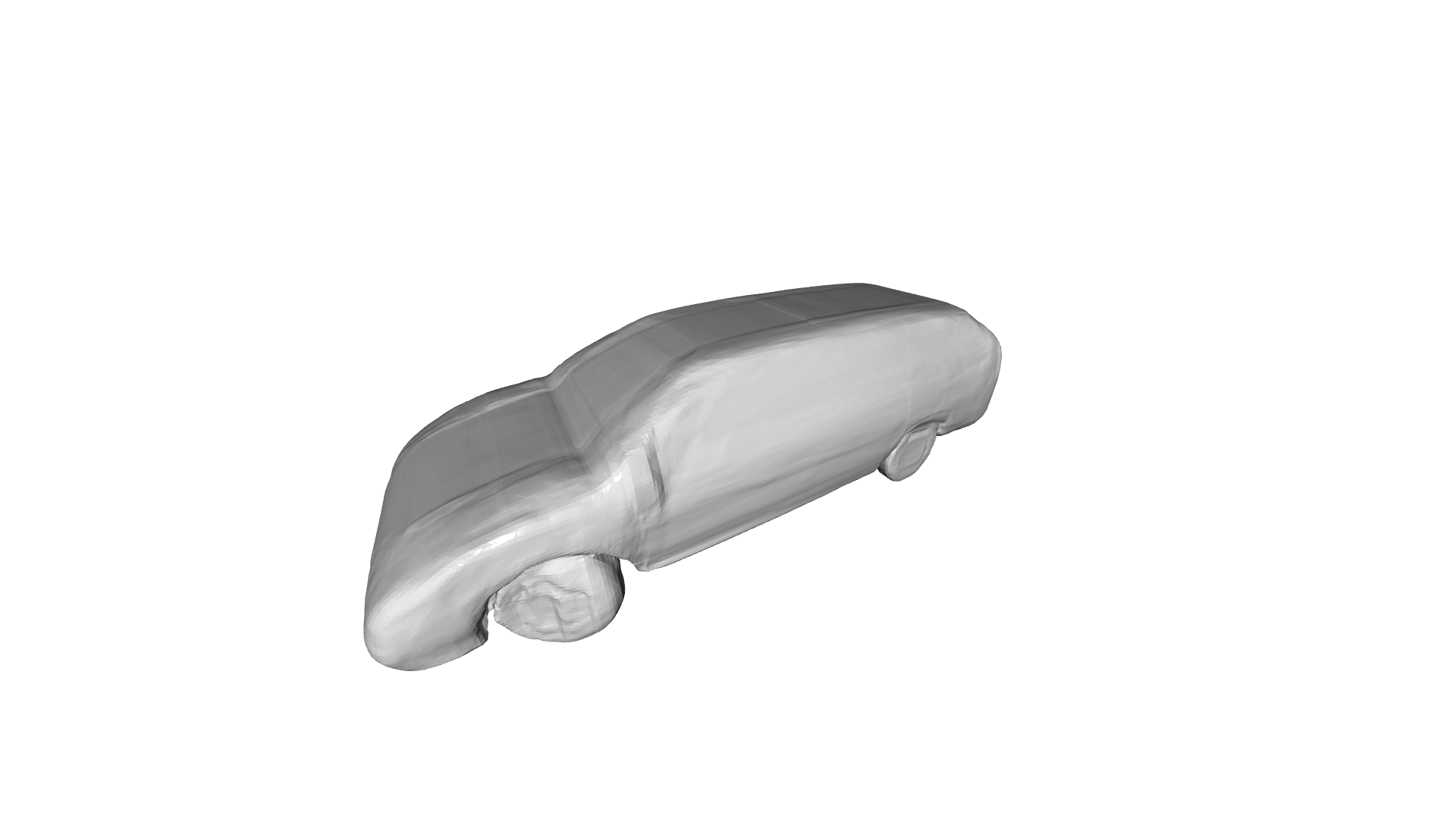}
    \end{subfigure}

&

\begin{subfigure}[b]{0.17\linewidth}
      \centering
      \includegraphics[width=\linewidth,trim={15cm 5cm 20cm 12cm},clip]{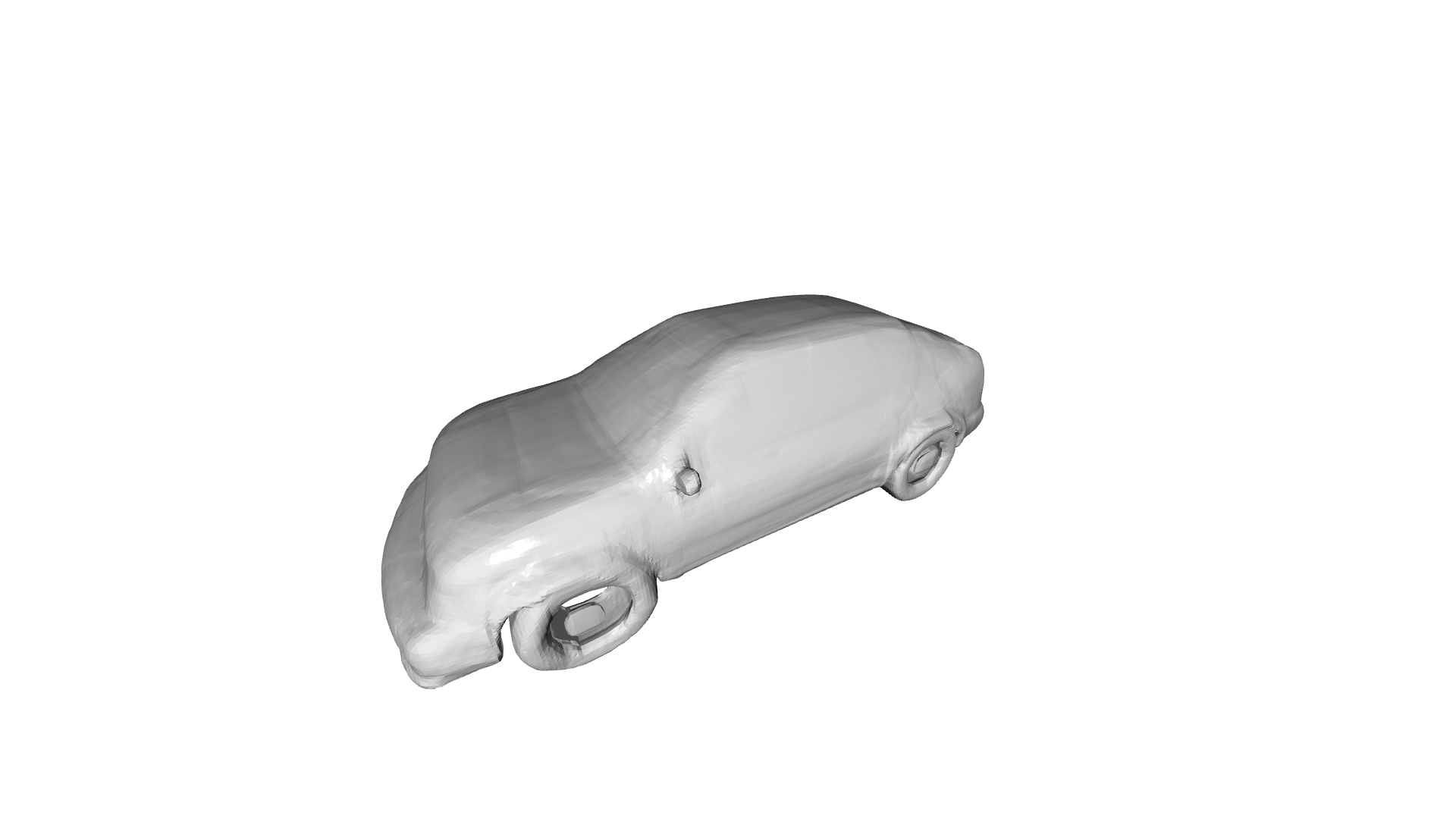}
    \end{subfigure}

\\
\begin{subfigure}[b]{0.17\linewidth}
      \centering
      \includegraphics[width=\linewidth,trim={15cm 5cm 20cm 12cm},clip]{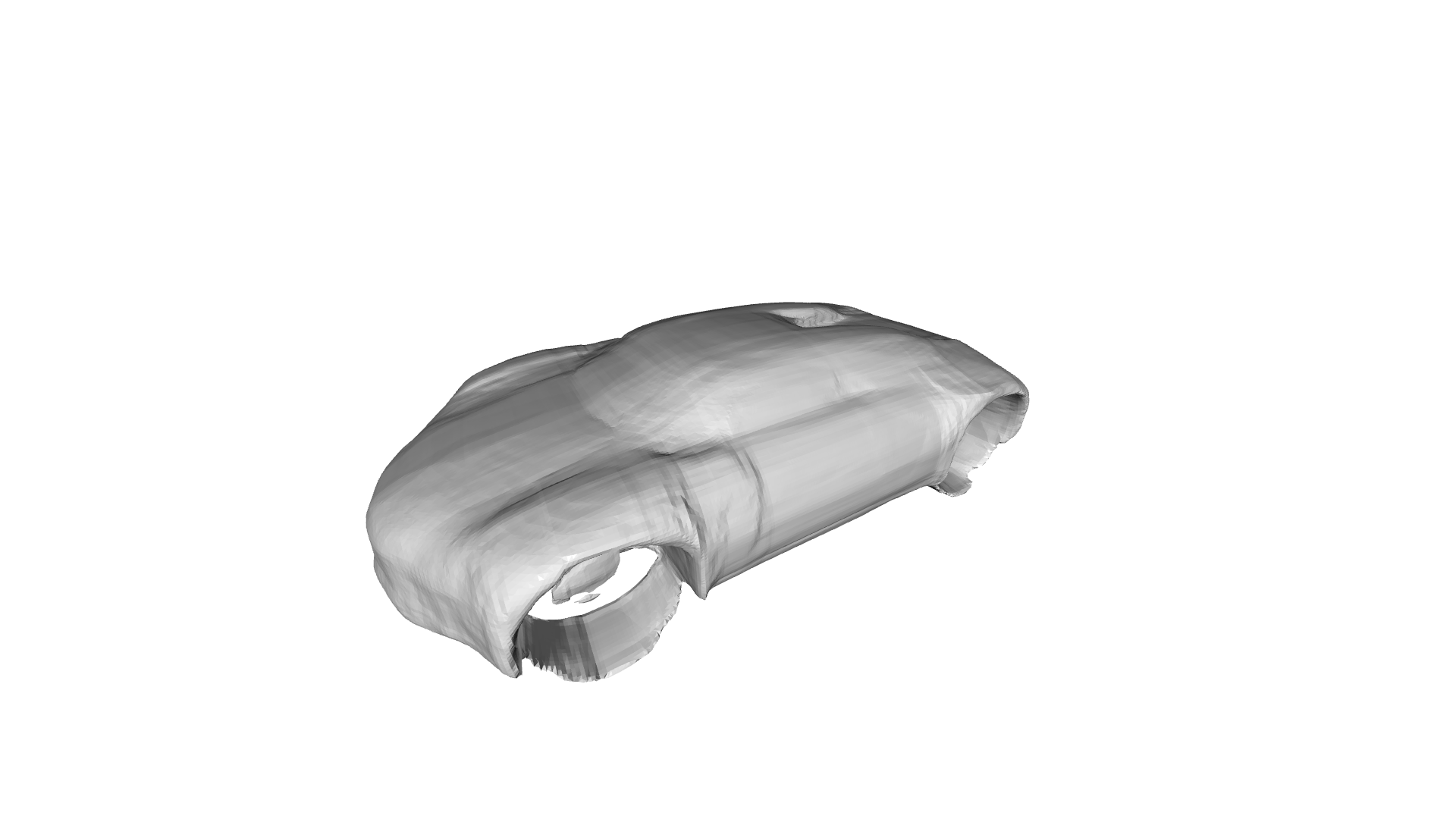}
    \end{subfigure}

&

\begin{subfigure}[b]{0.17\linewidth}
      \centering
      \includegraphics[width=\linewidth,trim={15cm 4cm 20cm 12cm},clip]{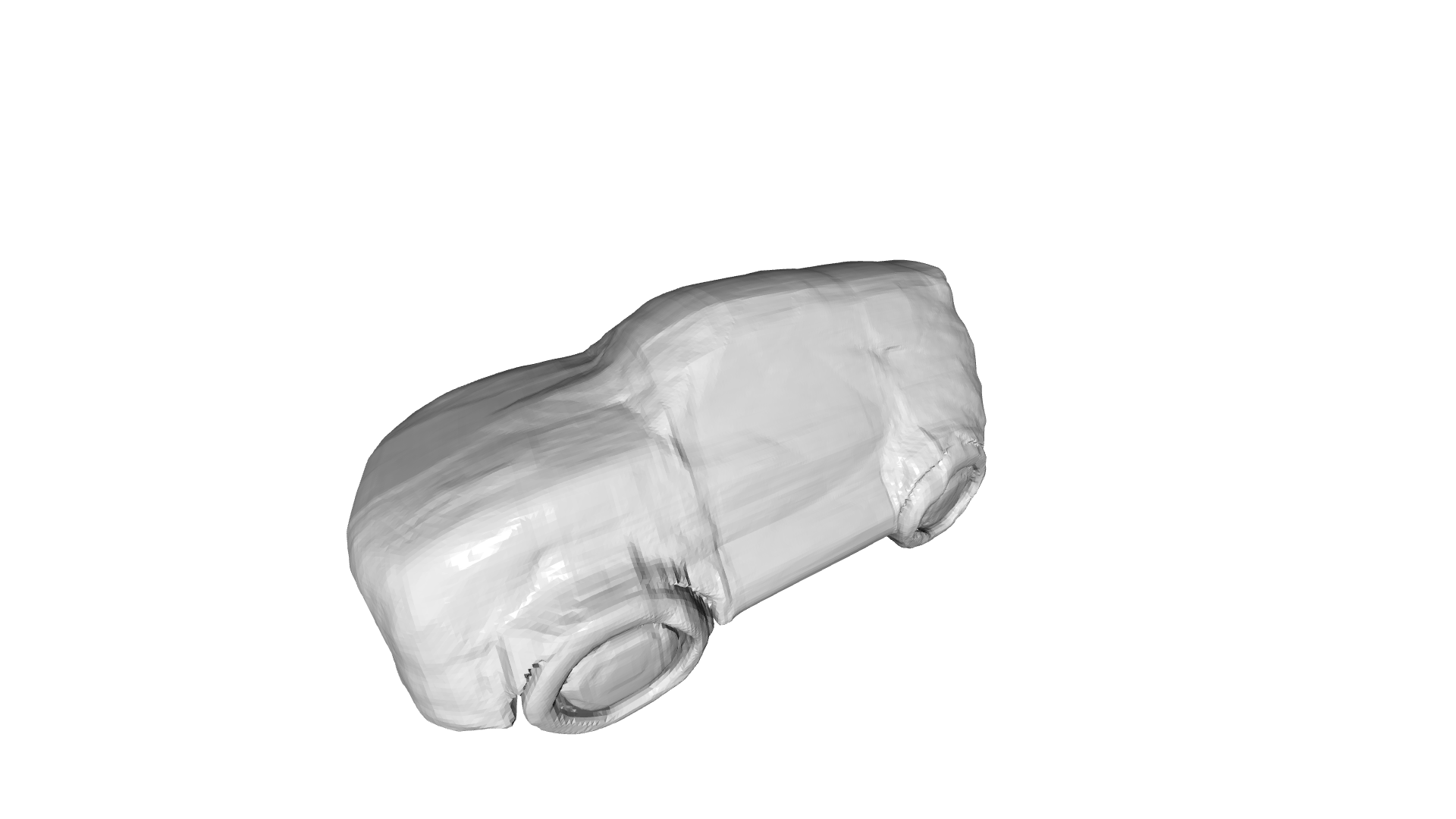}
    \end{subfigure}

&

\begin{subfigure}[b]{0.17\linewidth}
      \centering
      \includegraphics[width=\linewidth,trim={15cm 5cm 20cm 12cm},clip]{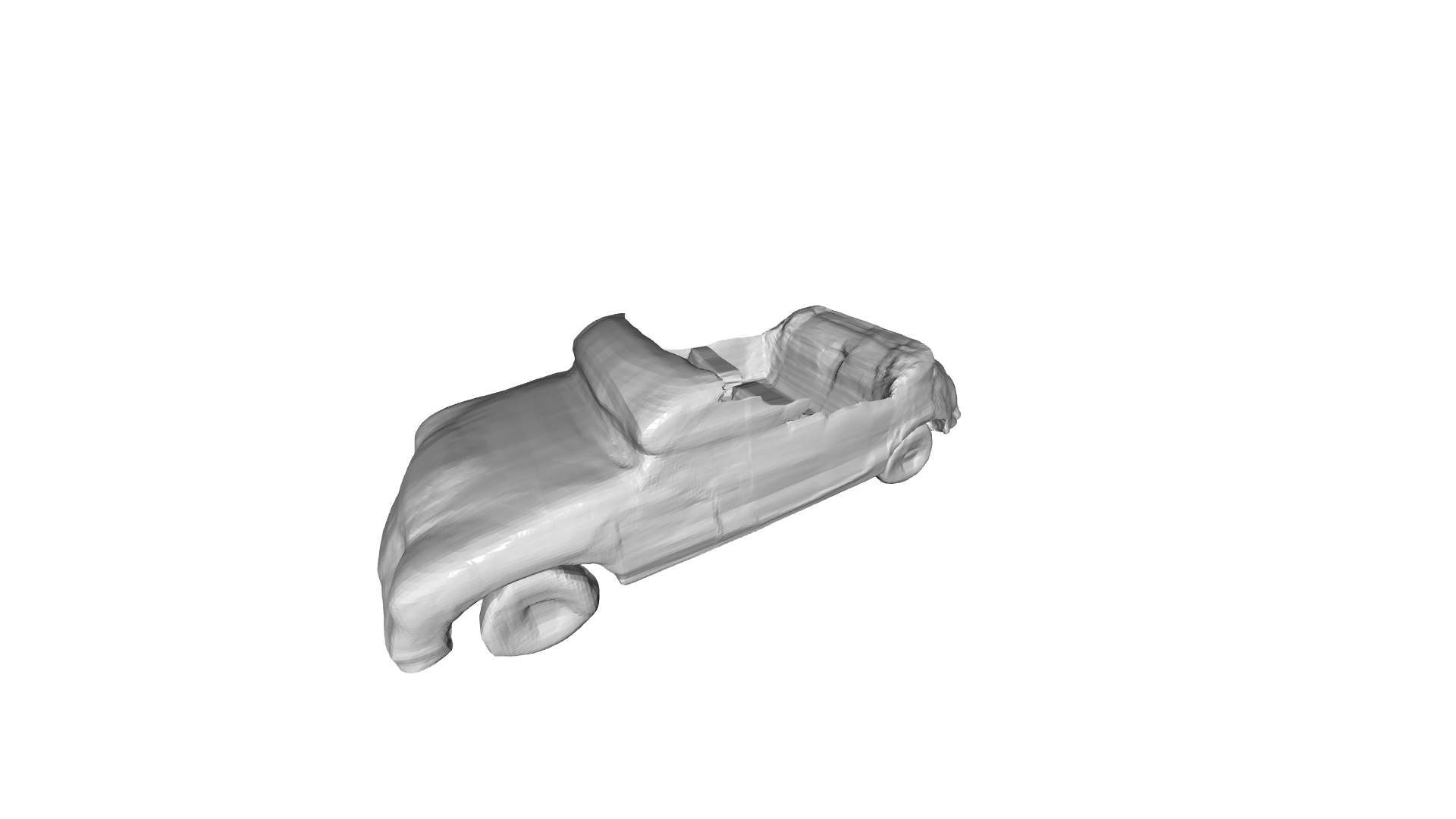}
    \end{subfigure}

&

\begin{subfigure}[b]{0.17\linewidth}
      \centering
      \includegraphics[width=\linewidth,trim={15cm 5cm 20cm 12cm},clip]{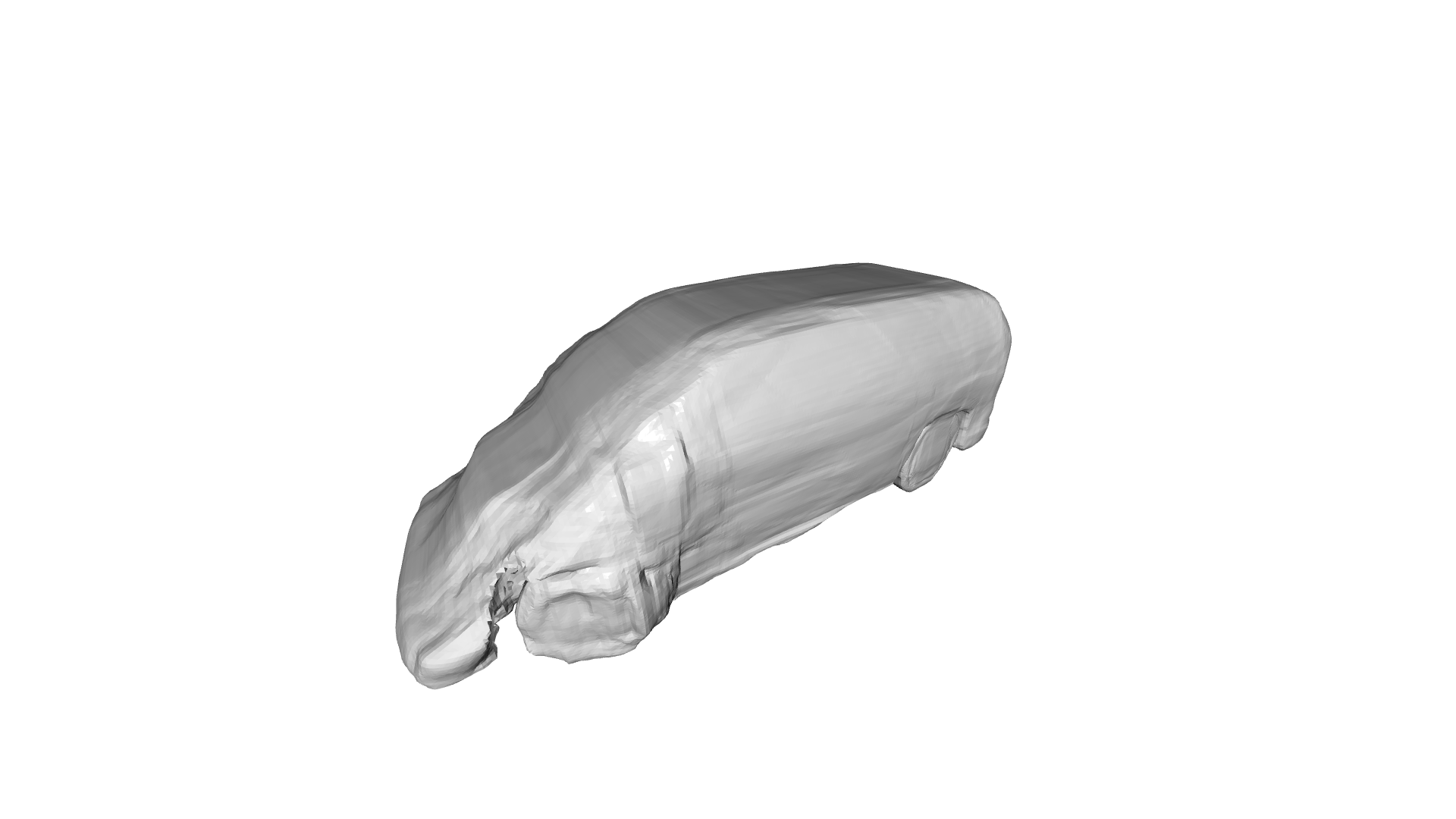}
    \end{subfigure}

&

\begin{subfigure}[b]{0.17\linewidth}
      \centering
      \includegraphics[width=\linewidth,trim={15cm 5cm 20cm 12cm},clip]{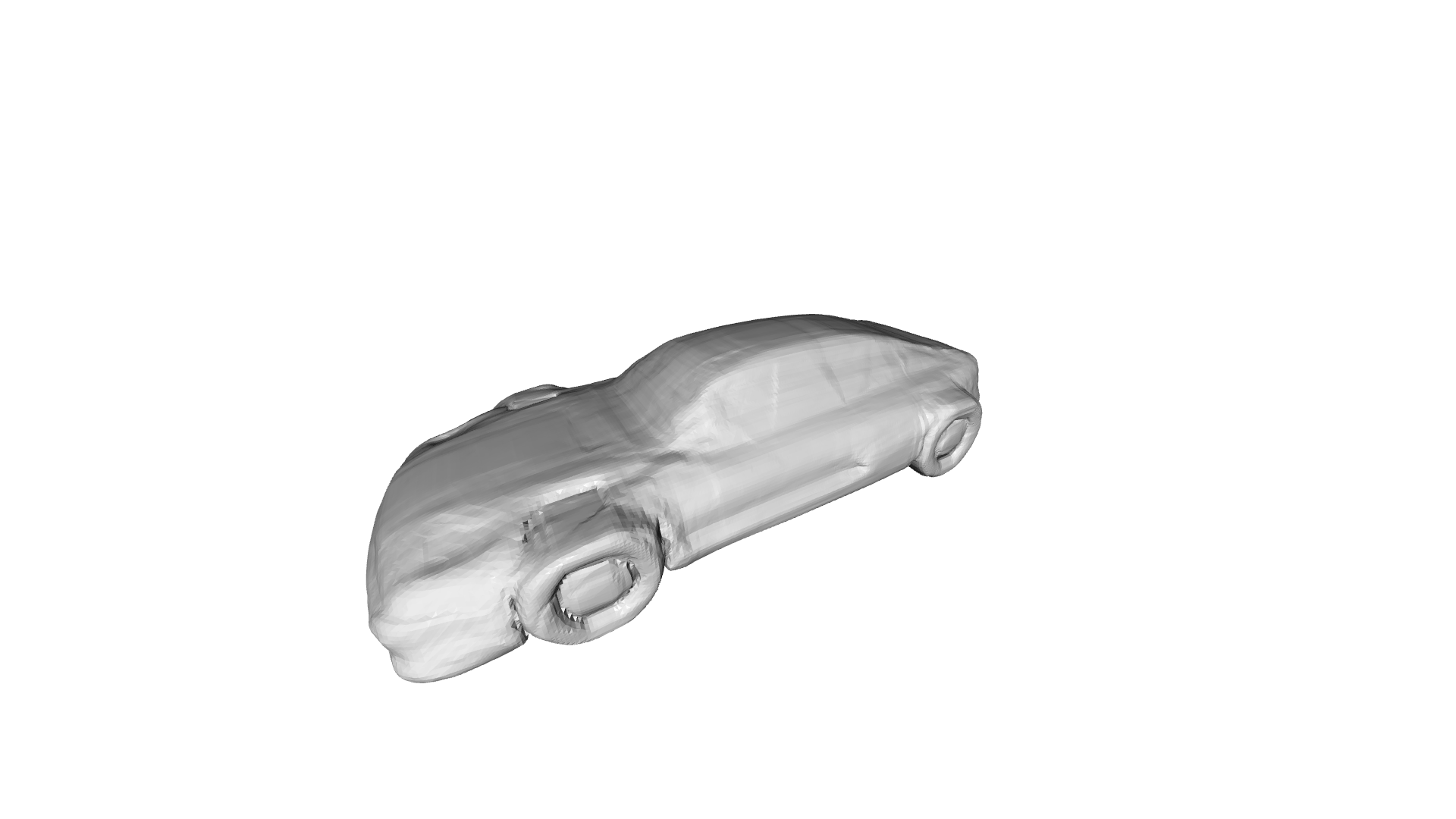}
    \end{subfigure}

\\
\begin{subfigure}[b]{0.17\linewidth}
      \centering
      \includegraphics[width=\linewidth,trim={15cm 5cm 20cm 12cm},clip]{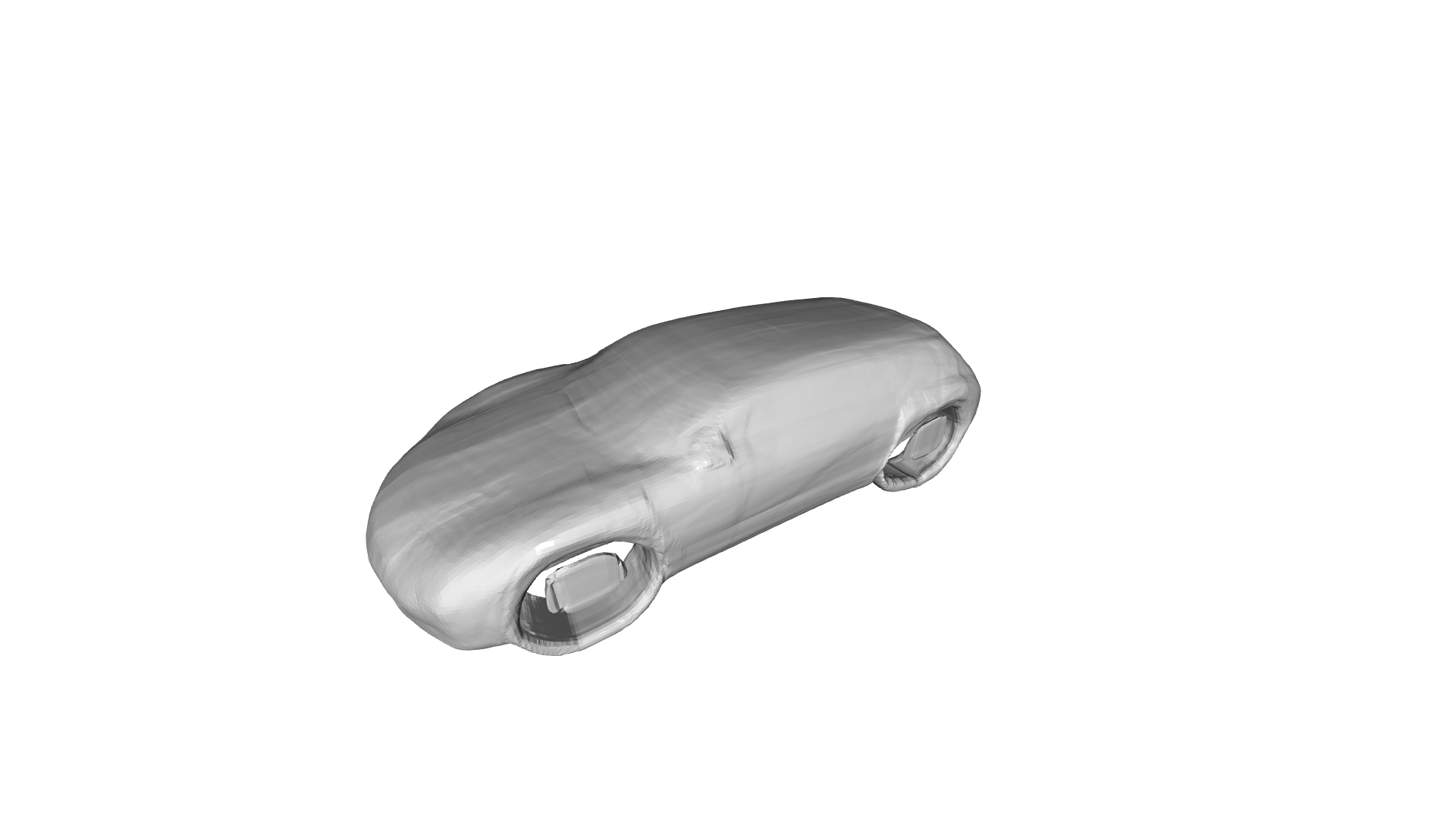}
    \end{subfigure}

&

\begin{subfigure}[b]{0.17\linewidth}
      \centering
      \includegraphics[width=\linewidth,trim={15cm 5cm 20cm 11cm},clip]{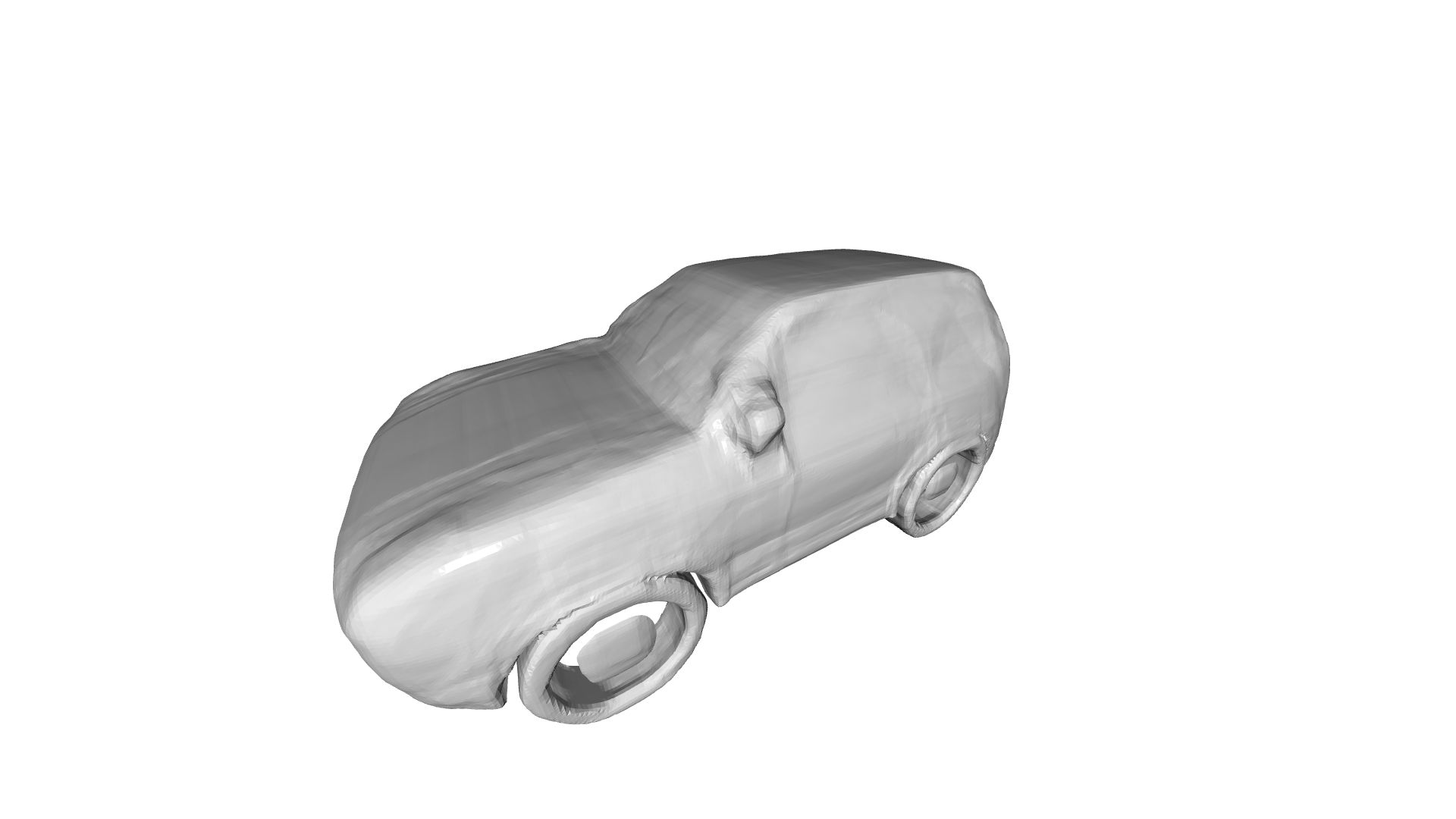}
    \end{subfigure}

&

\begin{subfigure}[b]{0.17\linewidth}
      \centering
      \includegraphics[width=\linewidth,trim={15cm 5cm 20cm 12cm},clip]{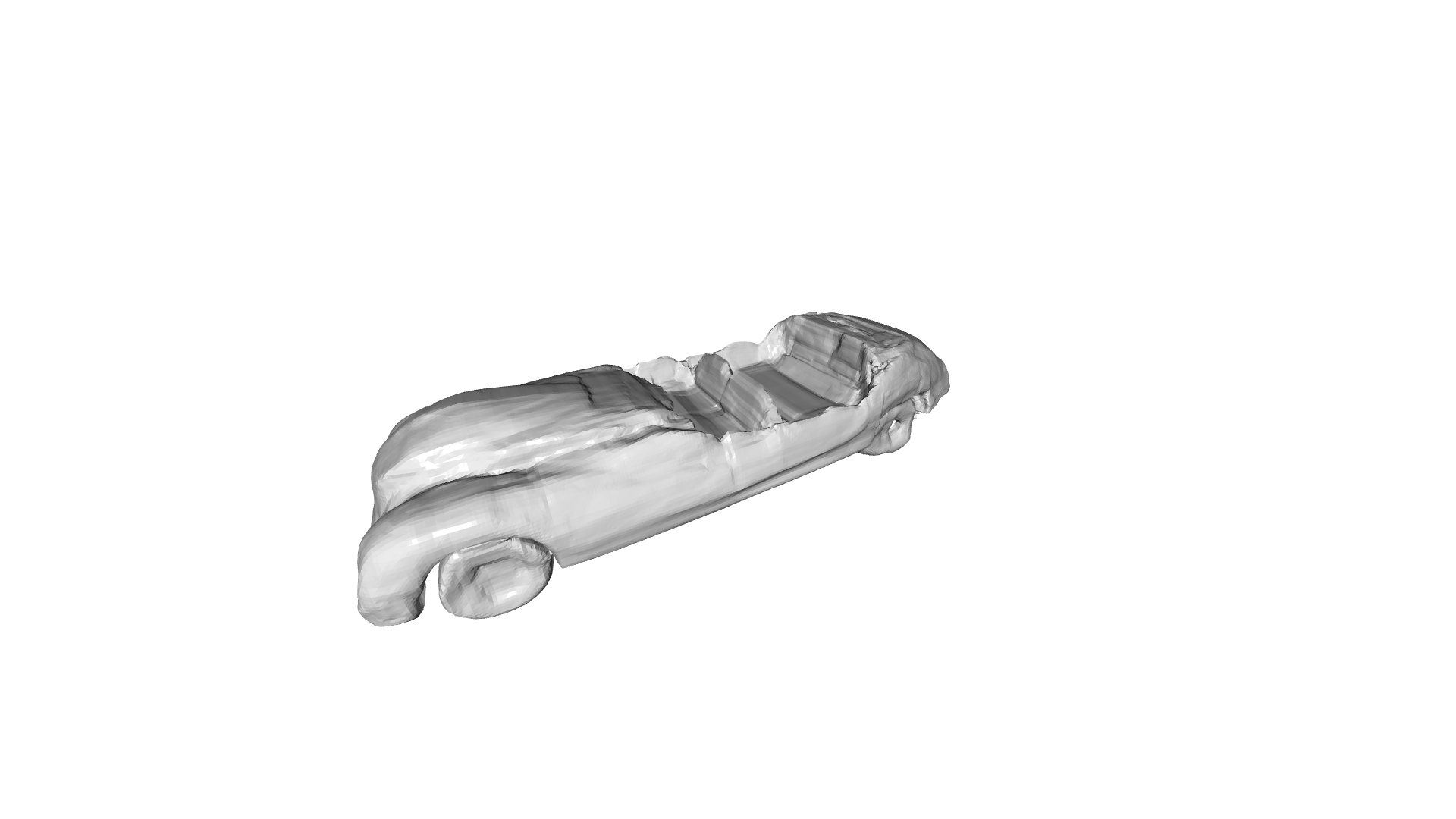}
    \end{subfigure}

&

\begin{subfigure}[b]{0.17\linewidth}
      \centering
      \includegraphics[width=\linewidth,trim={15cm 5cm 20cm 12cm},clip]{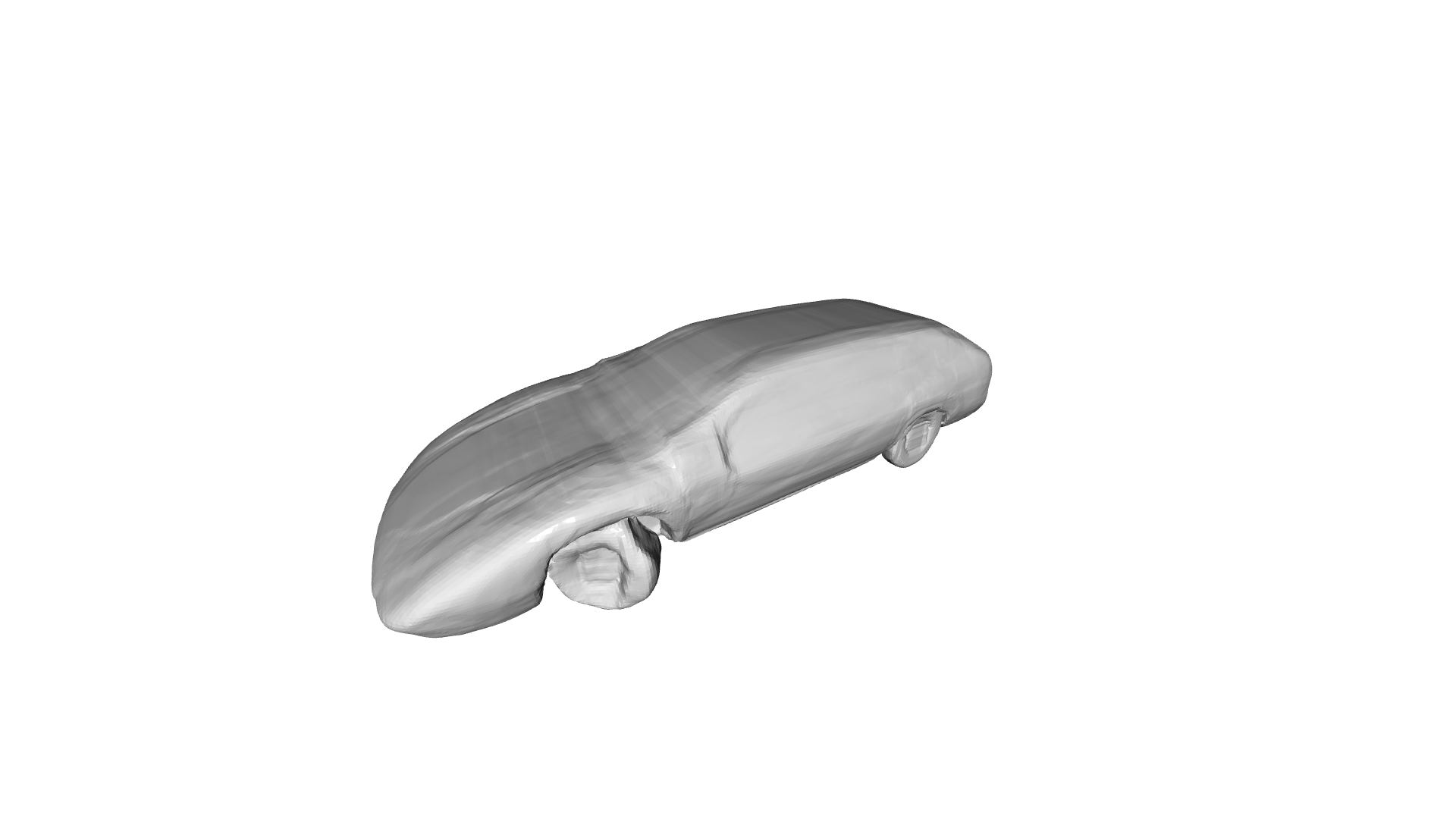}
    \end{subfigure}

&

\begin{subfigure}[b]{0.17\linewidth}
      \centering
      \includegraphics[width=\linewidth,trim={15cm 5cm 20cm 12cm},clip]{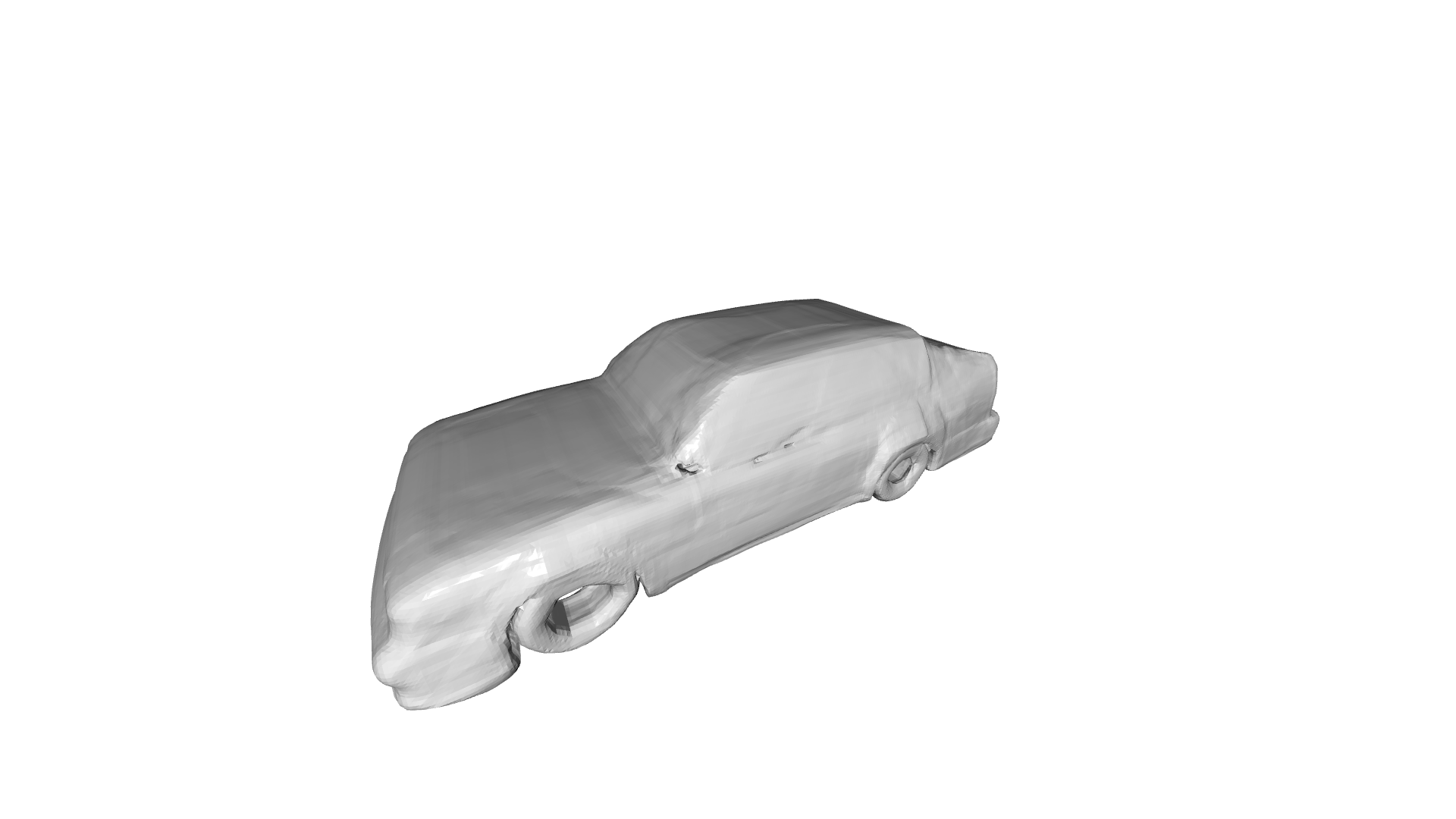}
    \end{subfigure}

\\
\begin{subfigure}[b]{0.17\linewidth}
      \centering
      \includegraphics[width=\linewidth,trim={15cm 5cm 20cm 12cm},clip]{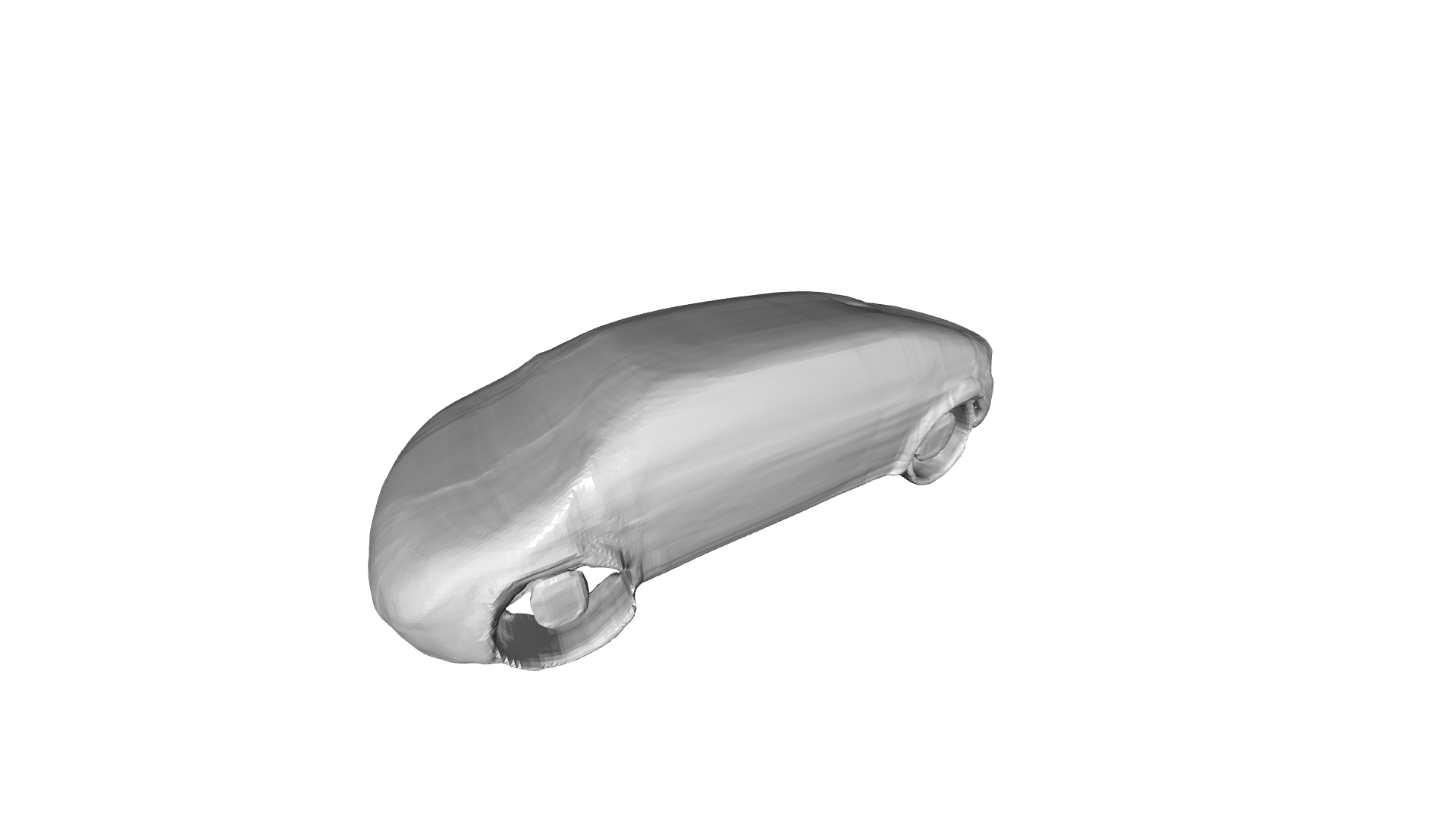}
    \end{subfigure}

&

\begin{subfigure}[b]{0.17\linewidth}
      \centering
      \includegraphics[width=\linewidth,trim={15cm 5cm 20cm 10cm},clip]{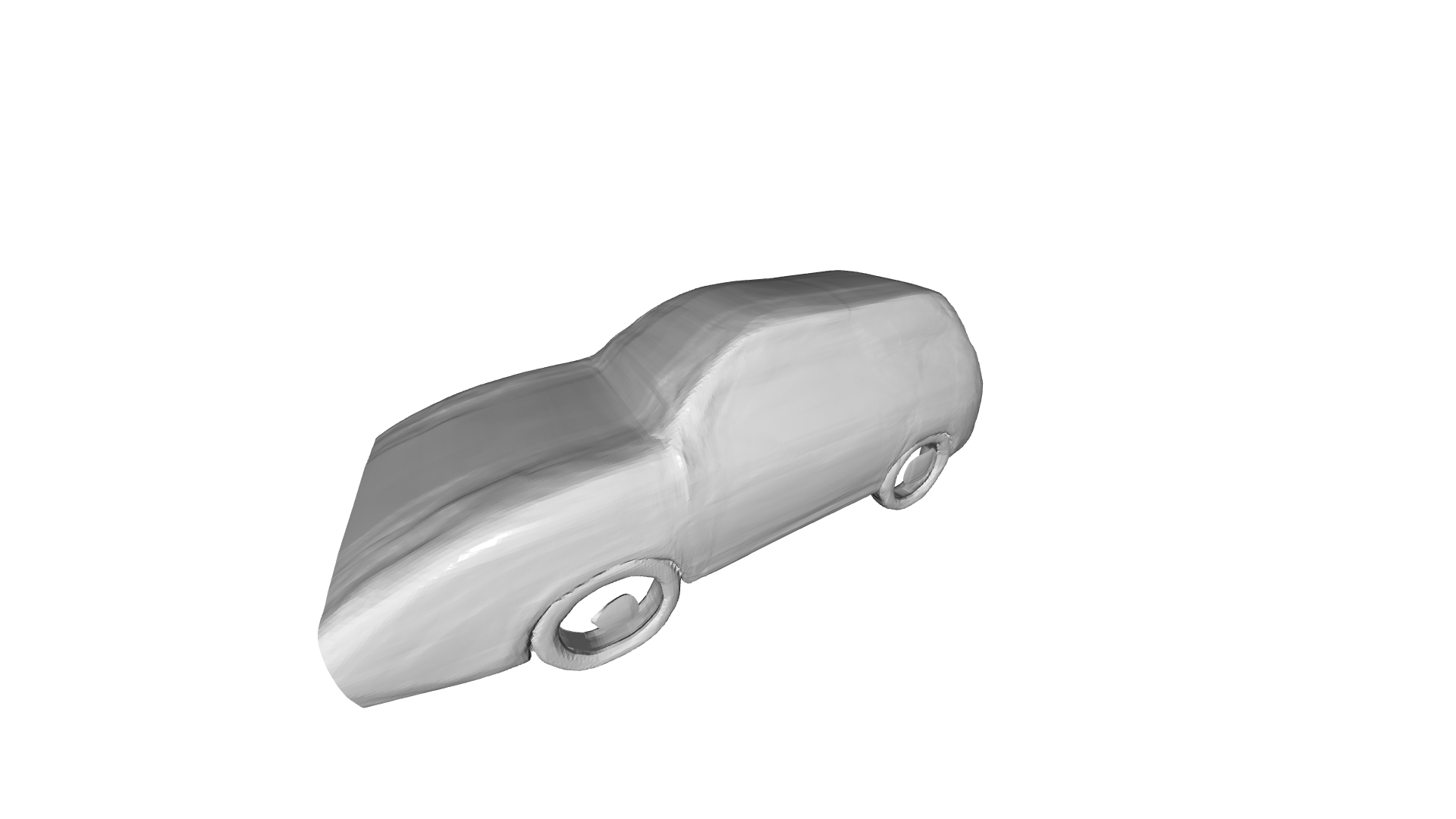}
    \end{subfigure}

&

\begin{subfigure}[b]{0.17\linewidth}
      \centering
      \includegraphics[width=\linewidth,trim={15cm 5cm 20cm 12cm},clip]{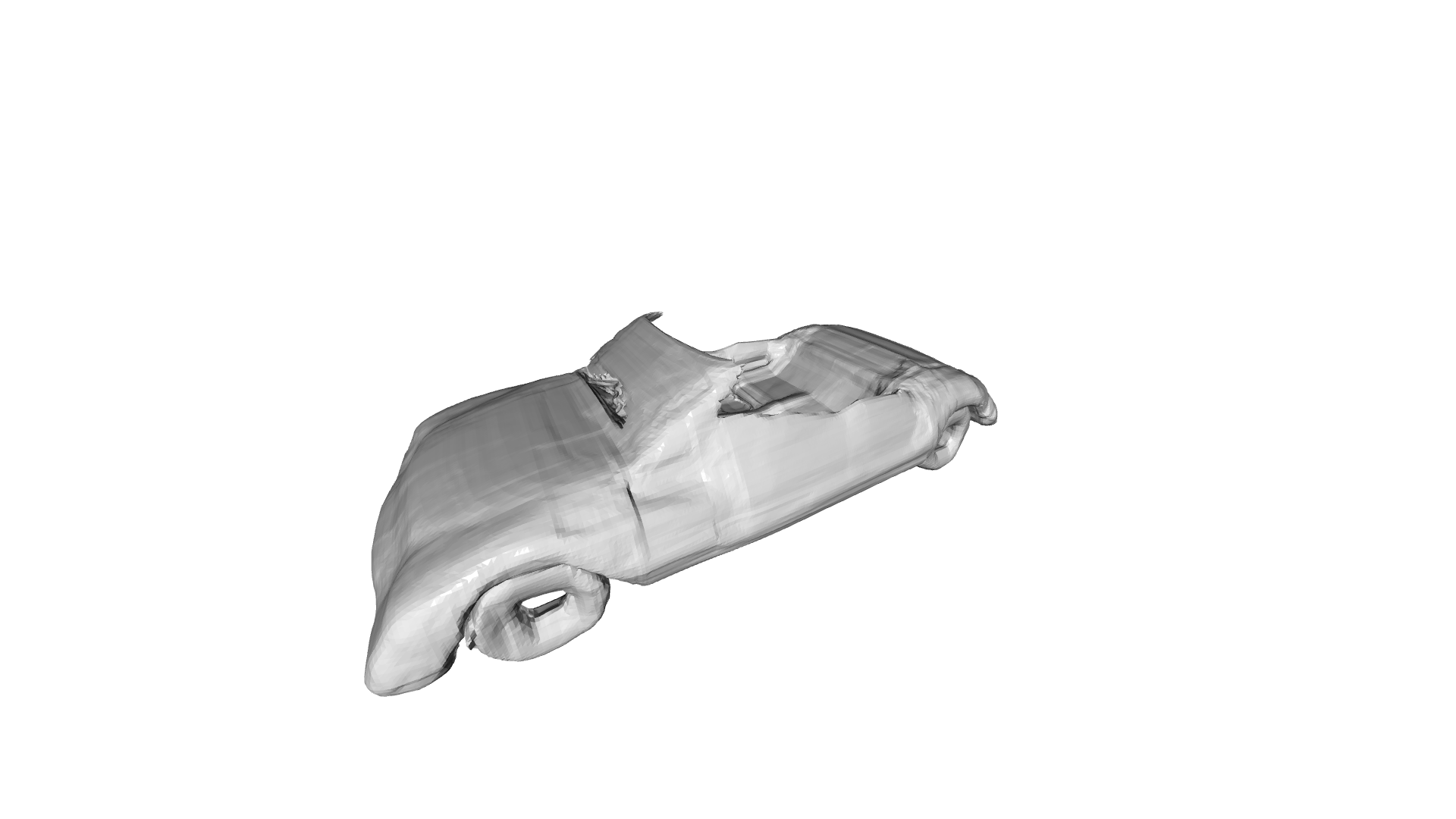}
    \end{subfigure}

&

\begin{subfigure}[b]{0.17\linewidth}
      \centering
      \includegraphics[width=\linewidth,trim={15cm 5cm 20cm 12cm},clip]{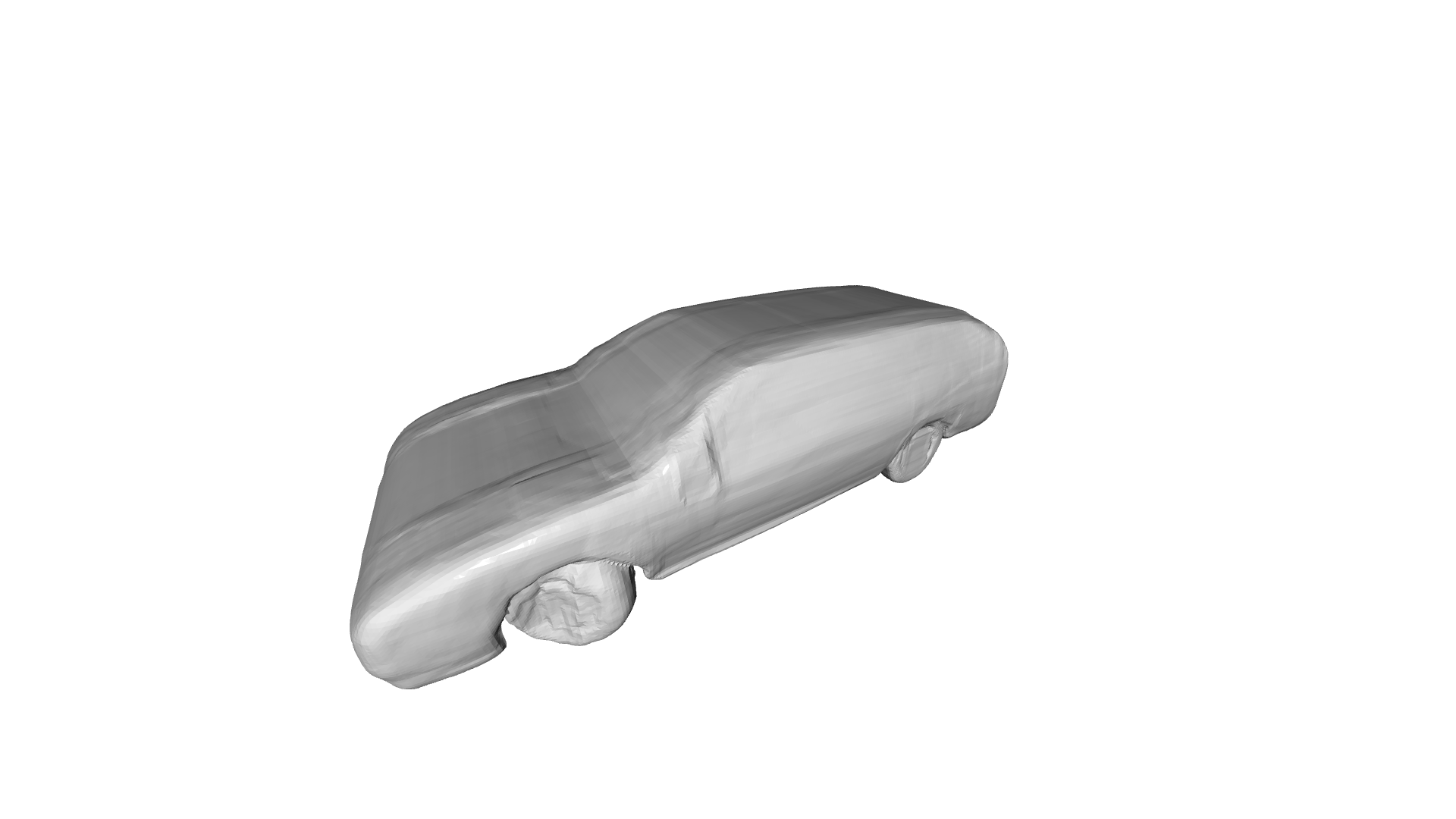}
    \end{subfigure}

&

\begin{subfigure}[b]{0.17\linewidth}
      \centering
      \includegraphics[width=\linewidth,trim={15cm 5cm 20cm 12cm},clip]{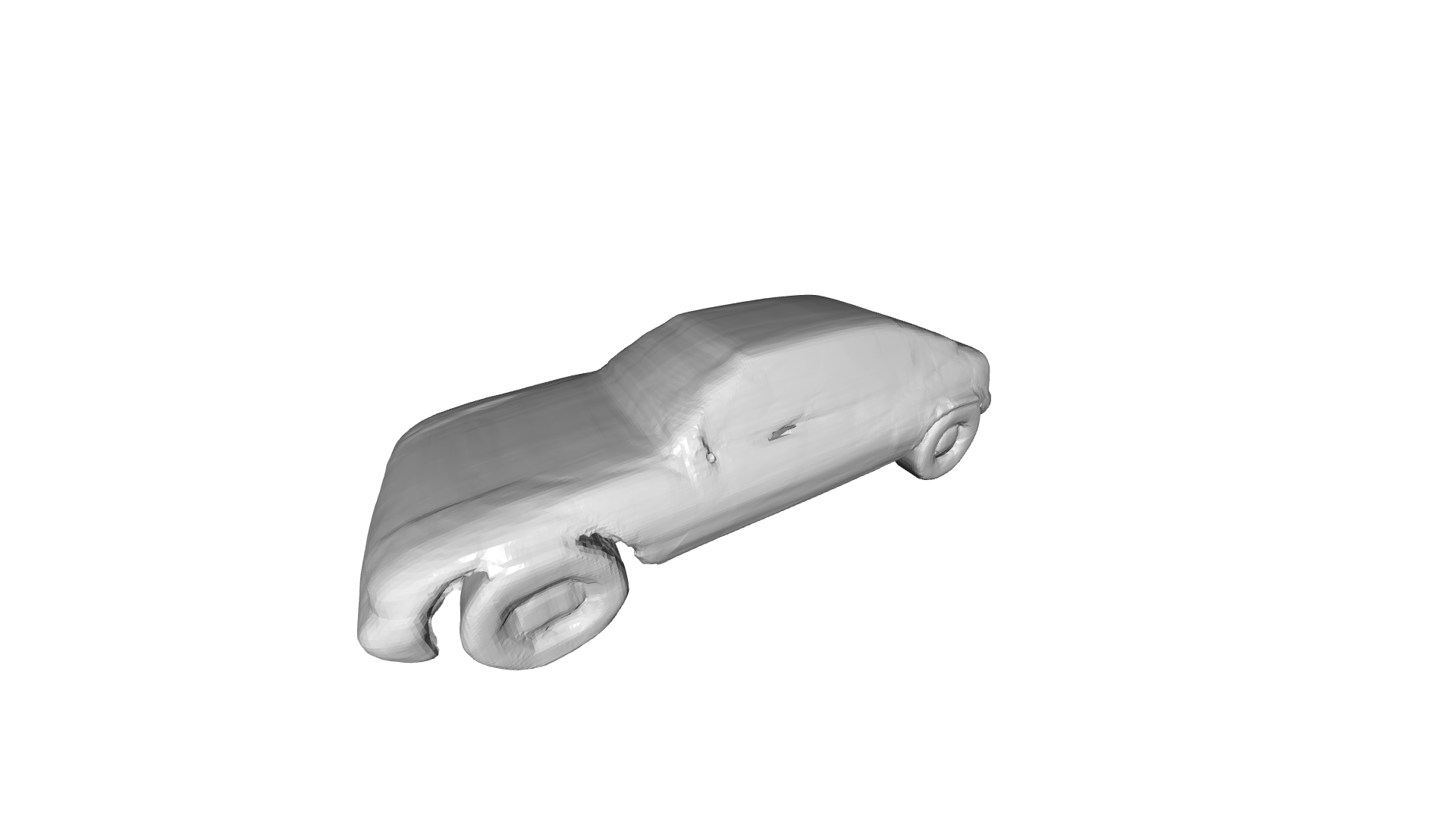}
    \end{subfigure}

\\
\begin{subfigure}[b]{0.17\linewidth}
      \centering
      \includegraphics[width=\linewidth,trim={15cm 5cm 20cm 12cm},clip]{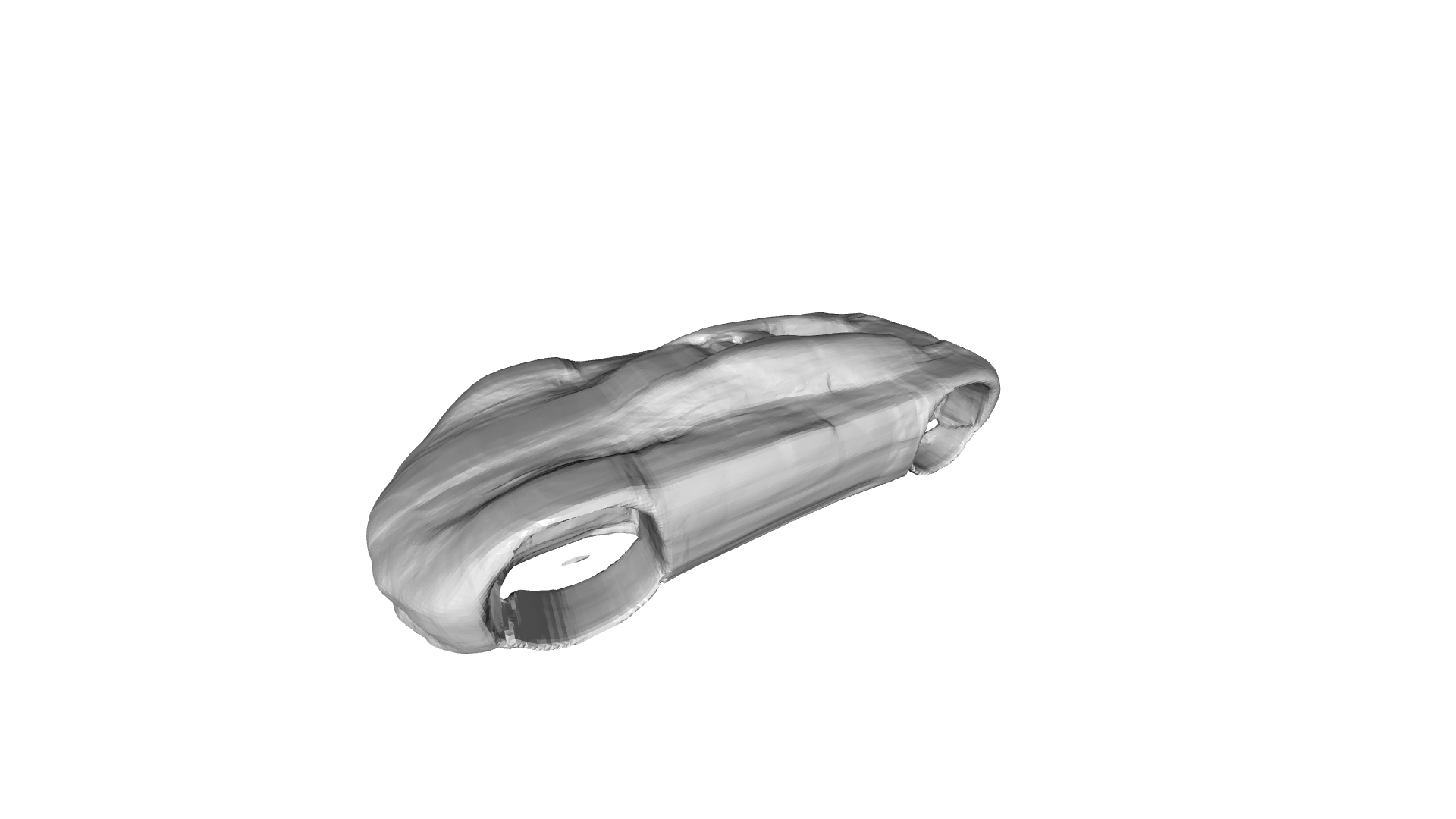}
    \end{subfigure}

&

\begin{subfigure}[b]{0.17\linewidth}
      \centering
      \includegraphics[width=\linewidth,trim={15cm 5cm 20cm 12cm},clip]{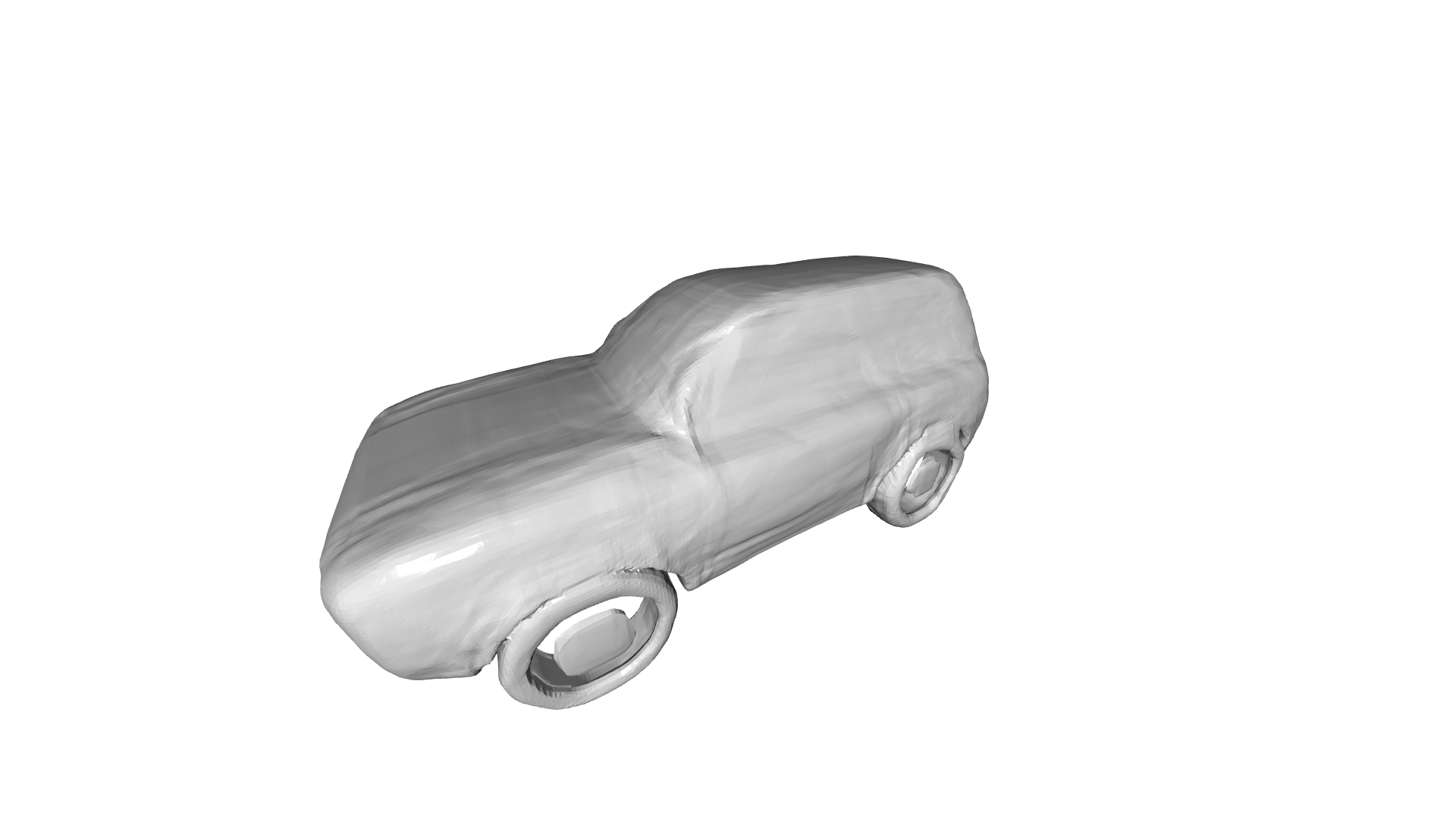}
    \end{subfigure}

&

\begin{subfigure}[b]{0.17\linewidth}
      \centering
      \includegraphics[width=\linewidth,trim={15cm 5cm 20cm 12cm},clip]{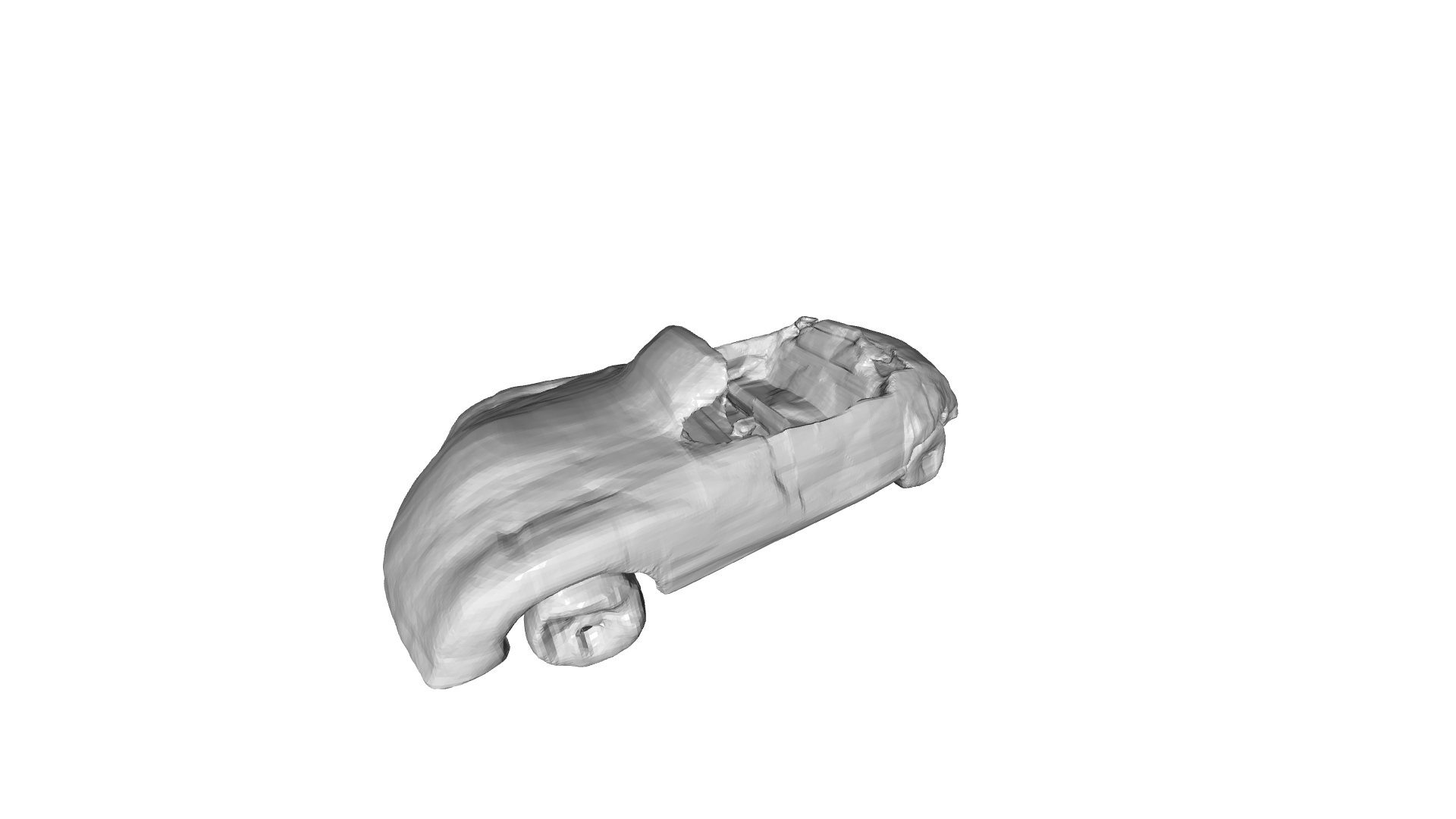}
    \end{subfigure}

&

\begin{subfigure}[b]{0.17\linewidth}
      \centering
      \includegraphics[width=\linewidth,trim={15cm 5cm 20cm 12cm},clip]{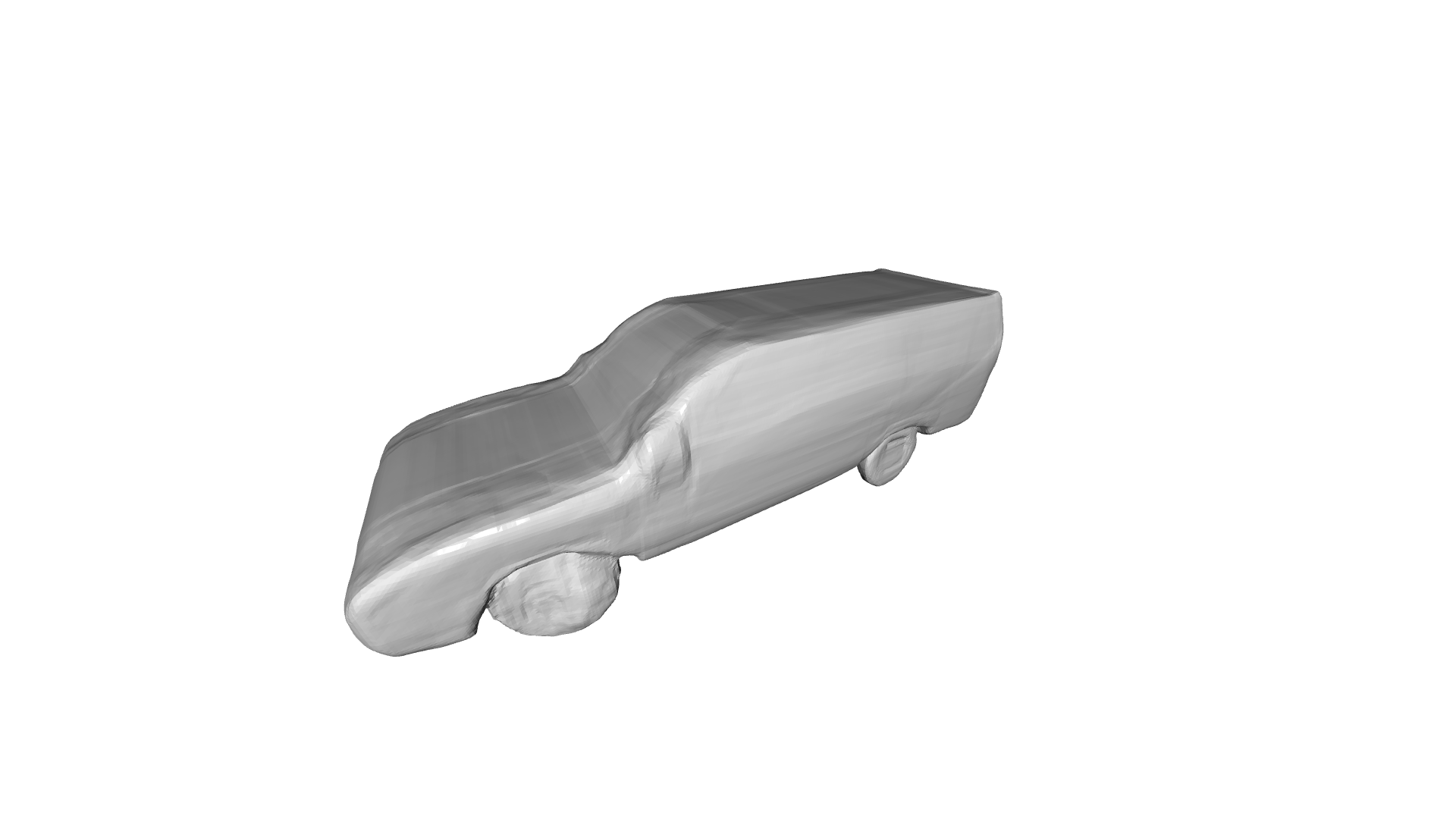}
    \end{subfigure}

&

\begin{subfigure}[b]{0.17\linewidth}
      \centering
      \includegraphics[width=\linewidth,trim={15cm 5cm 20cm 12cm},clip]{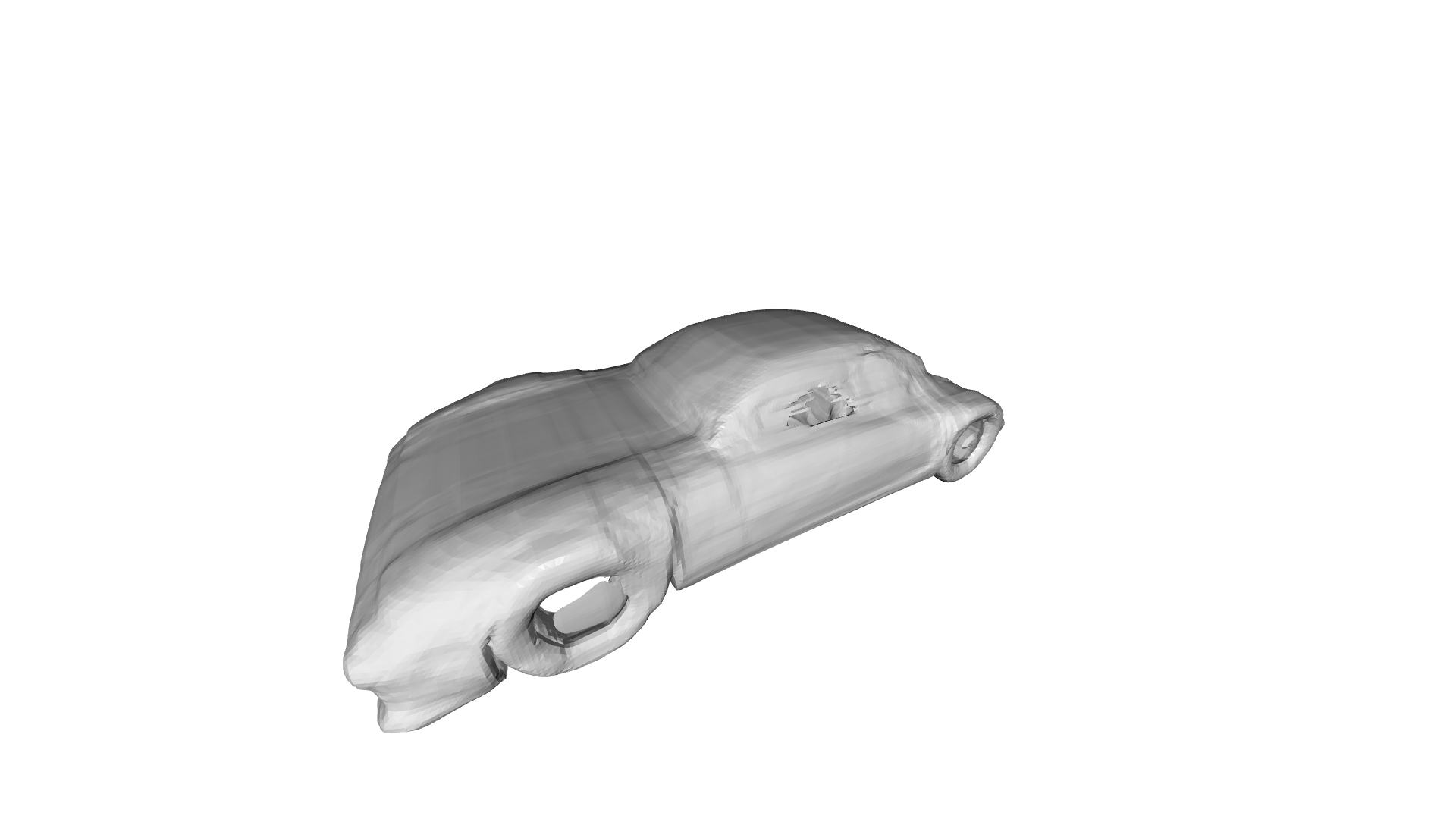}
    \end{subfigure}

\\
\begin{subfigure}[b]{0.17\linewidth}
      \centering
      \includegraphics[width=\linewidth,trim={15cm 5cm 20cm 12cm},clip]{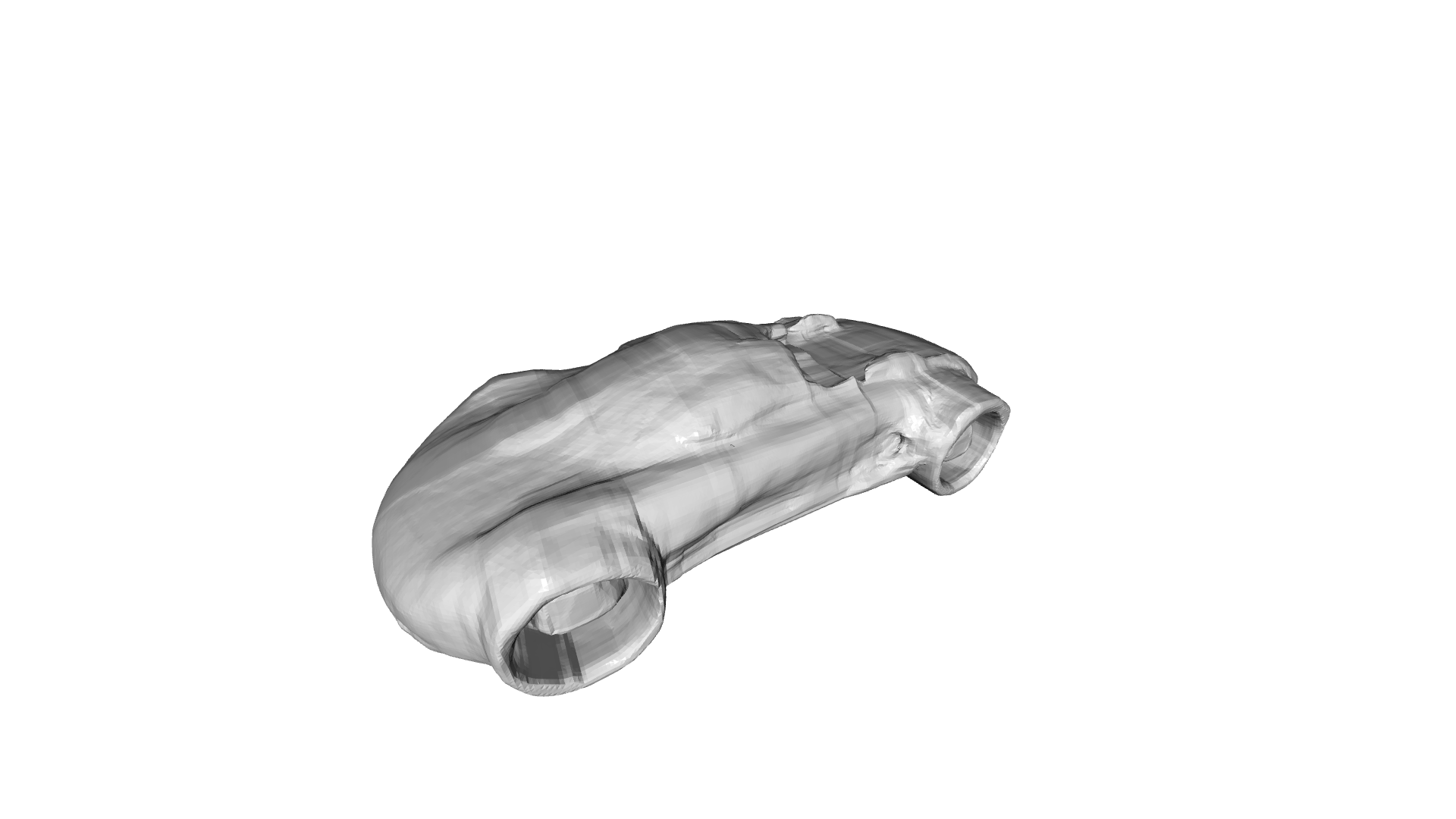}
    \end{subfigure}

&

\begin{subfigure}[b]{0.17\linewidth}
      \centering
      \includegraphics[width=\linewidth,trim={15cm 5cm 20cm 12cm},clip]{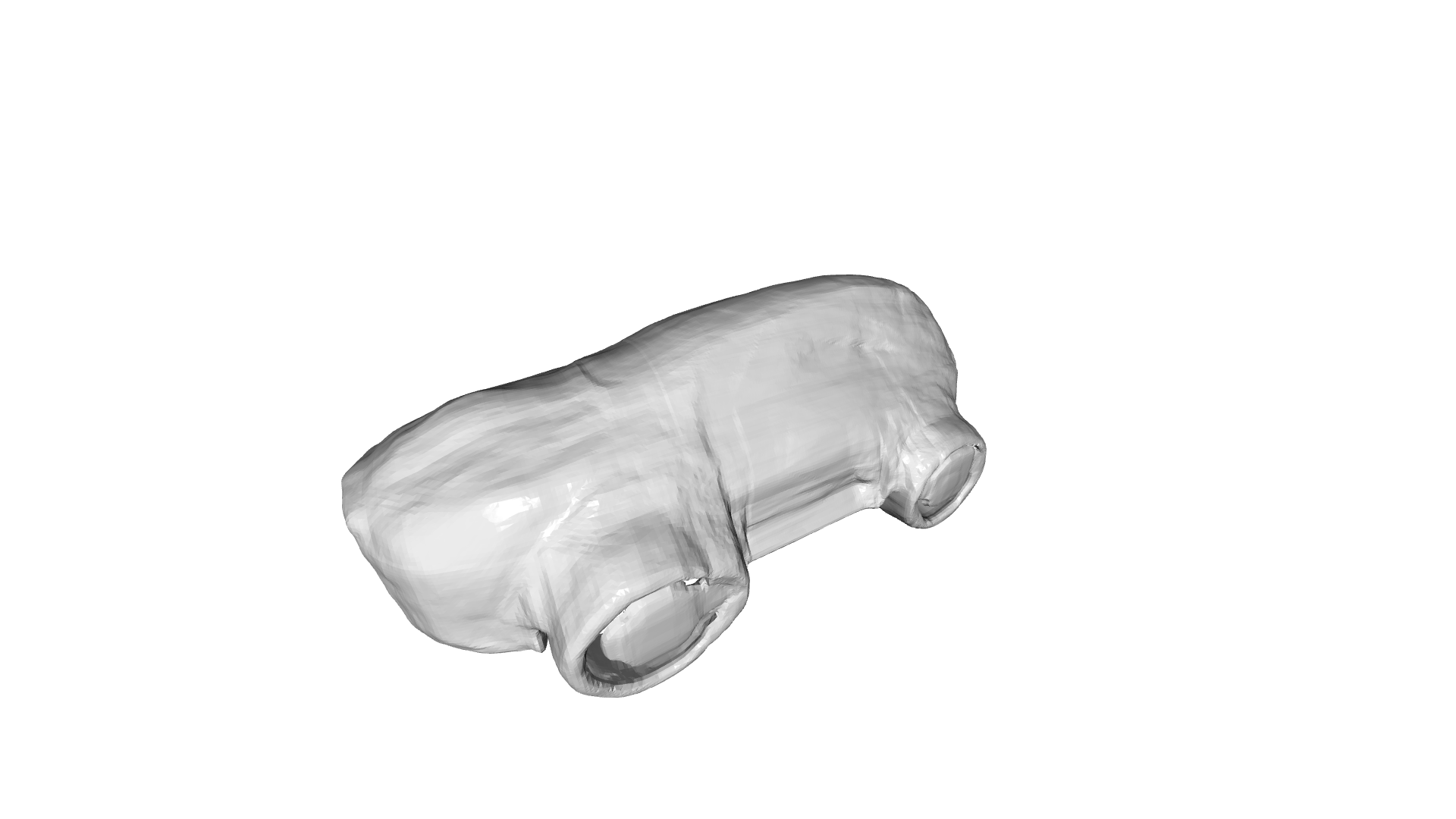}
    \end{subfigure}

&

\begin{subfigure}[b]{0.17\linewidth}
      \centering
      \includegraphics[width=\linewidth,trim={15cm 5cm 20cm 12cm},clip]{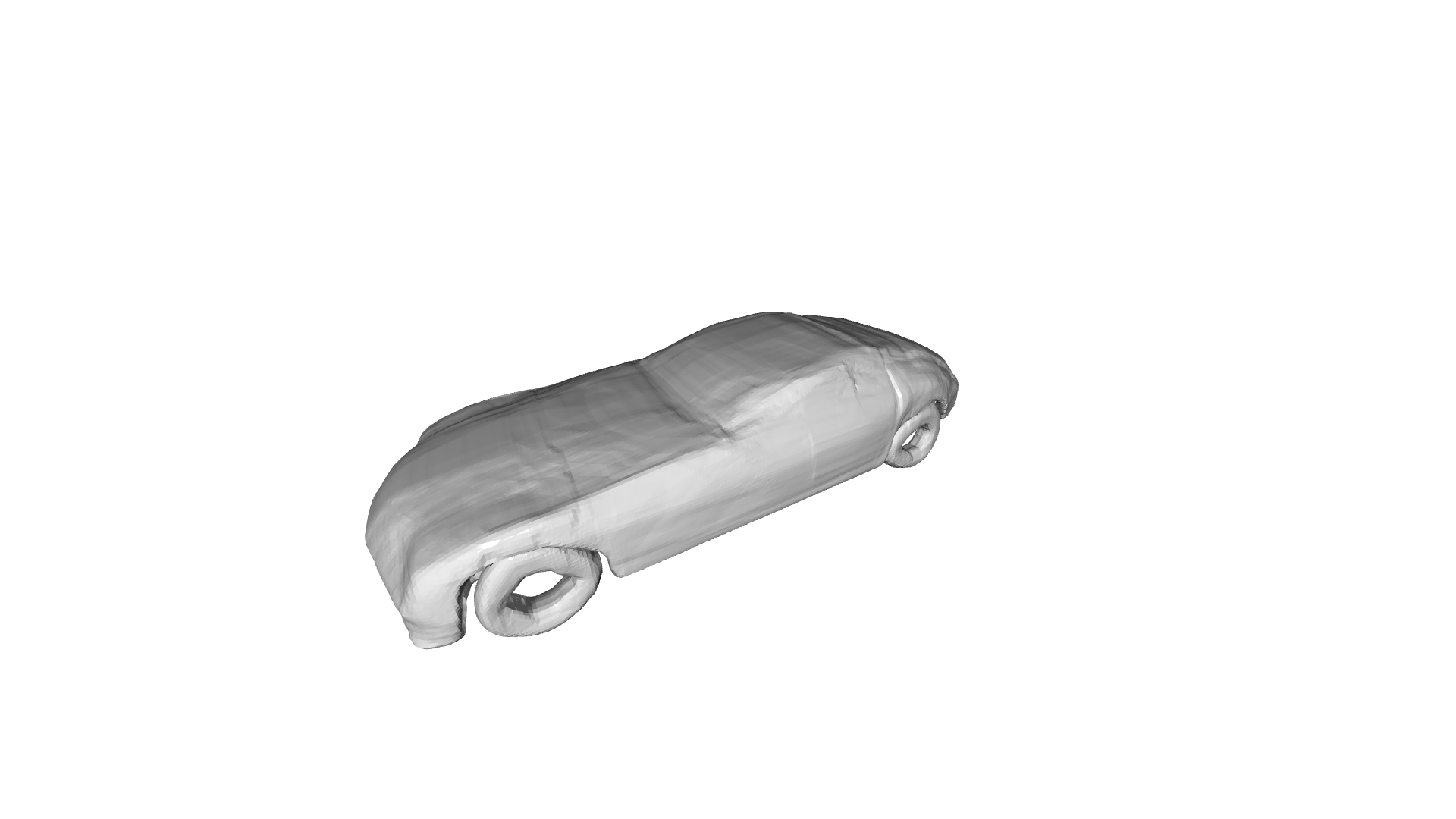}
    \end{subfigure}

&

\begin{subfigure}[b]{0.17\linewidth}
      \centering
      \includegraphics[width=\linewidth,trim={15cm 5cm 20cm 12cm},clip]{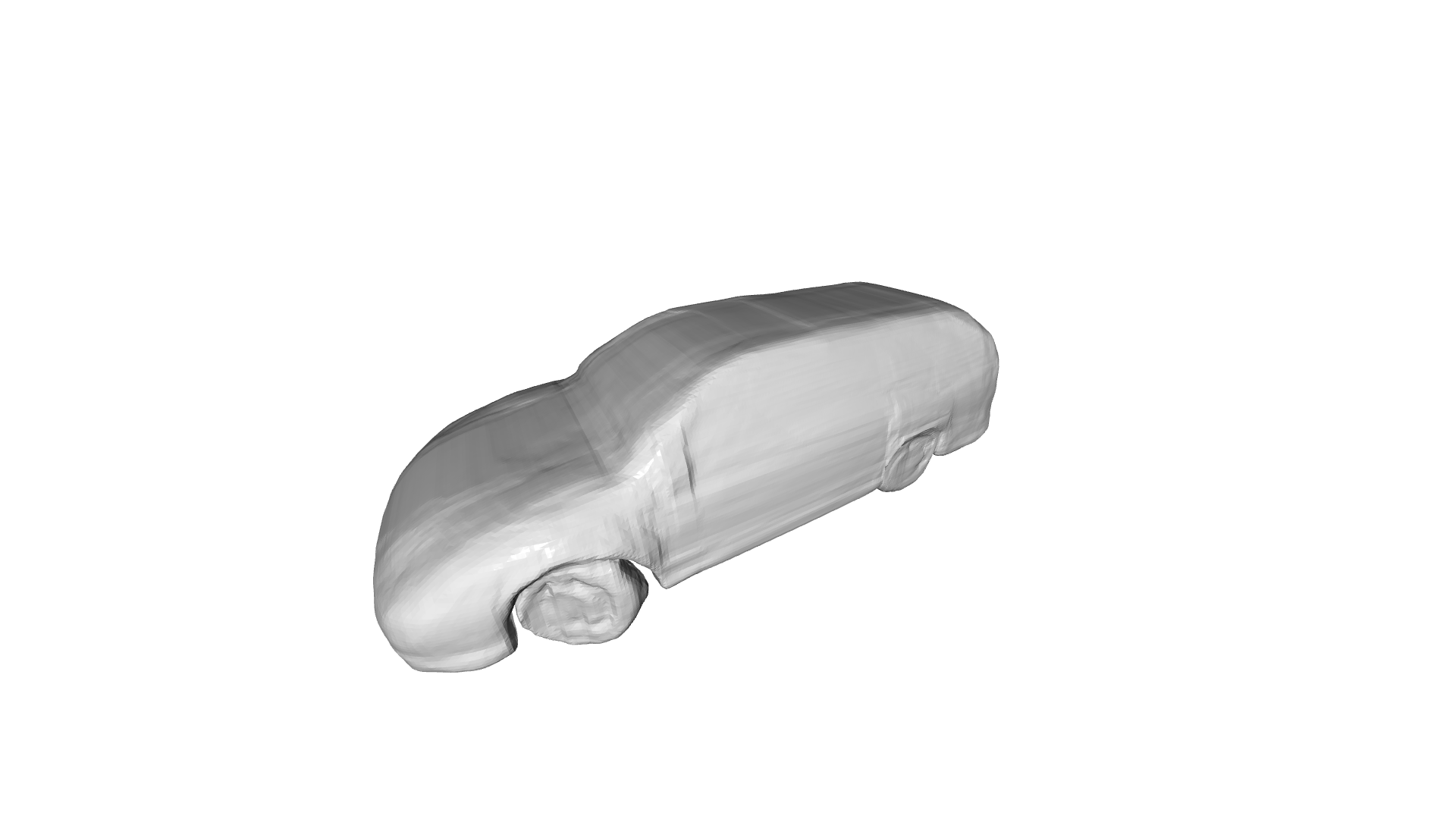}
    \end{subfigure}

&

\begin{subfigure}[b]{0.17\linewidth}
      \centering
      \includegraphics[width=\linewidth,trim={15cm 5cm 20cm 12cm},clip]{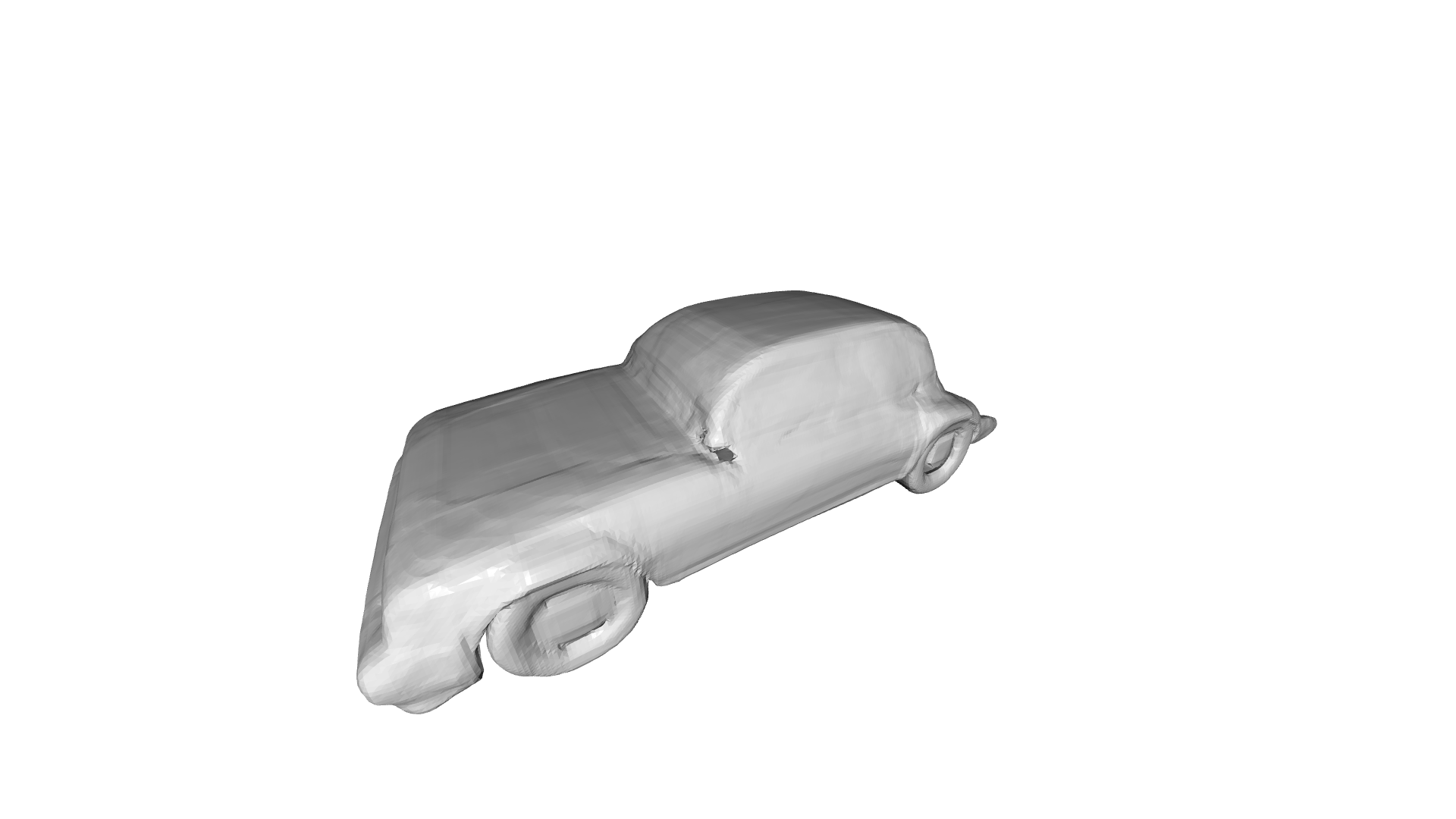}
    \end{subfigure}

\\
\bottomrule
  \end{tabular}
\end{figure*}

We randomly select 50 adversarial shapes generated by \ourmethod{} and present them in Figure~\ref{fig:shape_vis_big_table}. We group them by the baseline objects used for generation. We observe that most of the shapes generated are semantically meaningful. 

\begin{figure*}
\centering
\begin{subfigure}[b]{0.19\textwidth}
\includegraphics[width=\textwidth]{figures/threshold_recall/df7SUV.png}
\caption*{(1.a) Score = 0.82}
\end{subfigure}
\begin{subfigure}[b]{0.19\textwidth}
\includegraphics[width=\textwidth]{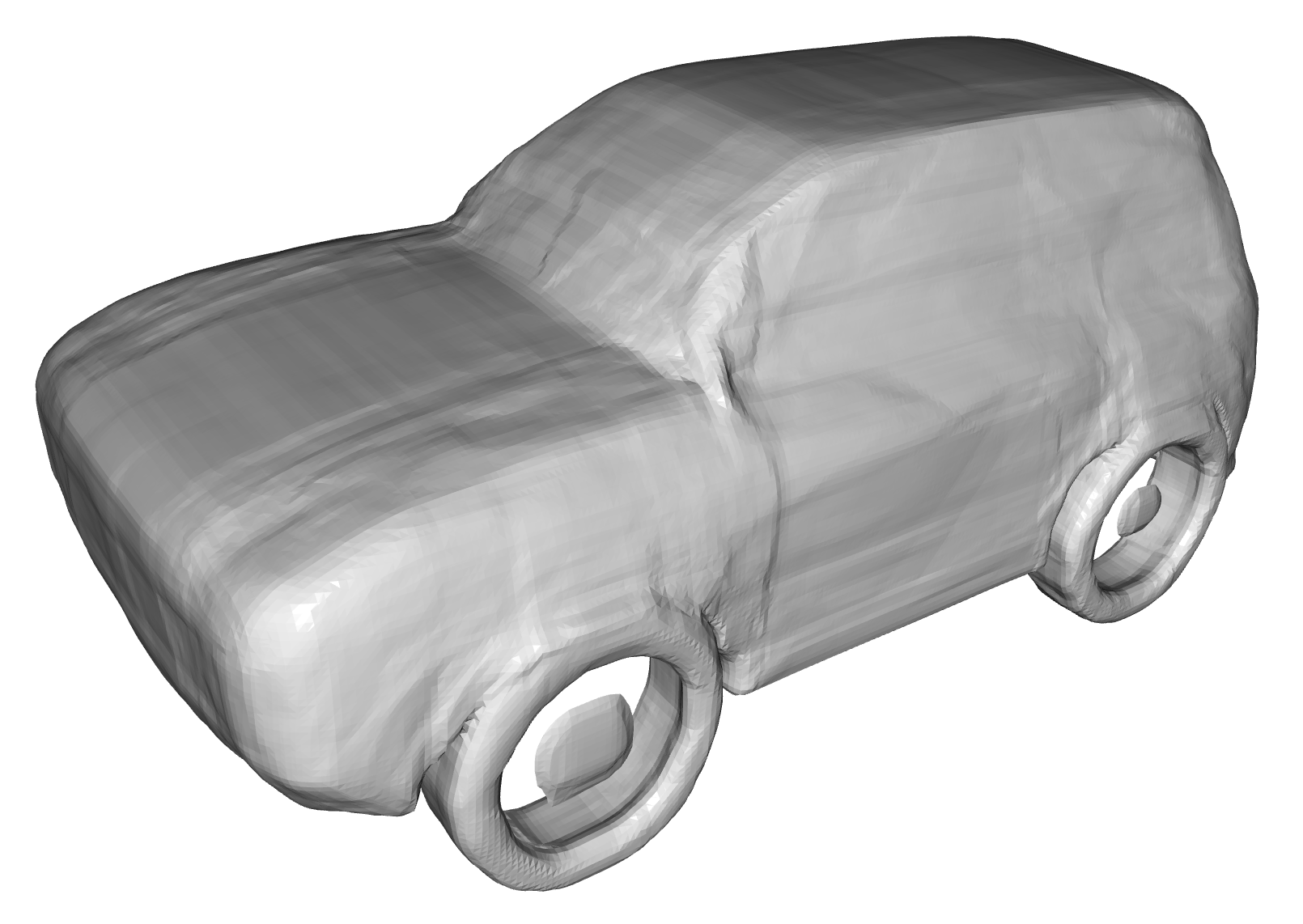}
\caption*{(1.b) Score = 0.60}
\end{subfigure}
\begin{subfigure}[b]{0.19\textwidth}
\includegraphics[width=\textwidth]{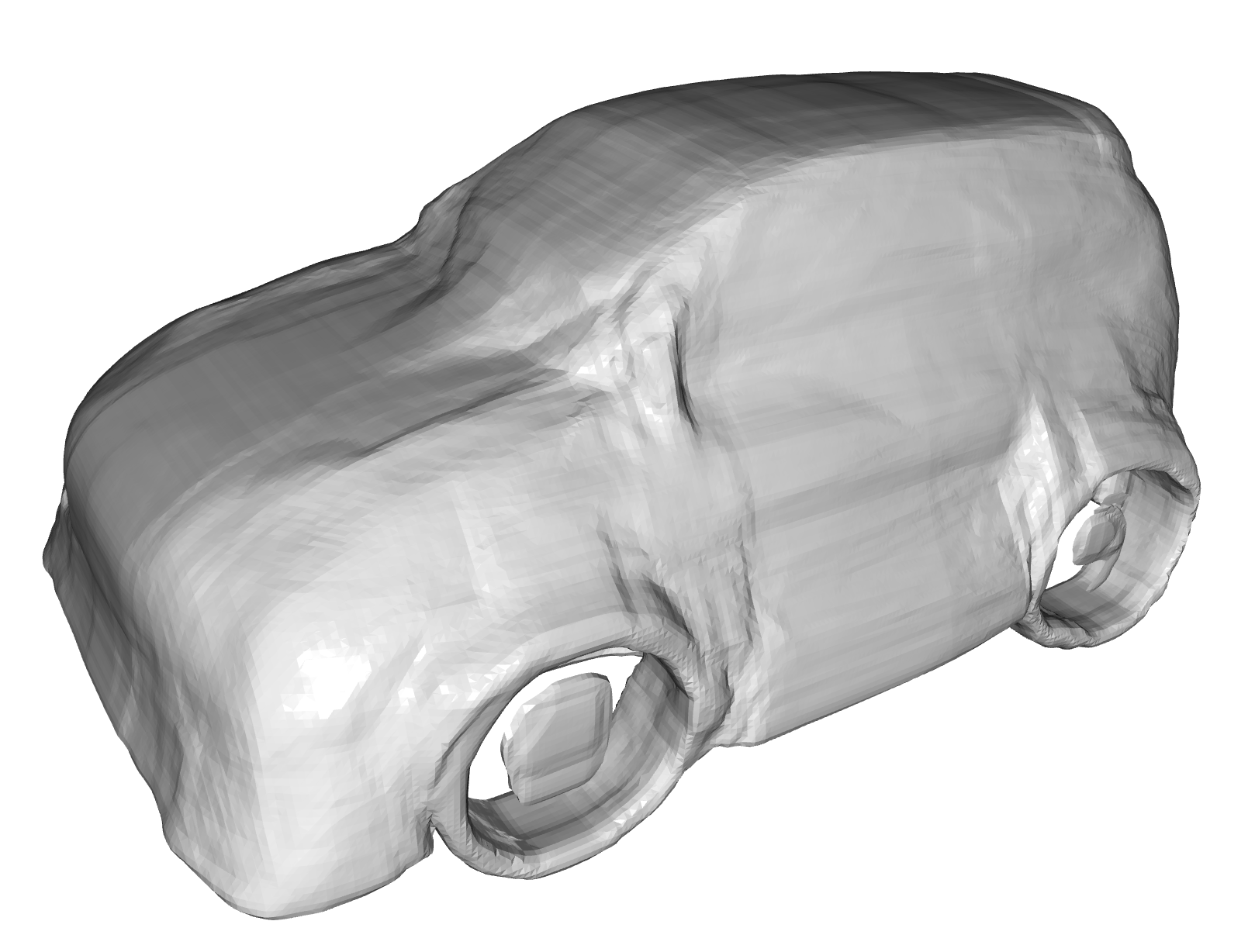}
\caption*{(1.c) Score = 0.43}
\end{subfigure}
\begin{subfigure}[b]{0.19\textwidth}
\includegraphics[width=\textwidth]{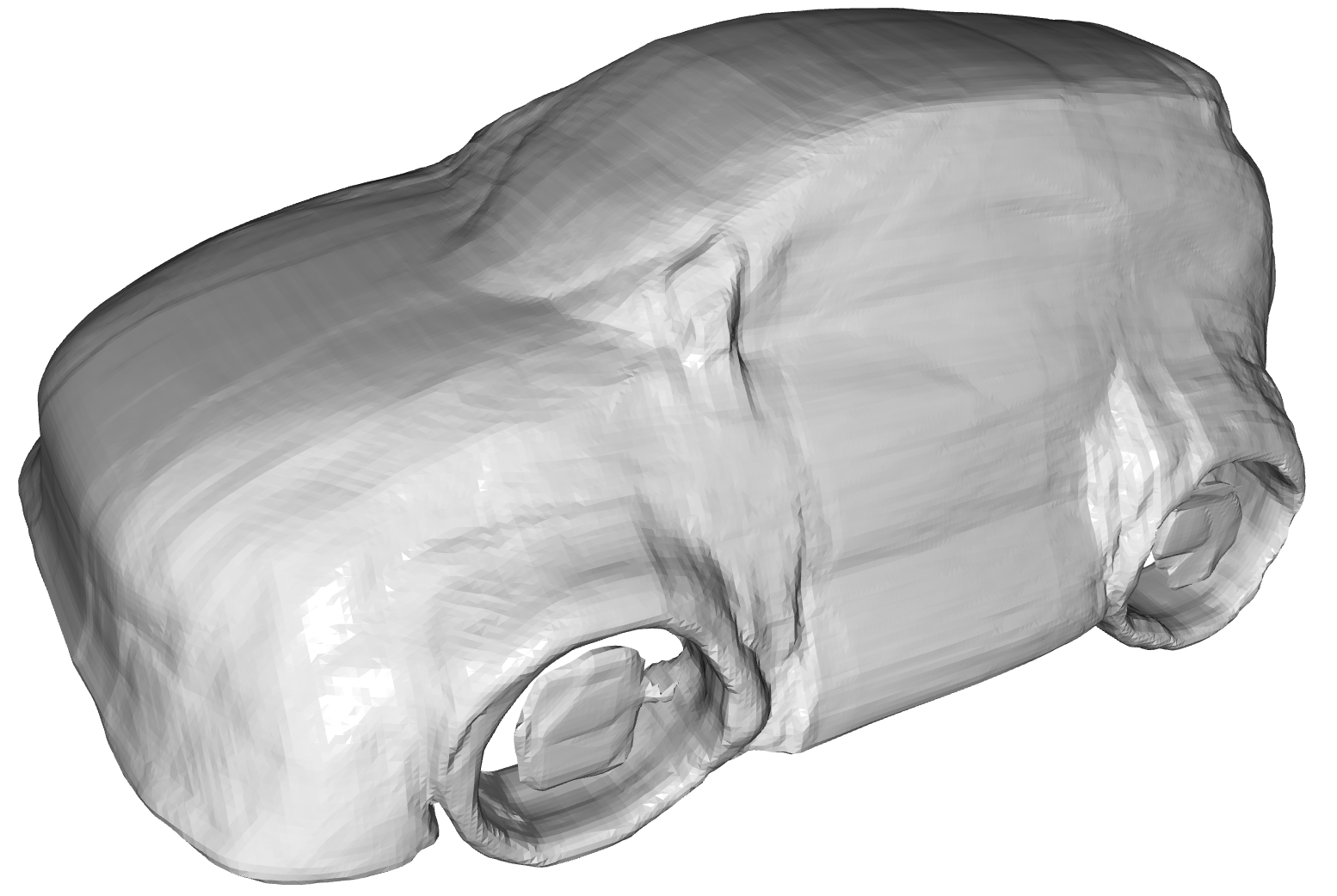}
\caption*{(1.d) Score = 0.35}
\end{subfigure}
\begin{subfigure}[b]{0.19\textwidth}
\includegraphics[width=\textwidth]{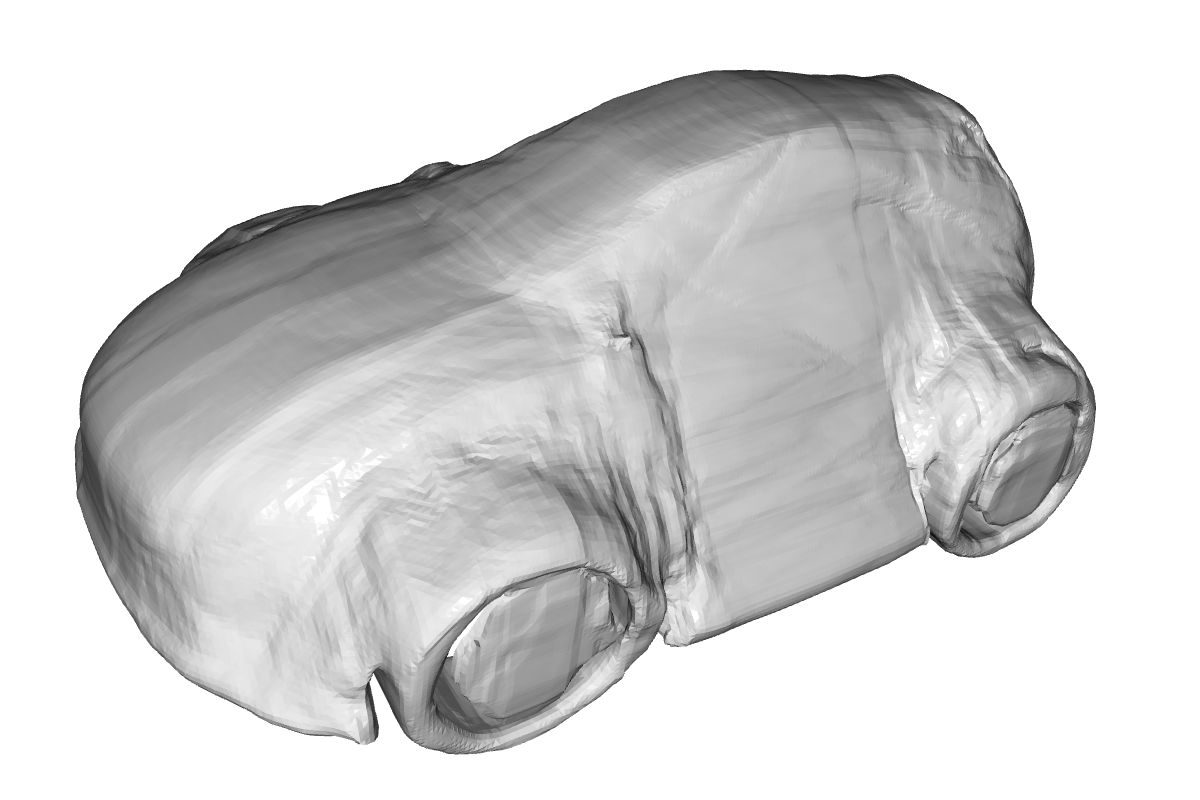}
\caption*{(1.e) Score = 0.03}
\end{subfigure}
\begin{subfigure}[b]{0.19\textwidth}
\includegraphics[width=\textwidth]{figures/threshold_recall/dcaSportsCar.png}
\caption*{(2.a) Score = 0.58}
\end{subfigure}
\begin{subfigure}[b]{0.19\textwidth}
\includegraphics[width=\linewidth,trim={15cm 5cm 20cm 11cm},clip]{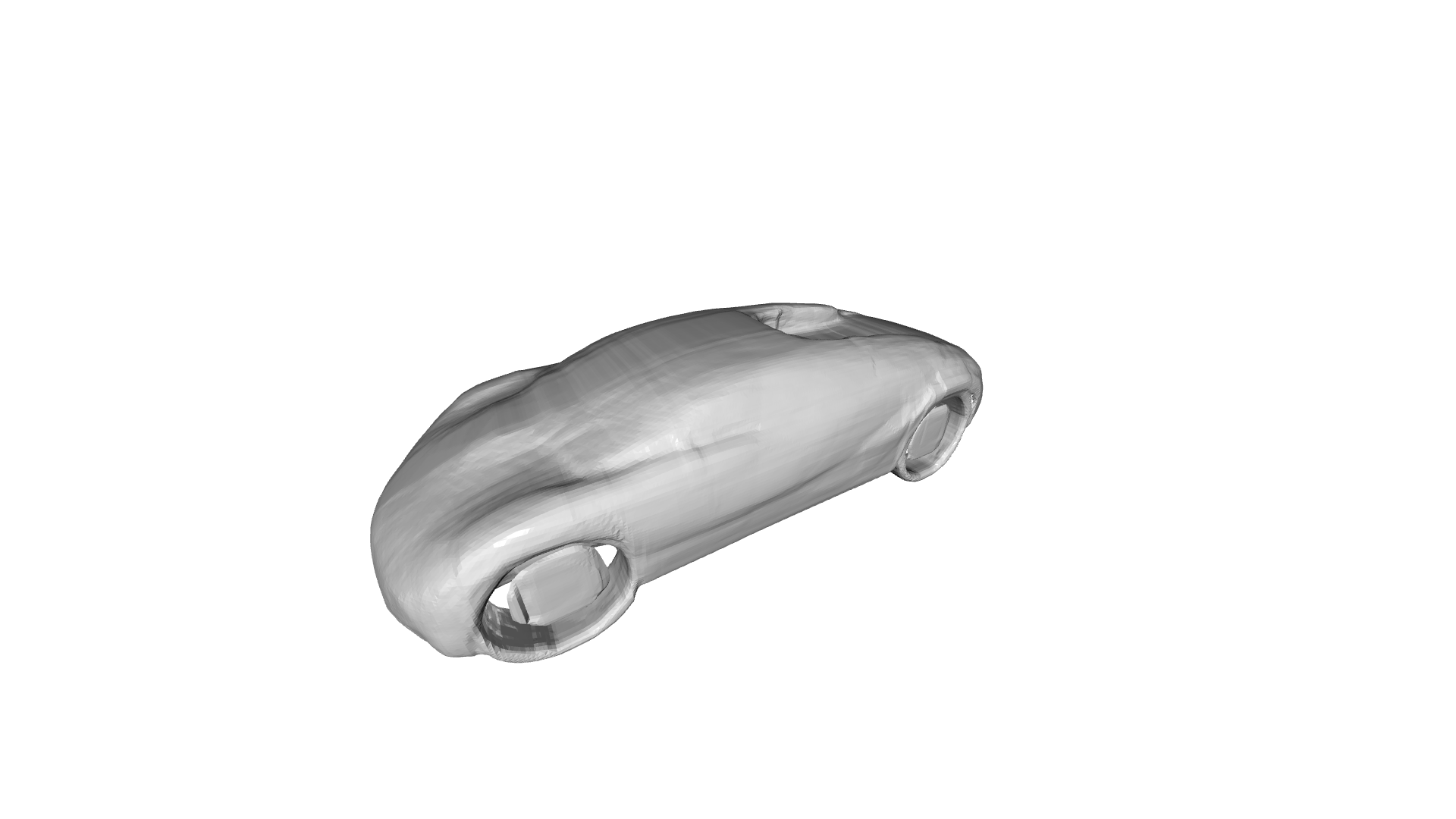}
\caption*{(2.b) Score = 0.42}
\end{subfigure}
\begin{subfigure}[b]{0.19\textwidth}
\includegraphics[width=\linewidth,trim={15cm 5cm 20cm 11cm},clip]{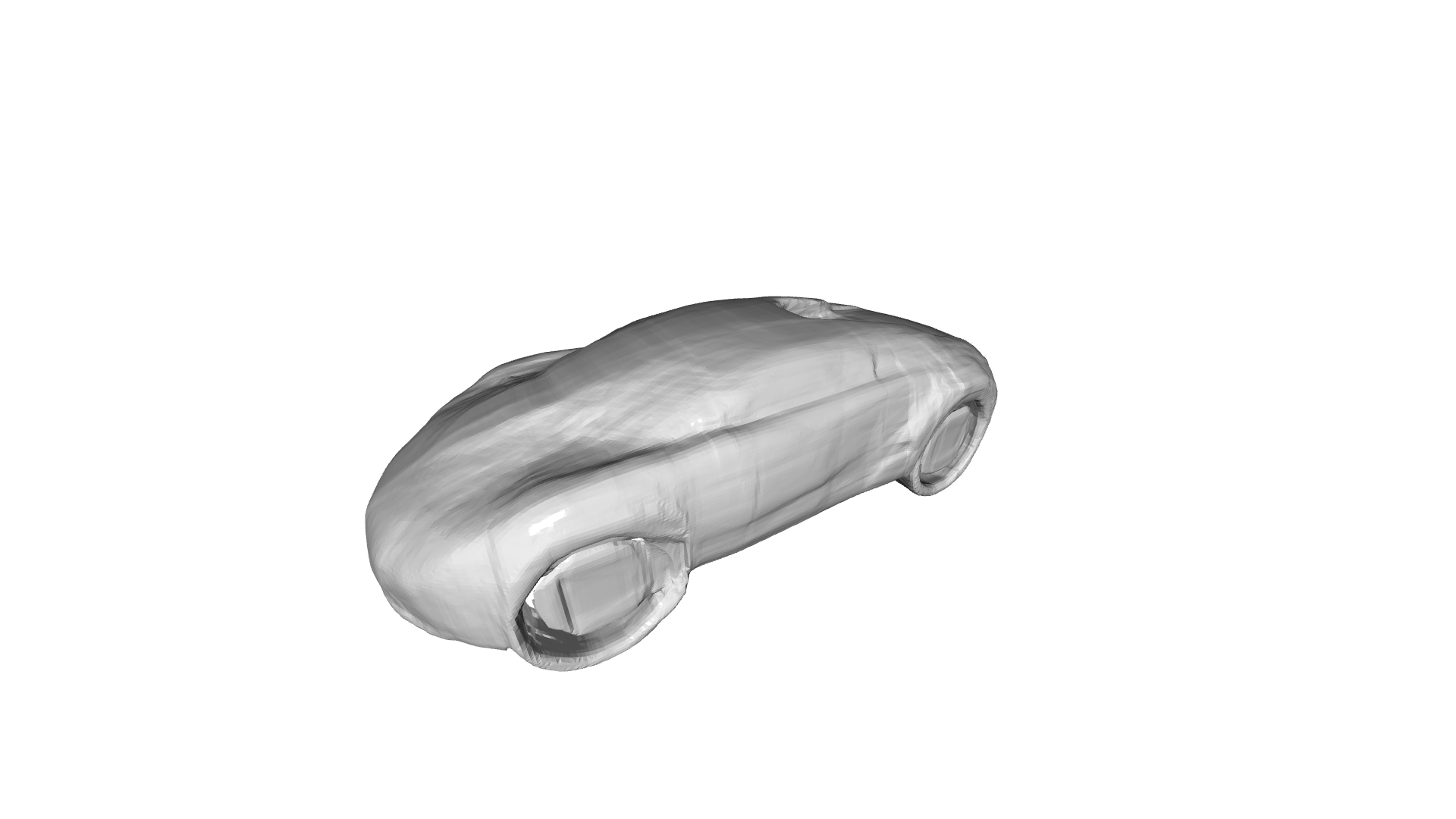}
\caption*{(2.c) Score = 0.35}
\end{subfigure}
\begin{subfigure}[b]{0.19\textwidth}
\includegraphics[width=\linewidth,trim={15cm 5cm 20cm 11cm},clip]{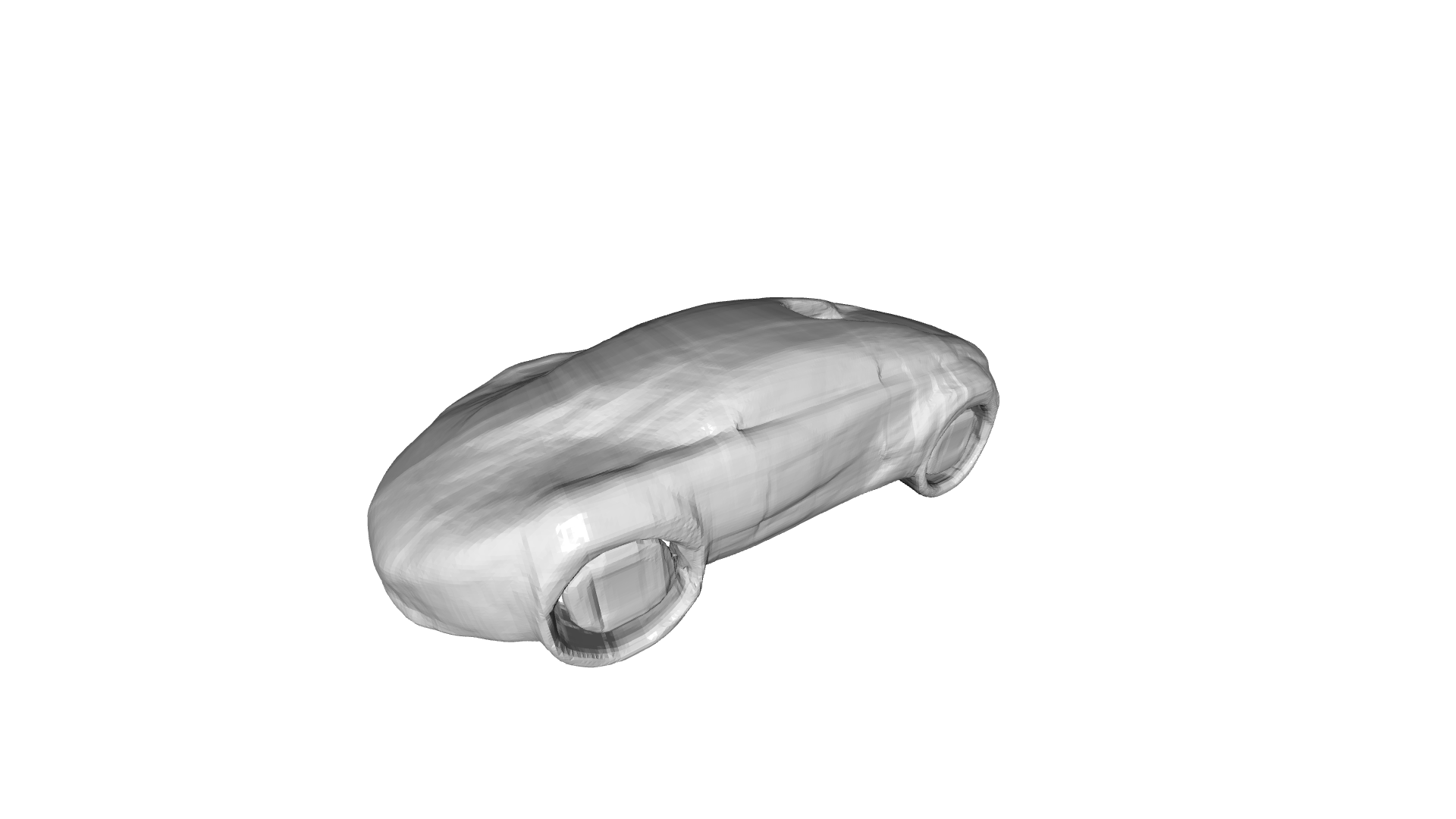}
\caption*{(2.d) Score = 0.17}
\end{subfigure}
\begin{subfigure}[b]{0.19\textwidth}
\includegraphics[width=\linewidth,trim={15cm 5cm 20cm 11cm},clip]{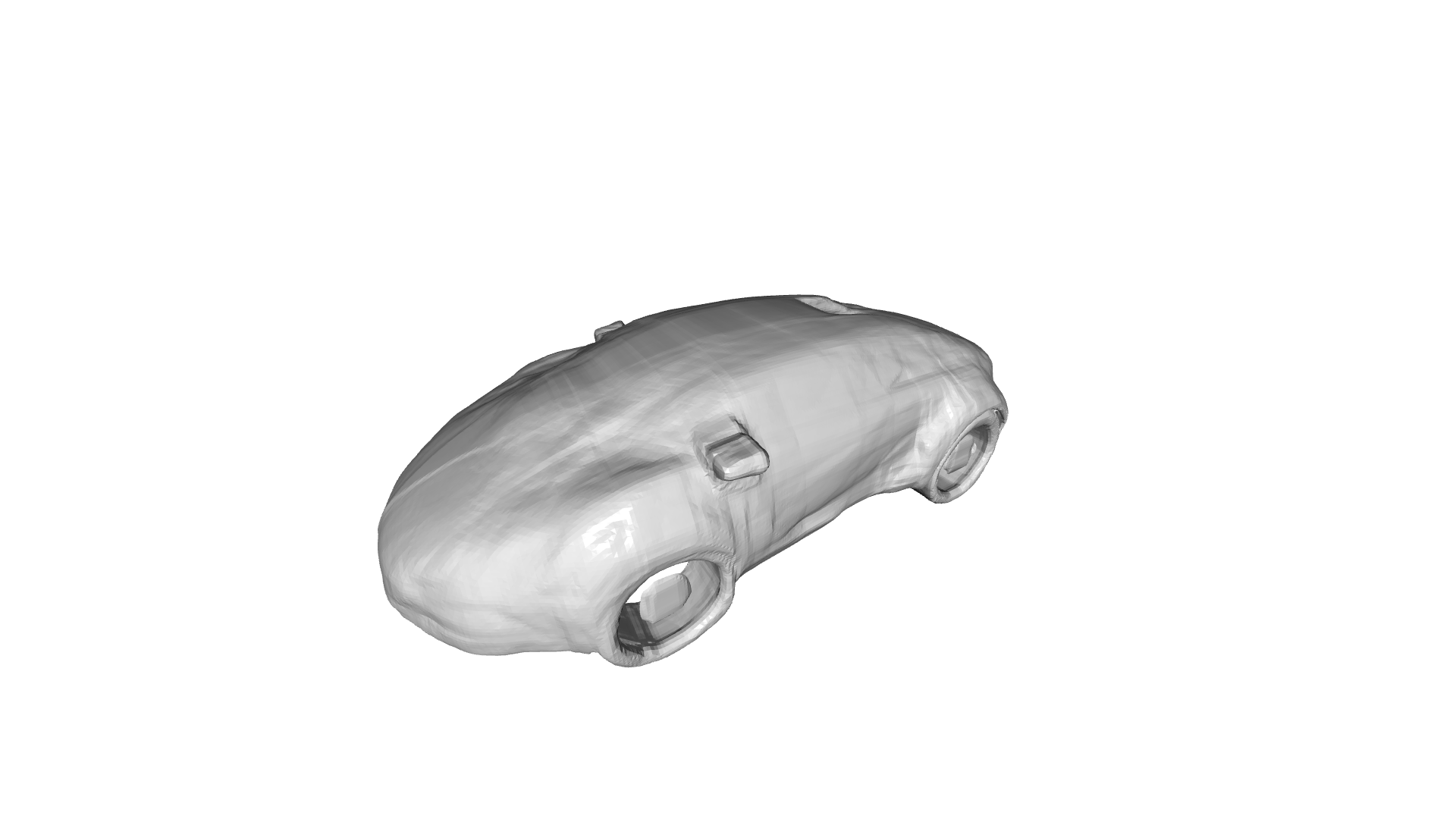}
\caption*{(2.e) Score = 0.00}
\end{subfigure}

\caption{Adversarial shapes and their corresponding detection scores at different steps of the optimization process. 
(a) shows the initial shape corresponding to the baseline object, while (b)-(d) depict intermediate shapes at various stages of the optimization. 
(e) shows the final adversarial shape at the end of the optimization.}
\label{fig:coupe_interpolation}

\end{figure*}

We also present the adversarial objects and the corresponding detection scores at various steps of the optimization process in Figure~\ref{fig:coupe_interpolation}. 
Notably, we observe that \ourmethod{} object detection scores smoothly decrease, which demonstrates that \ourmethod{} can be used to explore challenging examples for a detection model in a continuous space of increasing difficulty, rather than at just a singular point.

\mycomment{
\begin{subfigure}[b]{0.19\textwidth}
\includegraphics[width=\textwidth]{supp_figures/interpolation/Coupe0.png}
\caption*{(2.a) Score = 0.86}
\end{subfigure}
\begin{subfigure}[b]{0.19\textwidth}
\includegraphics[width=\textwidth]{supp_figures/interpolation/Coupe1.png}
\caption*{(2.b) Score = 0.79}
\end{subfigure}
\begin{subfigure}[b]{0.19\textwidth}
\includegraphics[width=\textwidth]{supp_figures/interpolation/Coupe2.png}
\caption*{(2.c) Score = 0.74}
\end{subfigure}
\begin{subfigure}[b]{0.19\textwidth}
\includegraphics[width=\textwidth]{supp_figures/interpolation/Coupe3.png}
\caption*{(2.d) Score = 0.50}
\end{subfigure}
\begin{subfigure}[b]{0.19\textwidth}
\includegraphics[width=\textwidth]{supp_figures/interpolation/Coupe4.png}
\caption*{(2.e) Score = 0.46}
\end{subfigure}
}
\section{Additional Qualitative Results for Adversarial Pose Generation}
\label{sec:pose_visualization}
\begin{figure*}[htbp]
  \centering
  \caption{Additional visualizations of challenging poses and their scenes. \textcolor{red}{Red} boxes are detector's output.}
  \label{fig:theta_supp_figures}
  \begin{tabular}{c|c|cc}
    \toprule
  Input Scene  & \thead{Baseline Object\\ (Shape Not Changed)}  &  \thead{Baseline Pose \\ in the Scene}   & \thead{\ourmethod{} Pose \\ in the Scene}\\  
    \midrule
    \begin{subfigure}[b]{0.18\linewidth}
      \centering
      \includegraphics[width=\linewidth,trim={82cm 80cm 78cm 85cm},clip]{supp_figures/theta_figures/bc1fba46-060b-4c39-96ec-b5d393bc204f_input.pdf}
    \end{subfigure}
    &
    \begin{subfigure}[b]{0.18\linewidth}
      \centering
      \includegraphics[width=\linewidth]{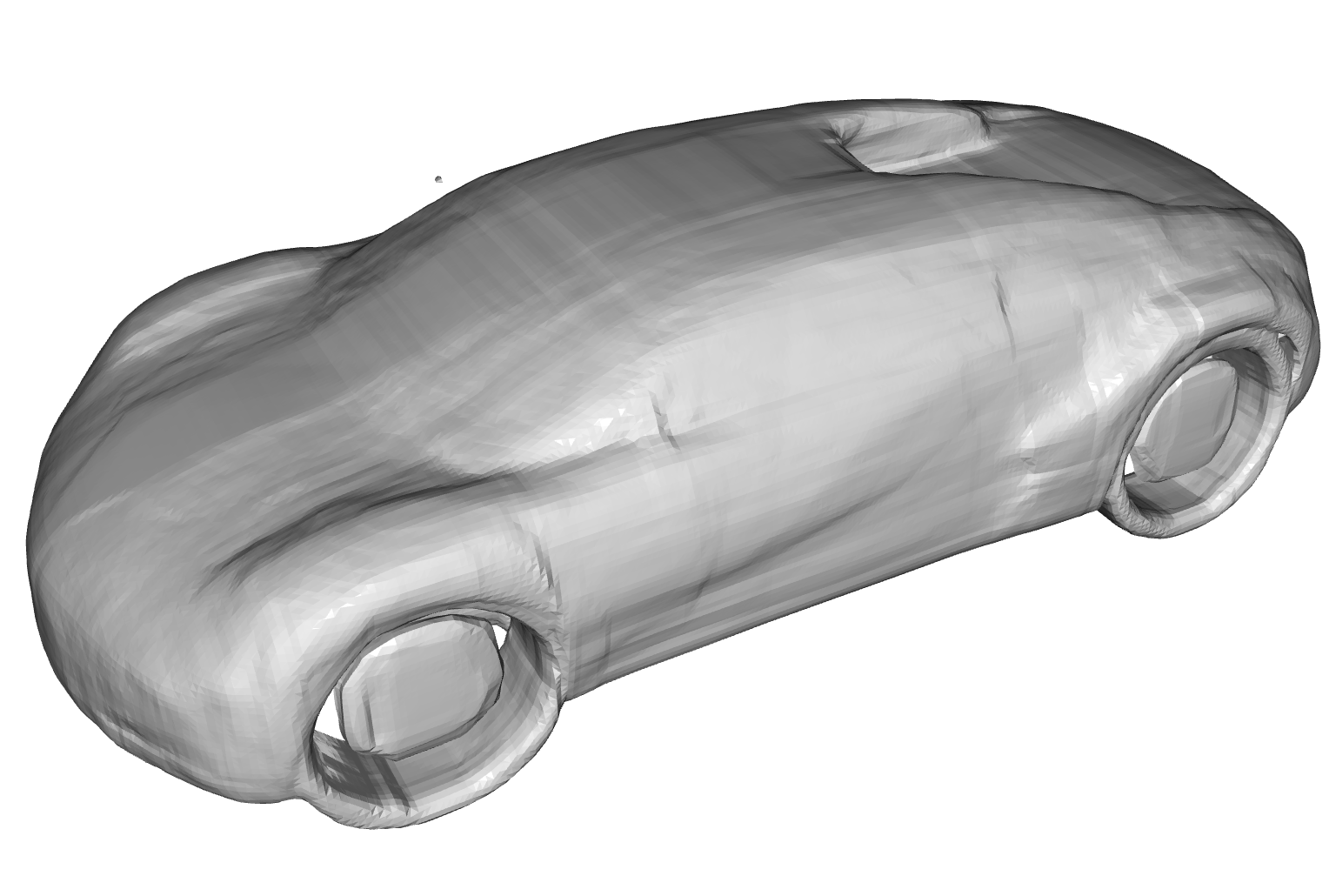}
    \end{subfigure}
    &
    \begin{subfigure}[b]{0.18\linewidth}
      \centering
      \includegraphics[width=\linewidth,trim={82cm 80cm 78cm 85cm},clip]{supp_figures/theta_figures/bc1fba46-060b-4c39-96ec-b5d393bc204f_ori.pdf}
    \end{subfigure}
    &
    \begin{subfigure}[b]{0.18\linewidth}
      \centering
      \includegraphics[width=\linewidth,trim={82cm 80cm 78cm 85cm},clip]{supp_figures/theta_figures/bc1fba46-060b-4c39-96ec-b5d393bc204f_adv_7.pdf}
    \end{subfigure}
    \\
    &  Sports Car & \textcolor{ForestGreen}{Detection Score: 0.95} & \textcolor{OrangeRed}{Detection Score: 0.04}\\
\midrule
    \begin{subfigure}[b]{0.18\linewidth}
      \centering
      \includegraphics[width=\linewidth,trim={82cm 90cm 78cm 75cm},clip]{supp_figures/theta_figures/09262e96-fc34-4f7f-bd7b-8c2c634baa54_input.pdf}
    \end{subfigure}
    &
    \begin{subfigure}[b]{0.18\linewidth}
      \centering
      \includegraphics[width=\linewidth]{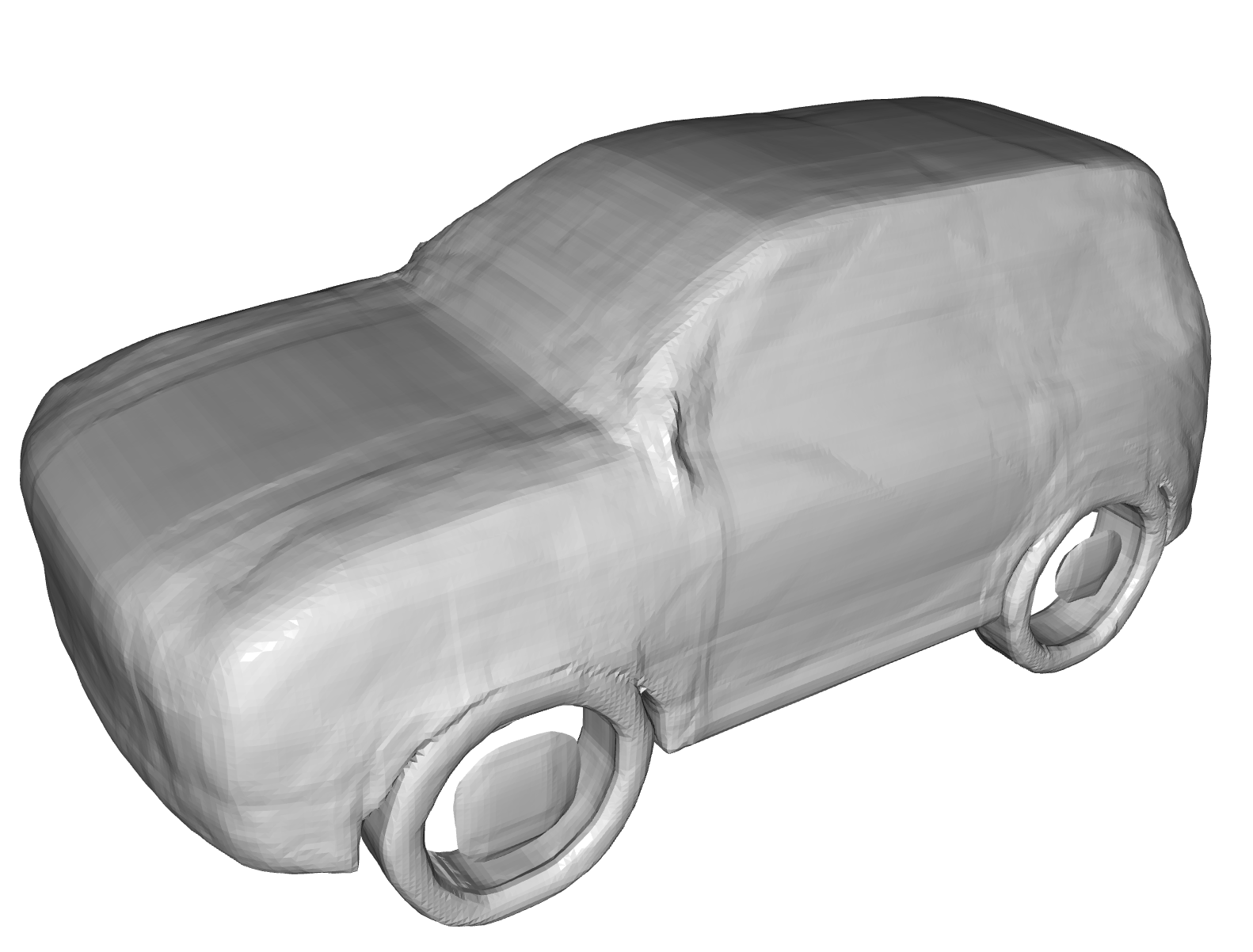}
    \end{subfigure}
    &
    \begin{subfigure}[b]{0.18\linewidth}
      \centering
      \includegraphics[width=\linewidth,trim={82cm 90cm 78cm 75cm},clip]{supp_figures/theta_figures/09262e96-fc34-4f7f-bd7b-8c2c634baa54_ori.pdf}
    \end{subfigure}
    &
    \begin{subfigure}[b]{0.18\linewidth}
      \centering
      \includegraphics[width=\linewidth,trim={82cm 90cm 78cm 75cm},clip]{supp_figures/theta_figures/09262e96-fc34-4f7f-bd7b-8c2c634baa54_adv_3.pdf}
    \end{subfigure}
    \\
    &  SUV & \textcolor{ForestGreen}{Detection Score: 0.80} & \textcolor{OrangeRed}{Detection Score: 0.04}\\
\midrule
    \begin{subfigure}[b]{0.18\linewidth}
      \centering
      \includegraphics[width=\linewidth,trim={82cm 90cm 78cm 75cm},clip]{supp_figures/theta_figures/bd2b4b46-c8e5-4e11-b015-6c4af423163c_input.pdf}
    \end{subfigure}
    &
    \begin{subfigure}[b]{0.18\linewidth}
      \centering
      \includegraphics[width=\linewidth]{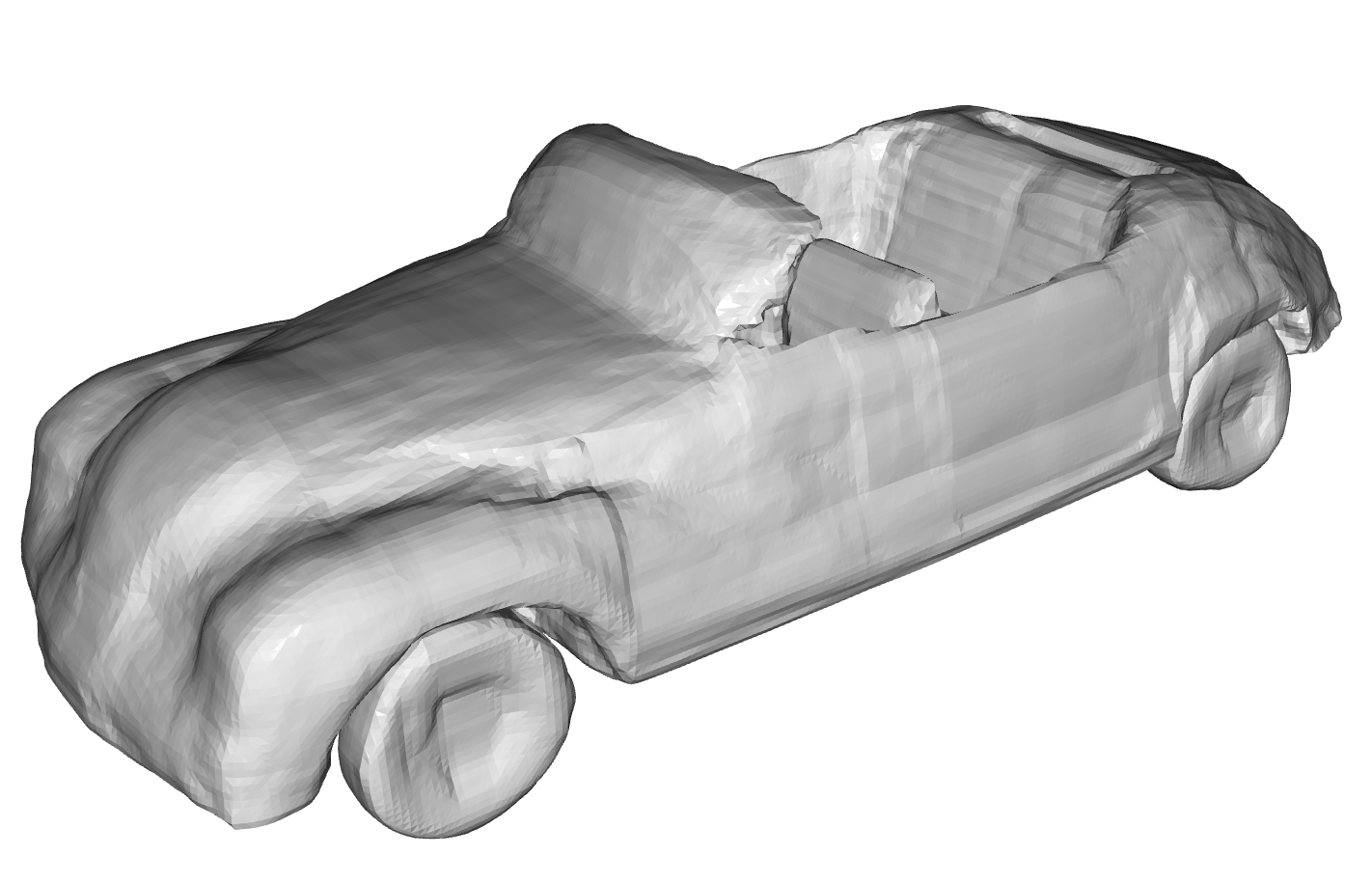}
    \end{subfigure}
    &
    \begin{subfigure}[b]{0.18\linewidth}
      \centering
      \includegraphics[width=\linewidth,trim={82cm 90cm 78cm 75cm},clip]{supp_figures/theta_figures/bd2b4b46-c8e5-4e11-b015-6c4af423163c_ori.pdf}
    \end{subfigure}
    &
    \begin{subfigure}[b]{0.18\linewidth}
      \centering
      \includegraphics[width=\linewidth,trim={82cm 90cm 78cm 75cm},clip]{supp_figures/theta_figures/bd2b4b46-c8e5-4e11-b015-6c4af423163c_adv_1.pdf}
    \end{subfigure}
    \\
    &  Convertible Car & \textcolor{ForestGreen}{Detection Score: 0.75} & \textcolor{OrangeRed}{Detection Score: 0.06}\\
\midrule
    \begin{subfigure}[b]{0.18\linewidth}
      \centering
      \includegraphics[width=\linewidth,trim={90cm 80cm 70cm 85cm},clip]{supp_figures/theta_figures/ff89fb29-5ff1-4d89-b79b-a8bbf930c615_input.pdf}
    \end{subfigure}
    &
    \begin{subfigure}[b]{0.18\linewidth}
      \centering
      \includegraphics[width=\linewidth]{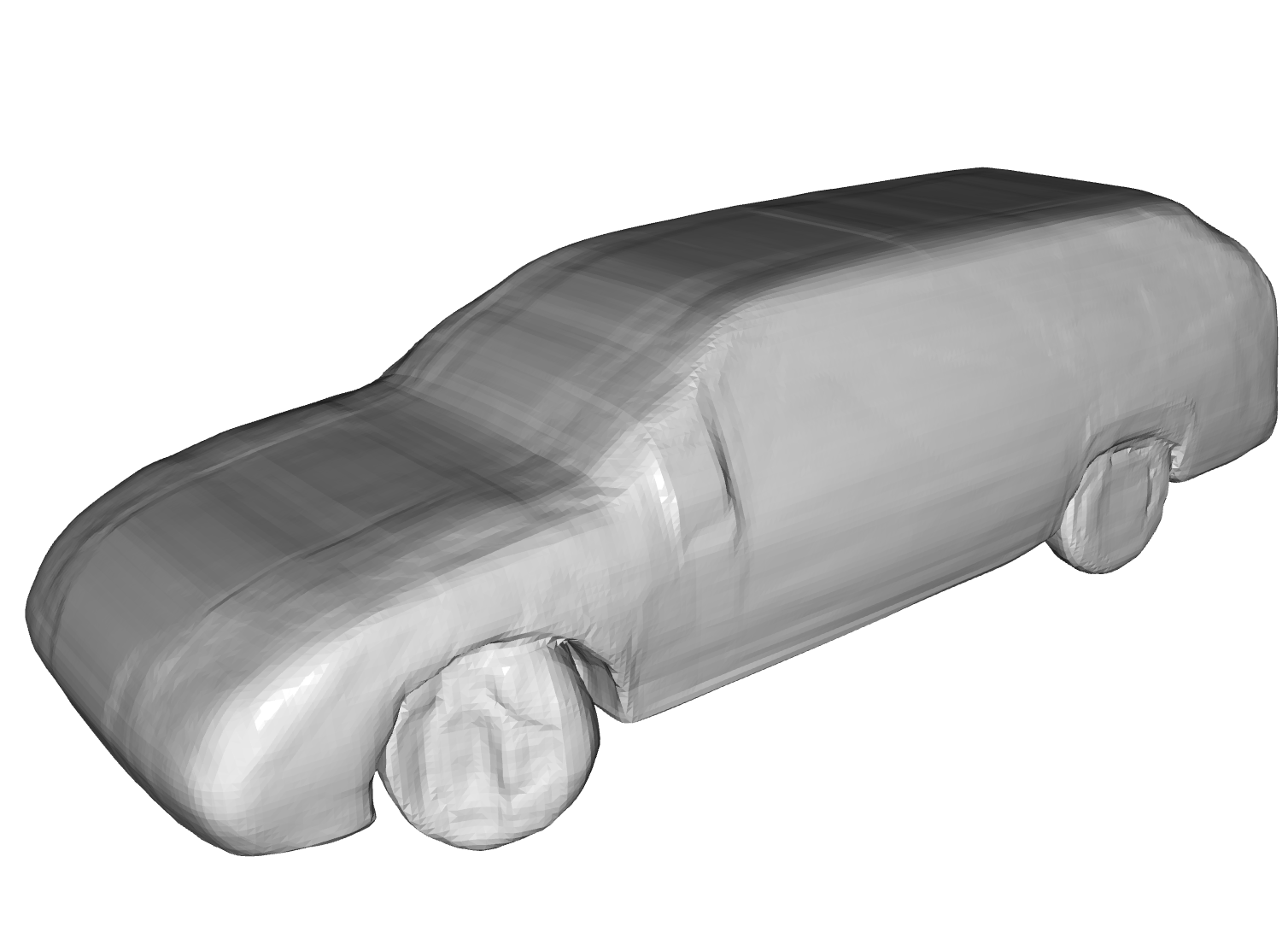}
    \end{subfigure}
    &
    \begin{subfigure}[b]{0.18\linewidth}
      \centering
      \includegraphics[width=\linewidth,trim={90cm 80cm 70cm 85cm},clip]{supp_figures/theta_figures/ff89fb29-5ff1-4d89-b79b-a8bbf930c615_ori.pdf}
    \end{subfigure}
    &
    \begin{subfigure}[b]{0.18\linewidth}
      \centering
      \includegraphics[width=\linewidth,trim={90cm 80cm 70cm 85cm},clip]{supp_figures/theta_figures/ff89fb29-5ff1-4d89-b79b-a8bbf930c615_adv_1.pdf}
    \end{subfigure}
    \\
    &  Beach Wagon & \textcolor{ForestGreen}{Detection Score: 0.78} & \textcolor{OrangeRed}{Detection Score: 0.06}\\
\midrule
    \begin{subfigure}[b]{0.18\linewidth}
      \centering
      \includegraphics[width=\linewidth,trim={70cm 80cm 90cm 85cm},clip]{supp_figures/theta_figures/79c941fd-071f-4ad0-866f-0d7897ac1c79_input.pdf}
    \end{subfigure}
    &
    \begin{subfigure}[b]{0.18\linewidth}
      \centering
      \includegraphics[width=\linewidth]{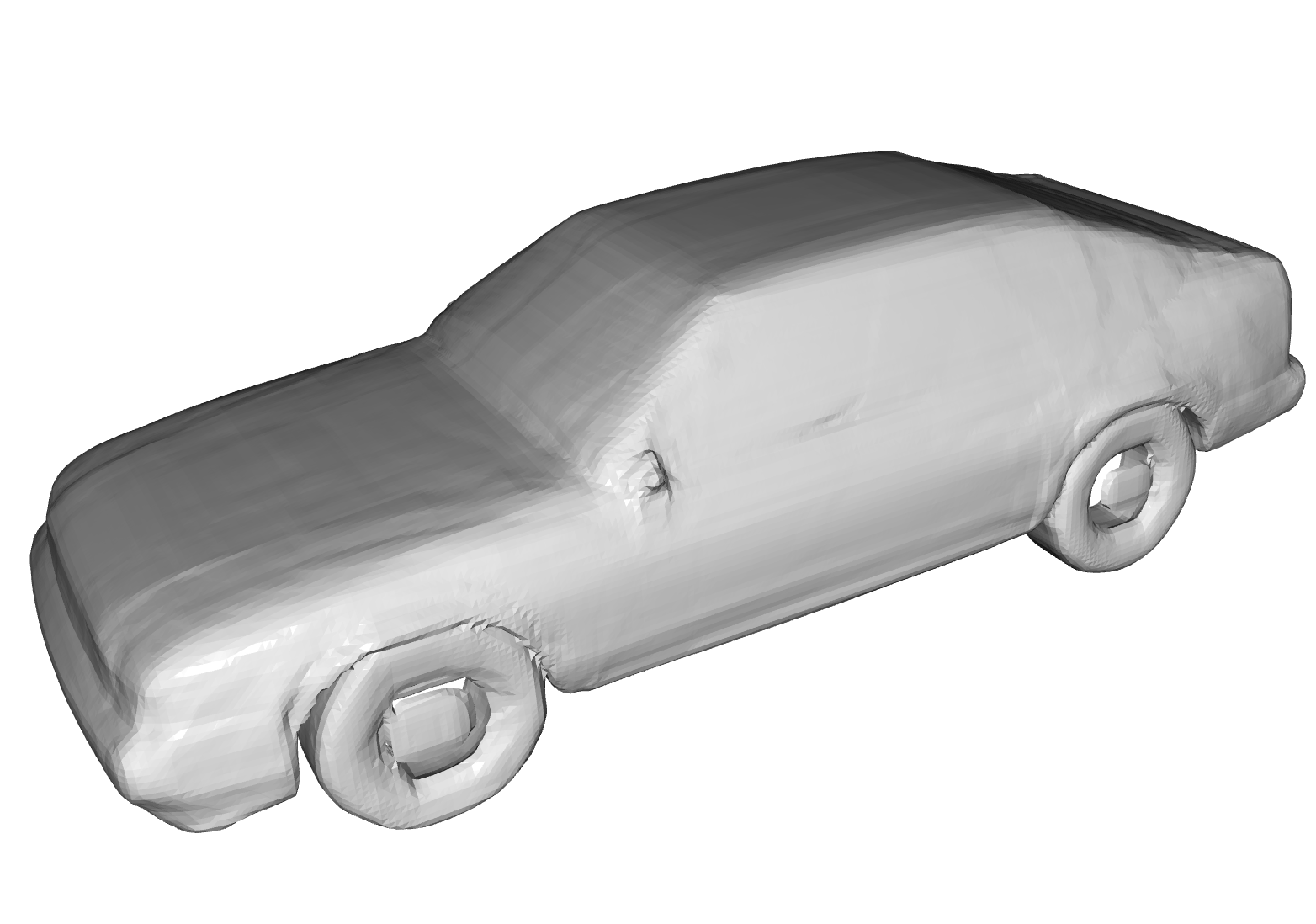}
    \end{subfigure}
    &
    \begin{subfigure}[b]{0.18\linewidth}
      \centering
      \includegraphics[width=\linewidth,trim={70cm 80cm 90cm 85cm},clip]{supp_figures/theta_figures/79c941fd-071f-4ad0-866f-0d7897ac1c79_ori.pdf}
    \end{subfigure}
    &
    \begin{subfigure}[b]{0.18\linewidth}
      \centering
      \includegraphics[width=\linewidth,trim={70cm 80cm 90cm 85cm},clip]{supp_figures/theta_figures/79c941fd-071f-4ad0-866f-0d7897ac1c79_adv_4.pdf}
    \end{subfigure}
    \\
    &  Coupe & \textcolor{ForestGreen}{Detection Score: 0.89} & \textcolor{OrangeRed}{Detection Score: 0.06}\\
\midrule
    \begin{subfigure}[b]{0.18\linewidth}
      \centering
      \includegraphics[width=\linewidth,trim={70cm 80cm 90cm 85cm},clip]{supp_figures/theta_figures/a519dd83-b08c-4dfa-93c6-11cd78ded805_input.pdf}
    \end{subfigure}
    &
    \begin{subfigure}[b]{0.18\linewidth}
      \centering
      \includegraphics[width=\linewidth]{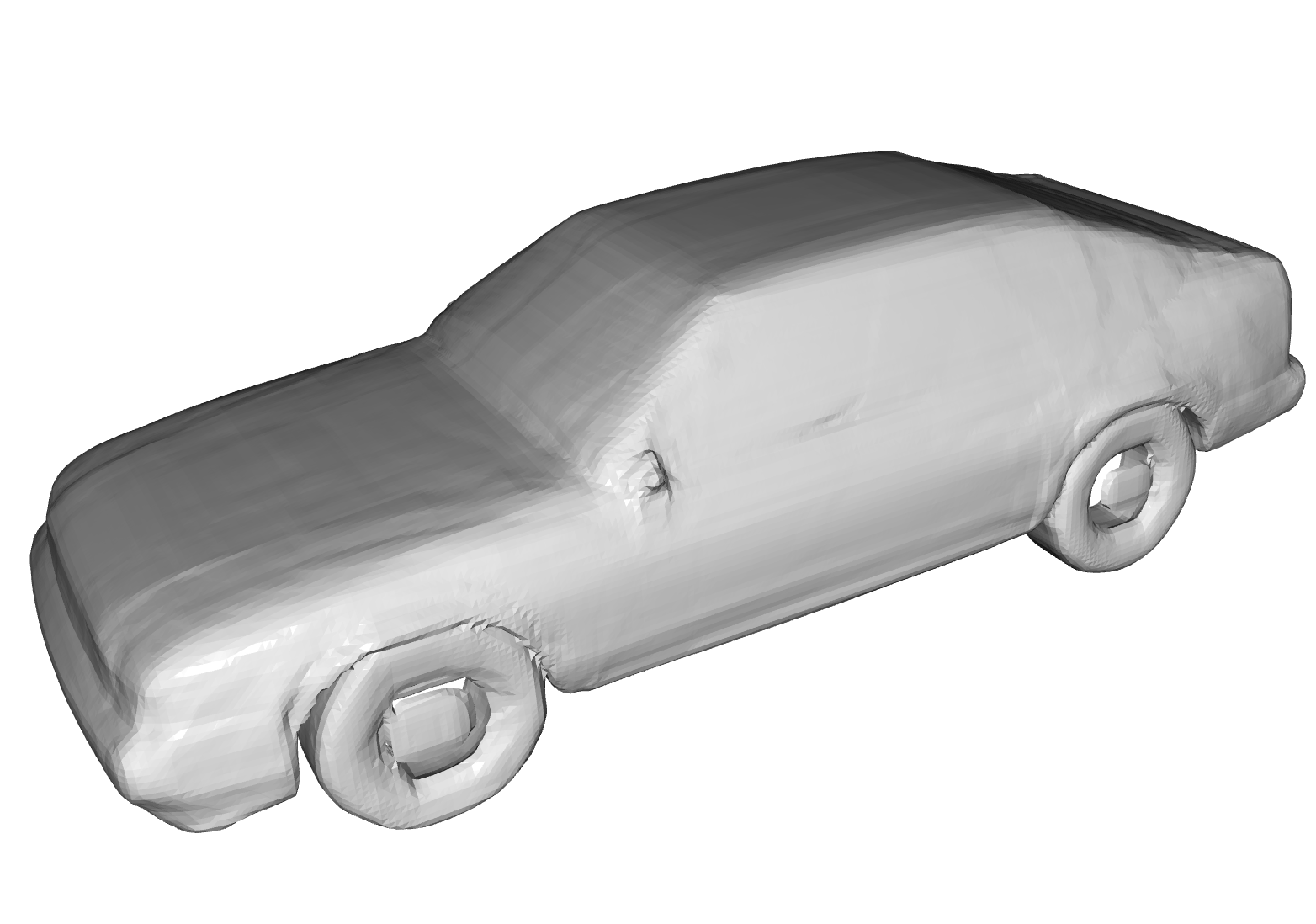}
    \end{subfigure}
    &
    \begin{subfigure}[b]{0.18\linewidth}
      \centering
      \includegraphics[width=\linewidth,trim={70cm 80cm 90cm 85cm},clip]{supp_figures/theta_figures/a519dd83-b08c-4dfa-93c6-11cd78ded805_ori.pdf}
    \end{subfigure}
    &
    \begin{subfigure}[b]{0.18\linewidth}
      \centering
      \includegraphics[width=\linewidth,trim={70cm 80cm 90cm 85cm},clip]{supp_figures/theta_figures/a519dd83-b08c-4dfa-93c6-11cd78ded805_adv_1.pdf}
    \end{subfigure}
    \\
    &  Coupe & \textcolor{ForestGreen}{Detection Score: 0.86} & \textcolor{OrangeRed}{Detection Score: 0.05}\\
   \bottomrule
  \end{tabular}
\end{figure*}

In Figure~\ref{fig:theta_supp_figures} we present more visualizations for the baseline and challenging poses produced by \ourmethod{} in their scenes. Interestingly, we observe that the detection model tends to fail when the inserted \ourmethod{} vehicle is closed to or partially occluded by other objects in the scenes, such as trees, bushes, fences, or other vehicles.

\section{Experiments on SST detectors}
\label{sec:sst}
In this section, we present the results of generating adversarial shape and pose on an SST detector, in order to demonstrate that \ourmethod{} can be used with various network architectures. Our SST detector's performance metrices on natural WOD vehicles are shown in Table~\ref{tab:natural_metrics}. We can see that it overperforms the baseline in~\cite{sun2020scalability}.

\subsection{Adversarial Shape Generation Results}

\begin{figure*}[h!]
       \centering

\begin{subfigure}[b]{0.196\textwidth}
\includegraphics[width=\textwidth,]{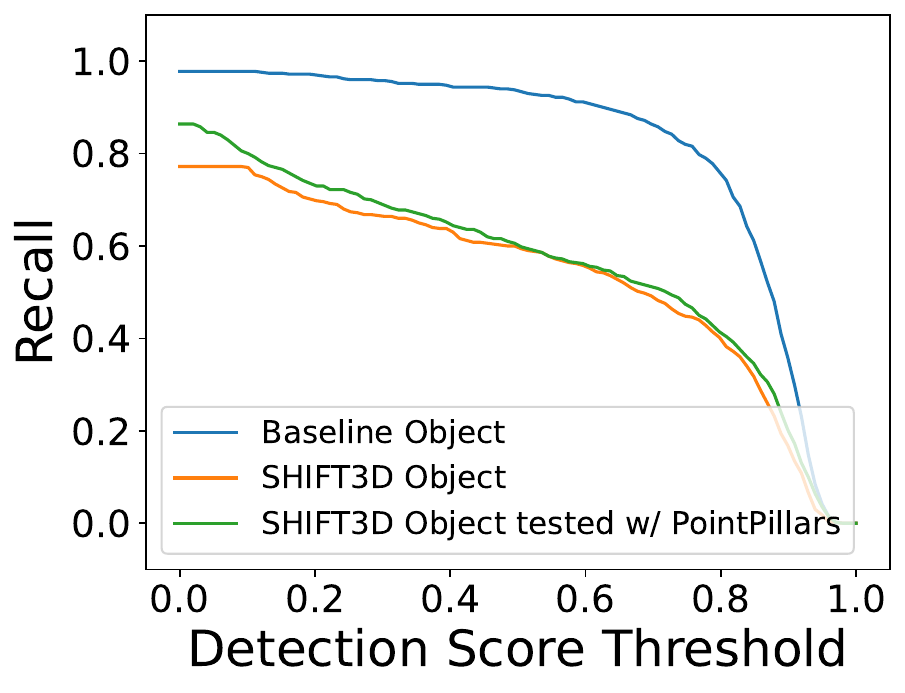}
\caption{Coupe}
\end{subfigure}
\begin{subfigure}[b]{0.196\textwidth}
\includegraphics[width=\textwidth,]{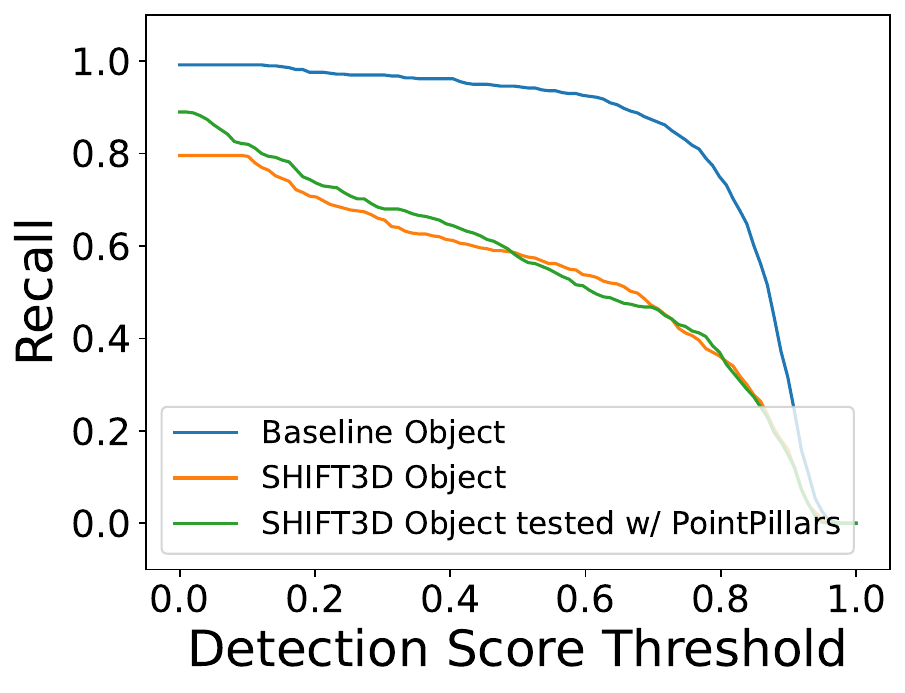}
\caption{Sports Car}
\end{subfigure}
\begin{subfigure}[b]{0.196\textwidth}
\includegraphics[width=\textwidth,]{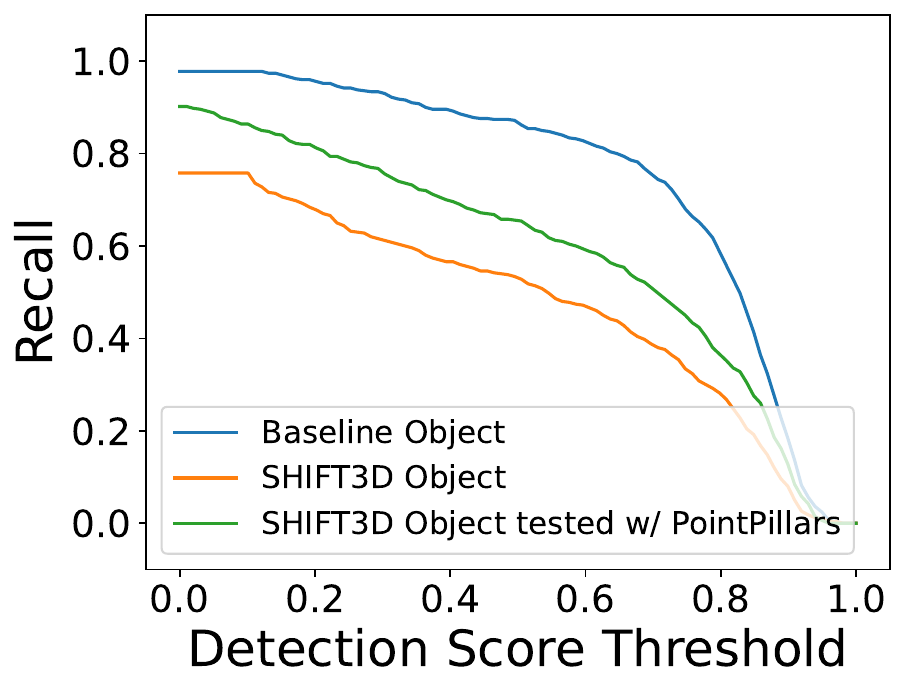}
\caption{SUV}
\end{subfigure}
\begin{subfigure}[b]{0.196\textwidth}
\includegraphics[width=\textwidth,]{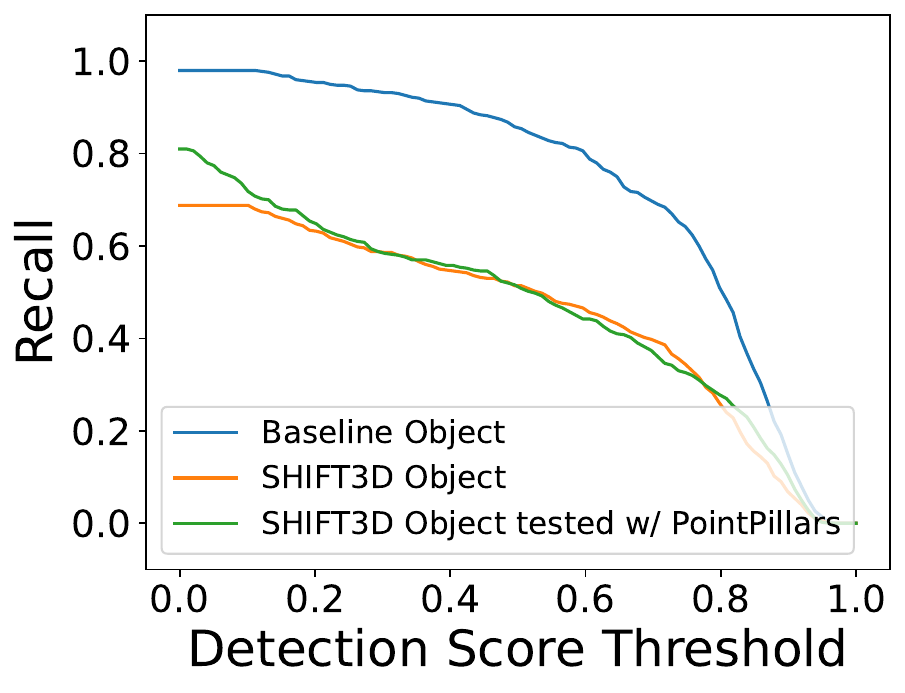}
\caption{Convertible Car}
\end{subfigure}
\begin{subfigure}[b]{0.196\textwidth}
\includegraphics[width=\textwidth,]{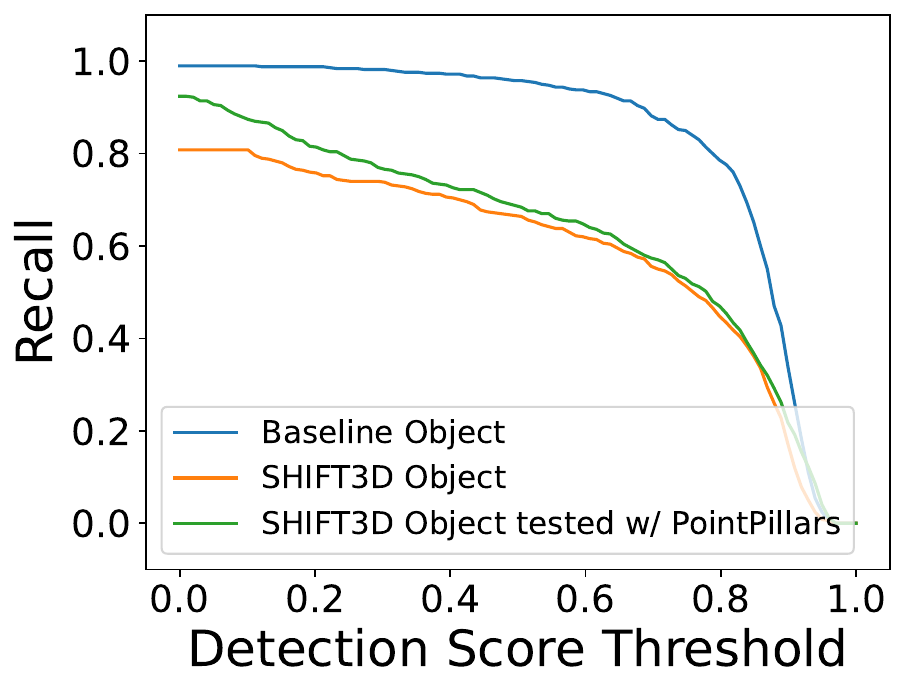}
\caption{Beach Wagon}
\end{subfigure}
\caption{ Threshold-recall curves to evaluate adversarial \emph{shape} generation for different vehicles in the ``Automobile" category with the SST object detector. 
Each curve represents the recall rate of the detector at different detection threshold values, computed from $500$ scenes with the corresponding vehicle present. 
\ourmethod{} demonstrates its effectiveness in deceiving the SST detector. We also evaluated the PointPillars's performance on these objects generated with the SST detector to show the transferability of \ourmethod{}.
}\label{fig:sst_recall_threshold_curve}
    \end{figure*}

Figure~\ref{fig:sst_recall_threshold_curve} presents the adversarial shape generation results for an SST detector, with threshold-recall curves for each
baseline object and its corresponding adversarial objects
across the 500 scenes. To quantify the overall reduction in recall, we compute the area under the threshold-recall curves (AUC) and
report the corresponding numerical values in Table~\ref{tab:sst_auc_z}.

    \begin{table}[h!]
  \centering
  \begin{tabular}{cccccc}
  \toprule
    &\multicolumn{4}{c}{Area Under Curve (AUC)}\\
    Method & Coupe & \thead{Sports \\ Car} & SUV & \thead{Conv. \\ Car} & \thead{Beach \\ Wagon} \\\midrule

    Baseline & 0.812 & 0.817 &  0.743&  0.726 &   0.830\\
    \text{\ourmethod{}} & \textbf{0.533} & \textbf{0.524}&    \textbf{0.471}& \textbf{0.448}  &\textbf{0.585}\\

    \thead{\text{\ourmethod{}} evaluated \\ w/ PointPillars} &\textbf{0.559}&\textbf{0.538}&\textbf{0.581}&\textbf{0.461}&\textbf{0.623}\\
    
    \bottomrule
  \end{tabular}
  \caption{The AUC for the curves in Figure~\ref{fig:sst_recall_threshold_curve}, which demonstrates that \ourmethod{} produces challenging shapes that confuse an SST detector and \ourmethod{} shows high transferability.
  }
  \label{tab:sst_auc_z}
\end{table}
\subsection{Adversarial Pose Generation Results}
\begin{figure*}[h!]
       \centering

\begin{subfigure}[b]{0.196\textwidth}
\includegraphics[width=\textwidth,]{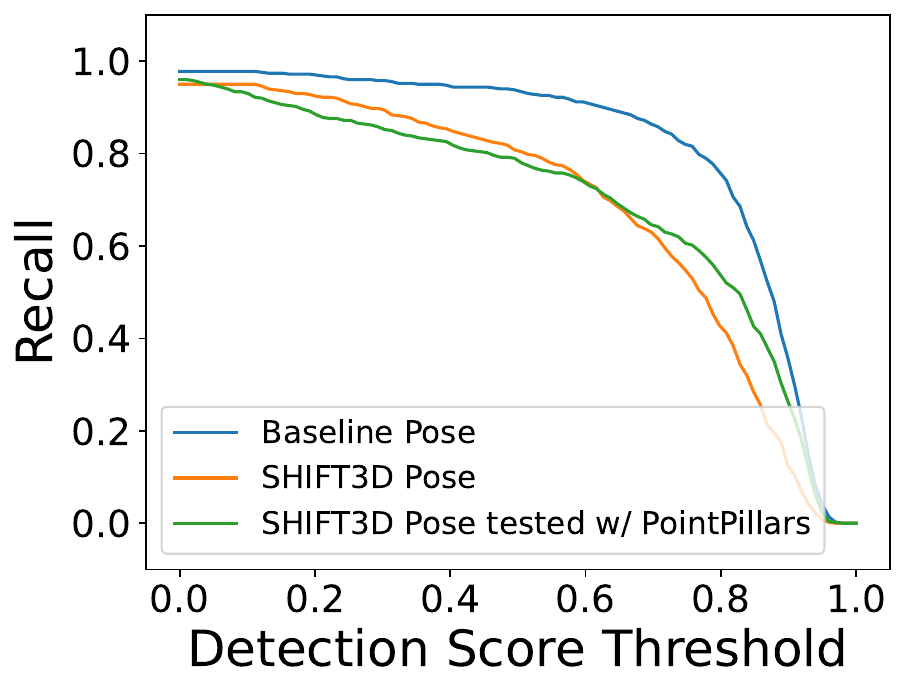}
\caption{Coupe}
\end{subfigure}
\begin{subfigure}[b]{0.196\textwidth}
\includegraphics[width=\textwidth,]{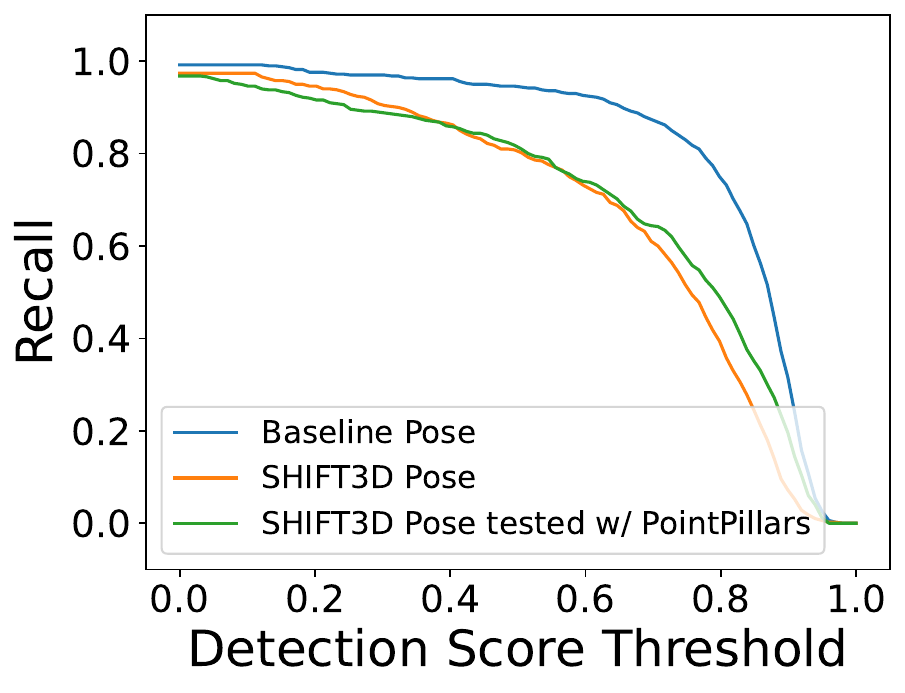}
\caption{Sports Car}
\end{subfigure}
\begin{subfigure}[b]{0.196\textwidth}
\includegraphics[width=\textwidth,]{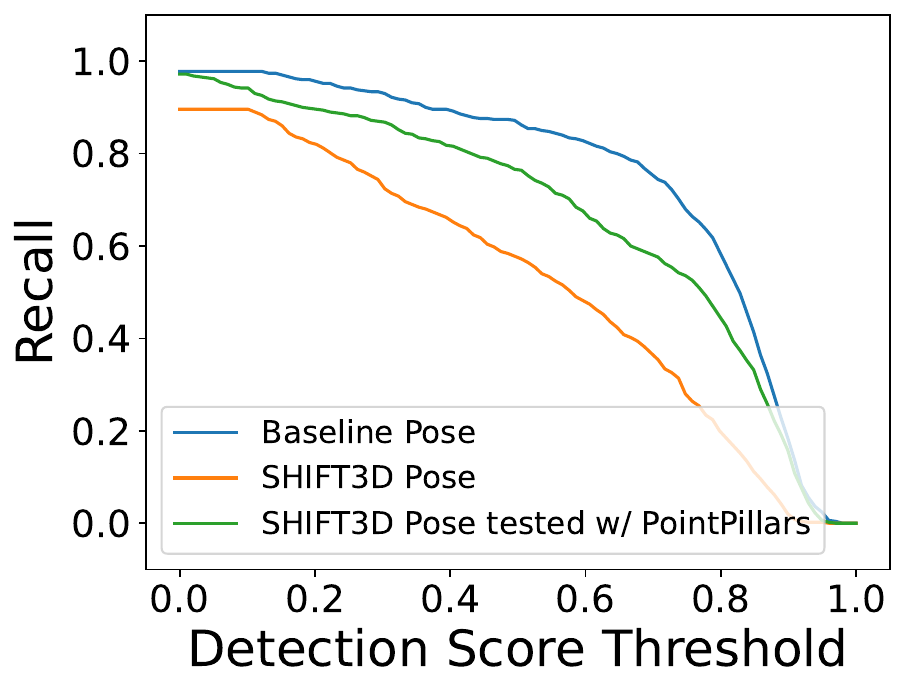}
\caption{SUV}
\end{subfigure}
\begin{subfigure}[b]{0.196\textwidth}
\includegraphics[width=\textwidth,]{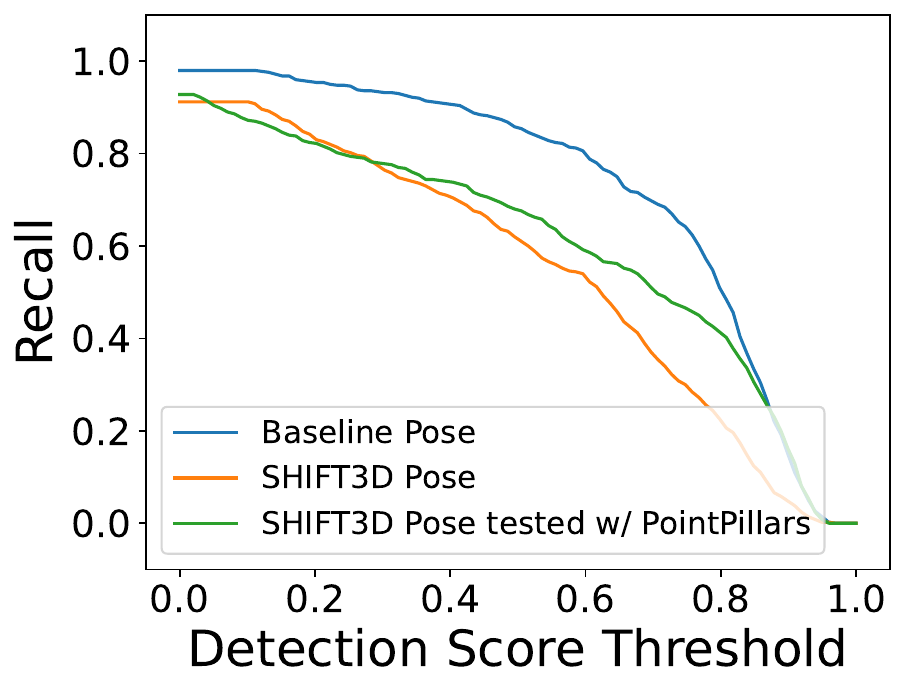}
\caption{Convertible Car}
\end{subfigure}
\begin{subfigure}[b]{0.196\textwidth}
\includegraphics[width=\textwidth,]{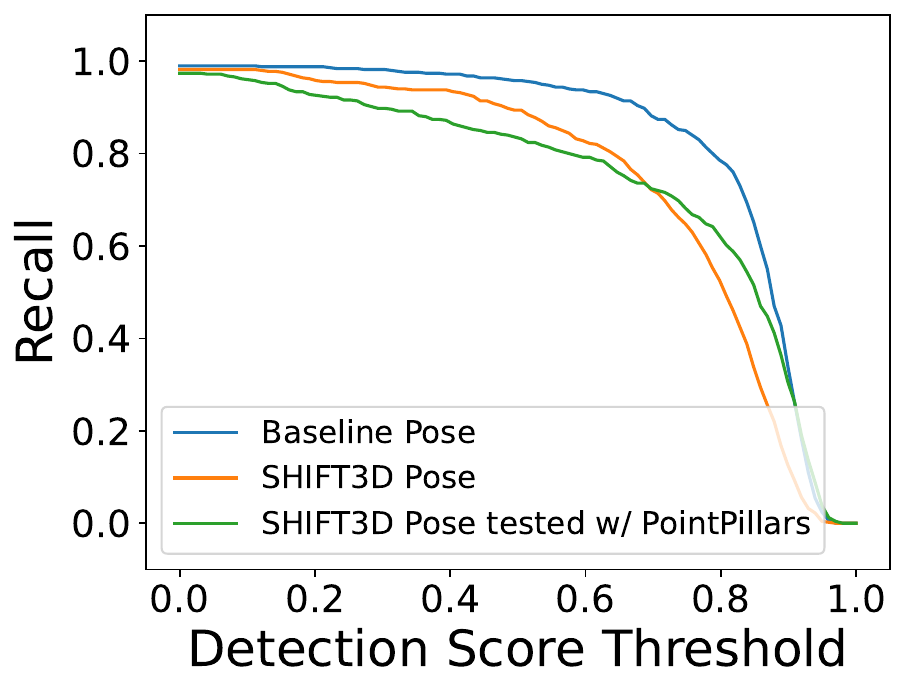}
\caption{Beach Wagon}
\end{subfigure}
\caption{ Threshold-recall curves evaluate adversarial pose for “Automobile” category vehicles with the SST detector. Similar to the curves in Figure~\ref{fig:sst_recall_threshold_curve}, each curve shows the SST detector recall rate at different thresholds, from 500 scenes with the vehicle present. SHIFT3D exhibits significantly lower recall rates and high transferability.
}\label{fig:sst_recall_threshold_curve_trans}
    \end{figure*}

Our results, presented in Figure~\ref{fig:sst_recall_threshold_curve_trans} and Table~\ref{tab:sst_auc_theta}, demonstrate
a significant reduction in recall performance for adversarial
poses, even when placed in alternate poses. 

    \begin{table}[h!]
  \centering
  \begin{tabular}{cccccc}
  \toprule
    &\multicolumn{4}{c}{Area Under Curve (AUC)}\\
    Method & Coupe & \thead{Sports \\ Car} & SUV & \thead{Conv. \\ Car} & \thead{Beach \\ Wagon} \\\midrule

    Baseline & 0.812 & 0.817 &  0.743&  0.726 &   0.830\\
    \text{\ourmethod{}} & \textbf{0.681} & \textbf{0.678}&    \textbf{0.518}& \textbf{0.544}  &\textbf{0.740}\\

    \thead{\text{\ourmethod{}} evaluated \\ w/ PointPillars} &\textbf{0.691} &\textbf{0.695}&\textbf{0.659}&\textbf{0.601}&\textbf{0.739}\\
    
    \bottomrule
  \end{tabular}
  \caption{The AUC for the curves in Figure~\ref{fig:sst_recall_threshold_curve_trans}, which demonstrates that \ourmethod{} produces challenging poses that confuse an SST detector and \ourmethod{} has high transferability.
  }
  \label{tab:sst_auc_theta}
\end{table}
\subsection{\ourmethod{} Transferability between PointPillars and SST}
\label{sec:supp_transfer}
In order to investigate the transferability of \ourmethod{} between different detector models, we also test our PointPillars detector on the shapes and poses generated by \ourmethod{} with the SST detector. As observed in Figures~\ref{fig:sst_recall_threshold_curve} and~\ref{fig:sst_recall_threshold_curve_trans} and Tables~\ref{tab:sst_auc_z} and~\ref{tab:sst_auc_theta}, both shape and pose generated by \ourmethod{} show high transferability, even between two completely different model structures.
\section{Retrieving the Nearest Match for SHIFT3D Queries in Natural Objects}
\label{sec:supp_retrieval}
To test the semantic features produced by \ourmethod{} on a set of natural objects, we examine objects within the WOD validation set that closely resemble the output of \ourmethod{}. These experiments will show that \ourmethod{} is not only useful for understanding 3D object detectors, but also for data discovery within large, possibly unlabeled datasets. 

First, we focus on reconstructing the shapes of natural vehicles from the WOD validation set from their partially occluded LiDAR point clouds. This process begins by cropping out the point cloud corresponding to each vehicle. Using DeepSDF, we then reconstruct the vehicle's shape. The goal is to optimize the shape latent parameters, $\B z$, so that all points are nearly equal to 0. However, a direct optimization of $\B z$ in its native 256-dimensional space often leads to overfitting, resulting in shapes that are not realistic. To counter this, we adopt a dimensionality reduction strategy, akin to the methods in~\cite{tu2020physically} and~\cite{engelmann2017samp}. We perform a PCA on the vehicle shapes' $\B z$ in our DeepSDF training set to create a 10-dimensional subspace $\Pi_{\text{PCA}}$.

Mathematically, for a set of points $\{\B x_i\}$ within a bounding box, where the ground truth pose is $\B\theta$, shape reconstruction is achieved by:
\begin{equation}
\label{eq:reconstruction}
\B z_{\text{natural}}=\argmin_{z\in\Pi_{\text{PCA}}}\sum_i\lVert g(\B z,T(\B x_i;\B \theta)) \rVert^2. 
\end{equation}

Subsequently, a retrieval pool is formulated by calculating $\{\B z_{\text{natural}}\}$ for 8000 natural vehicles in the WOD. For an object generated by \ourmethod{}, represented by $\B z_{\text{\ourmethod{}}}$, we identify the natural object whose $\B z_{\text{natural}}$ is the most similar by minimizing the $\ell_2$ distance:
\begin{equation}
\B z_{\text{natural}}^*=\argmin_{z_{\text{natural}}}\lVert \B z_{\text{natural}}-\B z_{\text{SHIFT3D}}\rVert_2
\end{equation}

To guarantee the reliability of the retrieval, we exclude instances where the computed $z_{\text{natural}}^*$ is significantly different from $\B z_{\text{\ourmethod{}}}$. For clarity, in our study, only objects with $\lVert \B z_{\text{natural}^*}-\B z_{\text{\ourmethod{}}}\rVert_2<\lVert \B z_{\text{\ourmethod{}}}\rVert_2$ are considered.

\begin{figure*}[h!]
       \centering

\begin{subfigure}[b]{0.196\textwidth}
\includegraphics[width=\textwidth,]{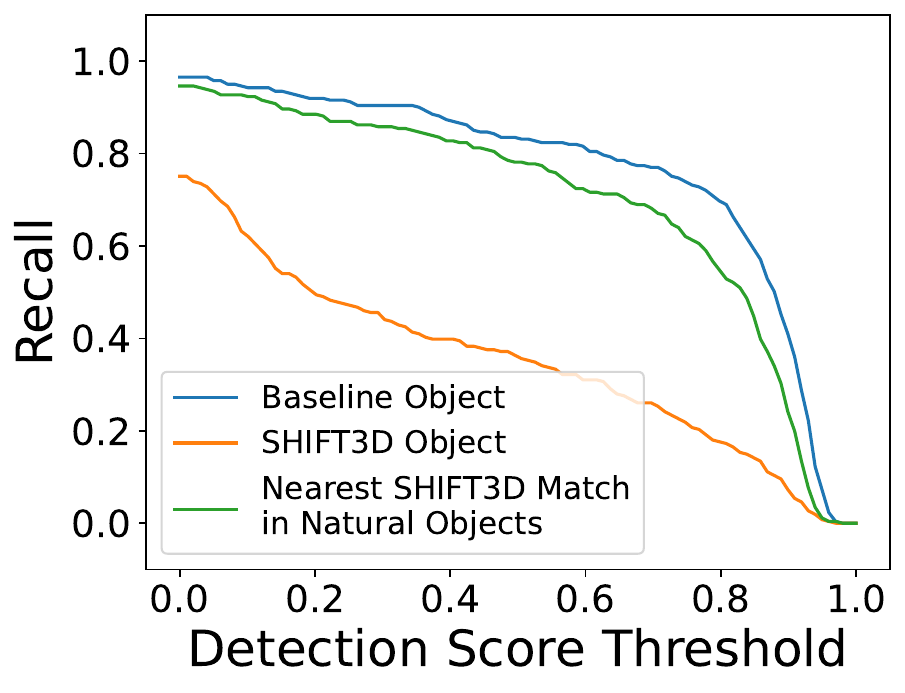}
\caption{Coupe}
\end{subfigure}
\begin{subfigure}[b]{0.196\textwidth}
\includegraphics[width=\textwidth,]{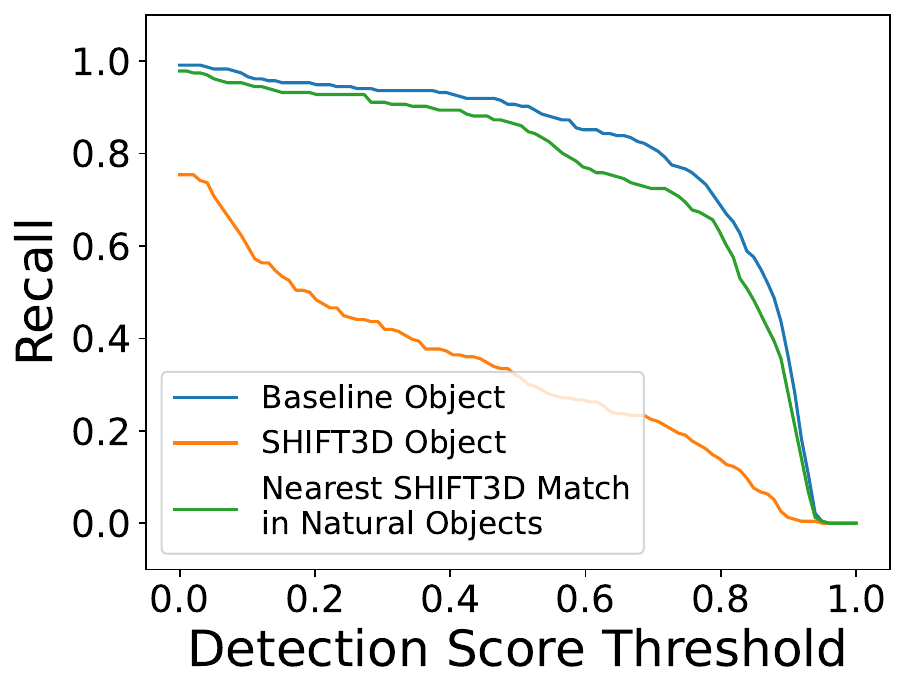}
\caption{Sports Car}
\end{subfigure}
\begin{subfigure}[b]{0.196\textwidth}
\includegraphics[width=\textwidth,]{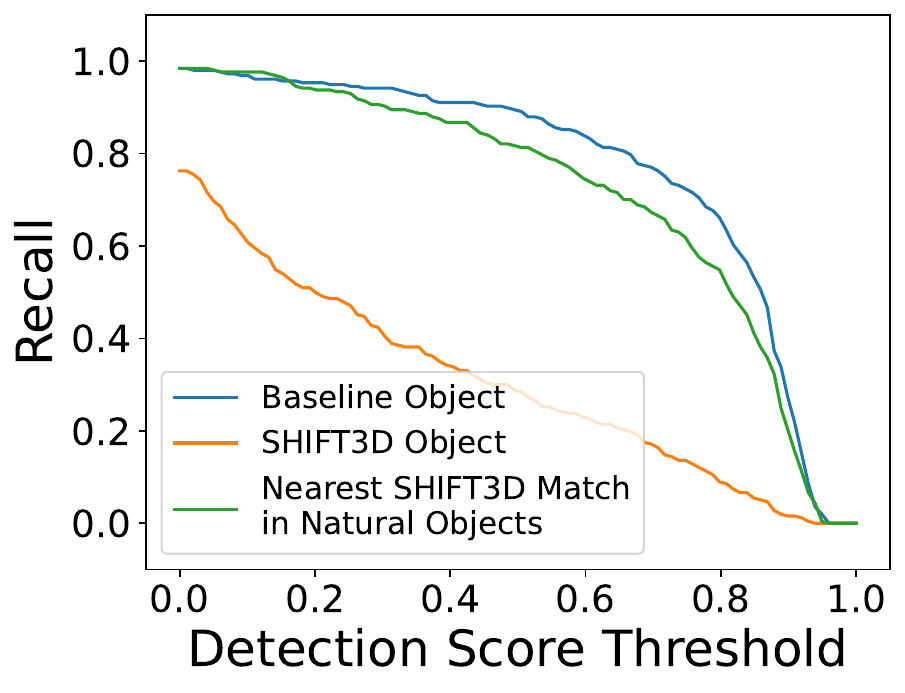}
\caption{SUV}
\end{subfigure}
\begin{subfigure}[b]{0.196\textwidth}
\includegraphics[width=\textwidth,]{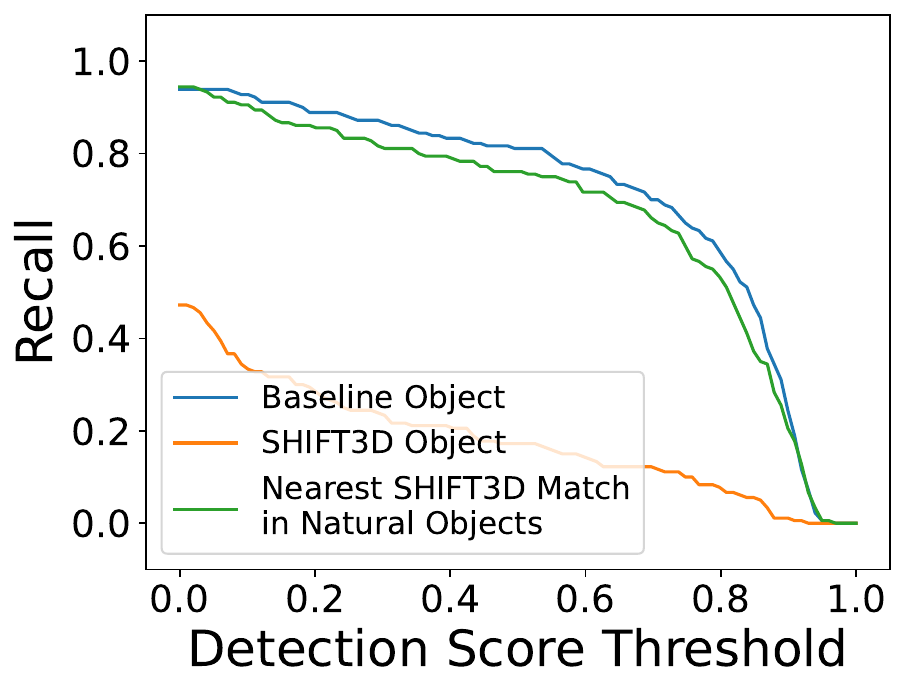}
\caption{Convertible Car}
\end{subfigure}
\begin{subfigure}[b]{0.196\textwidth}
\includegraphics[width=\textwidth,]{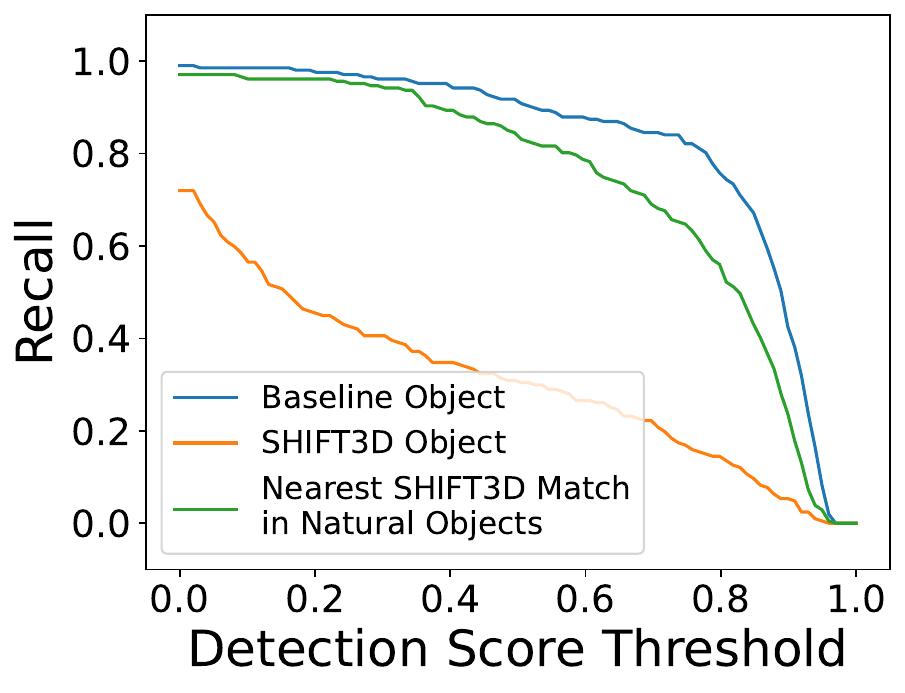}
\caption{Beach Wagon}
\end{subfigure}
\caption{ Threshold-recall curves to evaluate the retrieved nearest matches of
adversarial \emph{shape} generation for different vehicles in the ``Automobile" category. We only plot objects such that $\lVert \B z_{\text{natural}^*}-\B z_{\text{SHIFT3D}}\rVert_2<\lVert \B z_{\text{SHIFT3D}}\rVert_2$
}
\label{fig:retrieval_recall_threshold_curve}
    \end{figure*}

\begin{figure*}[htbp]
  \centering
  \caption{Visualizations of the retrieved nearest matches of \ourmethod{} objects.}
  \label{fig:retrieval_vis}
  \begin{tabular}{c|c|c|c|c|c}
    \toprule
     &  Sports Car   & SUV &    \thead{Convertible \\ Car} & \thead{Beach \\ Wagon} & Coupe\\  
    \midrule
    \thead{Baseline \\ Objects\\~\\~\\~}&
    \begin{subfigure}[b]{0.15\linewidth}
      \centering
      \includegraphics[width=\linewidth]{figures/threshold_recall/dcaSportsCar.png}
    \end{subfigure}
    &
    \begin{subfigure}[b]{0.15\linewidth}
      \centering
      \includegraphics[width=\linewidth]{figures/threshold_recall/df7SUV.png}
    \end{subfigure}
    &
    \begin{subfigure}[b]{0.15\linewidth}
      \centering
      \includegraphics[width=\linewidth]{figures/threshold_recall/e9aConvertible.png}
    \end{subfigure}
    &
    \begin{subfigure}[b]{0.15\linewidth}
      \centering
      \includegraphics[width=\linewidth]{figures/threshold_recall/ee9BeachWagon.png}
    \end{subfigure}
    &
    \begin{subfigure}[b]{0.15\linewidth}
      \centering
      \includegraphics[width=\linewidth]{figures/threshold_recall/ef6Coupe.png}
    \end{subfigure}
    \\
    &  \textcolor{ForestGreen}{Score: 0.91} & \textcolor{ForestGreen}{Score: 0.41} & \textcolor{ForestGreen}{Score: 0.62}&\textcolor{ForestGreen}{Score: 0.50}&\textcolor{ForestGreen}{Score: 0.60}\\\midrule
    \thead{\ourmethod{} \\ Objects\\~\\~\\~}&
    \begin{subfigure}[b]{0.15\linewidth}
      \centering
      \includegraphics[width=\linewidth,trim={15cm 5cm 20cm 12cm},clip]{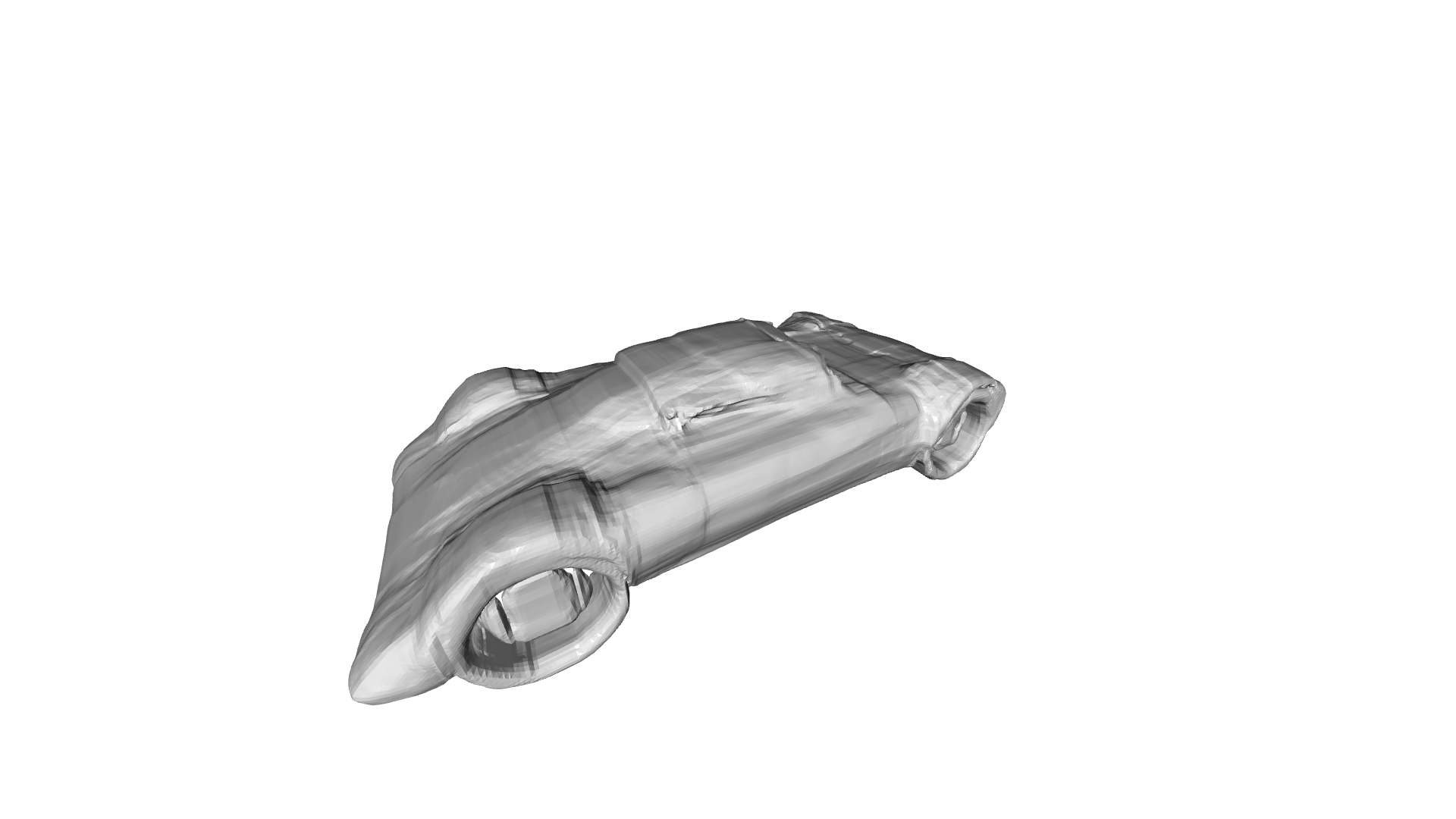}
    \end{subfigure}
    &
    \begin{subfigure}[b]{0.15\linewidth}
      \centering
      \includegraphics[width=\linewidth,trim={15cm 5cm 20cm 12cm},clip]{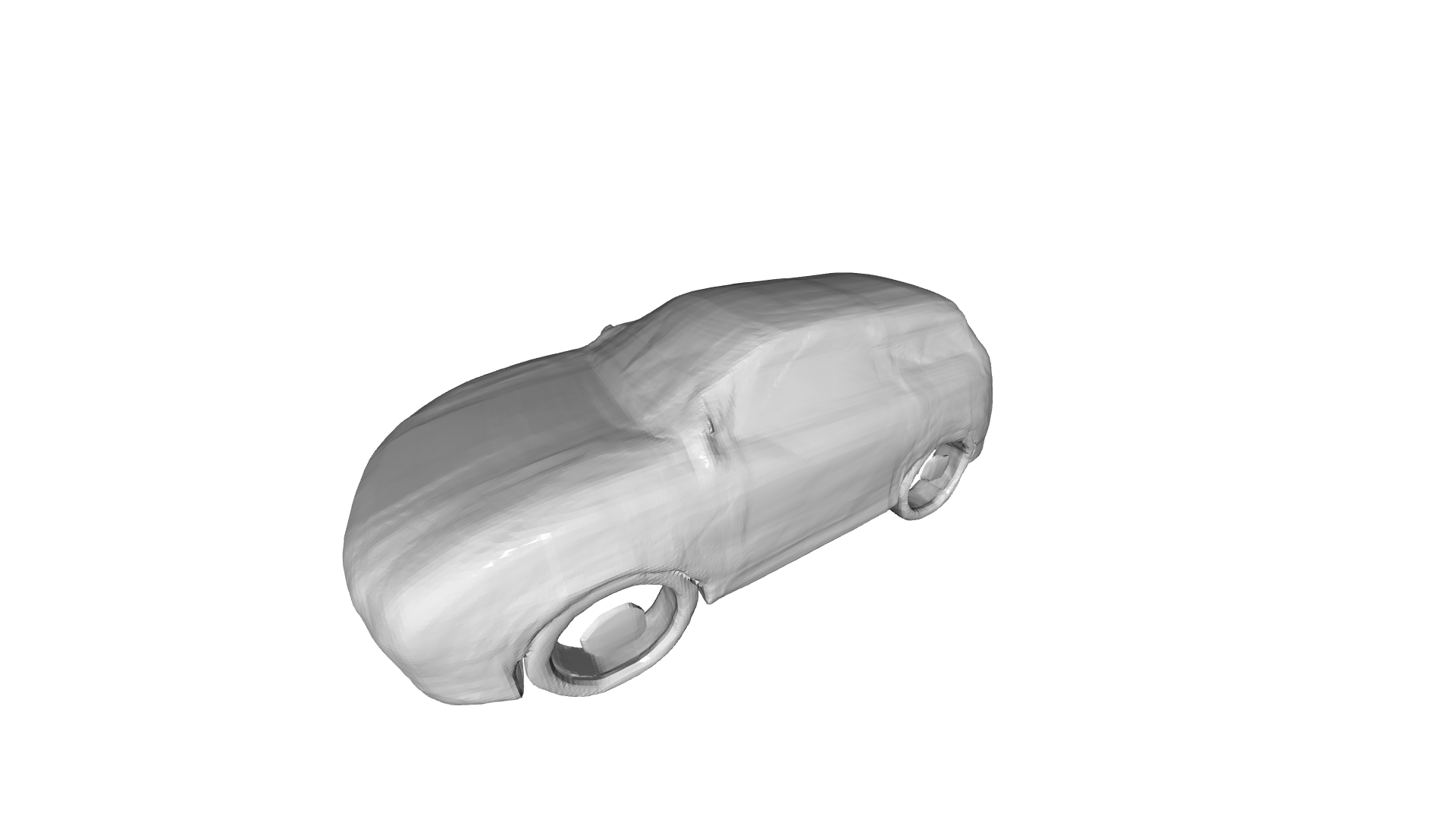}
    \end{subfigure}
    &
    \begin{subfigure}[b]{0.15\linewidth}
      \centering
      \includegraphics[width=\linewidth,trim={15cm 6.5cm 20cm 12cm},clip]{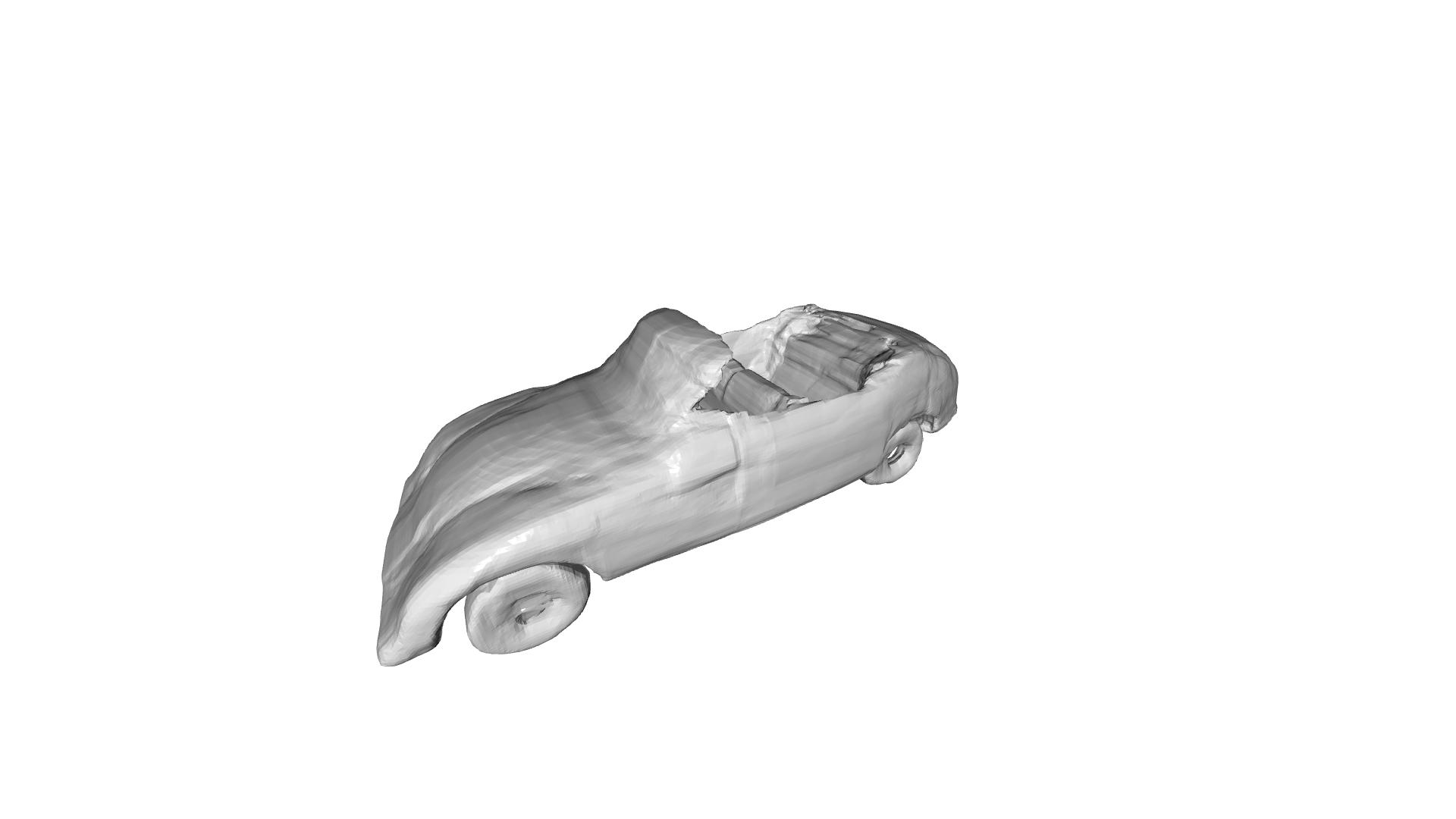}
    \end{subfigure}
    &
    \begin{subfigure}[b]{0.15\linewidth}
      \centering
      \includegraphics[width=\linewidth,trim={15cm 6.5cm 20cm 12cm},clip]{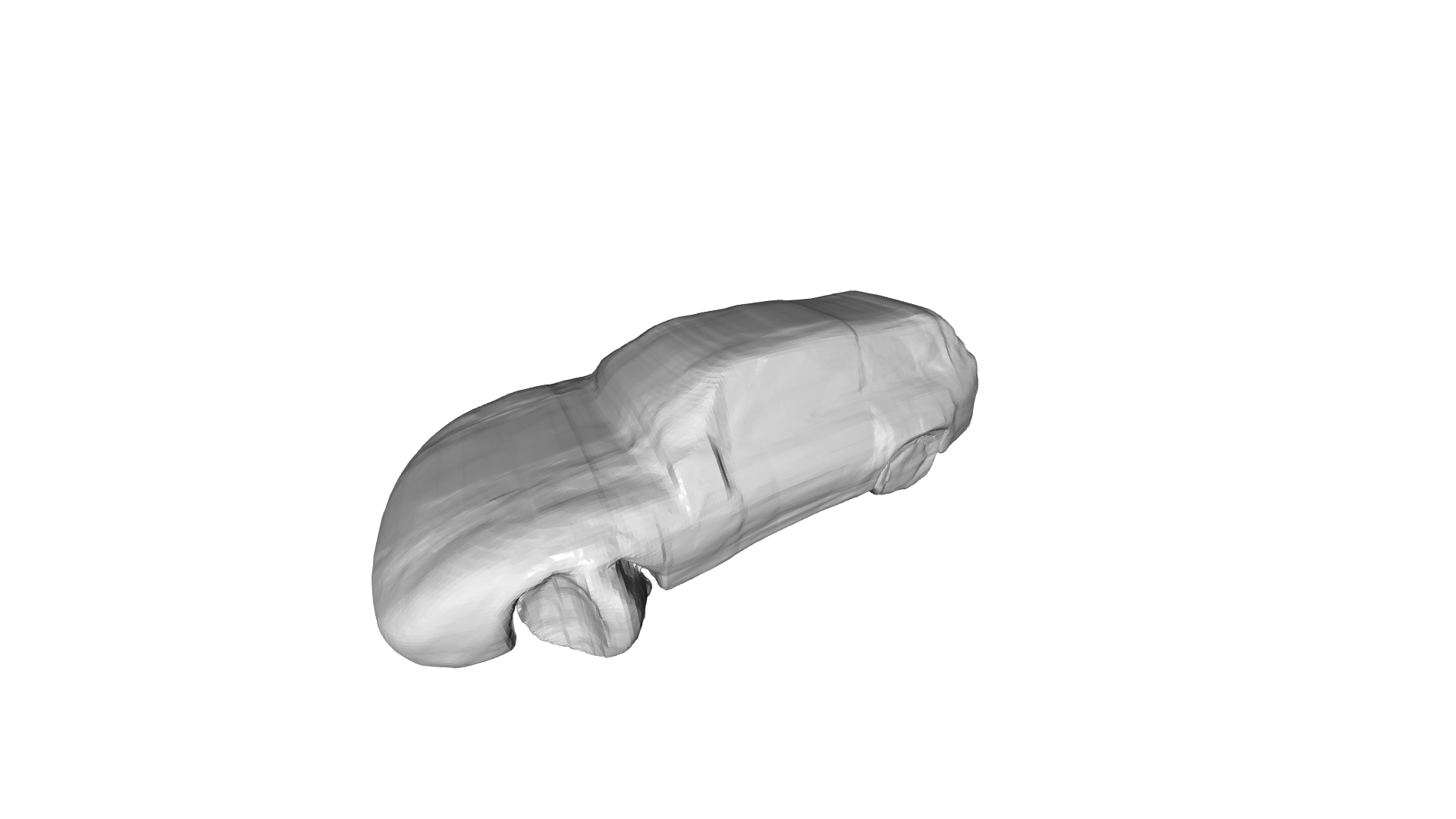}
    \end{subfigure}
    &
    \begin{subfigure}[b]{0.15\linewidth}
      \centering
      \includegraphics[width=\linewidth,trim={15cm 4.5cm 20cm 12cm},clip]{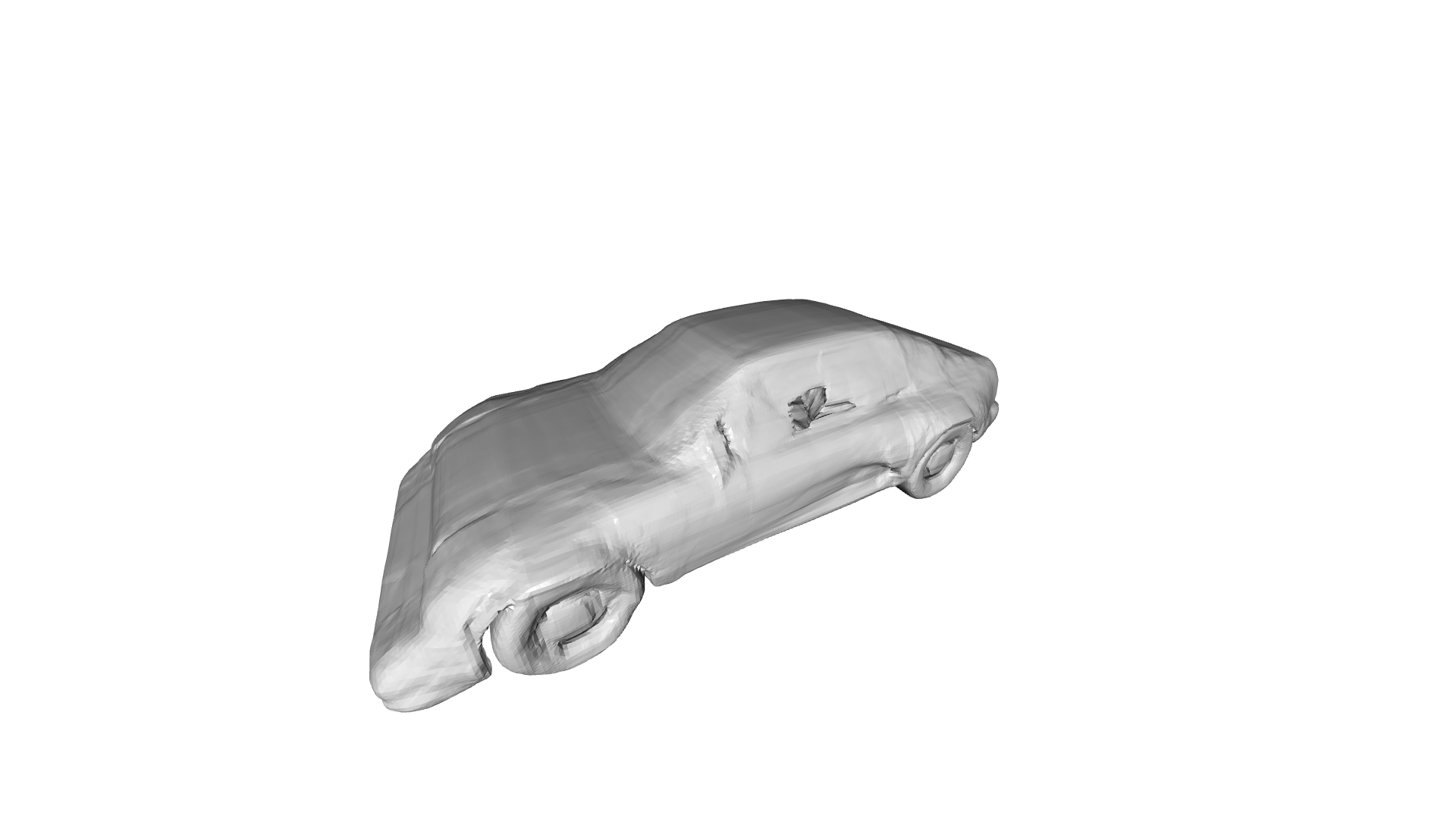}
    \end{subfigure}
    \\
    &  \textcolor{OrangeRed}{Score: 0.04} & \textcolor{OrangeRed}{Score: 0.0} & \textcolor{OrangeRed}{Score: 0.14}& \textcolor{OrangeRed}{Score: 0.10}& \textcolor{OrangeRed}{Score: 0.11}\\\midrule
    \thead{Retrieved \\ Natural Objects\\~\\~\\~}&
    \begin{subfigure}[b]{0.15\linewidth}
      \centering
      \includegraphics[width=\linewidth,trim={15cm 6cm 20cm 12cm},clip]{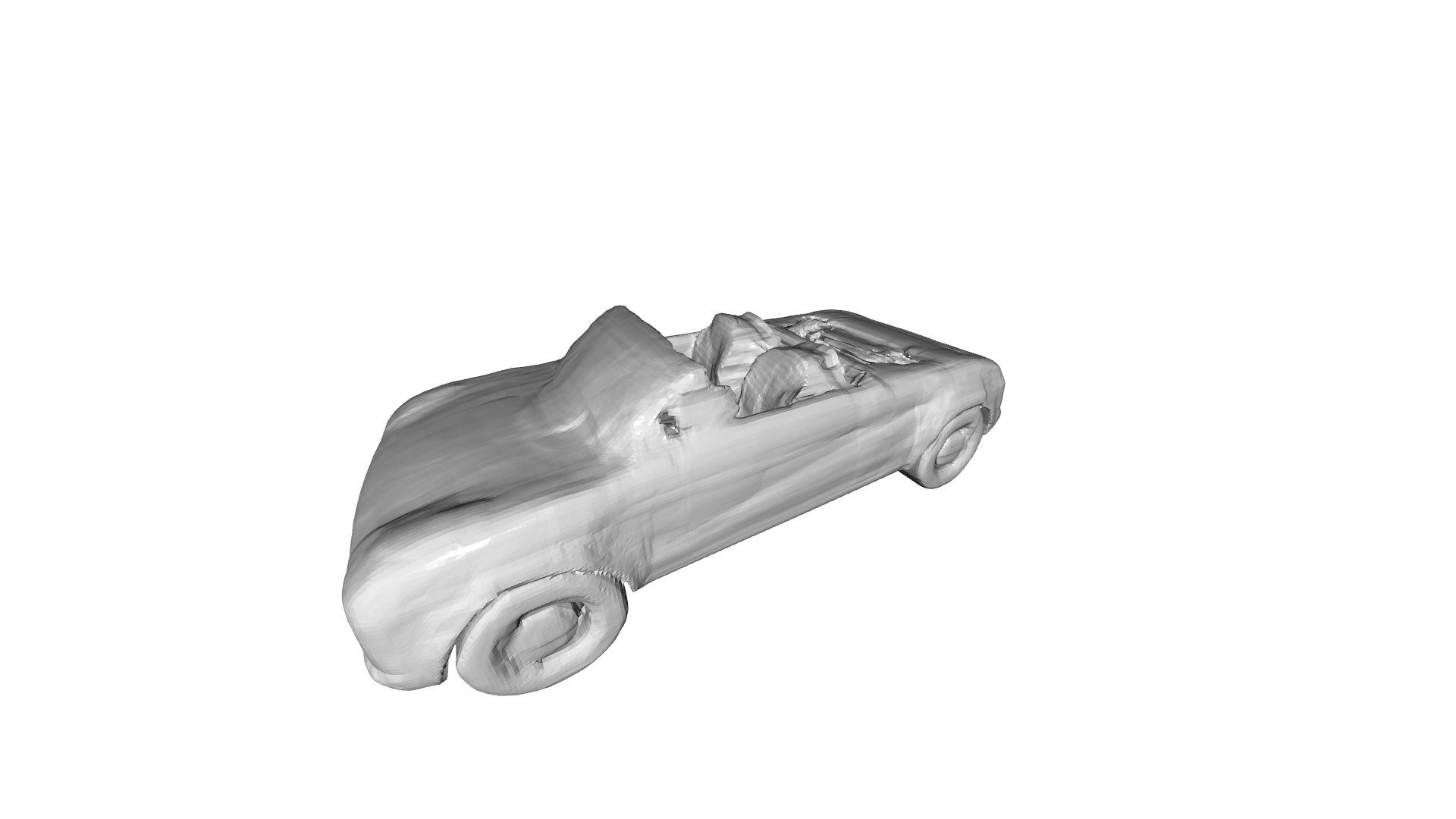}
    \end{subfigure}
    &
    \begin{subfigure}[b]{0.15\linewidth}
      \centering
      \includegraphics[width=\linewidth,trim={15cm 5cm 20cm 12cm},clip]{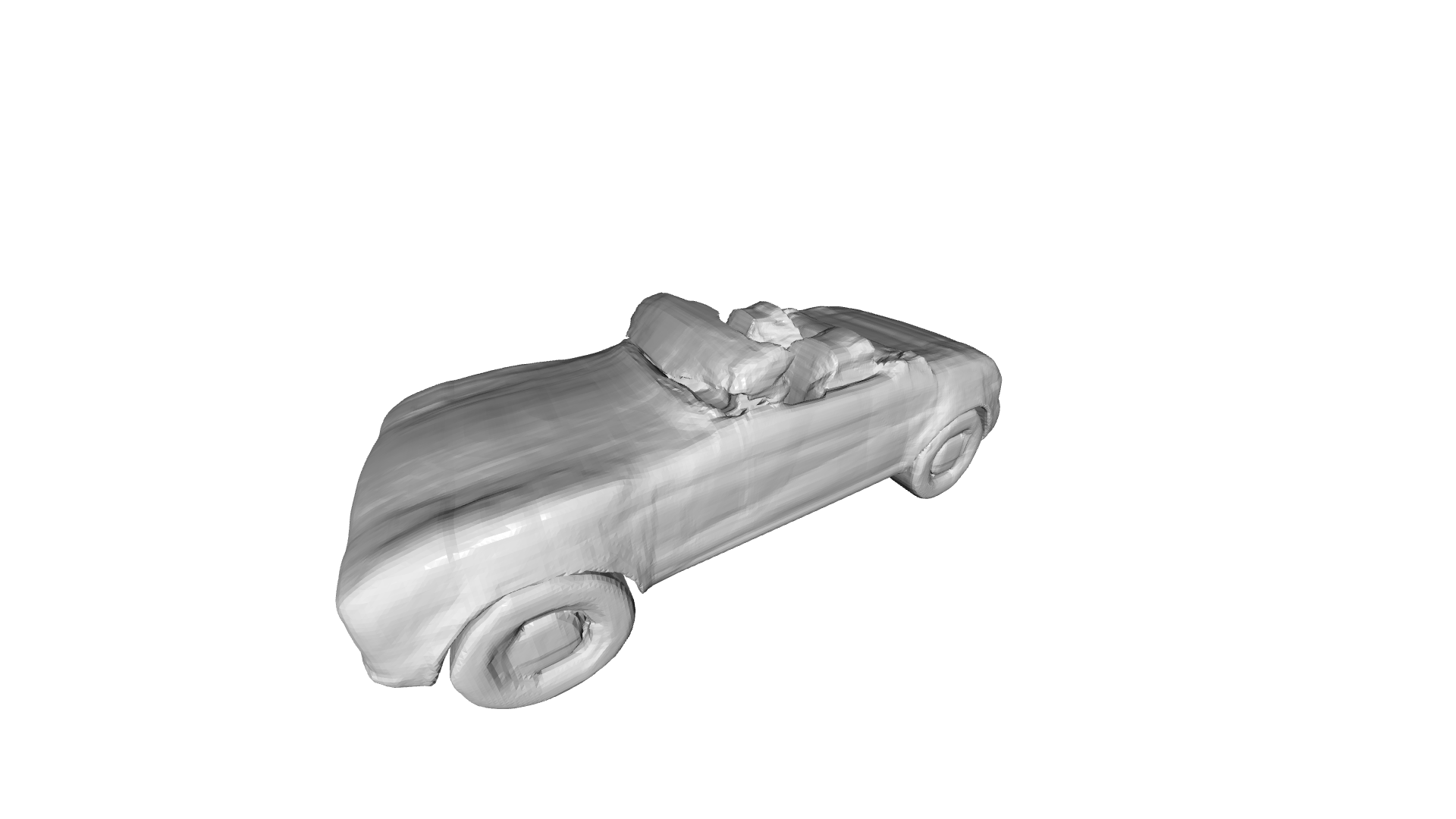}
    \end{subfigure}
    &
    \begin{subfigure}[b]{0.15\linewidth}
      \centering
      \includegraphics[width=\linewidth,trim={15cm 6.5cm 20cm 12cm},clip]{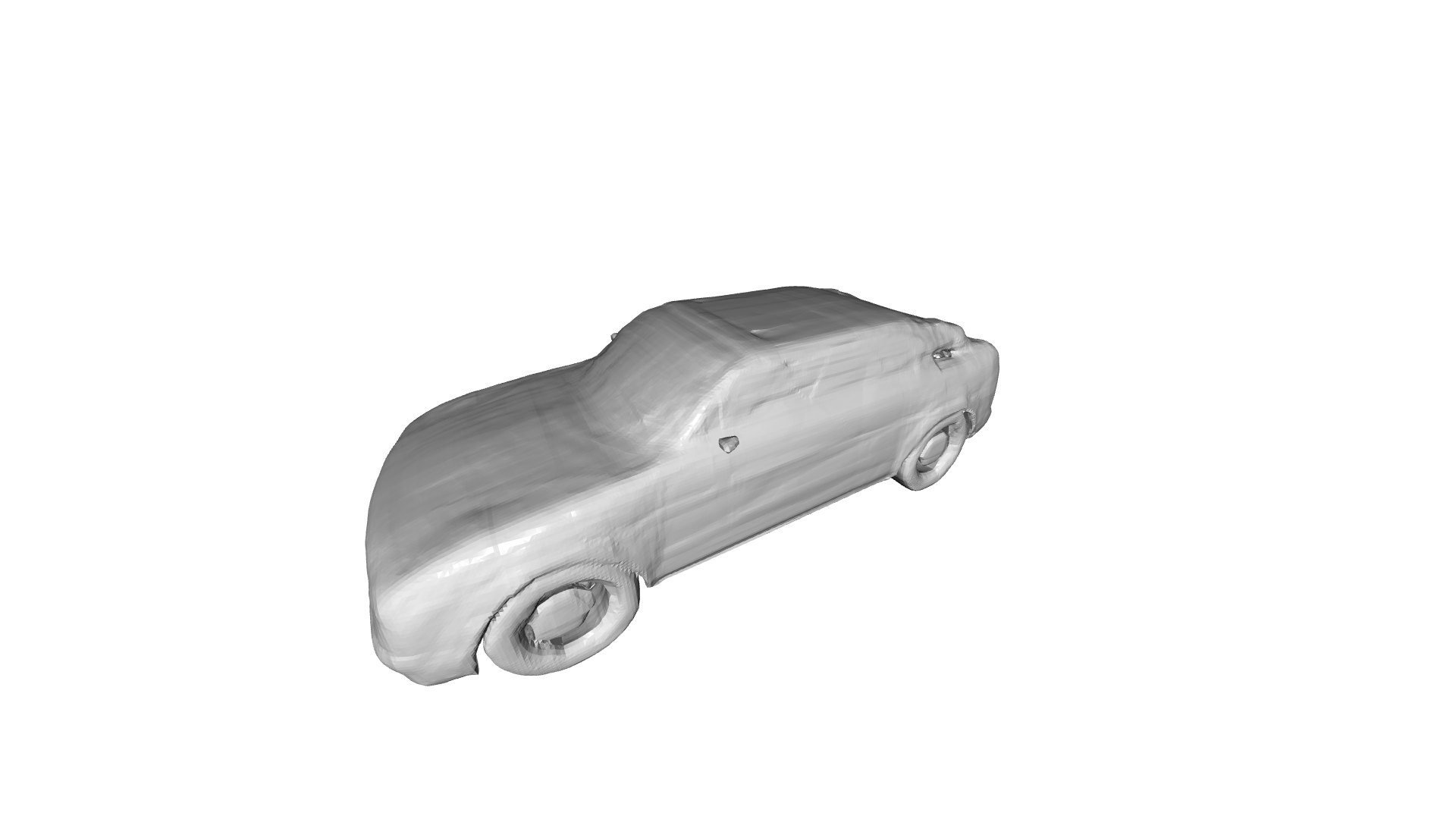}
    \end{subfigure}
    &
    \begin{subfigure}[b]{0.15\linewidth}
      \centering
      \includegraphics[width=\linewidth,trim={15cm 6.5cm 20cm 12cm},clip]{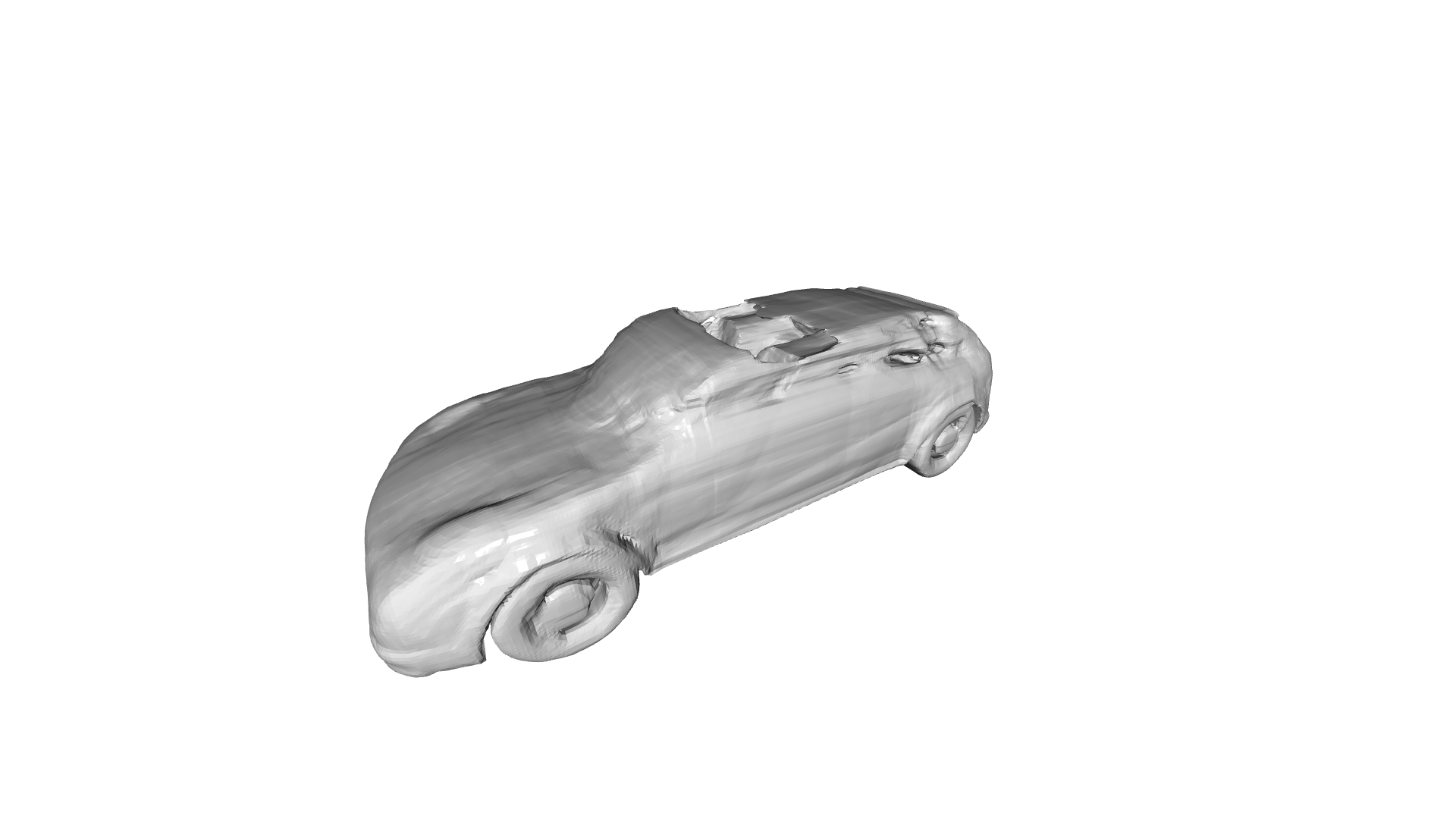}
    \end{subfigure}
    &
    \begin{subfigure}[b]{0.15\linewidth}
      \centering
      \includegraphics[width=\linewidth,trim={15cm 6cm 20cm 12cm},clip]{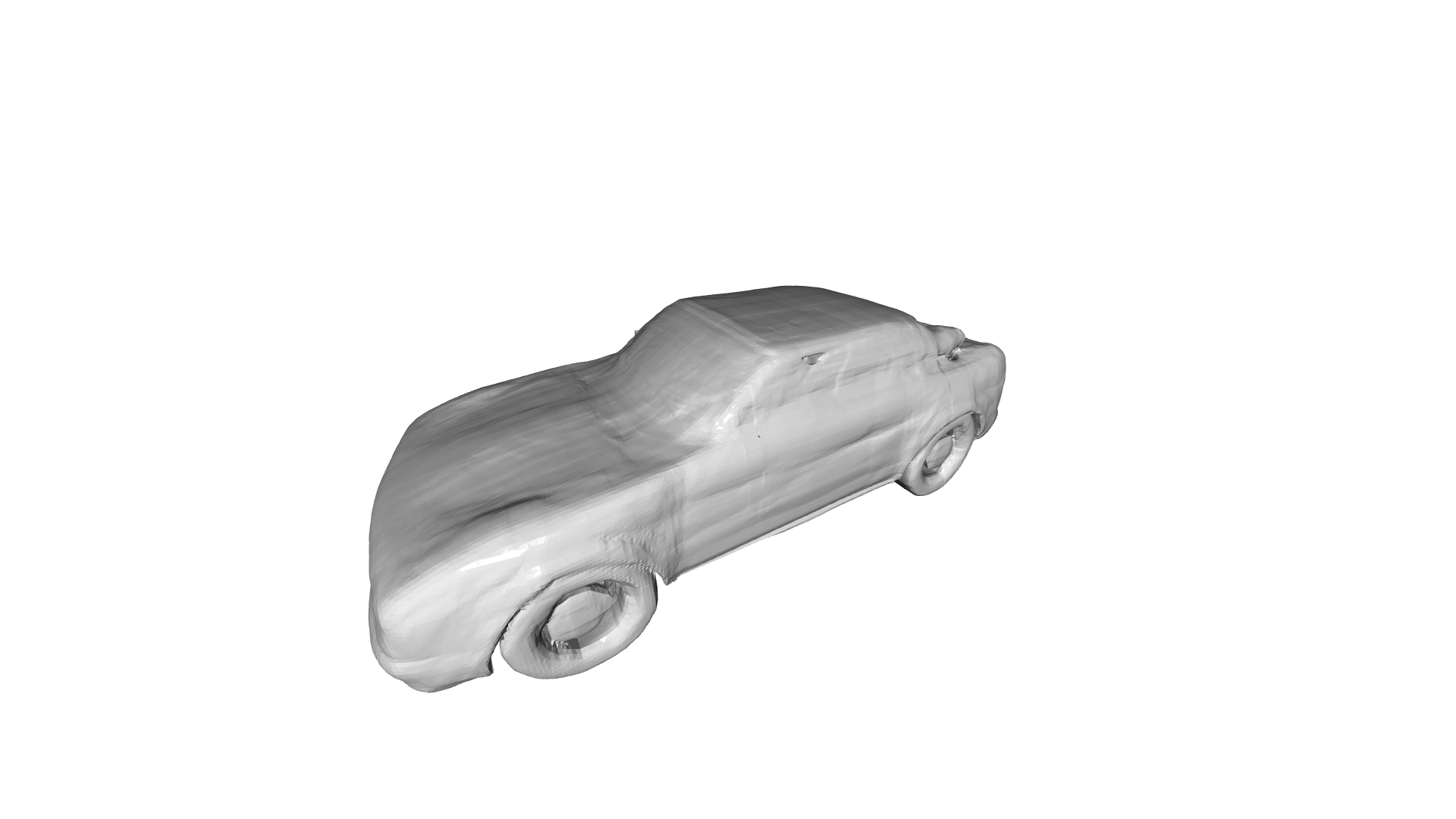}
    \end{subfigure}
    \\
    &  \textcolor{OrangeRed}{Score: 0.82} & \textcolor{OrangeRed}{Score: 0.07} & \textcolor{OrangeRed}{Score: 0.23}& \textcolor{OrangeRed}{Score: 0.40}& \textcolor{OrangeRed}{Score: 0.33}\\
   \bottomrule
  \end{tabular}
\end{figure*}

For evaluation purposes, we position the retrieved objects in identical poses to the \ourmethod{} objects to eliminate the influence of pose variation. The results, depicted as a threshold-recall curves, can be found in Figure~\ref{fig:retrieval_recall_threshold_curve}. Additionally, visual comparisons between \ourmethod{} objects and their natural counterparts are showcased in Figure~\ref{fig:retrieval_vis}. 

We note that some of the retrieved objects (e.g. in the latter columns) do look visually similar to our \ourmethod{} object, but some (e.g. the former columns) look quite visually different. We attribute this discrepancy to the limited nature of the retrieval dataset; \ourmethod{} will often generate examples that are not close to any example in the retrieval dataset, either because the retrieval dataset is small or not diverse enough, or because the \ourmethod{} object falls outside the real data distribution. However, generating OOD samples is very much one of the clear advantages of \ourmethod{}, as we can use these objects that are rare or impossible to find in the real world to learn lessons about our models' failure modes that would be hidden from us otherwise. More crucially, the retrieved objects always produce lower detection scores from our detector than the baseline objects. In other words, despite some discrepancies in visual similarity, using \ourmethod{} to retrieve real objects seems to consistently provide us with interesting examples that tend to fool our detector. And this application of \ourmethod{} is agnostic to these objects being labeled, so it can be applied to large, unlabeled datasets. 

\end{document}